\documentclass[notitlepage]{article}
\usepackage[T1]{fontenc}
\usepackage[utf8]{inputenc}

\usepackage[a4paper, total={6in, 9in}]{geometry}

\usepackage{authblk}

\usepackage{mathpazo}
\usepackage[scaled=0.90]{helvet}
\usepackage[scaled=0.85]{beramono}

\date{}

\usepackage[english]{babel}
\usepackage{booktabs}
\usepackage{appendix}
\usepackage[format=plain]{caption}
\usepackage{subcaption}
\usepackage{blkarray}
\usepackage{mleftright}
\usepackage{amsmath}
\usepackage{amssymb}
\usepackage{amsthm}
\usepackage[authoryear]{natbib}
\usepackage{multicol}
\usepackage{array}
\usepackage{bm}
\makeatletter
\newcommand{\pushright}[1]{\ifmeasuring@#1\else\omit\hfill$\displaystyle#1$\fi\ignorespaces}
\newcommand{\pushleft}[1]{\ifmeasuring@#1\else\omit$\displaystyle#1$\hfill\fi\ignorespaces}
\makeatother
\usepackage{cancel}
\usepackage{hyperref}
\usepackage{cleveref}
\usepackage{graphicx}
\usepackage[dvipsnames]{xcolor}
\usepackage{url}
\usepackage{adjustbox}
\usepackage{enumitem}
\def\eqref#1{equation~(\ref{#1})}
\def\Eqref#1{Equation~(\ref{#1})}

\newcommand{\ie}{\emph{i.e.}}
\newcommand{\eg}{\emph{e.g.}}
\newcommand{\etc}{\emph{etc}}

\renewcommand{\paragraph}[1]{\vspace{.5em}\noindent\textbf{#1.}}
\newcommand{\Paragraph}[1]{\noindent\textbf{#1.}}

\usepackage{dsfont}
\newcommand{\ind}{\mathds{1}}

\usepackage{mdframed}
\usepackage{nowidow}

\newtheoremstyle{ftheorem}%
  {-\topsep}%
  {}%
  {\normalfont}%
  {}%
  {\bfseries}%
  {.}%
  {.5em}%
  {}%
\newtheoremstyle{flemma}%
  {-\topsep}%
  {}%
  {\normalfont}%
  {}%
  {\bfseries}%
  {.}%
  {.5em}%
  {}%
\newtheoremstyle{fcorollary}%
  {-\topsep}%
  {}%
  {\normalfont}%
  {}%
  {\bfseries}%
  {.}%
  {.5em}%
  {}%
\newtheoremstyle{fproposition}%
  {-\topsep}%
  {}%
  {\normalfont}%
  {}%
  {\bfseries}%
  {.}%
  {.5em}%
  {}%
\newtheoremstyle{fremark}%
  {-\topsep}%
  {}%
  {\normalfont}%
  {}%
  {\bfseries}%
  {.}%
  {.5em}%
  {}%
\newtheoremstyle{fdefinition}%
  {-\topsep}%
  {}%
  {\normalfont}%
  {}%
  {\bfseries}%
  {.}%
  {.5em}%
  {}%
\newtheoremstyle{fexample}%
  {-\topsep}%
  {}%
  {\normalfont}%
  {}%
  {\bfseries}%
  {.}%
  {.5em}%
  {}%
\newtheoremstyle{fclaim}%
  {-\topsep}%
  {}%
  {\normalfont}%
  {}%
  {\bfseries}%
  {.}%
  {.5em}%
  {}%
\newtheoremstyle{ffact}%
  {-\topsep}%
  {}%
  {\normalfont}%
  {}%
  {\bfseries}%
  {.}%
  {.5em}%
  {}%
\newtheoremstyle{fassumption}%
  {-\topsep}%
  {}%
  {\normalfont}%
  {}%
  {\bfseries}%
  {.}%
  {.5em}%
  {}%
\newtheoremstyle{fsummary}%
  {-\topsep}%
  {}%
  {\normalfont}%
  {}%
  {\bfseries}%
  {.}%
  {.5em}%
  {}%

\theoremstyle{ftheorem}
\theoremstyle{flemma}
\theoremstyle{fcorollary}
\theoremstyle{fproposition}
\theoremstyle{fremark}
\theoremstyle{fdefinition}
\theoremstyle{fexample}
\theoremstyle{fclaim}
\theoremstyle{ffact}
\theoremstyle{fassumption}
\theoremstyle{fsummary}

\newmdtheoremenv[backgroundcolor=gray!17,linewidth=6pt,linecolor=gray!20]{ftheorem}{Theorem}[section]
\newmdtheoremenv[backgroundcolor=gray!17,linewidth=6pt,linecolor=gray!20]{flemma}{Lemma}[section]
\newmdtheoremenv[backgroundcolor=gray!17,linewidth=6pt,linecolor=gray!20]{fcorollary}{Corollary}[section]
\newmdtheoremenv[backgroundcolor=gray!17,linewidth=6pt,linecolor=gray!20]{fproposition}{Proposition}[section]
\newmdtheoremenv[backgroundcolor=gray!17,linewidth=6pt,linecolor=gray!20]{fremark}{Remark}[section]
\newmdtheoremenv[backgroundcolor=gray!17,linewidth=6pt,linecolor=gray!20]{fdefinition}{Definition}[section]
\newmdtheoremenv[backgroundcolor=gray!17,linewidth=6pt,linecolor=gray!20]{fexample}{Example}[section]
\newmdtheoremenv[backgroundcolor=gray!17,linewidth=6pt,linecolor=gray!20]{fclaim}{Claim}[section]
\newmdtheoremenv[backgroundcolor=gray!17,linewidth=6pt,linecolor=gray!20]{ffact}{Fact}[section]
\newmdtheoremenv[backgroundcolor=gray!17,linewidth=6pt,linecolor=gray!20]{fassumption}{Assumption}[section]
\newmdtheoremenv[backgroundcolor=gray!17,linewidth=6pt,linecolor=gray!20]{fsummary}{Summary}[section]

\title{An Introduction to Deep Survival Analysis Models for Predicting Time-to-Event Outcomes}

\author{George H.~Chen}
\affil[1]{Heinz College of Information Systems and Public Policy, Carnegie Mellon University}

\begin{document}

\sloppy

\maketitle{}

\begin{abstract}
Many applications involve reasoning about time durations before a critical event happens---also called \emph{time-to-event outcomes}. When will a customer cancel a subscription, a coma patient wake up, or a convicted criminal reoffend? Accurate predictions of such time durations could help downstream decision-making tasks. A key challenge is \emph{censoring}: commonly, when we collect training data, we do not get to observe the time-to-event outcome for every data point. For example, a coma patient has not woken up yet, so we do not know the patient's time until awakening. However, these data points should not be excluded from analysis as they could have characteristics that explain why they have yet to or might never experience the event.

Time-to-event outcomes have been studied extensively within the field of \emph{survival analysis} primarily by the statistical, medical, and reliability engineering communities, with textbooks already available in the 1970s and '80s. Recently, the machine learning community has made significant methodological advances in survival analysis that take advantage of the representation learning ability of deep neural networks. At this point, there is a proliferation of deep survival analysis models. How do these models work? Why? What are the overarching principles in how these models are generally developed? How are different models related?

This monograph aims to provide a reasonably self-contained modern introduction to survival analysis. We focus on predicting time-to-event outcomes at the individual data point level with the help of neural networks. Our goal is to provide the reader with a working understanding of precisely what the basic time-to-event prediction problem is, how it differs from standard regression and classification, and how key ``design patterns'' have been used time after time to derive new time-to-event prediction models, from classical methods like the Cox proportional hazards model to modern deep learning approaches such as deep kernel Kaplan-Meier estimators and neural ordinary differential equation models. We further delve into two extensions of the basic time-to-event prediction setup: predicting which of several critical events will happen first along with the time until this earliest event happens (the competing risks setting), and predicting time-to-event outcomes given a time series that grows in length over time (the dynamic setting). We conclude with a discussion of a variety of topics such as fairness, causal reasoning, interpretability, and statistical guarantees.

Our monograph comes with an accompanying code repository that implements every model and evaluation metric that we cover in detail: \mbox{\texttt{\url{https://github.com/georgehc/survival-intro}}}
\end{abstract}

\newpage

\tableofcontents

\newpage

\section{Introduction}
\label{chap:intro}

Predicting time durations before a critical event happens arises in numerous applications. These durations are called \emph{time-to-event outcomes}. For example, an e-commerce company may be interested in predicting a user's time until making a purchase (\eg, \citealt{chapelle2014modeling}). A video streaming service may be interested in predicting when a customer will stop watching a show (\eg, \citealt{hubbard2021beta}). In healthcare, hospitals may be interested in predicting when a patient's disease will relapse (\eg, \citealt{zupan2000machine}). In criminology, courts may be interested in predicting the time until a convicted criminal reoffends (\eg, \citealt{chung1991survival}). Accurate predictions for these time-to-event outcomes could help in decision-making tasks such as showing targeted advertisements or promotions in the e-commerce or video streaming examples, planning treatments to reduce a patient's risk of disease relapse in the healthcare example, and making bail decisions in the criminology example.

A defining feature of time-to-event prediction problems is that commonly, when we collect training data to learn a model from, we do not get to see the true time-to-event outcome for every data point, \ie, their time-to-event outcome is \emph{censored}. As an example, some training points (\eg, a coma patient) might not have experienced the critical event of interest yet (\eg, waking up). Discarding the points that have not experienced the event would be unwise: they could have characteristics that make them much less likely to experience the event (\eg, the patient's brain activity is highly abnormal).

\subsection{Survival Analysis and Time-to-Event Outcomes: Some History and Commentary on Naming}
\label{sec:history}

Time-to-event outcomes have been studied for hundreds of years if not longer, where the initial focus was on predicting time until death. Early analyses introduced the use of ``life tables'', which in a nutshell contain counts such as numbers of births and deaths over time. \citet{graunt1662natural} published what might be the first life table and looked at the chance of survival for different age groups.\footnote{As noted by \citet{glass1963john} and \citet{bacaer2011short} among others, there has been some debate as to whether Graunt or his friend William Petty wrote the book but regardless, Graunt's book had five editions published between 1662 and 1676 (for which our citation just uses the earliest year).} This particular dataset from London was challenging since the survival times (age at the time of death) were not actually recorded, corresponding to a censoring problem. Instead, Graunt largely guessed survival times based on causes of death, which were recorded (albeit they were not necessarily accurate). A few decades later, \citet{halley1693estimate} analyzed a life table collected from modern day Wroc\l{}aw and computed the probability of dying within the next year. Halley used these probabilities to determine how to price an annuity (roughly, an expected payout over a person's remaining lifetime). For a historical account of life tables and more generally reasoning about survival times, see for instance the book by \citet{bacaer2011short} and the bibliographical notes accompanying the different chapters of \citet{namboodiri2013life}---these readings altogether walk through highlights from hundreds of years of research on survival times leading up to modern time.

So much of the pioneering research on time-to-event outcomes was on time until death that the enterprise of modeling time-to-event outcomes is now commonly called \emph{survival analysis}, with textbooks already available decades ago (\eg, \citealt{mann1974methods,kalbfleisch1980statistical,cox1984analysis,fleming1991counting}). Countless other (text)books on survival analysis have since been written and have mainly originated from statistical, medical, and reliability engineering communities (\eg, \citealt{klein2003survival,machin2006survival,selvin2008survival,kleinbaum2012survival,li2013survival,harrell2015regression,klein2016handbook,ebeling2019introduction,prentice2019statistical,gerds2021medical,collett2023modelling}), and at this point, there is also a book tailored to social scientists \citep{box2004event}.

We want to emphasize though that as the examples we opened the monograph with showed, the critical event need not be death, meaning that we might not be reasoning about ``survival'' literally. In fact, some researchers work on survival analysis but in titling their papers choose to opt for more general phrasing such as ``time-to-event modeling'' (\eg, \citealt{chapfuwa2018adversarial}). We further emphasize that the ``time'' in ``time-to-event outcome'' does not literally have to measure time. For example, a survival analysis model could be used to predict how many units of an inventory item (\eg, a newspaper) to stock the next day given past days' sales counts, so that the ``time-to-event outcome'' here measures an integer number of items \citep{huh2011adaptive}. Ultimately, ``survival'' analysis or ``time-to-event'' models have been broadly applied to numerous applications far beyond reasoning about either ``survival'' or ``time-to-event'' outcomes in a literal sense.

\subsection{Machine Learning Models for Survival Analysis}

The phrase ``machine learning'' was only coined in 1959 \citep{samuel1959some}, the year after the highly influential paper by \citet{kaplan1958nonparametric} came out that analyzed a survival model based on life tables using what is called the ``product-limit'' estimator \citep{bohmer1912theorie}. (Kaplan and Meier's estimator remains one of the major workhorses of modern time-to-event data analysis; we will see it and deep learning versions of it later in this monograph.) Suffice it to say, machine learning as a field is young compared to survival analysis. Precisely when the first machine learning survival analysis model came about is perhaps not entirely straightforward to trace, in part because nowadays, what is considered a ``machine learning model'' depends on who one asks. While we may consider $k$ nearest neighbor and kernel survival analysis \citep{beran1981nonparametric} and survival trees \citep{ciampi1981approach,gordon1985tree} to be machine learning models, would the authors of these original papers?

Fast-forwarding to present time, there is now an explosion in the number of machine learning survival analysis models available.
For much larger lists of models than what we cover in this monograph, see the excellent surveys by \citet{wang2019machine} and \citet{wiegrebe2023deep}. As part of their survey, Wiegrebe \emph{et al.}~provide an online catalog of over 60 deep-learning-based survival models (which we will just abbreviate throughout this monograph as \emph{deep survival models}).\footnote{\texttt{\url{https://survival-org.github.io/DL4Survival/}}} This catalog is not exhaustive!

With so many machine learning models for survival analysis, what exactly are the major innovations? When and why do different models work? How do they relate to each other? What are overarching patterns in model development? In answering these questions, we think that it is extremely important to distinguish between innovations that are specific to time-to-event prediction vs ones that are not. For the purposes of this monograph, we want to focus on the former as they could help us better understand what is special about time-to-event prediction that helps us build better models.

\subsection{The Motivation for This Monograph}

We set out to write this monograph for two key reasons:
\begin{itemize}

\item
First, we wanted to provide a reasonably self-contained introductory text that covers the key concepts of survival analysis with a focus on time-to-event prediction \emph{at the individual data point level} and that also exploits the availability of now standard neural network software. We focus on neural network survival models (\ie, deep survival models) because these models are easy to modify (\eg, to accommodate different data modalities, add loss terms, set a custom learning rate schedule, \etc). Note that every model that we present in detail has publicly available source code (we discuss software shortly in Section~\ref{sec:software}). For readers who are new to survival analysis but are already very comfortable working with standard neural network software at the level of writing custom models and loss functions, we hope that our monograph provides enough survival analysis background to make implementing deep survival models from ``scratch'' using standard neural network software fairly straightforward.

\item
Second, we wanted to clearly convey how several major categories of deep survival models are related, and how in deriving these different survival models, we use some of the same key design patterns or derivation techniques over and over again. We hope that by leading the reader through many examples, these patterns will become apparent.

\end{itemize}
To the best of our knowledge, no existing text provides the sort of introduction to survival analysis that our monograph aims to be. The surveys of machine learning survival models \citep{wang2019machine,wiegrebe2023deep} are not written nor intended for the purpose of giving the reader a working knowledge of how to actually derive survival models from first principles. Meanwhile, the vast majority of survival analysis (text)books do not cover neural networks or deep learning due to how new these are (an example of a textbook that covers neural networks for survival analysis can be found in Chapter~11 of \citet{dybowski2001clinical}, but this book pre-dates the invention of nearly all the deep survival models we cover).

Overall, we hope that our introduction to survival analysis provides the reader with a solid understanding of what precisely the time-to-event problem setup is, why it is different from standard regression and classification, and how to build survival models with the help of neural networks. We also hope that the reader learns a little bit about where the state-of-the-art is in terms of a variety of other topics that we mention but do not discuss in detail, such as how fairness, causal reasoning, and interpretability play into survival models, and what progress has been made on theoretically analyzing some of these models.

\subsection{Monograph Overview and Outline}
\label{sec:scope}

\begin{figure}[!p]
\centering
\includegraphics[width=.7\linewidth]{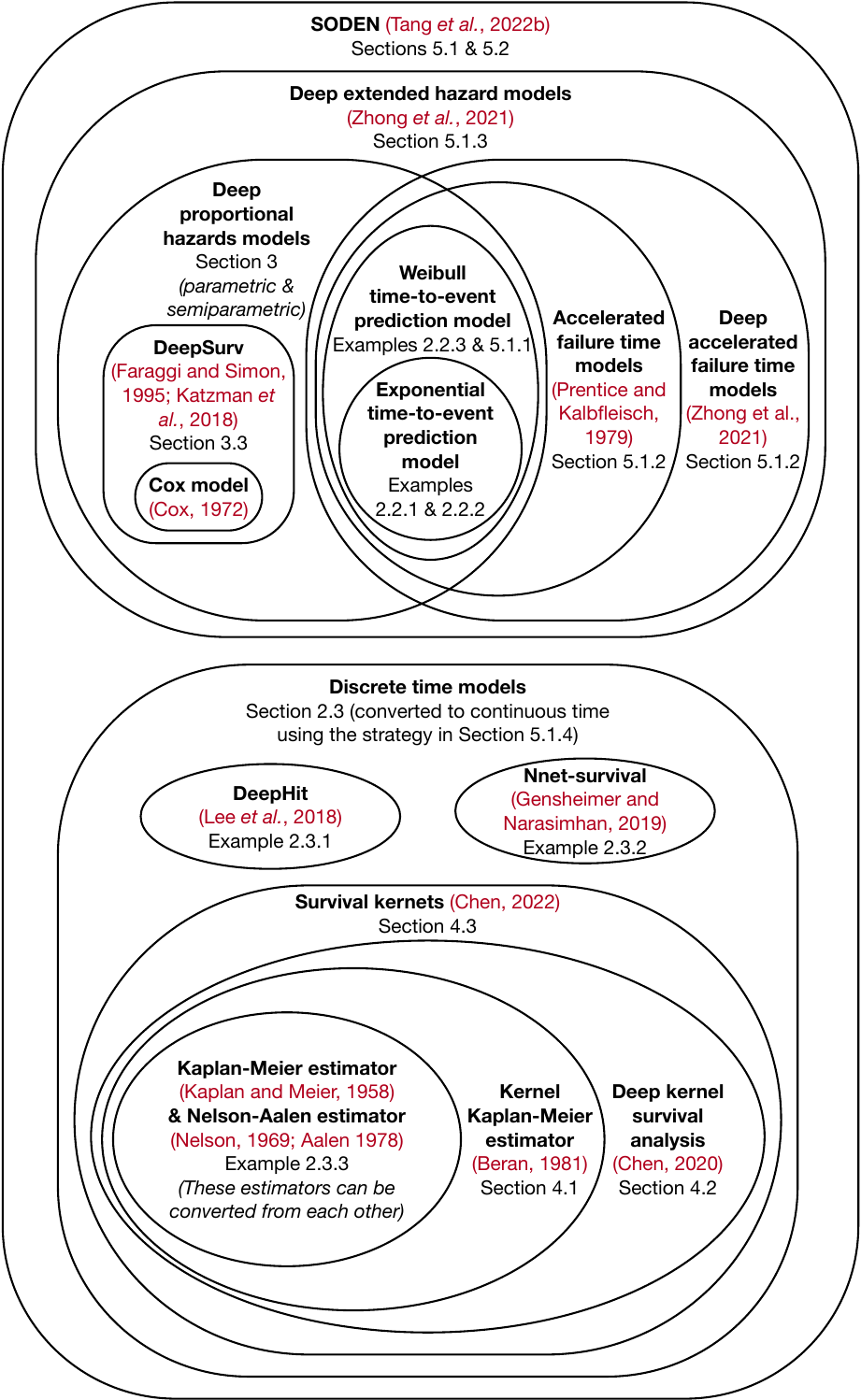}
\caption{An overview of the models we cover in detail in Sections~\ref{chap:setup} to~\ref{chap:ode}. One model being the child of another means that the child model could be represented (possibly with a known approximation) by the parent model. Note that when interpreting this diagram, two non-overlapping models could still possibly represent the same underlying time-to-event outcome distribution. For example, deep extended hazard models \citep{zhong2021deep} and survival kernets \citep{chen2024survival} are capable of modeling many of the same time-to-event outcome distributions. Note that we also cover \mbox{Cox-Time} \citep{kvamme2019time}, which does not easily fit in the diagram; \mbox{Cox-Time} is a generalization of the semiparametric model called DeepSurv \citep{faraggi1995neural,katzman2018deepsurv}, but \mbox{Cox-Time} can also represent models that are not deep extended hazard models.}
\label{fig:overview-of-monograph-models}
\end{figure}

Our coverage is not meant to remotely be exhaustive in showcasing how deep survival models have been used for time-to-event prediction.
We specifically cover the following:
\begin{itemize}

\item \textbf{Basic Time-to-Event Prediction Setup (Section~\ref{chap:setup}).} We first go over the standard time-to-event prediction problem setup. We state its statistical framework, its prediction task, common ways of writing a likelihood function to be maximized (maximum likelihood is the standard way of learning time-to-event prediction models), and how to evaluate prediction accuracy. Along the way, we lead the reader through various example models to help solidify concepts, all of which could be related to maximizing likelihood functions: exponential and Weibull time-to-event prediction models, DeepHit \citep{lee2018deephit}, Nnet-survival \citep{gensheimer2019scalable}, the Kaplan-Meier estimator \citep{kaplan1958nonparametric}, and the Nelson-Aalen estimator \citep{nelson1969hazard,aalen1978nonparametric}. Importantly, we distinguish between modeling time as continuous vs discrete since the math involved is a bit different. This section also discusses how time-to-event prediction relates to classification and regression.

\item \textbf{Deep Proportional Hazards Models (Section~\ref{chap:proportional-hazards}).} We next cover perhaps the most widely used family of time-to-event prediction models in practice, which are called \emph{proportional hazards models}. We define proportional hazards models in a general manner in terms of neural networks. Special cases include the exponential and Weibull models from Section~\ref{chap:setup}, the classical Cox model \citep{cox1972regression}, and DeepSurv \citep{faraggi1995neural,katzman2018deepsurv}. Proportional hazards models make a strong assumption that, in some sense, decouples how time contributes to a prediction and how a data point's features contribute to a prediction. This assumption often does not hold in practice. We present a generalization of the DeepSurv model called Cox-Time \citep{kvamme2019time} that removes this proportional hazards assumption.

\item \textbf{Deep Conditional Kaplan-Meier Estimators (Section~\ref{chap:deep-kaplan-meier}).} One of the standard models we encounter in Section~\ref{chap:setup} is the Kaplan-Meier estimator, which is extremely popular in practice and also different from deep proportional hazards models because it is \emph{nonparametric} (\ie, it does not assume the time-to-event outcome's distribution has a parametric form). However, it only works to describe a population and does not provide predictions for individual data points. We present deep learning versions of the Kaplan-Meier estimator that can make predictions at the individual level. Namely, we cover deep kernel survival analysis \citep{chen2020deep} and its generalization called survival kernets \citep{chen2024survival}; the latter can scale to large datasets, can in some sense be interpreted in terms of clusters, and has a statistical guarantee on accuracy for a special case of the model.

\item \textbf{Neural Ordinary Differential Equation Formulation of Time-to-Event Prediction (Section~\ref{chap:ode}).} We then present a model that can encode all the models we presented in preceding sections, where we phrase the standard time-to-event prediction problem in terms of a \emph{neural ordinary differential equation} model. We specifically go over the neural ODE time-to-event prediction model by \citet{tang2022soden} called SODEN. In presenting SODEN, we also mention some model classes that we did not previously point out, such as deep accelerated failure time models and deep extended hazard models \citep{zhong2021deep}.

\item \textbf{Beyond the Basic Time-to-Event Prediction Setup: Multiple Critical Events and Time Series as Raw Inputs (Section~\ref{chap:extensions}).} Whereas all the previous sections used the basic time-to-event prediction setup from Section~\ref{chap:setup}, we now consider two generalizations. First we consider the so-called \emph{competing risks setting} where there are multiple critical events of interest (\eg, for a coma patient, we consider the patient waking up and the patient dying as two different critical events; note that censoring could still happen but is not considered as one of the critical events). We aim to predict \emph{which critical event will happen first} and also the \emph{time until this earliest critical event happens}. The example model we use here is DeepHit \citep{lee2018deephit}. Note that the special case of there being one critical event reduces the problem to the one from Section~\ref{chap:setup}. We then generalize the competing risks setting further by considering what happens when we want to make predictions as we see more and more of a given a time series (the dynamic setting). The example model we use here is Dynamic-DeepHit \citep{lee2019dynamic}.

\item \textbf{Discussion (Section~\ref{chap:discussion}).} We end the monograph by discussing a variety of topics that we either only briefly glossed over or that we did not mention at all. For example, we discuss different kinds of censoring, ways to encourage a survival model to be ``fair'', causal reasoning with survival models, interpretability of deep survival models, issues of statistical guarantees, and more.

\end{itemize}
Specifically for the example models we cover in Sections~\ref{chap:setup} to~\ref{chap:ode}, we show how these models relate in Figure~\ref{fig:overview-of-monograph-models}. When one model is a child of another in this figure, it means that the child model could be represented (possibly with a known approximation) by the parent model. Note that in the figure, just because two models do not overlap does \emph{not} mean that they cannot represent the same underlying true time-to-event outcome distribution. For example, even though deep extended hazard models \citep{zhong2021deep} and survival kernets \citep{chen2024survival} do not overlap in the figure, they can represent many of the same time-to-event outcome distributions.

We emphasize that just because SODEN \citep{tang2022soden} can in principle represent all the other models we cover in Sections~\ref{chap:setup} to~\ref{chap:deep-kaplan-meier} (possibly with an approximation), that does not mean that one is best off just using SODEN. An important point is that many of these models are trained in different ways. SODEN's training procedure may not work the best for some of the simpler model classes that it can represent. In particular, it invokes calls to an ordinary differential equation (ODE) solver, which could be overkill if we just want to use one of the simpler models (that has its own simpler training procedure which typically is faster to run). By relying on an ODE solver, we could also run into numerical stability issues that occasionally arise with ODE solvers.

Separately, it is good to keep in mind that deep survival models that we cover allow the modeler to flexibly choose a ``base'' neural network to use with the model. For example, when working with tabular data, the modeler could choose the base neural network to be a multilayer perceptron. When working with images, the modeler could choose the base neural network to be a convolutional neural network or a vision transformer. And so forth. Such base neural networks could be chosen to be arbitrarily complicated (\eg, we could use as many layers and as many hidden nodes as we would like). Consequently, many deep survival models could in theory be considered equally expressive in what sorts of time-to-event outcomes they can model. However, in practice, training these deep models often requires using standard neural network tricks such as using early stopping, weight decay, dropout, \etc. Roughly, we would train these models with some regularization to prevent overfitting, and how this regularization impacts different models then depends on their different modeling assumptions.

We remark that the example models we chose to present in this monograph are not necessarily the ``best''. Instead, they were chosen largely for pedagogical considerations and also to showcase some classes of models that are quite different from one another. There are plenty of other time-to-event prediction models (deep-learning-based or not) that work well! By understanding the fundamental concepts in our monograph, the reader should be well-equipped to understand many of these other models. As a reminder, the surveys by \citet{wang2019machine} and \citet{wiegrebe2023deep} provide fairly extensive listings of many existing machine learning models for time-to-event prediction.

\subsection{Examples of Topics Beyond the Scope of Our Monograph}
To elaborate a bit more on our scope of coverage, our monograph focuses on survival models for \emph{prediction} that are estimated via \emph{maximum likelihood estimation} in a \emph{neural network framework}. We acknowledge that many survival analysis methods were originally derived for the purpose of statistical inference (\eg, reasoning about population-level quantities and constructing confidence intervals for these quantities) rather than for prediction, including the classical Kaplan-Meier estimator \citep{kaplan1958nonparametric} and the Cox model \citep{cox1972regression}. Our emphasis in this monograph, however, is on learning survival models for prediction, as this is what neural survival models currently are well-suited for. As such, we typically will not cover how to address questions of statistical inference. For readers interested in learning more about statistical inference with survival models, we point out that the various (text)books mentioned in Section~\ref{sec:history} cover statistical inference results for classical models.\footnote{As a concrete example, the textbook by \citet{klein2003survival} routinely explains how to construct confidence intervals for various estimated quantities, such as confidence intervals for survival functions obtained from the Kaplan-Meier estimator (see Section 4.3 of their book) and the Cox model (see Section 8.8 of their book).}

Next, many survival models are learned in a manner that is fundamentally not based on maximum likelihood estimation. For example, countdown regression \citep{avati2020countdown} defines a score function to optimize that does not correspond to the usual survival likelihood used in the literature. Meanwhile, \citet{chapfuwa2018adversarial} train a survival model using adversarial learning (with a generative adversarial network \citep{goodfellow2014generative}) rather than maximum likelihood. As another example, random survival forests \citep{ishwaran2008random} are trained in a greedy manner, where one cannot easily write down a global objective function that is being optimized.

In fact, many decision tree survival models are not optimized in a neural network framework at all, such as the aforementioned random survival forests \citep{ishwaran2008random} as well as XGBoost \citep{chen2016xgboost} (note that the official implementation of XGBoost supports survival analysis), optimal survival trees \citep{bertsimas2022optimal}, or optimal sparse survival trees \citep{zhang2024optimal}. While it is possible to set up learning a decision tree survival model in a neural network framework \citep{sun2023nsotree}, at the time of writing, this line of research appears to still be in early development.

For ease of exposition, we do not cover latent variable models for survival analysis (\eg, \citealt{nagpal2021dsm,nagpal2021deep,manduchi2022deep,moon2022survlatent,chen2024neural}). These particular models build on the ideas we present in this monograph and further use tools not covered in this monograph, notably that of variational inference (\eg, \citealt{blei2017variational}). We think that a reader who understands the fundamentals of our monograph and of variational inference should be well-versed in understanding latent variable models for survival analysis.

\subsection{Preliminaries}

Before we move onto other sections, we go over some prerequisite knowledge that we will assume that the reader is familiar with. We then explain how we view neural networks and the notation that we use throughout the rest of the monograph. At the end of this section, we provide links to some available software packages and to our companion code repository for this monograph.

\subsubsection{Prerequisites}

We assume that the reader knows calculus, introductory probability and statistics, and the basics of machine learning, especially neural networks, including how to code them up and optimize them in standard neural network software (\eg, PyTorch \citep{paszke2019pytorch}, TensorFlow \citep{tensorflow2015-whitepaper}, JAX \citep{jax2018github}). For example, we assume that the reader knows how to run minibatch gradient descent using a standard neural network optimizer (\eg, Adam \citep{kingma2015adam}). For a primer on neural networks, see, for instance, the interactive textbook \emph{Dive into Deep Learning} by \citet{zhang2023dive}.\footnote{\url{https://D2L.ai}}

In terms of neural network architectures that the reader should already know to understand our monograph, we have intentionally tried to keep this listing short:
\begin{itemize}

\item (Sections~\ref{chap:setup} to~\ref{chap:ode} and the first half of Section~\ref{chap:extensions})
The reader should know multilayer perceptrons for classification and regression (corresponding to the case where raw input data are fixed-length feature vectors). For example, the reader should know that softmax activation yields a probability distribution, and that the function defined by an inner product~$\mathbf{f}(x;\theta) := x^\top \theta$ for~$x,\theta\in\mathbb{R}^d$ is a special case of a multilayer perceptron. (Note that we present the material in a general manner where the raw inputs need not be fixed-length feature vectors.)

\item (Second half of Section~\ref{chap:extensions})
In the latter half of Section~\ref{chap:extensions}, in addition to multilayer perceptrons, the reader should also know recurrent neural networks (RNNs). RNNs enable us to work with variable-length time series as raw inputs.\footnote{While we do not explicitly cover nor assume that the reader knows transformers, we point out that transformers can also handle variable-length inputs (so that in our coverage, RNNs can actually be replaced by transformers).}
\end{itemize}

\subsubsection{How We View Neural Networks}

As we mentioned in Section~\ref{sec:scope}, deep survival models that we cover all depend on a base neural network. By analogy, if we were tackling a classification problem with $k$ classes using deep learning, then the standard strategy is to specify a base neural network (such as a multilayer perceptron) and then we feed the output of the base neural network to a linear layer (also called a full-connected layer or a dense layer) with $k$ output nodes and softmax activation (so that the output of the overall network consists of predicted probabilities of the $k$ classes).\footnote{This strategy would require the base neural network to output some number of nodes that should be at least $k$ (if it is less than~$k$, then we would have trouble representing all $k$ classes).} Then when we learn the network, we use a classification loss function (\eg, cross entropy loss). The final linear layer added with $k$ output nodes and softmax activation is referred to as a ``prediction head''. If instead of classification, we were looking at a regression problem (predicting a single real number), then we could set the prediction head to be a linear layer with 1 output node and no nonlinear activation.

When working with deep survival models for time-to-event prediction, the idea is the same. We first specify a base neural network. Afterward, to get the overall network to predict a time-to-event outcome, it is as simple as choosing a ``survival layer'' at the end (to serve as the prediction head) and using an appropriate survival loss. Depending on the survival layer chosen, there are restrictions on what the output of the base neural network is. For example, when we cover the Cox proportional hazards model, we will see that the base neural network should be set to output a single real number (which could be interpreted as a risk score), and there is actually no additional layer to add. The loss function is then specified a particular way for model training using these ``risk scores''.

\emph{Every survival model we cover could be thought of as a different possible survival layer to use as the prediction head. Each model comes with a loss function. For the models we cover, the loss function will always be a negative log likelihood loss with possibly some other loss terms added, depending on the model.}

Extremely importantly, we will typically not spell out details of how to set the base neural network aside from what we require of its output, meaning that we usually intentionally leave the specific architecture choices up to the modeler. We do this precisely because standard tricks can be used for how to choose the base neural network (as we mentioned above, we could choose a multilayer perceptron when working with tabular data, a convolutional neural network or a vision transformer when working with images, \etc). This also means that advances in neural network technology that are not specific to time-to-event prediction could also trivially be incorporated. For example, if we were to work with multimodal data such as the raw inputs being both images and text, then we could choose the base neural network to be based off a model such as CLIP \citep{radford2021learning}. \emph{An important implication is that when we cover an existing deep survival model in detail, even if the original authors of the model provided architecture details in their paper, we omit the architecture details that are not essential to understanding the design of their overall model.}

Another reason why we do not state very specific neural network architectures to use is because the technology has rapidly been changing! The latest trends in neural network architectures today might be out of fashion tomorrow. To complicate matters, depending on the dataset used in a time-to-event prediction task, which specific architecture works the best might vary, and also which neural network optimizer we should use and with what learning schedule might also vary. Our monograph does not dwell on these engineering details, which are important in practice but are not needed in understanding the core high-level concepts.

\subsubsection{Notation}

We typically use uppercase letters (\eg, $X$) to denote random variables and lowercase letters (\eg,~$x$) to denote deterministic quantities such as constants or specific realized values of random variables. Functions could either be uppercase or lowercase, where we have tried to stick to common conventions used in survival analysis literature (\eg, the so-called conditional survival function is represented by uppercase $S$). Bold letters (\eg, $\mathbf{f}$) are usually used to represent parametric functions such as neural networks. We also frequently use the notation~$[m]:=\{1,2,\dots,m\}$, where~$m$ is a positive integer. When we use the ``log'' function, we always mean natural log.

Optimization problems regularly appear in the monograph. When we write $\widehat{\theta}:=\arg\min_\theta \mathbf{L}(\theta)$, where $\mathbf{L}$ is a loss function with parameter variable~$\theta$, this minimization would be carried out using (some variant of) minibatch gradient descent and, technically, we are usually not finding a solution that achieves the global minimum.

\begin{table}[!t]
\caption{Some software packages used for survival analysis/time-to-event prediction. Models in \textcolor{RoyalBlue}{blue} and evaluation metrics in \textcolor{BrickRed}{red} are ones that we cover in detail in this monograph.}
\label{tab:software}
\vspace{-.5em}
\centering
\adjustbox{max width=\textwidth}{\renewcommand{\arraystretch}{1.4}
\begin{tabular}{>{\raggedright}p{0.37\linewidth} p{.74\linewidth} p{0.7\linewidth}}
\toprule
Package & Link & Some supported methods (not exhaustive) \\ \midrule
\texttt{scikit-survival} \citep{polsterl2020scikit} & \texttt{\url{https://github.com/sebp/scikit-survival}} & \textcolor{RoyalBlue}{Kaplan-Meier estimator}$^1$, \textcolor{RoyalBlue}{Nelson-Aalen estimator}$^2$, \textcolor{RoyalBlue}{Cox model}$^3$, various survival tree ensemble methods including random survival forests$^4$, \textcolor{BrickRed}{concordance index}$^5$, \textcolor{BrickRed}{time-dependent concordance index (truncated)}$^6$, \textcolor{BrickRed}{time-dependent AUC}$^7$, \textcolor{BrickRed}{Brier score}$^8$ \\
\texttt{lifelines} \citep{davidson2019lifelines} & \texttt{\url{https://github.com/CamDavidsonPilon/lifelines}} & \textcolor{RoyalBlue}{Kaplan-Meier estimator}$^1$, \textcolor{RoyalBlue}{Nelson-Aalen estimator}$^2$, \textcolor{RoyalBlue}{Cox model}$^3$\textcolor{RoyalBlue}{~and regularized variants}, \textcolor{RoyalBlue}{accelerated failure time (AFT) models}$^9$, \textcolor{BrickRed}{concordance index}$^5$ \\
\texttt{xgboost} \hspace{10em} \citep{chen2016xgboost} & \texttt{\url{https://github.com/dmlc/xgboost}} & XGBoost supports using Cox and accelerated failure time loss functions \\
\texttt{glmnet\_python} \hspace{10em} \citep{simon2011regularization} & \texttt{\url{https://github.com/bbalasub1/glmnet_python}} & \textcolor{RoyalBlue}{Cox model$^3$ and regularized variants}; this is the official port of \texttt{glmnet} from R \\
\texttt{pycox} \hspace{10em} \citep{kvamme2019time} & \texttt{\url{https://github.com/havakv/pycox}} & unified PyTorch implementations of \textcolor{RoyalBlue}{\mbox{DeepSurv}}$^{10}$, \textcolor{RoyalBlue}{Cox-Time}$^{11}$, \textcolor{RoyalBlue}{Nnet-survival}$^{12}$, \textcolor{RoyalBlue}{DeepHit}$^{13}$, \mbox{N-MTLR}$^{14}$, \textcolor{BrickRed}{time-dependent concordance index (not truncated)}$^{15}$, \textcolor{BrickRed}{Brier score}$^8$ \\
\texttt{pysurvival} \hspace{10em} \citep{pysurvival_cite} & \texttt{\url{https://github.com/square/pysurvival}} & \mbox{N-MTLR} implementation by original author$^{14}$, random survival forests$^4$ \\
\texttt{auton-survival} \hspace{10em} \citep{nagpal2022auton} & \texttt{\url{https://github.com/autonlab/auton-survival}} & \textcolor{RoyalBlue}{DeepSurv}$^{10}$, Deep Survival Machines$^{16}$, Deep Cox Mixtures$^{17}$ \\
\texttt{SurvivalEVAL} \hspace{10em} \citep{qi2024survivaleval} & \texttt{\url{https://github.com/shi-ang/SurvivalEVAL}} & \textcolor{BrickRed}{concordance index}$^5$, \textcolor{BrickRed}{Brier score}$^8$, \textcolor{BrickRed}{D-calibration}$^{18}$, \textcolor{BrickRed}{margin}$^{18}$\textcolor{BrickRed}{~and pseudo-observation}$^{19}$\textcolor{BrickRed}{~MAE scores} \\
\texttt{torchsurv} \hspace{10em} \citep{monod2024torchsurv} & \texttt{\url{https://github.com/Novartis/torchsurv}} & \textcolor{RoyalBlue}{Cox model}$^3$, \textcolor{RoyalBlue}{Weibull AFT model}$^9$, \textcolor{BrickRed}{concordance index}$^5$, \textcolor{BrickRed}{time-dependent AUC}$^7$, \textcolor{BrickRed}{Brier score}$^8$ \\
\bottomrule
\end{tabular}
}

\vspace{.25em}
\adjustbox{max width=\textwidth}{
\begin{tabular}{c}
$^1$\citet{kaplan1958nonparametric}
$\qquad^2$\citet{nelson1969hazard,aalen1978nonparametric}
$\qquad^3$\citet{cox1972regression} \\
$^4$\citet{ishwaran2008random}
$\qquad^5$\citet{harrell1982evaluating}
$\qquad^6$\citet{uno2011c} \\
$^7$\citet{uno2007evaluating,hung2010estimation}
$\qquad^8$\citet{graf1999assessment}
$\qquad^9$\citet{prentice1979hazard} \\
$^{10}$\citet{faraggi1995neural,katzman2018deepsurv}
$\qquad^{11}$\citet{kvamme2019time} \\
$^{12}$\citet{gensheimer2019scalable}
$\qquad^{13}$\citet{lee2018deephit}
$\qquad^{14}$\citet{fotso2018deep} \\
$^{15}$\citet{antolini2005time}
$\qquad^{16}$\citet{nagpal2021dsm}
$\qquad^{17}$\citet{nagpal2021deep} \\
$^{18}$\citet{haider2020effective}
$\qquad^{19}$\citet{qi2023effective}
\end{tabular}
}
\end{table}

\subsubsection{Software Packages and Datasets}
\label{sec:software}

As our exposition assumes that the reader is familiar with standard neural network software that have developer communities that primarily work in Python, we list some Python survival analysis packages in Table~\ref{tab:software}. This list is not exhaustive. We list packages for both deep and non-deep survival models since we think that trying both is important in practice. Per package, we list some (not all) of the models and evaluation metrics implemented. We anticipate that over time, many of these packages will add functionality. Overall, the current state of software packages that support deep survival models is a bit scattered: no single package is---in our opinion---sufficiently comprehensive, and at the time of writing, some packages have not been regularly maintained.

\begin{table}[!t]
\caption{Some models that we cover that are not currently implemented in the packages in Table~\ref{tab:software}.}
\label{tab:more-software}
\vspace{-.5em}
\centering
\adjustbox{max width=.85\textwidth}{\renewcommand{\arraystretch}{1.4}
\begin{tabular}{p{0.7\linewidth} p{0.74\linewidth}}
\toprule
Model & Link \\ \midrule
Deep kernel survival analysis \citep{chen2020deep} & \texttt{\url{https://github.com/georgehc/dksa}} \\
Survival kernets \citep{chen2024survival} & \texttt{\url{https://github.com/georgehc/survival-kernets}} \\
SODEN \citep{tang2022soden} & \texttt{\url{https://github.com/jiaqima/SODEN}} \\
Dynamic-DeepHit \citep{lee2019dynamic} & \texttt{\url{https://github.com/chl8856/Dynamic-DeepHit}} \\
\bottomrule
\end{tabular}
}
\end{table}

Currently, the packages in Table~\ref{tab:software} do not implement all the models that we cover in detail. SODEN \citep{tang2022soden}, deep kernel survival analysis \citep{chen2020deep}, survival kernets \citep{chen2024survival}, and Dynamic-DeepHit \citep{lee2019dynamic} are not currently included in the software packages in Table~\ref{tab:software}, but their code is available from the original authors; see the links in Table~\ref{tab:more-software}.

In terms of publicly available survival datasets, the \texttt{pycox} software package comes with datasets that are all sufficiently large for learning neural network models (mostly in the thousands of data points along with one dataset with roughly~3 million points). The \texttt{scikit-survival} and \texttt{lifelines} packages also come with datasets; some are a bit small though (a few hundred or fewer points).

\newpage
\paragraph{Companion code repository}
To help readers with starting to work with deep survival analysis models in Python, we provide Python code that accompanies our monograph in the following code repository:
\begin{quote}
\texttt{\url{https://github.com/georgehc/survival-intro}}
\end{quote}
This repository includes sample code for every model and every evaluation metric that we discuss in detail. Our code shows how to train different deep survival models, use them to predict time-to-event outcomes, and evaluate the quality of the predictions using some standard evaluation metrics. Our code is primarily in the form of Jupyter notebooks, which include a mix of code cells and explanations for different parts of the code. As we progress through the monograph, we point to specific Jupyter notebooks in our code repository for readers interested in seeing how concepts we cover get translated into code.

Our code has been written with pedagogy in mind. We stick to using standard PyTorch conventions, and we have written our notebooks at a level that exposes the main neural net optimization loop (minibatch gradient descent) and highlights where base neural networks appear in various deep survival models. Our code aims to make various preprocessing and model training steps more transparent, so that if one wants to modify any part of these, doing so should be straightforward.

Moreover, for ease of exposition, our notebooks that accompany Sections~\ref{chap:setup} through~\ref{chap:ode} all use the same standard dataset SUPPORT \citep{knaus1995support}, for which we predict the time until death of severely ill hospitalized patients with various diseases.\footnote{For these particular code notebooks, we also provide an example of how to modify the code to work with different data, with the concrete example being training on the Rotterdam tumor bank dataset \citep{foekens2000urokinase} and then testing on the German Breast Cancer Study Group dataset \citep{schumacher1994randomized}; these two datasets are on predicting survival times of breast cancer patients.} Our notebooks that accompany Section~\ref{chap:extensions} use the PBC dataset \citep{fleming1991counting}, which is on predicting the time until death and the time until transplantation of patients with primary biliary cirrhosis of the liver; here, death and transplantation are viewed as competing events where we only observe whichever one happens first for a training patient (or alternatively, if neither has happened for a training patient, then we at least know the last check-up time with the patient).

Importantly, in our Jupyter notebooks, we do \emph{not} extensively optimize hyperparameters for any particular deep survival model to try to push the prediction performance of the model to be as good as possible. Thus, the final evaluation scores obtained in our notebooks should not be interpreted as the best possible scores achievable by the different models we implement. Furthermore, our code is not written to be ``production-grade'' with, for instance, extensive sanity checks or unit tests.

Lastly, we anticipate occasionally updating our code notebooks to accommodate updates to software packages, to improve exposition or clarity, or to fix bugs that are discovered. The latest version will be available at the GitHub link provided above.

\section{Basic Time-to-Event Prediction Setup}
\label{chap:setup}

We now describe the standard problem setup in time-to-event prediction. We use this problem setup for much of the rest of the monograph. Our goal in this section is to give a reasonably self-contained introduction to the math involved for time-to-event prediction. By the end of this section, the reader should have an understanding of what the standard problem setup is, how prediction tasks are defined, what the general strategy is for learning many time-to-event prediction models (maximum likelihood), how to measure prediction accuracy, and how the time-to-event prediction setup relates to the more mainstream prediction tasks in machine learning of classification and regression.

This section is organized as follows:
\begin{itemize}

\item (Section~\ref{sec:setup}) We first go over the statistical framework for the standard time-to-event prediction problem setup.

\item (Section~\ref{sec:setup-continuous}) We then go over time-to-event prediction when modeling time as continuous. Note that we separate our coverage of continuous vs discrete time as the math involved is a bit different. Beginning by modeling time as continuous, we go over common prediction tasks that correspond to estimating a few different ``target'' functions. By relating the statistical framework from Section~\ref{sec:setup} to these target functions, we can state a general likelihood function. This likelihood function is important because most time-to-event prediction models we know of can be stated as maximizing a likelihood function, possibly accounting for additional regularization or loss terms, or constraints. Along the way, we provide a couple illustrative examples of simple parametric time-to-event prediction models (special cases of what are called proportional hazards models \citep{cox1972regression} and accelerated failure time models \citep{prentice1979hazard}).

\item (Section~\ref{sec:setup-discrete}) Turning to modeling time as discrete, we again go over prediction tasks in terms of target functions, followed by going over the standard likelihood function used. The example models we showcase are DeepHit \citep{lee2018deephit}, Nnet-survival \citep{gensheimer2019scalable}, the Kaplan-Meier estimator \citep{kaplan1958nonparametric}, and the Nelson-Aalen estimator \citep{nelson1969hazard,aalen1978nonparametric}. In the discrete time setting (unlike in continuous time), the likelihood function relates to classification at different discretized time points.

\item (Section~\ref{sec:evaluation}) Next, we discuss some approaches to evaluating the quality of predictions. Specifically, we go over ranking-based accuracy metrics, various mean absolute error and mean squared error metrics, and a calibration metric.

\item (Section~\ref{sec:connections-regression-classification}) We provide additional commentary on how the standard time-to-event prediction setup relates to classification and regression. Notably, there is an approach called \emph{survival stacking} \citep{craig2021survival} that enables one to use existing probabilistic binary classifiers for time-to-event prediction, but this reduction comes at a potentially steep computational cost.

\end{itemize}
We occasionally mention technicalities with details that we defer to Section~\ref{sec:setup-technical}. Understanding these details is not essential for understanding the rest of the monograph.

For ease of exposition, we phrase terminology in terms of time until death. Of course, as we already pointed out in Section~\ref{chap:intro}, the critical event of interest in real applications need not be death.

\subsection{Standard Right-Censored Statistical Framework}
\label{sec:setup}

We assume that we have $n$ training points $(X_1,Y_1,\Delta_1),\dots,(X_n,Y_n,\Delta_n)$, where training point $i\in[n]$ has raw input $X_i\in\mathcal{X}$ (such as a fixed-length feature vector, an image, a text document, \etc), observed time $Y_i\in[0,\infty)$, and ``event indicator'' $\Delta_i\in\{0,1\}$: if $\Delta_i=1$ (the critical event happened for the $i$-th point), then $Y_i$ is the true survival time; otherwise, $Y_i$ is the ``censoring'' time (which could be thought of as the last time we checked on the $i$-th point, when it was still alive). Classically, each data point corresponds to a different person.

We use $X$ to denote the random variable corresponding to a generic raw input, $T$ to denote the random variable corresponding to the true (possibly unobserved) survival time corresponding to raw input~$X$, and~$C$ to denote the random variable corresponding to the true (possibly unobserved) censoring time corresponding to raw input $X$. For these three random variables, we assume that there are the following three probability distributions that are unknown:
\begin{itemize}

\item $\mathbb{P}_X$ is the probability distribution for random raw input $X$, where we assume that the support of this distribution is the raw input space~$\mathcal{X}$ (we formally define the support in Section~\ref{sec:feature-support}). In this monograph, we take~$\mathcal{X}$ to be any input space that standard neural network software can work with.

\item $\mathbb{P}_{T|X}(\cdot|x)$ is the conditional probability distribution of survival time~$T$ given $X=x$.

\item $\mathbb{P}_{C|X}(\cdot|x)$ is the conditional probability distribution of censoring time~$C$ given $X=x$.
\end{itemize}
The training points $(X_1,Y_1,\Delta_1),\dots,(X_n,Y_n,\Delta_n)$ are assumed to be independent and identically distributed (i.i.d.), where each point $(X_i,Y_i,\Delta_i)$ for $i\in[n]$ is generated as follows:
\begin{enumerate}
\item Sample raw input $X_i$ from $\mathbb{P}_X$.
\item Sample true survival time $T_i$ from $\mathbb{P}_{T|X}(\cdot|X_i)$.
\item Sample true censoring time $C_i$ from $\mathbb{P}_{C|X}(\cdot|X_i)$.
\item If $T_i\le C_i$ (death happens before censoring): output $Y_i=T_i$ and $\Delta_i=1$ (no censoring).

Otherwise: output $Y_i=C_i$ and $\Delta_i=0$ (the true survival time is censored).
\end{enumerate}
Note that conditioned on $X_i$, the two random variables $T_i$ and $C_i$ are independent (this assumption is commonly referred to as ``independent censoring''). Moreover, even though $T_i$ and $C_i$ show up in the generative procedure, we do not observe both of them. Instead, we observe exactly one of them. The one we observe is stored in the variable $Y_i$, and the variable $\Delta_i$ tells us whether we observed the survival time (when $\Delta_i=1$) or the censoring time (when $\Delta_i=0$) for the $i$-th training point.

We point out that step~4 of the generative procedure above could equivalently be written in a more concise manner: we set the observed time to be $Y_i = \min\{T_i, C_i\}$ and the event indicator to be $\Delta_i = {\ind\{T_i \le C_i\}}$, where $\ind\{\cdot\}$ is the indicator function which is equal to 1 if its argument is true and is equal to 0 otherwise. Notationally, for a generic random raw input $X$ with true survival time $T$ and true censoring time $C$, we denote the observed time as $Y=\min\{T,C\}$ and the event indicator $\Delta=\ind\{T\le C\}$. In other words, each training point $(X_i,Y_i,\Delta_i)$ is i.i.d.~with the same distribution as $(X,Y,\Delta)$.

Technically, the statistical framework that we have described is referred to as \emph{right-censored}, which just means that for the censored data (the data for which $\Delta_i=0$), the true survival time is \emph{after} the observed censoring time. We discuss other types of censored data in Section~\ref{sec:truncation-other-censoring-cure} (namely, where the survival time is some time \emph{before} the observed censoring time or, separately, where the survival time is known to be within an interval).

\begin{fremark}[Defining time~0]\label{rem:time-of-origin}
Extremely importantly, we have to be precise what we mean by time~0 across the training data (sometimes, this time is referred to as the ``time of origin''). Put another way, for the $i$-th point, exactly what time is~$Y_i$ measured starting from? For example, for a video streaming service that wants to predict the time until a customer stops watching a show, time~0 could be defined to mean when each the customer first starts streaming the show. Thus, for different customers, their time~0 could correspond to different actual world times. In a healthcare example, if we are looking at coma patients in an intensive care unit (ICU) and we want to predict the time until they awaken, then we may want to define each patient's time~0 to mean when they were first admitted to the ICU. In general, time~0 is typically defined to correspond to what is called a ``synchronization event'' (in the two aforementioned examples, this synchronization event would be a user starting to stream a show, and a patient getting admitted to the ICU).
\end{fremark}

\subsection{Time-to-Event Prediction in Continuous Time}
\label{sec:setup-continuous}

Modeling time as continuous, we formally define a few common time-to-event prediction tasks, which could be thought of as estimating specific \emph{prediction target} functions (Section~\ref{sec:estimands-continuous}). Using the statistical framework from Section~\ref{sec:setup}, we can then derive a likelihood function that we can maximize to learn a time-to-event prediction model (Section~\ref{sec:likelihood}).

\paragraph{Key assumptions}
For test raw input~$x\in\mathcal{X}$, we assume that the survival time $T$ conditioned on $X=x$ is a continuous random variable with probability density function (PDF) $f(t|x)$ and a cumulative distribution function (CDF) $F(t|x)=\int_{0}^{t}f(u|x)\textrm{d}u$; either of these functions fully characterizes the distribution~$\mathbb{P}_{T|X}(\cdot|x)$.

\subsubsection{Prediction Targets}
\label{sec:estimands-continuous}

\Paragraph{Survival function}
The first prediction target (also called an \emph{estimand} in statistics terminology) that we present is called the \emph{conditional survival function}, given by
\begin{align}
S(t|x)
&:=\mathbb{P}(\text{survive beyond time }t\mid\text{raw input is }x) \nonumber\\
&\phantom{:}=\mathbb{P}(T>t|X=x) \nonumber\\
&\phantom{:}=1-\mathbb{P}(T\le t|X=x) \nonumber\\
&\phantom{:}=1-F(t|x),\label{eq:surv-function}
\end{align}
where $t\ge0$ and $x\in\mathcal{X}$. In this monograph, we refer to the conditional survival function $S(\cdot|x)$ simply as the \emph{survival function} since our notation already indicates that we are conditioning on~$x$. Predicting $S(\cdot|x)$ means estimating an entire function (\ie, a curve)---\emph{not} just a single number (survival time)---for test raw input~$x$. In literature, this function is sometimes called the \emph{survivor function} (\eg, \citealt{kalbfleisch1980statistical}), the \emph{reliability function} (\eg, \citealt{ebeling2019introduction}), or the \emph{complementary~CDF} (\eg, \citealt{downey2011think}).

\begin{figure}
\centering
\includegraphics[scale=.63]{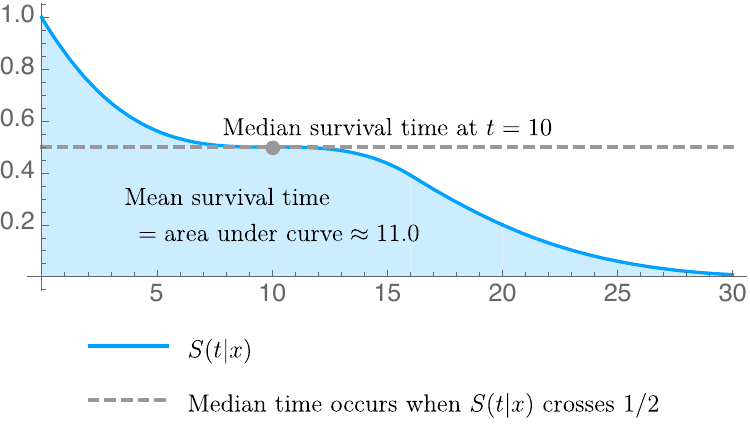}\vspace{-.5em}
\caption{Example of a survival function and its median and mean survival times.}
\label{fig:example-survival-curve}
\end{figure}

As \eqref{eq:surv-function} indicates, the true survival function $S(\cdot|x)$ is 1 minus the CDF $F(\cdot|x)$. A few implications are as follows:
\begin{itemize}

\item[(a)] $S(\cdot|x)=1-F(\cdot|x)$ monotonically decreases from 1 to 0 since any CDF monotonically increases from 0 to~1.

\item[(b)] Estimating the function $S(\cdot|x)$ is equivalent to estimating the CDF $F(\cdot|x)$, which means that we aim to estimate the conditional survival time distribution $\mathbb{P}_{T|X}(\cdot|x)$.

\item[(c)] If we want a single number survival time estimate for raw input~$x$, we can back one out if we know (an estimate of) $S(\cdot|x)$. We give two ways of doing this:
\begin{itemize}

\item[$\bullet$] \emph{Median survival time.} Where a CDF crosses 1/2 corresponds to a median of a distribution, so finding a time $t$ for which $S(t|x)=1-F(t|x)=1/2$ gives a \emph{median} survival time of raw input~$x$. In practice, we only have an estimate $\widehat{S}(\cdot|x)$ of $S(\cdot|x)$ so we find $t$ such that $\widehat{S}(t|x)\approx1/2$.

\item[$\bullet$]\emph{Mean survival time.} For any nonnegative random variable $A$, recall that $\mathbb{E}[A]=\int_{0}^{\infty}{\mathbb{P}(A>u)}\textrm{d}u$. Thus, the mean survival time of raw input $x$ is $\mathbb{E}[T|X=x]=\int_{0}^{\infty}{\mathbb{P}(T>u|X=x)}\textrm{d}u=\int_{0}^{\infty}S(u|x)\textrm{d}u$, the area under the survival function. In practice, we numerically integrate the survival function estimate $\widehat{S}(\cdot|x)$.

\end{itemize}
See Figure~\ref{fig:example-survival-curve} for an example survival function and its corresponding median and mean survival times.\footnote{In practice, we may only know or have an estimate of $S(t|x)$ when $t$ is not too large. For instance, we might only know $S(t|x)$ for all $t\in[0, t_{\max}]$ where $t_{\max}$ is some finite time horizon. In this case, it is possible that $S(t_{\max}|x) > 1/2$, so that we would only know that the median survival time is some time after $t_{\max}$. In this scenario, if we would still like a median survival time estimate that is a single number, then we would need some extrapolation strategy (which could mean having some parametric model for what $S(t|x)$ looks like for $t>t_{\max}$). Similarly, if we only know $S(t|x)$ up to some time horizon $t_{\max}$, then we would not be able to exactly compute the integral $\int_0^\infty S(t|x)dt$ to come up with a mean survival time estimate; we would need some way to extrapolate $S(t|x)$ for $t > t_{\max}$ to assist us in computing the integral.}

\end{itemize}
Different time-to-event prediction models make different assumptions on $S(\cdot|x)$ and often predict transformed variants of $S(\cdot|x)$ rather than predicting $S(\cdot|x)$ directly. The next two prediction targets we discuss are both transformed versions of $S(\cdot|x)$.

\paragraph{Hazard function} A transformed version of $S(\cdot|x)$ that is commonly predicted is the so-called \emph{hazard function}:
\begin{equation}
h(t|x)
:=-\frac{\textrm{d}}{\textrm{d} t}\log S(t|x)
=-\frac{\frac{\textrm{d}}{\textrm{d} t}S(t|x)}{S(t|x)}
=-\frac{\frac{\textrm{d}}{\textrm{d} t}[1-F(t|x)]}{S(t|x)}
=\frac{f(t|x)}{S(t|x)},
\label{eq:hazard}
\end{equation}
where, as a reminder, $f(\cdot|x)$ is the PDF of distribution $\mathbb{P}_{T|X}(\cdot|x)$. \emph{The right-most expression of \eqref{eq:hazard} says that the hazard function is the instantaneous rate of death conditioned on surviving up to time~$t$ for raw input~$x$.} Importantly, for the same reason why PDFs are nonnegative but could otherwise have arbitrarily large positive values, the hazard function is also only nonnegative and could have arbitrarily large positive values. Perhaps the most widely used time-to-event prediction model, the Cox proportional hazards model \citep{cox1972regression}, is stated in terms of the hazard function.

If we know $h(\cdot|x)$, then we can recover $S(\cdot|x)$ since
\begin{align}
h(t|x) = -\frac{\textrm{d}}{\textrm{d} t}\log S(t|x)
\quad
& \Longleftrightarrow\quad \int_0^t h(u|x)\textrm{d}u = -\log S(t|x) \nonumber\\
& \Longleftrightarrow\quad S(t|x) = \exp\Big(-\int_0^t h(u|x)\textrm{d}u\Big).
\label{eq:hazard-to-surv-conversion}
\end{align}

\paragraph{Cumulative hazard function} Another commonly predicted transformed version of $S(\cdot|x)$ is the \emph{cumulative hazard function}:
\begin{equation}
H(t|x):=\int_0^t h(u|x)\textrm{d}u.
\label{eq:cumulative-hazard}
\end{equation}
From \eqref{eq:hazard-to-surv-conversion}, we see that $S(t|x) = \exp(-H(t|x))$, so if we know $H(\cdot|x)$, then we can recover $S(\cdot|x)$. Meanwhile, \eqref{eq:cumulative-hazard} implies that $h(t|x) = \frac{\textrm{d}}{\textrm{d} t} H(t|x)$, so if we know $H(\cdot|x)$, then we can recover~$h(\cdot|x)$.
Some time-to-event prediction models directly estimate the cumulative hazard function such as random survival forests \citep{ishwaran2008random}.

\begin{figure}[!t]
\centering
\includegraphics[scale=0.63]{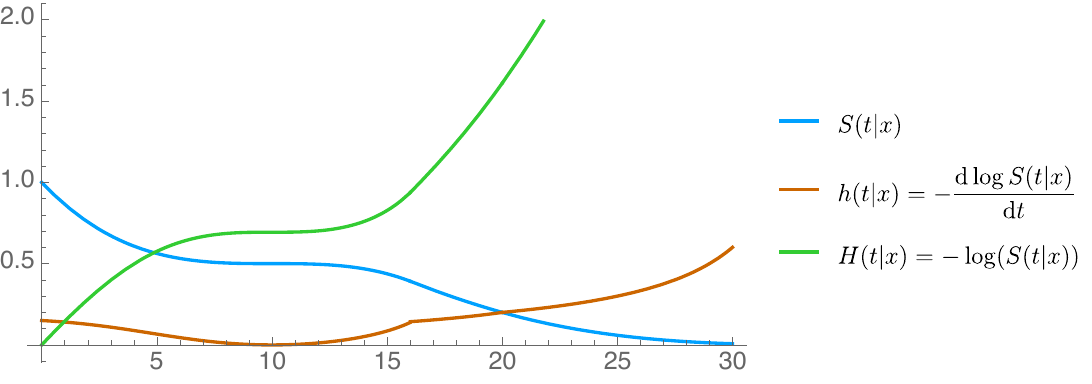}
\caption{The survival function $S(\cdot|x)$ from Figure \ref{fig:example-survival-curve} along with its hazard $h(\cdot|x)$ and cumulative hazard $H(\cdot|x)$ functions.}
\label{fig:survival-curve-variants}
\end{figure}

An example survival function $S(\cdot|x)$ and its corresponding hazard function $h(\cdot|x)$ and cumulative hazard function $H(\cdot|x)$ are shown in Figure \ref{fig:survival-curve-variants}. Note that whereas $S(\cdot|x)$ and $H(\cdot|x)$ are monotonic functions, $h(\cdot|x)$ need not be monotonic.

Because we will often convert between $f(\cdot|x)$, $F(\cdot|x)$, $S(\cdot|x)$, $h(\cdot|x)$, and $H(\cdot|x)$ later in the monograph, we summarize the relationship between these functions below.
\begin{fsummary}[Continuous time prediction targets]
\label{sum:conversions}
Suppose that the key assumptions stated at the start of Section~\ref{sec:setup-continuous} hold. Let $x\in\mathcal{X}$. The following equations show how the PDF $f(\cdot|x)$, the CDF $F(\cdot|x)$, the survival function $S(\cdot|x)$, the hazard function $h(\cdot|x)$, and the cumulative hazard function $H(\cdot|x)$ are related (where we have time $t\ge0$):
\begin{align*}
f(t|x) &= \frac{\textrm{d}}{\textrm{d} t}F(t|x) = \frac{\textrm{d}}{\textrm{d} t}(1-S(t|x)) = h(t|x)S(t|x), \\
F(t|x) &= \int_0^t f(u|x)\textrm{d}u = 1 - S(t|x), \\
S(t|x) &= 1-F(t|x) = \int_t^\infty f(u|x)\textrm{d}u = e^{-H(t|x)} = e^{-\int_0^t h(u|x)\textrm{d}u}, \\
h(t|x) &= \frac{\textrm{d} H(t|x)}{\textrm{d} t}=-\frac{\textrm{d}}{\textrm{d} t}\log S(t|x)=\frac{f(t|x)}{S(t|x)}, \\
H(t|x) &= -\log S(t|x)=\int_0^t h(u|x)\textrm{d}u.
\end{align*}
Importantly, any of these functions fully specifies the conditional survival time distribution $\mathbb{P}_{T|X}(\cdot|x)$.
\end{fsummary}

\subsubsection{Likelihood and Example Models (Parametric Proportional Hazards and Accelerated Failure Time Models)}
\label{sec:likelihood}

Now that we have presented the standard right-censored statistical framework and defined some common prediction targets, we state the likelihood function commonly used in deriving many time-to-event prediction models. Specifically, across the $n$ i.i.d.~training data, we define the following likelihood that does not depend on the censoring distribution:
\begin{equation}
\mathcal{L}
:=\prod_{i=1}^n
    \big\{f(Y_i|X_i)^{\Delta_i}
          S(Y_i|X_i)^{1-\Delta_i}\big\}.
\label{eq:likelihood-pdf}
\end{equation}
This likelihood could be parsed in an intuitive manner: for the $i$-th point, if $\Delta_i=1$ (so the time-to-event outcome is not censored), then the contribution to the product is the PDF of $\mathbb{P}_{T|X}(\cdot|X_i)$ evaluated at the observed time $Y_i$. Otherwise if $\Delta_i=0$, the contribution to the product is the probability of seeing a true survival time larger than $Y_i$ given raw input $X_i$. This latter case crucially uses the observation that, under the statistical framework of Section~\ref{sec:setup}, conditioned on $\Delta_i=0$ and on~$X_i$, the only information we know about the true survival time $T_i$ is that it is after~$Y_i$.

In practice, many time-to-event prediction models are derived by setting one of the conditional functions in Summary~\ref{sum:conversions} to have some parametric form. For instance, we could parameterize the hazard function by setting $h(t|x)=\mathbf{h}(t|x;\theta)$ for some user-specified function~$\mathbf{h}$ (such as a multilayer perceptron where the final output is a single number constrained to be nonnegative) with parameter $\theta$. In general, the variable $\theta$ could consist of multiple parameters, as we see in the following example.

\begin{fexample}[Exponential time-to-event prediction model]
\label{ex:parametric-hazard-exp-survival}
Suppose that the raw input space is $\mathcal{X}=\mathbb{R}^d$, and we set $h(t|x)$ to be equal to
\begin{equation}
\mathbf{h}(t|x;\theta):=e^{\beta^\top x + \psi}
\quad\text{for }t\ge0,x\in\mathcal{X},
\label{eq:hazard-linear-exp}
\end{equation}
where $\beta\in\mathbb{R}^d$ and $\psi\in\mathbb{R}$ are parameters, and $\beta^\top x = \sum_{j=1}^d \beta_j x_j$ is the Euclidean dot product between $\beta$ and $x$. Here, $\theta$ (which includes all the parameters) would be given by $\theta=(\beta,\psi)\in\mathbb{R}^d\times\mathbb{R}$. In this toy example, the model for the hazard does not depend on the input time~$t$.

We point out that for this toy example, the survival function corresponds to an exponential distribution, which is why we refer to this model as an exponential time-to-event prediction model. In particular, to show why the survival function is for an exponential distribution, we calculate the cumulative hazard function $H(t|x)$ from $h(t|x)$ and then we calculate $S(t|x)$ from $H(t|x)$ (using the conversions from Summary~\ref{sum:conversions}):
\begin{align}
H(t|x)
&= \int_0^t \mathbf{h}(u|x;\theta) \textrm{d}u
 = \int_0^t e^{\beta^\top x + \psi} \textrm{d}u
 = t e^{\beta^\top x + \psi},
\label{eq:cumulative-hazard-linear-exp}\\
S(t|x)
&= \exp(-H(t|x)) = \exp(- t e^{\beta^\top x + \psi}).
\label{eq:surv-linear-exp}
\end{align}
Here, $S(t|x)$ corresponds to an exponential distribution with rate parameter $e^{\beta^\top x + \psi}$.
\end{fexample}

Before we can plug in a parametric form of $h(t|x)$ (such as \eqref{eq:hazard-linear-exp}) into the likelihood function (\eqref{eq:likelihood-pdf}), we first rewrite \eqref{eq:likelihood-pdf} in terms of the hazard function. Recall from Summary~\ref{sum:conversions} that: (i) $f(t|x)=h(t|x)S(t|x)$), and (ii) $S(t|x) = e^{-\int_0^t h(u|x)\textrm{d}u}$. Using (i) and (ii), we rewrite \eqref{eq:likelihood-pdf} as
\begin{align}
\mathcal{L}
&=
  \prod_{i=1}^n
    \big\{f(Y_i|X_i)^{\Delta_i}
          S(Y_i|X_i)^{1-\Delta_i}\big\} \nonumber\\
&\overset{\text{(i)}}{=}
  \prod_{i=1}^n
    \big\{h(Y_i|X_i)^{\Delta_i}
          S(Y_i|X_i)\big\} \nonumber\\
&\hspace{-1.1pt}\overset{\text{(ii)}}{=}
  \prod_{i=1}^n
    \bigg\{h(Y_i|X_i)^{\Delta_i}
          \exp\Big(-\int_0^{Y_i} h(u|X_i)\textrm{d}u\Big)\bigg\}.
\label{eq:likelihood-pdf-hazard}
\end{align}
Next, we plug in $h(t|x)=\mathbf{h}(t|x;\theta)$ to get the following likelihood function (that we now emphasize to be a function of $\theta$):
\begin{equation*}
\mathcal{L}(\theta)
= \prod_{i=1}^n
    \bigg\{
      \mathbf{h}(Y_i|X_i;\theta)^{\Delta_i}
      \exp\Big(
        -\int_0^{Y_i} \mathbf{h}(u|X_i;\theta)\textrm{d}u
      \Big)
    \bigg\}.
\end{equation*}
We then estimate $\theta$ by solving the maximum likelihood optimization problem $\widehat{\theta} := \arg\max_{\theta} \mathcal{L}(\theta)$ using, for instance, some variant of gradient ascent. Commonly, the log likelihood is used in the optimization instead, which of course yields the same solution, \ie, $\widehat{\theta} = \arg\max_{\theta} \log \mathcal{L}(\theta)$, where
\begin{align}
\log\mathcal{L}(\theta)
&=\log
  \prod_{i=1}^n
    \bigg\{
      \mathbf{h}(Y_i|X_i;\theta)^{\Delta_i}
      \exp\Big(
        -\int_0^{Y_i} \mathbf{h}(u|X_i;\theta)\textrm{d}u
      \Big)
    \bigg\} \nonumber\\
&=\sum_{i=1}^n
    \bigg\{
      \Delta_i \log \mathbf{h}(Y_i|X_i;\theta)
      -\int_0^{Y_i} \mathbf{h}(u|X_i;\theta)\textrm{d}u
    \bigg\}.
\label{eq:log-likelihood-pdf-hazard-theta}
\end{align}
Especially as our monograph emphasizes deep time-to-event prediction models, we point out that usually when working with standard neural network software, we phrase learning a neural network in terms of minimizing a \emph{loss function}. Specifically, we commonly set the loss function to be the negative log likelihood (NLL), averaged across training data:
\begin{align}
\textbf{L}_{\text{Hazard-NLL}}(\theta)
&:= -\frac{1}{n} \log\mathcal{L}(\theta) \nonumber\\
&\phantom{:}=
 -
 \frac{1}{n}
 \sum_{i=1}^n
    \bigg\{
      \Delta_i \log \mathbf{h}(Y_i|X_i;\theta)
      -\int_0^{Y_i} \mathbf{h}(u|X_i;\theta)\textrm{d}u
    \bigg\}.
\label{eq:nll-loss-hazard-form}
\end{align}
The averaging helps normalize the resulting loss function's values as we vary the number of training points $n$.\footnote{For example, commonly, the training data are split into minibatches for minibatch gradient descent, where we can tune how large these minibatches are (\ie, the batch size). We can look at how the loss function values change as we increase the number of optimization steps and as we vary the batch size. Normalizing helps make sure that the loss function values are comparable across different batch sizes.} We use a standard neural network optimizer to minimize this loss in minibatches.

\begin{fexample}[Exponential time-to-event prediction model, continued]
\label{ex:parametric-hazard-exp-survival-maximum-likelihood}
Continuing with Example~\ref{ex:parametric-hazard-exp-survival}, where $\mathcal{X}=\mathbb{R}^d$, $\theta=(\beta,\psi)\in\mathbb{R}^d\times\mathbb{R}$, and $\mathbf{h}(t|x;\theta)=e^{\beta^\top x + \psi}$, we plug this parametric form of $\mathbf{h}(t|x;\theta)$ into \eqref{eq:nll-loss-hazard-form} to get
\begin{align*}
\mathbf{L}_{\text{Hazard-NLL}}(\beta,\psi)
&=-\frac{1}{n}
   \sum_{i=1}^n
     \bigg\{
       \Delta_i (\beta^\top X_i + \psi)
       -\int_0^{Y_i} e^{\beta^\top X_i + \psi}\textrm{d}u
     \bigg\} \nonumber\\
&=-\frac{1}{n}
   \sum_{i=1}^n
     \big\{
       \Delta_i (\beta^\top X_i + \psi)
       - Y_i e^{\beta^\top X_i + \psi}
     \big\}.
\end{align*}
We then numerically minimize the loss $\mathbf{L}_{\text{Hazard-NLL}}(\beta,\psi)$ with respect to $\beta$ and $\psi$ using a neural network optimizer:
\[
(\widehat{\beta},\widehat{\psi})
:= \arg\min_{(\beta,\psi)\in\mathbb{R}^d\times\mathbb{R}}
\mathbf{L}_{\text{Hazard-NLL}}(\beta,\psi).
\]
For any test raw input $x\in\mathcal{X}$, we can predict the survival, hazard, or cumulative functions by just plugging in the estimates~$\widehat{\beta}$ and~$\widehat{\psi}$ into equations~(\ref{eq:surv-linear-exp}), (\ref{eq:hazard-linear-exp}), and (\ref{eq:cumulative-hazard-linear-exp}) respectively.

Our companion code repository includes a Jupyter notebook that implements this exponential time-to-event prediction model applied to the SUPPORT dataset \citep{knaus1995support}.\footnote{\texttt{\url{https://github.com/georgehc/survival-intro/blob/main/S2.2.2_Exponential.ipynb}}} Note that this is the first Jupyter notebook of the monograph and includes an explanation of the basic experimental setup used for all of our Jupyter notebooks from Sections~\ref{chap:setup} to~\ref{chap:ode}. The notebooks accompanying Section~\ref{chap:extensions} also largely build off this first Jupyter notebook. Thus, for the reader interested in learning how to code with deep survival models, we highly recommend going over this first Jupyter notebook in detail prior to looking at the later notebooks. The overall structure of the notebooks is the same: we load and preprocess data, we learn a survival model using training data (typically using minibatch gradient descent), we predict survival functions of test data, and finally we compute evaluation metrics (discussed later in Section~\ref{sec:evaluation}) on test data.

We also provide a modified version of the first notebook that, instead of using the SUPPORT dataset, trains on the Rotterdam tumor bank dataset \citep{foekens2000urokinase} and tests on the German Breast Cancer Study Group dataset \citep{schumacher1994randomized}.\footnote{\texttt{\url{https://github.com/georgehc/survival-intro/blob/main/S2.2.2_Exponential_RotterdamGBSG.ipynb}}} This second notebook aims to show what code changes are needed to work with different data but is otherwise the same as the first notebook. The rest of the code notebooks for Sections~\ref{chap:setup} to~\ref{chap:ode} use the SUPPORT dataset.
\end{fexample}

\begin{fexample}[Weibull time-to-event prediction model]
\label{ex:parametric-hazard-weibull-survival}
Suppose that $\mathcal{X}=\mathbb{R}^d$, and we set $h(t|x)$ to be
\begin{equation}
\mathbf{h}(t|x;\theta)
:= t^{e^\phi-1} e^{(\beta^\top x) e^\phi + \psi + \phi}
\quad\text{for }t\ge0,x\in\mathcal{X}.
\label{eq:hazard-weibull}
\end{equation}
This is a generalization of the exponential time-to-event prediction model (from Examples~\ref{ex:parametric-hazard-exp-survival} and~\ref{ex:parametric-hazard-exp-survival-maximum-likelihood}), which corresponds to the special case where $\phi=0$. In this more general case, the collection of parameters is $\theta=(\beta,\psi,\phi)\in\mathbb{R}^d\times\mathbb{R}\times\mathbb{R}$, and the survival function corresponds to a Weibull distribution. To see this, we calculate the cumulative hazard and survival functions:
\begin{align}
H(t|x)
&= \int_0^t \mathbf{h}(u|x;\theta) \textrm{d}u \nonumber\\
&= \int_0^t t^{e^\phi-1} e^{(\beta^\top x) e^\phi + \psi + \phi} \textrm{d}u
 = t^{e^\phi} e^{(\beta^\top x) e^\phi + \psi},
\label{eq:cumulative-hazard-weibull} \\
S(t|x)
&= \exp(-H(t|x)) = \exp(-t^{e^\phi} e^{(\beta^\top x) e^\phi + \psi}) \nonumber \\
&= \exp\Bigg(-\bigg(\frac{t}{\exp(-\beta^{\top}x-\psi e^{-\phi})}\bigg)^{e^\phi}\Bigg).
\label{eq:surv-weibull}
\end{align}
Here, $S(t|x)$ corresponds to a Weibull distribution with shape parameter~$e^\phi$ and scale parameter~$\exp(-\beta^{\top}x-\psi e^{-\phi})$. In particular, $S(t|x)$ is of the form $\exp\big(-(\frac{t}{\text{scale}})^{\text{shape}}\big)$. Note that we use the definition of the Weibull scale parameter found in standard software packages such as SciPy \citep{virtanen2020scipy} and R \citep[``stats'' package]{r2021r}. Some texts define the Weibull scale parameter differently (\eg, \citealt{collett2023modelling}) although typically the Weibull shape parameter is defined the same way.\footnote{As a technical remark, we point out that our exposition of the Weibull model is not standard compared to what is in existing literature (see, for instance, Section 12.2 of~\citet{klein2003survival}) in that we have intentionally stated the model so that the parameters~$\beta$,~$\psi$, and~$\phi$ are unconstrained. We use this unconstrained parameterization because commonly neural network optimizers treat parameters as unconstrained. Standard tricks are used to constrain parameters. For example, to constrain a scalar parameter~$\lambda\in\mathbb{R}$ to be nonnegative, we could define~$\lambda = \exp(\varphi)$, treating~$\varphi\in\mathbb{R}$ to be an unconstrained parameter that we optimize over (instead of optimizing over $\lambda$). More generally, we could set $\lambda = g(\varphi)$, where $g:\mathbb{R}\rightarrow[0,\infty)$ is any activation function that outputs a nonnegative value (such as softplus or ReLU). To constrain a parameter to be between~0 and~1, a sigmoid activation could be used. To constrain a collection of parameters to form a probability distribution, the softmax activation function could be used. Etc.}

In terms of learning model parameters, we can use a neural network optimizer to minimize the loss from \eqref{eq:nll-loss-hazard-form}, which in this case is equal to
\begin{align*}
&\mathbf{L}_{\text{Hazard-NLL}}(\beta,\psi,\phi) \\
&\;
 = -\frac{1}{n}\sum_{i=1}^n
     \{\Delta_i \log (Y_i^{e^\phi-1} e^{(\beta^\top X_i) e^\phi + \psi + \phi})
     - (Y_i)^{e^\phi} e^{(\beta^\top X_i) e^\phi + \psi} \} \\
&\;
 = -\frac{1}{n}\sum_{i=1}^n
     \{\Delta_i [(e^\phi-1)\log Y_i + (\beta^\top X_i) e^\phi + \psi + \phi] \\[-1em]
&\phantom{\;=-\frac{1}{n}\sum_{i=1}^n\{}
     - (Y_i)^{e^\phi} e^{(\beta^\top X_i) e^\phi + \psi} \}.
\end{align*}
Similar to how we proceeded in Example~\ref{ex:parametric-hazard-exp-survival-maximum-likelihood}, we would obtain the estimates
$(\widehat{\beta},\widehat{\psi},\widehat{\phi}):=\arg\min_{(\beta,\psi,\phi)\in\mathbb{R}^d\times\mathbb{R}\times\mathbb{R}} \mathbf{L}_{\text{Hazard-NLL}}(\beta,\psi,\phi)$, and plug the estimates $\widehat{\beta}$, $\widehat{\psi}$, $\widehat{\phi}$ into equations~(\ref{eq:surv-weibull}), (\ref{eq:hazard-weibull}), and (\ref{eq:cumulative-hazard-weibull}) to predict survival, hazard, and cumulative hazard functions. We provide a Jupyter notebook that implements this Weibull time-to-event prediction model.\footnote{\texttt{\url{https://github.com/georgehc/survival-intro/blob/main/S2.2.2_Weibull.ipynb}}}
\end{fexample}

The exponential and Weibull time-to-event prediction models are special cases of what are called \emph{proportional hazards models}, the main topic of Section~\ref{chap:proportional-hazards}. As a preview, proportional hazards models assume that the hazard function has the factorization
\[
h(t|x) = \mathbf{h}_0(t;\theta) e^{\mathbf{f}(x;\theta)}
\quad\text{for }t\ge0,x\in\mathcal{X},
\]
for some functions (\eg, neural networks) $\mathbf{h}_0(\cdot;\theta):[0,\infty)\rightarrow[0,\infty)$ and $\mathbf{f}(\cdot;\theta):\mathcal{X}\rightarrow\mathbb{R}$ with parameter variable $\theta$. This factorization makes it clear that time $t$ and raw input $x$ contribute to different multiplicative factors of $h(t|x)$. Regardless of what $x$ is, $h(\cdot|x)$ must be proportional to $\mathbf{h}_0(\cdot;\theta)$. Meanwhile, $\mathbf{f}(x;\theta)\in\mathbb{R}$ could be interpreted as a ``risk score'' for raw input $x$. When $\mathbf{f}(x;\theta)$ is larger, then this corresponds to the hazard $h(t|x)$ being larger, which in turn corresponds to $x$ tending to have a shorter survival time.

In the exponential time-to-event prediction model, we have $\mathcal{X}=\mathbb{R}^d$, $\mathbf{h}_0(t;\theta) = e^\psi$ and $\mathbf{f}(x;\theta) = \beta^\top x$, where $\theta=(\beta,\psi)\in\mathbb{R}^d\times\mathbb{R}$. In the Weibull time-to-event prediction model, we have $\mathcal{X}=\mathbb{R}^d$, $\mathbf{h}_0(t;\theta) = t^{e^\phi - 1} e^{\psi + \phi}$ and $\mathbf{f}(x;\theta) = (\beta^\top x) e^\phi$, where $\theta=(\beta,\psi,\phi)\in\mathbb{R}^d\times\mathbb{R}\times\mathbb{R}$.

As it turns out, the exponential and Weibull time-to-event prediction models are also special cases of what are called \emph{accelerated failure time} (AFT) models, which we discuss more in Section~\ref{sec:aft}. In a nutshell, AFT models assume that the survival function $S(\cdot|x)$ has the same shape across different raw inputs $x$ except that for different $x$, this basic shape of the survival function could be stretched along the time axis as to either accelerate or decelerate when death is likely to occur. The neural ordinary differential equation model in Section~\ref{chap:ode} encompasses both deep proportional hazards and deep AFT models.\footnote{This monograph does not cover classical (\ie, non-neural-network-based) AFT models in much detail, largely because the deep survival models that we cover do not require the reader to know results regarding classical AFT models. However, we would like to mention that classical AFT models are very commonly used (for example, the survival analysis losses supported by XGBoost \citep{chen2016xgboost} include proportional hazards and AFT losses). For the reader interested in learning more about classical AFT models that do not use neural networks, please see, for instance, Chapter~12 of the textbook by \citet{klein2003survival} or Chapter~3 of the textbook by \citet{box2004event}.\label{foot:classical-aft}}

\subsection{Time-to-Event Prediction in Discrete Time}
\label{sec:setup-discrete}

A key challenge of working in continuous time is that computing the (log) likelihood requires evaluating integrals. Earlier, when we parameterized the continuous time likelihood in terms of the hazard function $\mathbf{h}(\cdot|x;\theta)$, the likelihood involved terms of the form $\int_0^{Y_i} \mathbf{h}(u|X_i;\theta)\textrm{d}u$. For the exponential and Weibull models, this integral could be evaluated in closed-form. However, to model time-to-event outcomes in as flexible of a manner as possible, there would in general not be a closed-form expression. To this end, many time-to-event prediction models discretize time into a finite grid, converting integrals into easier-to-compute finite sums. Of course, some time-to-event prediction problems are stated in discrete time to begin with (\eg, \citealt{huh2011adaptive}).

In this section, we state the discrete time versions of the survival, hazard, and cumulative hazard functions (Section~\ref{sec:estimands-discrete}). These discrete time versions do not behave quite the same way as their continuous time analogues. In the case where time actually is continuous but we are discretizing it, we explain some common discretization approaches and also point out some interpolation strategies (Section~\ref{sec:how-to-discretize}). Afterward, we state the standard discrete time likelihood function used in practice, provide some example time-to-event prediction models, and relate the likelihood function to classification (Section~\ref{sec:likelihood-discrete}).

\paragraph{Key assumptions}
Suppose that we discretize time into a user-specified grid of $L$ time points $\tau_{(1)},\tau_{(2)},\dots,\tau_{(L)}\in[0,\infty)$ such that $\tau_{(1)}<\tau_{(2)}<\cdots<\tau_{(L)}$ (we point out how time can be discretized in Section~\ref{sec:how-to-discretize}). We assume that all training $Y_i$ values have been discretized to take on values among $\tau_{(1)},\dots,\tau_{(L)}$. In terms of notation, we use uppercase $L$ since as we point out later, in practice, the maximum number of time steps could be chosen in a way that depends on the training data (which we view as random), in which case $L$ would be a random variable.

\subsubsection{Prediction Targets}
\label{sec:estimands-discrete}

We denote the survival, hazard, and cumulative hazard functions as $S[\cdot|x]$, $h[\cdot|x]$, and $H[\cdot|x]$ respectively (throughout the monograph, we use functions with square brackets to indicate that time is discrete). The CDF of distribution $\mathbb{P}_{T|X}(\cdot|x)$ is denoted as $F[\cdot|x]$ and its corresponding probability mass function (PMF) is denoted as $f[\cdot|x]$. Namely,
\[
f[\ell|x] := \mathbb{P}(T=\tau_{(\ell)}|X=x)
\quad\text{for }\ell\in[L],
\]
and
\[
F[\ell|x] := \mathbb{P}(T\le\tau_{(\ell)}|X=x) = \sum_{m=1}^\ell f[m|x].
\]
Since $f[\cdot|x]$ is a PMF, this means that $f[\ell|x]\ge0$ for each $\ell\in[L]$, and $\sum_{\ell=1}^L f[\ell|x] = 1$. Moreover, we have $f[\ell|x]=F[\ell|x] - F[\ell-1|x]$ with the convention that $F[0|x]:=0$.

\paragraph{Survival function} For $x\in\mathcal{X}$, we define the discrete time survival function evaluated at time index $\ell\in[L]$ to be
\begin{equation}
S[\ell|x]
:= \mathbb{P}(T>\tau_{(\ell)}|X=x)
 = 1 - F[\ell|x] = 1 - \sum_{m=1}^\ell f[m|x].
\label{eq:survival-discrete}
\end{equation}
Thus, if we know either $F[\cdot|x]$ or $f[\cdot|x]$, then we can readily compute $S[\cdot|x]$ using the above equation. Meanwhile, if we know $S[\cdot|x]$, then we can easily compute $F[\cdot|x] = 1 - S[\cdot|x]$.

To compute $f[\cdot|x]$ given $S[\cdot|x]$, note that
\begin{align}
f[\ell|x]
&= F[\ell|x] - F[\ell-1|x] \nonumber\\
&= (1 - S[\ell|x]) - (1 - S[\ell-1|x]) \nonumber\\
&= S[\ell-1|x] - S[\ell|x]\qquad\text{for }\ell\in[L],
\label{eq:survival-pmf-in-terms-of-tail}
\end{align}
where we use the convention $S[0|x]:=1$.

\paragraph{Hazard function}
Next, for time index $\ell\in[L]$, we define the discrete time hazard function $h[\ell|x]$ in a similar manner as the continuous time version in \eqref{eq:hazard}. Namely, $h[\ell|x]$ is the probability of dying at time $\tau_{(\ell)}$ conditioned on still being alive at time $\tau_{(\ell)}$ for raw input~$x$:
\begingroup
\allowdisplaybreaks
\begin{align}
h[\ell|x]
&:=\mathbb{P}(T=\tau_{(\ell)}|X=x,\!\!\!\!\!\!\overbrace{T\ge\tau_{(\ell)}}^{\text{still alive at time }\tau_{(\ell)}}\!\!\!\!\!\!) \nonumber\\
&\phantom{:}=\frac{\mathbb{P}(T=\tau_{(\ell)},T\ge\tau_{(\ell)}|X=x)}{\mathbb{P}(T>\tau_{(\ell-1)}|X=x)}
\nonumber\\
&\phantom{:}=\frac{\mathbb{P}(T=\tau_{(\ell)}|X=x)}{\mathbb{P}(T\ge\tau_{(\ell)}|X=x)} \nonumber\\
&\phantom{:}=\frac{\mathbb{P}(T=\tau_{(\ell)}|X=x)}{\mathbb{P}(T>\tau_{(\ell-1)}|X=x)} \nonumber\\
&\phantom{:}=\frac{f[\ell|x]}{S[\ell-1|x]}.
\label{eq:hazard-discrete}
\end{align}
\endgroup
Importantly, whereas the continuous time version $h(t|x)$ is constrained to be nonnegative and could possibly be larger than~1, in discrete time, $h[\ell|x]$ is a probability so it cannot be larger than~1.

\Eqref{eq:hazard-discrete} tells us how to compute $h[\cdot|x]$ given both $f[\cdot|x]$ and $S[\cdot|x]$. To compute $h[\cdot|x]$ only using $S[\cdot|x]$, we note that by plugging \eqref{eq:survival-pmf-in-terms-of-tail} into \eqref{eq:hazard-discrete}, we obtain
\begin{equation}
h[\ell|x]
= \frac{S[\ell-1|x] - S[\ell|x]}{S[\ell-1|x]}.
\label{eq:hazard-in-terms-of-survival-discrete}
\end{equation}
This tells us how to convert from $S[\cdot|x]$ to $h[\cdot|x]$.

To convert from $h[\cdot|x]$ to $S[\cdot|x]$, we first derive the following recurrence relation:
\begin{align*}
S[\ell|x]
&=\mathbb{P}(T>\tau_{(\ell)}|X=x) \nonumber\\
&=\mathbb{P}(T>\tau_{(\ell-1)}|X=x)\mathbb{P}(T\ne\tau_{(\ell)}|X=x,T>\tau_{(\ell-1)}) \nonumber\\
&=\underbrace{\mathbb{P}(T>\tau_{(\ell-1)}|X=x)}_{S[\ell-1|x]}\big[1-\!\!\!\!\underbrace{\mathbb{P}(T=\tau_{(\ell)}|X=x,T>\tau_{(\ell-1)})}_{h[\ell|x]\text{ using the initial line of \eqref{eq:hazard-discrete}}}\!\!\!\!\big] \nonumber\\
&=S[\ell-1|x](1-h[\ell|x]).
\end{align*}
By plugging in $\ell=1,2,\dots$, we get that:
\begin{itemize}
\item $S[1|x] = 1 - h[1|x]$,
\item $S[2|x] = (1 - h[1|x])(1 - h[2|x])$,
\item $S[3|x] = (1 - h[1|x])(1 - h[2|x])(1 - h[3|x])$,
\end{itemize}
and so forth. In general, the pattern that emerges is that
\begin{equation}
S[\ell|x] = \prod_{m=1}^\ell (1 - h[m|x])\qquad\text{for }\ell\in[L],
\label{eq:survival-discrete-in-terms-of-hazard-discrete}
\end{equation}
which tells us how to convert an estimate of $h[\cdot|x]$ into one of $S[\cdot|x]$.

\Eqref{eq:survival-discrete-in-terms-of-hazard-discrete} has the following interpretation: $h[1|x]$ is the probability of dying at time index~1 for raw input~$x$, so surviving beyond time index~1 happens with probability $1 - h[1|x]$. Conditioned on surviving time index~1, then the probability of dying at time index~2 is $h[2|x]$. Thus, the probability of surviving beyond time index~2 conditioned on surviving past time index~1 for raw input~$x$ is $1 - h[2|x]$; to get the probability without conditioning, we multiply by the probability of the event we conditioned on: $(1 - h[1|x])(1 - h[2|x])$. The $m$-th term in the right-hand side of \eqref{eq:survival-discrete-in-terms-of-hazard-discrete} is the probability that we survive beyond time index~$m$ conditioned on surviving all previous time indices.

\paragraph{Cumulative hazard function}
We define the discrete time cumulative hazard function as
\begin{equation}
H[\ell|x] := \sum_{m=1}^\ell h[m|x].
\label{eq:cumulative-hazard-discrete}
\end{equation}
Thus,~\eqref{eq:cumulative-hazard-discrete} tells us how to convert from $h[\cdot|x]$ to $H[\cdot|x]$. Converting from $H[\cdot|x]$ to $h[\cdot|x]$ can be done in a straightforward manner:
\begin{equation}
h[\ell|x] = H[\ell|x] - H[\ell-1|x]
\label{eq:hazard-in-terms-of-cumulative-hazard-discrete}
\end{equation}
where $H[0|x] := 0$.

However, whereas in continuous time, we had $H(t|x) = -\log S(t|x)$, using the definition of $H[\ell|x]$ in \eqref{eq:cumulative-hazard-discrete}, it turns out that $H[\ell|x]$ is \emph{not} equal to $-\log S[\ell|x]$.

\begin{fproposition}
\label{prop:discrete-cumulative-hazard-not-neg-log-surv}
Let $x\in\mathcal{X}$. Suppose that $h[\ell|x]\in[0,1)$ for all $\ell\in[L]$. Then $H[\cdot|x]$ is a first-order Taylor approximation of $-\log S[\cdot|x]$. In particular, from using a Taylor expansion, we get
\[
-\log S[\ell|x]
= H[j|x]
  +
  \underbrace{
    \sum_{m=1}^\ell
      \sum_{p=2}^{\infty}
        \frac{(h[m|x])^p}{p}
  }_{\substack{\text{second and higher order}\\
               \text{Taylor expansion terms}}}
\quad\text{for }\ell\in[L].
\]
\end{fproposition}
We defer the proof of Proposition~\ref{prop:discrete-cumulative-hazard-not-neg-log-surv} to Section~\ref{sec:discrete-cumulative-hazard-not-neg-log-surv-pf}.

Instead, we can directly relate $H[\cdot|x]$ and $S[\cdot|x]$ in an exact manner by combining equations~(\ref{eq:cumulative-hazard-discrete}) and~(\ref{eq:hazard-in-terms-of-survival-discrete}) to get
\begin{equation*}
H[\ell|x]
= \sum_{m=1}^\ell h[m|x]
= \sum_{m=1}^\ell \frac{S[m-1|x] - S[m|x]}{S[m-1|x]},
\end{equation*}
which tells us how to convert from $S[\cdot|x]$ to $H[\cdot|x]$. Converting from $H[\cdot|x]$ to $S[\cdot|x]$ can be done in a two-step procedure: first compute $h[\cdot|x]$ based on $H[\cdot|x]$ using \eqref{eq:hazard-in-terms-of-cumulative-hazard-discrete}, and then compute $S[\cdot|x]$ based on $h[\cdot|x]$ using \eqref{eq:survival-discrete-in-terms-of-hazard-discrete}.

We summarize the relationship between $f[\cdot|x]$, $F[\cdot|x]$, $S[\cdot|x]$, $h[\cdot|x]$, and $H[\cdot|x]$ below.
\begin{fsummary}[Discrete time prediction targets]
\label{sum:conversions-discrete}
Suppose that the key assumptions stated at the start of Section~\ref{sec:setup-discrete} hold, where we discretize time into the grid $\tau_{(1)}<\tau_{(2)}<\cdots<\tau_{(L)}$. Let $x\in\mathcal{X}$. The following equations show how the PMF $f[\cdot|x]$, the CDF $F[\cdot|x]$, the survival function $S[\cdot|x]$, the cumulative hazard function $H[\cdot|x]$, and the hazard function $h[\cdot|x]$ are related (where time index $\ell\in[L]$):
\begin{align*}
f[\ell|x] &= F[\ell|x]-F[\ell-1|x] = S[\ell-1|x] - S[\ell|x] \\
&= h[\ell|x]S[\ell-1|x], \\
F[\ell|x] &= \sum_{m=1}^\ell f[m|x] = 1 - S[\ell|x], \\
S[\ell|x] &= 1-F[\ell|x] = \sum_{m=\ell+1}^L f[m|x] = \prod_{m=1}^\ell (1-h[m|x]), \\
h[\ell|x] &= H[\ell|x]-H[\ell-1|x] = \frac{S[\ell-1|x]-S[\ell|x]}{S[\ell-1|x]} \\
&= \frac{f[\ell|x]}{S[\ell-1|x]}, \\
H[\ell|x] &= \sum_{m=1}^\ell \frac{S[m-1|x] - S[m|x]}{S[m-1|x]} = \sum_{m=1}^\ell h[m|x].
\end{align*}
We use the convention that $F[0|x]=0$, $S[0|x]=1$, and ${H[0|x]=0}$. Importantly, any of the above functions fully specifies the conditional survival time distribution $\mathbb{P}_{T|X}(\cdot|x)$.
\end{fsummary}

\subsubsection{Time Discretization and Interpolation}
\label{sec:how-to-discretize}

\Paragraph{Time discretization}
Assuming that time is not already discretized, then we have to decide on a discretization method. There are many ways to do this. We begin with the classical example used by the Kaplan-Meier estimator \citep{kaplan1958nonparametric}: set $\tau_{(1)}<\tau_{(2)}<\cdots<\tau_{(L)}$ to be the unique observed times among the $Y_i$ variables for which death happened. In other words, take the set of times ${\{Y_i : i\in[n]\text{ such that }\Delta_i=1\}}$ and sort them (keeping only the unique values) to get $\tau_{(1)}<\tau_{(2)}<\cdots<\tau_{(L)}$.

Importantly, discretizing time can be done with the help of the training data. As another example, we could specify the number of time steps $L$ that we want to use (\eg, 100) and then set $\tau_{(1)}$ to be the smallest (\ie, earliest) time seen among the $Y_i$ variables for which death happened, and then set $\tau_{(L)}$ to be the largest (\ie, latest) time seen among the $Y_i$ variables for which death happened. Then, we could define the rest of the grid points so that they are evenly spaced apart (\ie, $\tau_{(\ell+1)} = \tau_{(\ell)} + \frac{\tau_{(L)}-\tau_{(1)}}{L - 1}$ for $\ell\in[L-1]$). Alternatively, we could use even spacing on a log scale (\ie, $\log\tau_{(1)},\log\tau_{(2)},\dots,\log\tau_{(L)}$ are evenly spaced), or even spacing in terms of percentiles (\eg, if $L=5$, we use the 0\%, 25\%, 50\%, 75\%, and 100\% percentile values among observed times of death). Of course, the time grid need not be set based on training data if the user has good intuition for how to manually choose~it.

As a technical remark, in \eqref{eq:survival-discrete}, since we stated that $f[\cdot|x]$ is a PMF, this implies that $S(\tau_{(L)}) = 0$. This is not a stringent assumption in that we could easily have set the time grid so that time step $L-1$ is the ``last'' time step and time step $L$ is a placeholder time step for times that are ``too large''. For instance, if we discretize time using the Kaplan-Meier approach stated above, we could instead set $L-1$ to be the unique number of times of death (and so $\tau_{(L-1)}$ is the largest time of death encountered in the training data), and we then add a final time step $L$, where any $Y_i$ value larger than the largest time of death gets discretized to be of the final time step $L$. During training of many discrete time models, we do not actually need to specify a precise time for the last time step (so that its time could just be considered ``$>\tau_{(L-1)}$''), but if for whatever reason an actual time is needed, some maximum time could be specified by the user.

\paragraph{Time interpolation}
Sometimes a discrete time model is used, but when making predictions, we may want to switch to using a different time grid from what was used during training (such as a more fine-grain time grid). Naturally, the issue of how to interpolate (or even extrapolate) arises. While a basic strategy like linear interpolation can be used, we point out that \citet{kvamme2021continuous} discuss more sophisticated interpolation strategies that assume either constant probability density or constant hazard values in between time grid points. Which one works best depends on the dataset at hand.

Importantly, if one wants to use some other interpolation strategy aside from the ones we mentioned already, we suggest checking that the interpolation method appropriately retains the monotonicity of $S(\cdot|x)$ (\ie, the interpolated version should also monotonically decrease). Also, we point out that commonly the earliest time grid point $\tau_{(1)}$ is larger than 0, in which case a standard assumption is that $S(0|x) = 1$ and $S(\cdot|x)$ decays from time 0 to time $\tau_{(1)}$ (how it decays depends on the choice of interpolation method).

\subsection[Likelihood, Connection to Classification, and\texorpdfstring{\\}{} Example Models (DeepHit, Nnet-survival,\texorpdfstring{\\}{} Kaplan-Meier, Nelson-Aalen)]{Likelihood, Connection to Classification, and Example Models (DeepHit, Nnet-survival, Kaplan-Meier, Nelson-Aalen)}
\label{sec:likelihood-discrete}

We now cover the discrete time log likelihood and also relate it to classification. Along the way, we present some example models. We introduce the notation $\kappa(Y_i)$ to denote the specific time index (from $1,2,\dots,L$) that time $Y_i$ corresponds to (as a reminder, we assume that we have discretized all $Y_i$ values to the values in $\tau_{(1)},\dots,\tau_{(L)}$). Then the likelihood function that does not depend on the censoring distribution~is
\begin{equation*}
\mathcal{L}
:=
  \prod_{i=1}^n
    \big\{
      f[\kappa(Y_i)|X_i]^{\Delta_i}
      S[\kappa(Y_i)|X_i]^{1-\Delta_i}
    \big\},
\end{equation*}
which is similar to the continuous time version from \eqref{eq:likelihood-pdf}.

\begin{fexample}[DeepHit]
\label{ex:deephit}
The DeepHit model \citep{lee2018deephit} specifies the survival time PMF $f[\cdot|x]$ in terms of a user-specified neural network $\mathbf{f}(\cdot;\theta):\mathcal{X}\rightarrow[0,1]^L$ with parameter variable $\theta$ so that:
\begin{equation*}
\begin{bmatrix}
f[1|x] \\
f[2|x] \\
\vdots \\
f[L|x]
\end{bmatrix}
=
\begin{bmatrix}
\mathbf{f}_1(x;\theta) \\
\mathbf{f}_2(x;\theta) \\
\vdots \\
\mathbf{f}_L(x;\theta)
\end{bmatrix}
=: \mathbf{f}(x;\theta).
\end{equation*}
Note that the output of $\mathbf{f}(\cdot;\theta)$ needs to be a valid probability distribution (so that $f[\cdot|x]$ is a valid PMF). As an example of how to enforce this, if $\mathbf{f}(\cdot;\theta)$ is a multilayer perceptron, then we could set the last linear layer to output $L$ numbers and have softmax activation.

We could learn $\theta$ by maximizing the likelihood function
\begingroup
\allowdisplaybreaks
\begin{align*}
\mathcal{L}(\theta)
&=
  \prod_{i=1}^n
    \big\{
      f[\kappa(Y_i)|X_i]^{\Delta_i}
      S[\kappa(Y_i)|X_i]^{1-\Delta_i}
    \big\} \nonumber\\
&=
  \prod_{i=1}^n
    \bigg\{
      f[\kappa(Y_i)|X_i]^{\Delta_i}
      \bigg[
        \sum_{m=\kappa(Y_i)+1}^L
          f[m|X_i]
      \bigg]^{1-\Delta_i}
    \bigg\} \nonumber\\
&=
  \prod_{i=1}^n
    \bigg\{[\mathbf{f}_{\kappa(Y_i)}(X_i;\theta)]^{\Delta_i}
           \bigg[
             \sum_{m=\kappa(Y_i)+1}^L
               \mathbf{f}_{m}(X_i;\theta)
           \bigg]^{1-\Delta_i}
    \bigg\},
\end{align*}
\endgroup
where the second equality uses Summary~\ref{sum:conversions-discrete} (namely that $S[\ell|x] = 1 - F[\ell|x] = 1 - \sum_{m=1}^\ell f[m|x] = \sum_{m=\ell+1}^L f[m|x]$). In practice, to maximize $\mathcal{L}(\theta)$, we could use a standard neural network optimizer to numerically minimize the negative log likelihood averaged across training data:
\begin{align*}
&\mathbf{L}_{\text{PMF-NLL}}(\theta)\nonumber\\
&\quad:= -\frac{1}{n}\log\mathcal{L}(\theta) \nonumber\\
&\phantom{\quad:}=
  -
  \frac{1}{n}
  \log
  \prod_{i=1}^n\!
    \bigg\{[\mathbf{f}_{\kappa(Y_i)}(X_i;\theta)]^{\Delta_i}
           \bigg[
             \sum_{m=\kappa(Y_i)+1}^L
               \mathbf{f}_m(X_i;\theta)
           \bigg]^{1-\Delta_i}
    \bigg\} \nonumber\\
&\phantom{\quad:}=
  -
  \frac{1}{n}
  \sum_{i=1}^n\!
    \bigg\{
      \Delta_i\log(\mathbf{f}_{\kappa(Y_i)}(X_i;\theta)) \nonumber\\
&\phantom{\quad:=-\frac{1}{n}\sum_{i=1}^n\!\bigg\{}
      +
      (1-\Delta_i)
      \log
        \bigg(
          \sum_{m=\kappa(Y_i)+1}^L
            \mathbf{f}_m(X_i;\theta)
        \bigg)
    \bigg\}.
\end{align*}
Specifically, we compute
\[
\widehat{\theta} := \arg\min_{\theta} \mathbf{L}_{\text{PMF-NLL}}(\theta).
\]
After obtaining estimate $\widehat{\theta}$ for $\theta$, we could predict the survival time PMF for any test raw input $x\in\mathcal{X}$ using
\[
\widehat{f}\!~[\ell|x] := \mathbf{f}_{\ell}(x; \widehat{\theta}\hspace{1.5pt}),
\]
from which we could back out estimates of the survival function $S[\ell|x]$, hazard function $h[\ell|x]$, or cumulative hazard function $H[\ell|x]$ using the conversions from Summary~\ref{sum:conversions-discrete}. In particular, we have:
\begin{align*}
\widehat{S}_{\text{DeepHit}}[\ell|x]
&= \sum_{m=\ell+1}^L \widehat{f}\!~[m|x]
 = \sum_{m=\ell+1}^L \mathbf{f}_m(x; \widehat{\theta}\hspace{1.5pt}), \\
\widehat{h}_{\text{DeepHit}}[\ell|x]
&= \frac{\widehat{f}\!~[\ell|x]}{\widehat{S}[\ell-1|x]}
 = \frac{\mathbf{f}_\ell(x; \widehat{\theta}\hspace{1.5pt})}{\sum_{m=\ell}^L \mathbf{f}_m(x; \widehat{\theta}\hspace{1.5pt})}, \\
\widehat{H}_{\text{DeepHit}}[\ell|x]
&= \sum_{m=1}^\ell \widehat{h}[m|x]
 = \sum_{m=1}^\ell
     \frac{\mathbf{f}_m(x; \widehat{\theta}\hspace{1.5pt})}{\sum_{p=m}^L \mathbf{f}_p(x; \widehat{\theta}\hspace{1.5pt})}.
\end{align*}
To see DeepHit in code, please see our accompanying Jupyter notebook.\footnote{\texttt{\url{https://github.com/georgehc/survival-intro/blob/main/S2.3.3_DeepHit_single.ipynb}}} This is the first Jupyter notebook of the monograph that goes over discretizing time prior to learning the model. The same discretization code shows up in a number of our later Jupyter notebooks that involve discrete time survival~models.

We remark that the full DeepHit model is more general than the special case of it that we present in this example. In particular, the full DeepHit model adds a second loss term related to ranking (the user has to tune a hyperparameter that trades off between the negative log likelihood loss and the ranking loss) and, furthermore, the full model can keep track of multiple kinds of critical events rather than only a single one such as death (this is referred to as the ``competing risks'' setup). We present the full DeepHit model in Section~\ref{sec:deephit-general} during our coverage of competing risks.

A special case of how to specify the neural network $\mathbf{f}(\cdot;\theta)$ results in a time-to-event prediction model called Multi-Task Logistic Regression \citep{yu2011learning,fotso2018deep}. Details on this connection are provided by \citet[Appendix~C]{kvamme2021continuous}.
\end{fexample}

\paragraph{Parameterizing the hazard function and connection to classification}
In the continuous case, we showed how we can parameterize the hazard function $h(\cdot|x)$ and minimize a negative log likelihood loss in terms of the hazard function. In discrete time, we could also choose to directly work with the hazard function $h[\cdot|x]$ (instead of working with the PMF $f[\cdot|x]$ as done by DeepHit). The resulting hazard-function-based likelihood function looks a bit different from that of continuous time (\eqref{eq:likelihood-pdf-hazard}) and relates to classification. Using Summary~\ref{sum:conversions-discrete}, we have: (i) $f[\ell|x] = h[\ell|x]S[\ell-1|x]$, and (ii) $S[\ell|x] = \prod_{m=1}^\ell (1-h[m|x])$. Then using (i) and (ii), we obtain:
\begingroup
\allowdisplaybreaks
\begin{align*}
\mathcal{L}
&=
  \prod_{i=1}^n\!
    \big\{
      f[\kappa(Y_i)|X_i]^{\Delta_i}
      S[\kappa(Y_i)|X_i]^{1-\Delta_i}
    \big\} \nonumber \\
&\overset{\text{(i)}}{=}
  \prod_{i=1}^n\!
    \big\{
      (h[\kappa(Y_i)|X_i] S[\kappa(Y_i)-1|X_i])^{\Delta_i}
      S[\kappa(Y_i)|X_i]^{1-\Delta_i}
    \big\} \nonumber \\
&\hspace{-1.1pt}\overset{\text{(ii)}}{=}
  \prod_{i=1}^n\!
    \bigg\{
      \bigg[
        h[\kappa(Y_i)|X_i]
        \!\!
        \prod_{m=1}^{\kappa(Y_i)-1}
        \!\!(1-h[m|X_i])
      \bigg]^{\Delta_i}
      \bigg[
        \prod_{m=1}^{\kappa(Y_i)}(1-h[m|X_i])
      \bigg]^{1-\Delta_i}
    \bigg\} \nonumber \\
&=
  \prod_{i=1}^n\!
    \bigg\{
      h[\kappa(Y_i)|X_i]^{\Delta_i}
      (1-h[\kappa(Y_i)|X_i])^{1 - \Delta_i}
      \bigg[
        \prod_{m=1}^{\kappa(Y_i)-1}
          \!\!\!\!(1-h[m|X_i])
      \bigg]
    \bigg\}.
\end{align*}
\endgroup
Take the log of both sides to get
\begin{align}
\log\mathcal{L}
&=
  \sum_{i=1}^n
    \Bigg\{
      \overbrace{
      \Delta_i \log(h[\kappa(Y_i)|X_i])
      +
      (1 - \Delta_i) \log(1-h[\kappa(Y_i)|X_i])}^{
        \substack{\text{Bernoulli log likelihood (the negative of the so-called}\\\text{binary cross entropy loss) at time index }\kappa(Y_i)}
      } \nonumber\\
&\phantom{=\sum_{i=1}^n\Bigg\{}
      + \underbrace{\sum_{m=1}^{\kappa(Y_i)-1} \log(1-h[m|X_i])}_{
        \text{death not encountered before time index }\kappa(Y_i)
      }
    \Bigg\}.
\label{eq:log-likelihood-pmf-hazard}
\end{align}
In particular, $h[\cdot|X_i]$ could be viewed as a probabilistic binary classifier for the $i$-th point where at each time index $\ell\in[L]$, the classifier has a different predicted probability of death conditioned on the $i$-th point still being alive at time index $\ell$. Ideally, $h[\cdot|X_i]$ should be low for all time indices prior to $\kappa(Y_i)$. Then at time index $\kappa(Y_i)$, if death happened, then we want $h[\cdot|X_i]$ to be high; otherwise, we want $h[\cdot|X_i]$ to be low.

Notice that $h[\cdot|X_i]$ could be thought of as a multi-time-horizon classifier: regardless of what time index we make a prediction for, we always condition on the same raw input $X_i$. In practice, we could think of $X_i$ as information collected prior to time index~1. Using this information, we predict hazard probabilities for all time indices (or time horizons)~$1,2,\dots,L$.

\begin{fexample}[Nnet-survival]
\label{ex:nnet-survival}
The Nnet-survival model \citep{gensheimer2019scalable} specifies the hazard function $h[\cdot|x]$ in terms of a user-specified neural network $\mathbf{g}(\cdot;\theta):\mathcal{X}\rightarrow\mathbb{R}^L$ with parameter variable $\theta$. In particular,
\[
\mathbf{g}(x;\theta)
:=
  \begin{bmatrix}
    \mathbf{g}_1(x;\theta) \\
    \mathbf{g}_2(x;\theta) \\
    \vdots \\
    \mathbf{g}_L(x;\theta)
  \end{bmatrix},
\]
so each output of $\mathbf{g}(\cdot;\theta)$ corresponds to a different time index. Then Nnet-survival sets the hazard function $h[\ell|x]$ equal to
\begin{equation}
\mathbf{h}[\ell|x;\theta]
:=
\frac{1}{1 + e^{-\mathbf{g}_\ell(x;\theta)}}
\quad\text{for }\ell\in[L],x\in\mathcal{X},
\label{eq:nnet-survival-hazard}
\end{equation}
which corresponds to applying the logistic function to each of the $L$ outputs of $\mathbf{g}(\cdot;\theta)$, ensuring that $\mathbf{h}[\ell|x;\theta]\in[0,1]$ for every $\ell\in[L]$.

To learn $\theta$, we maximize the log likelihood in \eqref{eq:log-likelihood-pmf-hazard}, which we instead state as minimizing the negative log likelihood loss, averaged across training data:
\begin{align}
\mathbf{L}_{\text{Nnet-survival}}(\theta)
&=-
  \frac{1}{n}
  \sum_{i=1}^n
    \Bigg\{
      \Delta_i \log(\mathbf{h}[\kappa(Y_i)|X_i;\theta]) \nonumber\\
&\phantom{=-\frac{1}{n}\sum_{i=1}^n\Bigg\{}
      +
      (1 - \Delta_i) \log(1-\mathbf{h}[\kappa(Y_i)|X_i;\theta]) \nonumber\\
&\phantom{=-\frac{1}{n}\sum_{i=1}^n\Bigg\{}
      +
      \sum_{m=1}^{\kappa(Y_i)-1}
        \log(1 - \mathbf{h}[m|X_i;\theta])
    \Bigg\}.
\label{eq:nnet-survival-loss-helper}
\end{align}
Plugging \eqref{eq:nnet-survival-hazard} into \eqref{eq:nnet-survival-loss-helper} and using the fact that $1-\mathbf{h}[\ell|x;\theta] = \frac{1}{1 + e^{\mathbf{g}_\ell(x;\theta)}}$, we get
\begin{align*}
\mathbf{L}_{\text{Nnet-survival}}(\theta)
&=
  \frac{1}{n}
  \sum_{i=1}^n
    \Bigg\{
      \Delta_i
      \log\!\bigg(\frac{1}{1 + e^{-\mathbf{g}_{\kappa(Y_i)}(X_i;\theta)}}\bigg) \\
&\phantom{=-\frac{1}{n}\sum_{i=1}^n\Bigg\{}
      +
      (1 - \Delta_i)
      \log\!\bigg(\frac{1}{1 + e^{\mathbf{g}_{\kappa(Y_i)}(X_i;\theta)}}\bigg) \\
&\phantom{=-\frac{1}{n}\sum_{i=1}^n\Bigg\{}
      +
      \sum_{m=1}^{\kappa(Y_i)-1}
        \log\!\bigg(\frac{1}{1 + e^{\mathbf{g}_m(X_i;\theta)}}\bigg)
    \Bigg\} \\
&=\frac{1}{n}
  \sum_{i=1}^n
    \Bigg\{
      \Delta_i
      \log(1 + e^{-\mathbf{g}_{\kappa(Y_i)}(X_i;\theta)}) \\
&\phantom{=\frac{1}{n}\sum_{i=1}^n\Bigg\{}
      +
      (1 - \Delta_i)
      \log(1 + e^{\mathbf{g}_{\kappa(Y_i)}(X_i;\theta)}) \\
&\phantom{=\frac{1}{n}\sum_{i=1}^n\Bigg\{}
       +
       \sum_{m=1}^{\kappa(Y_i)-1}
         \log(1 + e^{\mathbf{g}_m(X_i;\theta)})
     \Bigg\}.
\end{align*}
Thus, we numerically compute the estimate
\[
\widehat{\theta} := \arg\min_\theta \mathbf{L}_{\text{Nnet-survival}}(\theta),
\]
using a standard neural network optimizer.

Afterward, we can predict the hazard function and from that back out the cumulative hazard and survival functions:
\begin{align*}
\widehat{h}_{\text{Nnet-survival}}[\ell|x]
&:=
  \mathbf{h}[\ell|x;\widehat{\theta}]
=
\frac{1}{1 + e^{-\mathbf{g}_\ell(x;\widehat{\theta}\hspace{1.5pt})}}, \\
\widehat{H}_{\text{Nnet-survival}}[\ell|x]
&:=
  \sum_{m=1}^\ell \widehat{h}_{\text{Nnet-survival}}[m|x]
  = \sum_{m=1}^\ell \frac{1}{1 + e^{-\mathbf{g}_m(x;\widehat{\theta}\hspace{1.5pt})}}, \\
\widehat{S}_{\text{Nnet-survival}}[\ell|x]
&:=
  \prod_{m=1}^\ell (1 - \widehat{h}_{\text{Nnet-survival}}[m|x]) \\
&\phantom{:}=
  \prod_{m=1}^\ell \!\Big(1 - \frac{1}{1 + e^{-\mathbf{g}_m(x;\widehat{\theta}\hspace{1.5pt})}}\Big)\!
  =
  \prod_{m=1}^\ell \frac{1}{1 + e^{\mathbf{g}_m(x;\widehat{\theta}\hspace{1.5pt})}}.
\end{align*}
As discussed by \citet[Section 2.2]{kvamme2021continuous}, if the function $\mathbf{g}(\cdot;\theta)$ is a general parametric function that is not restricted to be a neural network, then a number of authors have discussed variants of this same model (\eg, \citealt{cox1972regression,brown1975use,allison1982discrete,tutz2016modeling}). Our companion code repository includes a Jupyter notebook that covers \mbox{Nnet-survival}.\footnote{\texttt{\url{https://github.com/georgehc/survival-intro/blob/main/S2.3.3_Nnet-survival.ipynb}}}
\end{fexample}
\paragraph{An alternative way to write the same log likelihood}
We now point out an equivalent way of writing the log likelihood \eqref{eq:log-likelihood-pmf-hazard} that is useful for a few reasons: first, this alternative way of writing the log likelihood more clearly shows how the log likelihood could be written as the sum of log likelihood terms from different time indices. In fact, this way of writing the log likelihood leads to a straightforward derivation of a method called the Kaplan-Meier estimator. Second, this alternative way of writing the log likelihood is what is implemented in the now-standard software package \texttt{pycox} \citep{kvamme2019time}.

\newcommand{\binaryeventmatrix}{B}

We proceed by defining the matrix $\binaryeventmatrix\in\{0,1\}^{n\times L}$, where the $i$-th row, $\ell$-th column entry is
\begin{align*}
\binaryeventmatrix_{i,\ell}
&:= \ind\{Y_i = \tau_{(\ell)}, \Delta_i = 1\} \nonumber\\
&\phantom{:}= \ind\{\kappa(Y_i) = \ell, \Delta_i = 1\} \nonumber\\
&\phantom{:}= \Delta_i \ind\{\kappa(Y_i) = \ell\}.
\end{align*}
In other words, if the $i$-th data point's survival time is censored, then the $i$-th row of $\binaryeventmatrix$ would be all zeros. Otherwise, the $i$-th row of $\binaryeventmatrix$ would consist of all zeros except for a 1 at column $\kappa(Y_i)$. Note that the matrix $\binaryeventmatrix$ in can readily be computed from $(Y_1,\Delta_1),\dots,(Y_n,\Delta_n)$.\footnote{However, we cannot in general recover $(Y_1,\Delta_1),\dots,(Y_n,\Delta_n)$ from only knowing the matrix $\binaryeventmatrix$. To see this, note that if the $i$-th point is censored, then the $i$-th row in~$\binaryeventmatrix$ would consist of all zeros, from which we cannot recover $Y_i$.}

Then \eqref{eq:log-likelihood-pmf-hazard} can be written as
\begin{align}
\log\mathcal{L}
&=
  \sum_{i=1}^n
    \Bigg\{
      \Delta_i \log(h[\kappa(Y_i)|X_i])
      +
      (1 - \Delta_i) \log(1-h[\kappa(Y_i)|X_i])
      \nonumber\\
&\phantom{=\sum_{i=1}^n\Bigg\{}
      + \sum_{m=1}^{\kappa(Y_i)-1} \log(1-h[m|X_i])
    \Bigg\} \nonumber\\
&=
  \sum_{i=1}^n
  \sum_{\ell=1}^{\kappa(Y_i)}
    \underbrace{
      \big\{
        \binaryeventmatrix_{i,\ell} \log h[\ell|X_i]
        + (1 - \binaryeventmatrix_{i,\ell}) \log (1-h[\ell|X_i])
      \big\}}_{\text{Bernoulli log likelihood}},
\label{eq:log-likelihood-pmf-hazard-alt}
\end{align}
where we see that every term being added can be thought of as another classification problem. For each data point $i\in[n]$, at each time index $\ell\in[\kappa(Y_i)]$, we check how well the probabilistic classifier $h[\ell|X_i]$ agrees with the data $\binaryeventmatrix_{i,\ell}$.

We can further rearrange \eqref{eq:log-likelihood-pmf-hazard-alt} by exchanging the inner and outer summations to show that each time index can be viewed as having its own loss term:
\begin{align}
&\log\mathcal{L} \nonumber\\
&=
  \sum_{i=1}^n
  \sum_{\ell=1}^{\kappa(Y_i)}
    \big\{
      \binaryeventmatrix_{i,\ell} \log h[\ell|X_i]
      + (1 - \binaryeventmatrix_{i,\ell}) \log (1-h[\ell|X_i])
    \big\} \nonumber\\
&=
  \sum_{\ell=1}^{L}
    \underbrace{
      \sum_{i=1}^n
        \ind\{\ell\le\kappa(Y_i)\}
        \big\{
          \binaryeventmatrix_{i,\ell} \log h[\ell|X_i]
          + (1 - \binaryeventmatrix_{i,\ell}) \log (1-h[\ell|X_i])
        \big\}}_{\text{likelihood term for time index }\ell}.
\label{eq:log-likelihood-pmf-hazard-alt2}
\end{align}
This way of writing the log likelihood makes it clear that we could view the problem as doing classification at each of the $L$ time indices (so that it is a multi-time-horizon classification problem). One could, for example, parameterize $h[\cdot|x]$ so that each time index has its own parameters and also there could be some parameters shared across time.

\begin{fexample}[Kaplan-Meier and Nelson-Aalen estimators]
\label{ex:kaplan-meier-nelson-aalen}
Consider a simple setup where for each time index $\ell\in[L]$, we predict the hazard probability to be $\theta_\ell\in[0,1]$ (completely disregarding what the raw input $x$ is). In particular, we use the model
\begin{equation}
h[\ell|x] := \theta_\ell\quad\text{for }\ell\in[L],x\in\mathcal{X}.
\label{eq:kaplan-meier-nelson-aalen-hazard-model}
\end{equation}
Because the hazard does not depend on the raw input~$x$, this simple model could be thought of as specifying a population-level discrete time hazard function rather than one that actually accounts for raw inputs. Then the log likelihood in this case, using~\eqref{eq:log-likelihood-pmf-hazard-alt2} and now emphasizing the dependence on $\theta:=(\theta_1,\dots,\theta_L)\in[0,1]^L$, is given by
\begin{equation*}
\mathcal{L}(\theta)
= \sum_{\ell=1}^L
   \underbrace{
     \sum_{i=1}^n
       \ind\{\ell\le\kappa(Y_i)\}
       \big\{
         \binaryeventmatrix_{i,\ell} \log \theta_\ell
         + (1 - \binaryeventmatrix_{i,\ell}) \log (1-\theta_\ell)
       \big\}}_{=:\mathcal{L}_{(\ell)}(\theta_\ell)\text{ (likelihood term for time index }\ell\text{)}}.
\end{equation*}
Because $\mathcal{L}(\theta)=\sum_{\ell=1}^L \mathcal{L}_{(\ell)}(\theta_\ell)$, to maximize $\mathcal{L}(\theta)$ with respect to $\theta$, it suffices to maximize each time index's likelihood term $\mathcal{L}_{(\ell)}(\theta_\ell)$ with respect to $\theta_\ell$. This maximization has a closed-form solution (we derive this solution in Section~\ref{sec:kaplan-meier-nelson-aalen-hazard-maximum-likelihood-derivation}):
\begin{equation}
\widehat{\theta}_\ell
:= \arg\max_{\theta_\ell} \mathcal{L}_{(\ell)}(\theta_\ell)
 = \frac{D[\ell]}{N[\ell]}
\quad\text{for }\ell\in[L],
\label{eq:kaplan-meier-nelson-aalen-hazard-maximum-likelihood}
\end{equation}
where~$D[\ell]$ is the number of deaths that occurred at time index~$\ell$ within the training data, and~$N[\ell]$ is the number of points ``at risk'' (that could possibly die) at time index~$\ell$ within the training data. Formally, recalling that $\tau_{(\ell)}$ is the actual time corresponding to time index $\ell\in[L]$:
\begin{align}
D[\ell]
&:= \sum_{j=1}^n\ind\{Y_j=\tau_{(\ell)}\}\Delta_j = \sum_{j=1}^n \binaryeventmatrix_{j,\ell},
\label{eq:death-counter} \\
N[\ell]
&:= \sum_{j=1}^n\ind\{Y_j\ge\tau_{(\ell)}\} = \sum_{j=1}^n \ind\{\kappa(Y_j)\ge\ell\}.
\label{eq:at-risk-counter}
\end{align}
Note that these were classically written out in a table referred to as a ``life table''. Of course, on a computer, we would typically store $D[\cdot]$ and $N[\cdot]$ each in 1D arrays/``tables''.

Using the conversions in Summary~\ref{sum:conversions-discrete}, the estimated hazard, cumulative hazard, and survival functions are (again, these are population-level estimates that do not depend on the test raw input $x$):
\begingroup
\allowdisplaybreaks
\begin{align}
\widehat{h}[\ell]
&:= \widehat{\theta}_\ell = \frac{D[\ell]}{N[\ell]},
  \label{eq:kaplan-meier-hazard-discrete} \\
\widehat{H}[\ell]
&:= \sum_{m=1}^\ell \widehat{h}[m]
  = \sum_{m=1}^\ell \frac{D[m]}{N[m]}, \nonumber\\
\widehat{S}[\ell]
&:= \prod_{m=1}^\ell (1 - \widehat{h}[m])
  = \prod_{m=1}^\ell \Big(1 - \frac{D[m]}{N[m]}\Big).
  \nonumber
\end{align}
\endgroup
In fact, $\widehat{H}[\ell]$ and $\widehat{S}[\ell]$ correspond to the discrete time versions of what are called the Nelson-Aalen estimator \citep{nelson1969hazard,aalen1978nonparametric} and the Kaplan-Meier estimator \citep{kaplan1958nonparametric}, respectively. These estimators are typically stated in continuous time, where the only difference is that we simply interpolate the discrete time estimator so that at any time $t\ge0$, we output the most recently seen discrete time index's value (also called \emph{forward filling} interpolation).

Specifically, using the convention that $\tau_{(0)}:=0$ and $\widehat{H}_{\text{NA}}[0] := 0$, the Nelson-Aalen estimator is:
\begin{align}
\widehat{H}_{\text{NA}}(t)
&:=
  \begin{cases}
    \widehat{H}[\ell-1] & \text{if }\tau_{(\ell-1)}< t\le\tau_{(\ell)}\text{ for }\ell\in[L], \\
    \widehat{H}[L] & \text{if }t > \tau_{(L)},
  \end{cases} \nonumber\\
&\phantom{:}=
  \sum_{m=1}^L \ind\{\tau_{(m)}\le t\} \frac{D[m]}{N[m]}
  \quad\text{for }t\ge0.
\label{eq:nelson-aalen}
\end{align}
Using the convention that $\tau_{(0)}:=0$ (again) and $\widehat{S}_{\text{KM}}[0] := 1$, the Kaplan-Meier estimator is:
\begin{align}
\widehat{S}_{\text{KM}}(t)
&:=
  \begin{cases}
    \widehat{S}[\ell-1] & \text{if }\tau_{(\ell-1)}< t\le\tau_{(\ell)}\text{ for }\ell\in[L], \\
    \widehat{S}[L] & \text{if }t > \tau_{(L)},
  \end{cases} \nonumber\\
&\phantom{:}=
  \prod_{m=1}^L \Big(1 - \frac{D[m]}{N[m]}\Big)^{\ind\{\tau_{(m)}\le t\}}
  \quad\text{for }t\ge0.
\label{eq:kaplan-meier}
\end{align}
To summarize, the Nelson-Aalen and Kaplan-Meier estimators are inherently discrete time methods for predicting population-level cumulative hazard and survival functions, and they correspond to using the population-level hazard function (\eqref{eq:kaplan-meier-nelson-aalen-hazard-model}) that is estimated using maximum likelihood. We provide a Jupyter notebook that covers both the Nelson-Aalen and Kaplan-Meier estimators, although we mostly focus on the latter.\footnote{\texttt{\url{https://github.com/georgehc/survival-intro/blob/main/S2.3.3_Kaplan-Meier_Nelson-Aalen.ipynb}}}
\end{fexample}

The Kaplan-Meier and Nelson-Aalen estimators are considered \emph{nonparametric} since they do not impose a parametric form on the survival, cumulative hazard, or hazard functions, and typically they are used in a manner where the time grid is taken to be all unique times of death (so that the parameter variable $\theta\in[0,1]^L$ in Example~\ref{ex:kaplan-meier-nelson-aalen} could grow in size since as we collect more training data, the number of unique times of death $L$ could increase).

In fact, under fairly general settings, as the amount of training data grows to $\infty$, the Kaplan-Meier estimator provably converges to $S_\text{pop}(t):=\mathbb{P}(T>t)$ for all times $t$ that are not too large \citep{foldes1981strong} (after all, we cannot expect the Kaplan-Meier estimator to predict survival probabilities well when $t$ is close to or exceeds the maximum observed time of death). This theory could be extended to the Nelson-Aalen estimator as well using the fact that the latter is a first-order Taylor series approximation of the former (using the same argument as in Proposition~\ref{prop:discrete-cumulative-hazard-not-neg-log-surv}).

In practice, the Kaplan-Meier estimator is extremely popular because it can easily be used to compare groups. For example, suppose that we have a group of patients who received a treatment and another group of patients who did not receive the treatment (the control group). We could collect the ground truth labels (the $Y_i$ and $\Delta_i$ variables) for all these patients and then compute a Kaplan-Meier survival function just for those who received treatment and, separately, another Kaplan-Meier survival function just for those in the control group. Plotting the two survival functions overlaid over each other could be informative. In fact, there are also statistical tests for comparing the two groups (\eg, testing whether they are the ``same'') such as the log-rank test \citep{mantel1966evaluation}. Of course, we can use this same idea provided that we have any approach for partitioning the complete collection of training data into different groups or clusters and, per cluster, fit a Kaplan-Meier survival function. We will revisit this idea in Section~\ref{sec:kernets}.

\subsection{Evaluation Metrics for Time-to-Event Prediction}
\label{sec:evaluation}

The fundamental challenge in measuring accuracy in time-to-event prediction problems is censoring. Consider a training point $X_i$ with observed time $Y_i$ and event indicator $\Delta_i=0$. This means that $Y_i$ is the censoring time $C_i$ and \emph{not} the true unobserved survival time $T_i$. Thus, even if a time-to-event prediction model could come up with an estimate $\widehat{T}_i$ of $T_i$, we would not have a ground truth value to compare $\widehat{T}_i$ with, and there is no guarantee that the observed censoring time $Y_i$ is close to the true unobserved survival time $T_i$.

We now cover many common evaluation metrics for time-to-event prediction models. Throughout this section, we define evaluation metrics just using the training data, mainly to avoid introducing new notation. Very importantly, these evaluation metrics could also be computed on validation or test data.

We want to emphasize that none of the evaluation metrics we present is perfect for every situation. In fact, many evaluation metrics we present are known to be ``improper'' \citep{gneiting2007strictly}, meaning that the best score possible is \emph{not} achieved by the true conditional survival distribution (\ie, an evaluation metric can assign a better score to an incorrect model compared to the correct model). With this cautionary note in mind, we generally recommend using multiple evaluation metrics.

Our Jupyter notebooks that accompany Sections~\ref{chap:setup} through~\ref{chap:ode} show how to compute all the evaluation metrics that we discuss in detail in this section. Our code specifically computes these evaluation metrics on held out test data.

\subsubsection{Ranking-Based Accuracy Metrics}
\label{sec:ranking-metrics}

We begin with accuracy metrics based on ranking. Sometimes, these ranking metrics are referred to as ``discrimination'' metrics.

\paragraph{Harrell's concordance index}
One of the most common accuracy metrics used in survival analysis is Harrell's \emph{concordance index} \citep{harrell1982evaluating}, often abbreviated as ``c-index''. The c-index is the fraction of pairs of data points that are correctly ranked by a prediction procedure among pairs of data points that can be ranked unambiguously, which as we will see shortly neatly handles censored data. As it is a fraction, c-index values range from~0 to~1, where~1 means the most accurate.

We first work out a simple example before we more formally state the definition of c-index.

\begin{fexample}[C-index calculation]
Consider if we have three devices:
\begin{itemize}[itemsep=0pt,parsep=0pt]

\item Device A fails after 2 days of use.

\item Device B fails after 10 days of use.

\item Device C is still working after 6 days of use.

\end{itemize}
In other words, their respective $(Y_i,\Delta_i)$ values are $(2,1)$, $(10,1)$, and $(6,0)$. (We take time~0 to be when a device first starts being used. For simplicity, our subsequent exposition is phrased as if we started using the devices at the same time.)

We know for sure that device A failed before device B, and that device A failed before device C. However, we do not know which of device B or C fails first. Thus, in this example, only two pairs (A \& B, A \& C) consist of data points that can be ranked unambiguously. If a prediction procedure ranks device~A as having a shorter survival time than device~B, and device~A as having a longer survival time than device~C, then only one out of the two pairs is correctly predicted, so the c-index is~1/2.
\end{fexample}

From this example, we see that computing c-index values requires that our time-to-event prediction model can rank different data points, which is a different task from what we had previously presented in predicting survival, hazard, or cumulative hazard functions of different data points. As we mentioned in Section~\ref{sec:likelihood} and will discuss in detail in Section~\ref{chap:proportional-hazards}, proportional hazards models assign a risk score $\mathbf{f}(x;\theta)\in\mathbb{R}$ to each raw input~$x$. Thus, we can rank data points by their risk scores.

We now formally define the c-index in terms of a risk score function (such as the one from a proportional hazards model that is learned using training data).

\begin{fdefinition}[C-index]
\label{def:c-index}
Suppose that we have a risk score function $r:\mathcal{X}\rightarrow\mathbb{R}$. For any $x,x'\in\mathcal{X}$, if $r(x)>r(x')$, then the risk score predicts $x$ to be ``worse off'' than $x'$ (\eg, $x$ tends to have a shorter survival time than~$x'$).

We first define the set of ``comparable pairs'' (\ie, pairs of points that could be unambiguously ordered):
\begin{equation}
\mathcal{E}
:= \{ (i,j)\in[n]\times[n] :
      \Delta_i = 1,
      Y_i < Y_j \}.
\label{eq:comparable-pairs}
\end{equation}
Thus, each pair $(i,j)\in\mathcal{E}$ has data point~$i$ unambiguously having shorter survival time than data point~$j$ since data point~$i$ died whereas data point~$j$ has a higher observed time (and it does not matter whether data point~$j$ died or not). This means that ideally, we should have $r(X_i) > r(X_j)$.

Then we define
\begin{equation*}
\text{c-index}
:= \frac{1}{|\mathcal{E}|}
     \sum_{(i,j)\in\mathcal{E}} \ind\{ r(X_i) > r(X_j) \},
\end{equation*}
which is a fraction between~0 and~1. Higher scores are better.
\end{fdefinition}
Note that the c-index as defined above aims to estimate
\[
\text{c-index}^*
:=
\mathbb{P}( r(X) > r(X') \mid \Delta = 1, Y < Y' ),
\]
where the two points $(X,Y,\Delta)$ and $(X',Y',\Delta')$ are i.i.d.~samples from the generative procedure in Section~\ref{sec:setup} (the random variable $\Delta'$ is not needed though in defining $\text{c-index}^*$). We use the superscript ``$^*$'' to indicate that $\text{c-index}^*$ is a population-level quantity that depends on the true underlying distributions $\mathbb{P}_X$, $\mathbb{P}_{T|X}$ and $\mathbb{P}_{C|X}$ that we do not know in practice.

\smallskip
\noindent
\emph{Connection to AUC:}
Similar to the area under the receiver operator characteristic curve (AUC) for binary classification, a c-index score of 1/2 is considered low and can be achieved via ``random guessing''. As an example, consider the random risk score function $r_{\text{random}}:\mathcal{X}\rightarrow\mathbb{R}$, where $r_{\text{random}}(x)$ just ignores the input~$x$ and outputs a random number sampled from a standard Gaussian $\mathcal{N}(0,1)$. One can show that the expected value of the c-index achieved by $r_{\text{random}}$ is~1/2. In fact, when there is no censoring, then the population-level quantity $\text{c-index}^*$ is equal to the AUC for an appropriately defined classification problem (see, for example, \citealt{harrell1996multivariable,koziol2009concordance}).

\smallskip
\noindent
\emph{Key limitations:}
One problem with the c-index metric is that it requires having a risk score function to rank points with. Many time-to-event prediction models do not explicitly learn such a function. One workaround is simple: as we mentioned in Section~\ref{sec:estimands-continuous}, given any estimated survival function~$\widehat{S}(\cdot|x)$, we could compute a median (or mean) survival time estimate of~$x$, which would enable us to rank points based on their estimated median (or mean) survival times. This workaround is somewhat unsatisfying, as we convert each point's predicted survival function into a single number, and we otherwise ignore the shape of the function.

Another problem with the c-index metric is that it is known to be improper for predicting the risk of a data point dying within a pre-specified time horizon (\eg, 10 years). For details on this result, see the paper by \citet{blanche2019c}.

\smallskip
\noindent
\emph{Handling ties in observed times:}
As a minor technical remark, we point out that the c-index calculation is sometimes defined slightly differently to address what happens if there are two points $i$ and $j$ (with $i\ne j$) that have the same observed time $Y_i=Y_j$ and at least one of them has died. For ease of exposition, we do not include such pairs of points in the set $\mathcal{E}$.

\paragraph{Time-dependent concordance index} \citet{antolini2005time}  proposed a time-dependent concordance index (denoted as $C^{\text{td}}$) that aims to make better use of any predicted survival function $\widehat{S}(\cdot|x)$. We state how to compute it before providing intuition for its definition.
\begin{fdefinition}[$C^{\text{td}}$ index]
\label{def:c-td-index}
Suppose that we have a survival function estimate $\widehat{S}(\cdot|x)$ for any $x\in\mathcal{X}$. Then using the set of comparable pairs $\mathcal{E}$ from \eqref{eq:comparable-pairs}, we define the~$C^{\text{td}}$ index~as
\[
C^{\text{td}}
:= \frac{1}{|\mathcal{E}|}
     \sum_{(i,j)\in\mathcal{E}} \ind\{ \widehat{S}(Y_i | X_i) < \widehat{S}(Y_i | X_j) \},
\]
which again is between 0 and 1. Higher scores are better.
\end{fdefinition}
To motivate this definition, consider two points $(i,j)\in\mathcal{E}$, so point~$i$ died and $Y_i<Y_j$. Then at the earlier time $Y_i$, we would like point~$i$ to be predicted as having \emph{higher} risk of death (in other words, \emph{lower} survival probability) than point~$j$. We can use survival probability $\widehat{S}(Y_i|\cdot)$ as a ranking function in this case that is specific to time $Y_i$, where $\widehat{S}(Y_i|x)$ being lower for~$x$ means that~$x$ is predicted to be at \emph{higher} risk of death at time $Y_i$. Thus, for $(i,j)\in\mathcal{E}$, we would like $\widehat{S}(Y_i | X_i) < \widehat{S}(Y_i | X_j)$.

Importantly, in the case of proportional hazards models (which we previewed at the end of Section~\ref{sec:likelihood} and cover in detail in Section~\ref{chap:proportional-hazards}), the~$C^{\text{td}}$ index is the same as the c-index defined earlier, so that the~$C^{\text{td}}$ index could be viewed as a generalization of the c-index to accommodate any time-to-event prediction model that predicts $\widehat{S}(\cdot|x)$.

Lastly, we point out two theoretical properties. First, the $C^\text{td}$ score is an empirical estimate of the population-level quantity
\[
C^{\text{td}\hspace{1.5pt}*}
:=
\mathbb{P}( \widehat{S}(Y | X) < \widehat{S}(Y | X') \mid \Delta = 1, Y < Y' ),
\]
where just as before, we assume that the two points $(X,Y,\Delta)$ and $(X',Y',\Delta')$ are i.i.d.~samples from the generative procedure in Section~\ref{sec:setup}. Second, we point out that the $C^{\text{td}}$ score is known to be improper \citep{rindt2022survival}.

\paragraph{Truncated time-dependent concordance index}
The c-index and $C^{\text{td}}$-index scores each give only a single number and do not provide an accuracy score at a specific user-specified time~$t\in[0,\infty)$. \citet{uno2011c} presented a ``truncated'' version of the time-dependent concordance index that does depend on~$t$. We begin by defining the time-dependent set of comparable pairs:
\[
\mathcal{E}(t)
:= \{ (i,j)\in[n]\times[n] : \Delta_i=1, Y_i < t, Y_j > Y_i\}.
\]
The only change compared to $\mathcal{E}$ is that $\mathcal{E}(t)$ has the additional constraint that $Y_i < t$. Now for $(i,j)\in\mathcal{E}(t)$, we would like $\widehat{S}(t|X_i)<\widehat{S}(t|X_j)$. Then Uno \emph{et al.}~defined the following evaluation metric (following \citet{tang2022soden}, we denote this metric as $C_t^{\text{td}}$).
\begin{fdefinition}[Truncated time-dependent concordance index]
\label{def:trunc-time-dep-c-index}
Let $t\ge0$. Suppose that we have a survival function estimate $\widehat{S}(\cdot|x)$ for any $x\in\mathcal{X}$. Then using the time-dependent set of comparable pairs $\mathcal{E}(t)$, we define the truncated time-dependent concordance index at time $t$ as
\[
C_t^{\text{td}}
:= \frac{\sum_{(i,j)\in\mathcal{E}(t)} w_i \ind\{ \widehat{S}(t | X_i) < \widehat{S}(t | X_j) \}}
        {\sum_{(i,j)\in\mathcal{E}(t)} w_i},
\]
where $w_1,w_2,\dots,w_n\in[0,\infty)$ are so-called \emph{inverse probability of censoring weights} to be defined in a moment (\eqref{eq:ipcw}). Values of $C_t^{\text{td}}$ are between 0 and 1, where higher is better.
\end{fdefinition}
If the weights $w_1,w_2,\dots,w_n$ are all set to be~1, then we would have a score function that looks quite similar to the~$C^{\text{td}}$ index except now looking at a specific time~$t$. By having weights be unequal, we are changing how much each pair $(i,j)\in\mathcal{E}(t)$ contributes to both the numerator and the denominator of $C_t^{\text{td}}$.

We state how Uno \emph{et al.}~set the weights before providing intuition for why. In what follows, we assume that censoring time~$C$ is independent of raw input~$X$, meaning that the conditional censoring distribution $\mathbb{P}_{C|X}(\cdot|x)$ is equal to a population-level censoring distribution $\mathbb{P}_C(\cdot)$. Define $S_{\text{censor}}(t):=\mathbb{P}(C > t)$ for $t\ge0$. To estimate this function, the standard approach is to fit the Kaplan-Meier estimator (\eqref{eq:kaplan-meier}) on the training labels $(Y_1, \ind\{{\Delta_1=0}\}), (Y_2, \ind\{{\Delta_2=0}\}), \dots, (Y_n, \ind\{{\Delta_n=0}\})$ (by switching censoring to be the critical event of interest, we reason about time until censoring instead of time until death); the resulting estimated Kaplan-Meier ``survival'' function is denoted~$\widehat{S}_{\text{censor}}(\cdot)$. Then we set
\begin{equation}
w_i := \frac{1}{(\widehat{S}_{\text{censor}}(Y_i))^2}
\quad\text{for }i\in[n].
\label{eq:ipcw}
\end{equation}
The intuition is that the set $\mathcal{E}(t)$ is a biased sample of possible pairs: a point $(i,j)\in\mathcal{E}(t)$ must have point~$i$ not be censored. All else equal, had point~$i$ been censored instead (but had the same true survival time), then its survival time would still be less than that of point~$j$, but this single change (whether point~$i$ was censored or not) would alter the value of $C_t^{\text{td}}$. In particular, $\mathcal{E}(t)$ ``over-emphasizes'' pairs $(i,j)$ where point~$i$ is not censored, so we ``up-weight'' point~$i$ if its observed time $Y_i$ is more likely to have been censored. \citet[Appendix~A]{uno2011c}~make this argument precise by formally showing that $C_t^{\text{td}}$ converges (as the amount of data goes to infinity) to
\[
C_t^{\text{td}\hspace{1.5pt}*}
:= \mathbb{P}\big(\widehat{S}(Y|X) < \widehat{S}(Y|X') \mid Y < t, Y < Y').
\]
Notice that this true population-level target does \emph{not} depend on the event indicator $\Delta$. This sort of weighting could be applied to the c-index and $C^{\text{td}}$ scores too if one wishes to modify them so that they estimate population-level quantities that do not depend on $\Delta$.

\begin{fremark}[Using validation or test data with $C_t^{\text{td}}$]
\label{rem:val-test-data-censoring-km}
As an important implementation detail, we had mentioned that the evaluation metrics we present can be computed using validation or test data. In the case of the $C_t^{\text{td}}$ metric, we would learn both the time-to-event prediction model (that can predict $\widehat{S}(\cdot|x)$ for any $x\in\mathcal{X}$) and the censoring distribution's right tail probability function $\widehat{S}_{\text{censor}}(\cdot)$ from training data. After learning these functions, we treat them as fixed and evaluate $C_t^{\text{td}}$ on validation or test data.
\end{fremark}
We caution that if $\widehat{S}_{\text{censor}}(\cdot)$ is a poor estimate of the true population-level $S_{\text{censor}}(t) = \mathbb{P}(C > t)$ function, then the $C_t^{\text{td}}$ score with weights given by \eqref{eq:ipcw} may be suspect. Note that we have made the simplifying assumption that censoring time $C$ is independent of raw input $X$. If this assumption does not hold, then we should replace $\widehat{S}_{\text{censor}}(\cdot)$ in \eqref{eq:ipcw} with a version that conditions on a given raw input (meaning that we instead aim to estimate the true population level function $S_{\text{censor}}(t|x) := \mathbb{P}(C > t | X = x)$). However, estimating $S_{\text{censor}}(\cdot|x)$ could be as difficult as estimating the conditional survival function $S(\cdot|x)$, so that it is possible that the weights $w_i$ used with the $C_t^{\text{td}}$ score in this case are unreliable.

\paragraph{Integrated truncated time-dependent concordance index}
We point out that we could integrate $C_t^{\text{td}}$ over time~$t$ to arrive at a single number. We would have to specify the limits of integration.
\begin{fdefinition}[Integrated truncated time-dependent concordance index]
\label{def:integrated-truncated-time-dependent-concordance-index}
Suppose that we have a survival function estimate $\widehat{S}(\cdot|x)$ for any $x\in\mathcal{X}$. Let $t_{\min}\ge0$ and $t_{\max}>t_{\min}$ be user-specified lower and upper limits of integration. Then we define the integrated truncated time-dependent concordance index as
\[
C_{[t_{\min},t_{\max}]}^{\text{td}}
:= \frac{1}{t_{\max} - t_{\min}}\int_{t_{\min}}^{t_{\max}} C_u^{\text{td}}\textrm{d}u.
\]
This score is also between 0 and 1, where higher is better.
\end{fdefinition}
In practice, we numerically evaluate this integral along a time grid.

\paragraph{Time-dependent AUC}
There are also time-dependent AUC scores \citep{uno2007evaluating,hung2010estimation}, which we will only briefly cover since they are very similar to the truncated time-dependent concordance index~$C_t^{\text{td}}$. The key idea is that for a fixed time $t$, we set up a binary classification problem of distinguishing between points who died no later than time $t$ (the ``positive'' class), and points who died after time $t$ (the ``negative'' class). We take $\widehat{S}(t|\cdot):\mathcal{X}\rightarrow[0,1]$ to be the probabilistic binary classifier that we aim to compute an AUC for. When this classifier predicts a lower survival probability for $x\in\mathcal{X}$, then this means that $x$ is predicted to have a higher chance of being in the positive class.

Then to estimate the AUC, \citet{hung2010estimation} showed that we can use the same equation as the~$C_t^{\text{td}}$ index (\eqref{eq:truncated-time-dependent-concordance-index}) except with two small changes. First, we replace $\mathcal{E}(t)$ with
\[
\widetilde{\mathcal{E}}(t)
:= \{ (i,j)\in[n]\times[n] : \Delta_i=1, Y_i\le t, Y_j > t\}.
\]
Notice that if $(i,j)$ is in $\widetilde{\mathcal{E}}(t)$, then it means that point~$i$ is an example point from the ``positive'' class, and point~$j$ is an example point from the ``negative'' class. These are points we evaluate the binary classifier on. The second change is that we set the weight $w_i := 1/[\widehat{S}_{\text{censor}}(Y_i)\widehat{S}_{\text{censor}}(t)]$ (where we are simplifying Hung and Chiang's equation~(3) for the case where the censoring distribution does not depend on the raw input).

Again, this approach uses weights $w_i$ that depend on $\widehat{S}_{\text{censor}}(t)$ estimating $S_{\text{censor}}(t) = \mathbb{P}(C>t)$ accurately. If $\widehat{S}_{\text{censor}}(\cdot)$ is a poor estimate of $S_{\text{censor}}(\cdot)$, then the time-dependent AUC score could be unreliable.

There are many other ways to define a time-dependent AUC evaluation metric though. See the surveys by \citet{blanche2013time} and \citet{lambert2016summary} for details.

\subsubsection{Squared Error of the Predicted Survival Function}
\label{sec:brier}

\Paragraph{Brier score}
There are accuracy metrics that more directly assess error of an estimated survival function $\widehat{S}(\cdot|x)$ without ranking. For example, the Brier score \citep{graf1999assessment} is defined for a specific time $t\ge0$ and aims to measure the error
\[
\text{BS}^*(t)
:= \mathbb{E}[(\ind\{T>t\}- \widehat{S}(t|X))^2],
\]
where the expectation is over sampling $X$ and $T$ using the generative procedure in Section~\ref{sec:setup}.

To empirically estimate $\text{BS}^*(t)$, we first consider if censoring never happens, so every point $i\in[n]$ has observed time $Y_i$ equal to the true survival time $T_i$. Then we could estimate $\text{BS}^*(t)$~with
\[
\text{BS}_{\text{no-censoring}}(t)
:=
\frac{1}{n}\sum_{i=1}^n [(\ind\{Y_i > t\} - \widehat{S}(t|X_i))^2].
\]
To account for censoring, \citet{graf1999assessment} proposed the following approach. Similar to how we had defined the $C_t^{\text{td}}$ index, we assume that censoring time~$C$ is independent of raw input~$X$ and estimate $S_{\text{censor}}(\cdot)$ with the Kaplan-Meier estimator to obtain $\widehat{S}_{\text{censor}}(\cdot)$. Then we define the Brier score error metric we use is as follows.
\begin{fdefinition}[Brier score]\label{def:brier}
Suppose that we have a survival function estimate $\widehat{S}(\cdot|x)$ for any $x\in\mathcal{X}$ and also an estimate $\widehat{S}_{\text{censor}}(\cdot)$ of $S_{\text{censor}}(\cdot)$. We define the Brier score at time $t\ge0$ by
\[
\text{BS}(t)
:=
  \frac{1}{n}
    \sum_{i=1}^n
      \bigg[
        \frac{\widehat{S}(t|X_i)^2 \Delta_i \ind\{ Y_i \le t\}}
             {\widehat{S}_{\text{censor}}(Y_i)}
        +
        \frac{(1 - \widehat{S}(t|X_i))^2 \ind\{ Y_i > t \}}
             {\widehat{S}_{\text{censor}}(t)}
      \bigg],
\]
which is nonnegative. Lower scores are better.
\end{fdefinition}
\citet{graf1998explained} showed that $\text{BS}(t)$ converges to $\text{BS}^*(t)$ as $n\rightarrow\infty$ with the additional assumption that $S_{\text{censor}}(t)>0$. Separately, as a reminder, if we are computing the Brier score on validation or test data, then $\widehat{S}_{\text{censor}}(\cdot)$ is estimated using \emph{training} data (as mentioned in Remark~\ref{rem:val-test-data-censoring-km}).

Just like with the truncated time-dependent concordance index and time-dependent AUC scores, Brier scores crucially depend on $\widehat{S}_{\text{censor}}(\cdot)$ estimating $S_{\text{censor}}(\cdot)$ well. As a reminder, we have assumed that censoring time $C$ is independent of raw input $X$. If this assumption does not hold, then Brier scores (as we have defined them in Definition~\ref{def:brier}) are not guaranteed to be proper \citep{rindt2022survival}.

\paragraph{Integrated Brier score}
We could of course integrate the Brier score across time to arrive at a single number.
\begin{fdefinition}[Integrated Brier score]\label{def:ibs}
Suppose that we have a survival function estimate $\widehat{S}(\cdot|x)$ for any $x\in\mathcal{X}$. Let $t_{\min}\ge0$ and $t_{\max}>t_{\min}$ be user-specified lower and upper limits of integration. The integrated Brier score is defined as
\[
\text{IBS}
:= \frac{1}{t_{\max} - t_{\min}} \int_{t_{\min}}^{t_{\max}} \text{BS}(u)\textrm{d}u.
\]
This score is also nonnegative, where lower scores are better.
\end{fdefinition}

\subsubsection{Distribution Calibration}

To assess how well-calibrated a predicted survival function $\widehat{S}(\cdot|x)$ is, \citet{haider2020effective} proposed a calibration metric called Distribution Calibration (abbreviated ``D-Calibration''). To explain how this works, Haider \emph{et al.}~consider $\widehat{S}(\cdot|x)$ to be perfectly calibrated if
\begin{equation}
\mathbb{P}\Big( \widehat{S}(T|X) \in [a,b] \Big) = b - a
\quad\text{for all intervals }[a,b]\subseteq[0,1],
\label{eq:perfect-calibration}
\end{equation}
where the probability is over randomness in sampling raw input~$X$ and its corresponding survival time $T$ as stated in the generative procedure in Section~\ref{sec:setup}.

Note that if we plug in the true survival function $S(\cdot|x)$ in place of the estimate $\widehat{S}(\cdot|x)$ into \eqref{eq:perfect-calibration}, then $S(\cdot|x)$ would be perfectly calibrated. This is a consequence of the \emph{probability integral transform} result that states that for any continuous real-valued random variable~$A$ with CDF~$G:\mathbb{R}\rightarrow[0,1]$, the random variable~$G(A)$ is uniformly distributed over $[0,1]$.\footnote{We are applying the probability integral transform result to the continuous random variable corresponding to $T$ conditioned on $X=x$, which has CDF $F(\cdot|x)$. Thus, if we evaluate the CDF at random survival time $T$ sampled from $\mathbb{P}_{T|X=x}$, \ie, we compute the random variable $F(T|x)$, then this random variable is uniform over $[0,1]$. Since $S(\cdot|x) = 1 - F(\cdot|x)$, we conclude that $S(T|x)$ is also uniformly distributed over $[0,1]$. This analysis holds for any realization $x$ that could be sampled from $\mathbb{P}_X$ so that accounting for randomness in sampling $X=x$, we still get that $S(T|X)$ is uniform over $[0,1]$.}

To turn condition~(\ref{eq:perfect-calibration}) into a calibration metric that we can compute, we empirically evaluate it for some user-specified intervals. We first do this when there is no censoring. Note that the D-Calibration procedure that we are about to describe is a bit more involved than the previous evaluation metrics. As we shall see, it ultimately leads to thresholding on a p-value of a statistical test to decide whether the distribution corresponding to $\widehat{S}(\cdot|x)$ should be deemed calibrated or~not.

\paragraph{The case without censoring} We define the following subset of data points for any interval $\mathcal{I}\subseteq[0,1]$:
\[
\mathcal{D}(\mathcal{I})
:= \{ i\in[n] : \Delta_i = 1, \widehat{S}(Y_i|X_i) \in \mathcal{I} \}.
\]
In particular, point $i$ being in $\mathcal{D}(a,b)$ means that point~$i$ died (so that~$Y_i = T_i$) and the predicted survival probability $\widehat{S}(Y_i|X_i)$ is in the interval $\mathcal{I}$. Then we would like
\[
\frac{|\mathcal{D}([a,b])|}{n} \approx b-a
\quad\text{for }[a,b]\subseteq[0,1].
\]
We evaluate this condition for some pre-specified equal-width intervals that cover the whole interval $[0,1]$. For example, we could use the 10 bins $\mathcal{I}_1:=[0,0.1], \mathcal{I}_2:=(0.1,0.2], \dots, \mathcal{I}_{10}:=(0.9, 1]$ (the intervals do not have to be closed on both ends). Here, we would like
\[
\frac{|\mathcal{D}(\mathcal{I}_{\ell})|}{n} \approx \frac{1}{10}
\quad\text{for all }\ell\in[10],
\]
meaning that we want to check that the 10 probabilities $\widehat{p}_1:=\frac{|\mathcal{D}(\mathcal{I}_1)|}{n}, \dots, \widehat{p}_{10}:=\frac{|\mathcal{D}(\mathcal{I}_{10})|}{n}$ are close to a uniform distribution. Haider \emph{et al.}~suggest using a standard chi-squared test to check for uniformity, which in this case corresponds to computing the test statistic
\[
\chi^2
:=
10 n
\sum_{\ell=1}^{10} \Big( \widehat{p}_\ell - \frac{1}{10} \Big)^2
= 10 n \sum_{\ell=1}^{10} \bigg( \frac{|\mathcal{D}(\mathcal{I}_{\ell})|}{n} - \frac{1}{10} \bigg)^2.
\]
We then compute
\[
\text{p-value}
:= \mathbb{P}(\text{chi-squared variable with $9$ degrees of freedom}\ge \chi^2).
\]
If the p-value is at least some user-specified threshold (\citet{haider2020effective} use 0.05), then we declare the distribution to be uniform, so the predicted survival function $\widehat{S}$ is considered calibrated. Otherwise, we consider $\widehat{S}$ to not be calibrated. For a continuous version that is not binary, we could, for instance, use the $\chi^2$ test statistic (\citet{goldstein2020x} use what we have written for the $\chi^2$ test statistic but they exclude the $10n$ scale factor).

\paragraph{Accounting for censoring} By how we have defined the set $\mathcal{D}(\mathcal{I})$ above, note that probabilities $\widehat{p}_1,\dots,\widehat{p}_{10}$ that we checked against a uniform distribution would not sum to 1 if there are censored data:
\[
\sum_{\ell=1}^m \widehat{p}_{(\ell)}
= \sum_{\ell=1}^m \frac{|\mathcal{D}(\mathcal{I}_{\ell})|}{n}
= \frac{1}{n} \sum_{i=1}^n \ind\{\Delta_i = 1\}
= 1 - \underbrace{\frac{1}{n} \sum_{i=1}^n \ind\{\Delta_i = 0\}}_{\substack{\text{fraction of data}\\\text{that are censored}}}.
\]
The idea then is that we will modify the estimated probabilities $\widehat{p}_1,\dots,\widehat{p}_{10}$, increasing them in a particular way so that they form a valid probability distribution and that they use information from the censored data.

Prior to doing any modification, remember that for each interval $\ell\in[10]$,
\[
\widehat{p}_\ell
= \frac{|\mathcal{D}(\mathcal{I}_\ell)|}{n}
= \sum_{i\in\mathcal{D}(\mathcal{I}_\ell)} \frac{1}{n}.
\]
The right-most expression suggests the following interpretation: each training point in $\mathcal{D}(\mathcal{I}_\ell)$ (i.e., each training point that is assigned to interval $\ell$) contributes a probability mass of $1/n$ to interval $\ell$.

We now view each censored point to also have probability mass $1/n$, but we want to figure out how to allocate this probability mass to the 10 different intervals. \citet{haider2020effective} proposed the following strategy. For each point $i\in[n]$ that is censored (\ie, $\Delta_i=0$), we look at which interval $\widehat{S}(Y_i|X_i)$ falls into among $\mathcal{I}_1,\dots,\mathcal{I}_{10}$. Denote the resulting interval's index as $\widetilde{\ell}\in[10]$. We then distribute the probability mass $1/n$ evenly among intervals $\widetilde{\ell},\widetilde{\ell}+1,\dots,10$. In other words, for censored point $i$, we update
\[
\widehat{p}_\ell \leftarrow \widehat{p}_{\ell} + \frac{1}{n}\cdot\Big(\frac{1}{10 - \widetilde{\ell} + 1}\Big)
\quad\text{for }\ell \in \{\widetilde{\ell},\widetilde{\ell}+1,\dots,10\}.
\]
After iterating through all censored points and distributing each of their $1/n$ probability mass to the intervals in the above manner, the resulting probabilities $\widehat{p}_1,\dots,\widehat{p}_{10}$ will indeed sum to~1, and we proceed with the chi-squared test as before to determine if the distribution is calibrated.

\subsubsection{Error Metrics for Survival Time Point Estimates}
\label{sec:eval-point-estimates}

Aside from the original c-index evaluation metric, all the other evaluation metrics we covered so far evaluated predicted survival functions $\widehat{S}(\cdot|\cdot)$ for different raw inputs at different times. Now we turn to evaluation metrics specific to when we predict a single survival time number per data point.

For point~$i$ with predicted survival function $\widehat{S}(\cdot|X_i)$, we denote its predicted survival time as $\widehat{T}_i$, which we could take to be the median survival time estimate (the time at which $\widehat{S}(\cdot|X_i)$ crosses probability 1/2) or the mean survival time estimate (area under the function $\widehat{S}(\cdot|X_i)$).

If point~$i$ is not censored, then we could easily use any standard regression error metric to compare $Y_i$ and $\widehat{T}_i$, such as squared error $(Y_i - \widehat{T}_i)^2$ or absolute error $|Y_i - \widehat{T}_i|$. If point~$i$ is censored, then we have no ground truth survival time to compare against. A naive solution is to use the so-called \emph{hinge} error, for which the squared error version is $(Y_i - \widehat{T}_i)^2\ind\{ Y_i > \widehat{T}_i \}$ and the absolute error version is $(Y_i - \widehat{T}_i)\ind\{ Y_i > \widehat{T}_i \}$; these only give nonzero error when the predicted survival time $\widehat{T}_i$ is less than the observed time $Y_i$ (which we know to be when the point is still~alive since it is a censoring time).

We state two strategies for dealing with censored data that have been shown to often work well. Both involve estimating a pseudo ``ground truth'' survival time to compare the predicted survival time against for each censored data point (put another way, we are imputing the ground truth survival times for censored data). Let $\widehat{S}_{\text{KM}}(\cdot)$ denote the Kaplan-Meier estimate of the population-level survival function $S_{\text{pop}}(t):=\mathbb{P}(T>t)$ (as given in \eqref{eq:kaplan-meier}). Note that this Kaplan-Meier estimator is fitted to the \emph{training} data, meaning that even if we are evaluating error on validation or test data that are not the same as the training data, in the equations that follow, $\widehat{S}_{\text{KM}}(\cdot)$ is fitted to the training data (this is similar to what we stated in Remark~\ref{rem:val-test-data-censoring-km} with the difference being that we are now estimating $S_{\text{pop}}(t)=\mathbb{P}(T>t)$ and not $S_{\text{censor}}(t)=\mathbb{P}(C>t)$).

The \emph{margin} \citep{haider2020effective} approach estimates the pseudo ground truth survival time of a censored point~$i\in[n]$ to be
\[
T_i^{\text{margin}}
:= Y_i + \frac{\int_{Y_i}^\infty \widehat{S}_{\text{KM}}(u)\textrm{d}u}
              {\widehat{S}_{\text{KM}}(Y_i)},
\]
where we numerically evaluate the integral. The right-hand side aims to estimate $\mathbb{E}[T_i | T_i > Y_i]$.\footnote{Recall from Section~\ref{sec:estimands-continuous} that integrating a survival function from time 0 to time $\infty$ yields the mean survival time. Integrating a survival function instead from time $Y_i$ to time $\infty$ and dividing by $\mathbb{P}(T_i > Y_i)$ yields the mean survival time conditioned on survival beyond time $Y_i$.}

Meanwhile, the \emph{pseudo-observation} (PO) approach \citep{qi2023effective} instead estimates the ground truth survival time of a censored point~$i\in[n]$ to be
\begin{equation}
T_i^{\text{PO}}
:= (n+1) \underbrace{\int_0^\infty \widehat{S}_{\text{KM}^{+i}}(u)\textrm{d}u}_{\substack{\text{estimate of mean survival}\\\text{time including evaluation}\\\text{point }i}} - n \underbrace{\int_0^\infty \widehat{S}_{\text{KM}}(u)\textrm{d}u}_{\substack{\text{estimate of mean survival}\\\text{time excluding evaluation}\\\text{point }i}},
\label{eq:T_i-PO}
\end{equation}
where $\widehat{S}_{\text{KM}^{+i}}(\cdot)$ refers to the Kaplan-Meier estimator fitted to the training data with evaluation point~$i$ included as an additional training point. Once again, the integrals are numerically evaluated. The first integral is an estimate of the mean survival time of survival function $\widehat{S}_{\text{KM}^{+i}}(\cdot)$, and the second integral is an estimate of the mean survival time of survival function $\widehat{S}_{\text{KM}}(\cdot)$. Taking this difference is based on the bias-corrected jackknife estimator.\footnote{As a technical remark, our exposition here of the PO approach follows how \citet{qi2024survivaleval} have currently implemented it in their GitHub repository rather than how they have originally stated it in their paper and in their earlier work \citep{qi2023effective}. The difference is that originally, \citet{qi2023effective} define $T_i^{\text{PO}} := n \int_0^\infty \widehat{S}_{\text{KM}}(u)\textrm{d}u - (n-1) \int_0^\infty \widehat{S}_{\text{KM}^{-i}}(u)\textrm{d}u$, where $\widehat{S}_{\text{KM}}$ is assumed to be fitted to a dataset that includes evaluation point $i$, and $\widehat{S}_{\text{KM}^{-i}}$ is the version of the Kaplan-Meier estimator fitted to the dataset excluding evaluation point $i$.}

After computing one of these pseudo ground truth labels, we could treat a censored point's pseudo ground truth survival time as if it were a real ground truth survival time and evaluate standard regression error metrics like squared error or absolute error. For example, the mean absolute error metric using the PO approach would be given by
\begin{equation}
\text{MAE-PO}
:=
\frac{1}{n}\sum_{i=1}^n
  \big(\Delta_i \!\!\!\!\underbrace{|\widehat{T}_i - Y_i|}_{\substack{\text{when uncensored,}\\\text{use observed time}}}\!\!\!\!
   + (1 - \Delta_i) \!\!\!\!\underbrace{|\widehat{T}_i - T_i^{\text{PO}}|}_{\substack{\text{when censored, use}\\\text{pseudo ground truth}}}\!\!\!\!\big),
\label{eq:MAE-PO}
\end{equation}
where, as a reminder, $\widehat{T}_i$ is the predicted survival time of the $i$-th point using the time-to-event prediction model that we are evaluating (possibly where we convert a predicted survival function into a point estimate by backing out a median or mean survival time estimate).

\citet{haider2020effective} explained that taking an equally weighted average as in \eqref{eq:MAE-PO} may not be a good idea, as we may be more confident in the (pseudo) ground truth values for some points vs others. The intuition is as follows. Suppose that we are measuring survival times of people in years, and that for the population under consideration, no one has a survival time greater than 130 years. Imagine that a data point (corresponding to a person) was censored at time~0 (and censoring times are independent of survival times). Then we know very little about what the true survival time should be, and the pseudo ground truth value computed (whether using the margin, PO, or some other approach altogether) would likely be unreliable. In contrast, suppose that a data point was censored at 110 years. For this data point, we would be much more confident about the pseudo ground truth value being close to the true value. With this intuition, Haider \emph{et al.}~suggested that for data points that are censored, we should give higher weights to points that are censored later. They operationalize this intuition by using a weighted mean absolute error metric
\begin{align*}
&\text{weighted-MAE-PO} \\
&\quad:=
\frac{1}{\sum_{i=1}^n w_i}\sum_{i=1}^n w_i
  \big(\Delta_i \!\!\!\!\underbrace{|\widehat{T}_i - Y_i|}_{\substack{\text{when uncensored,}\\\text{use observed time}}}\!\!\!\!
   + (1 - \Delta_i) \!\!\!\!\underbrace{|\widehat{T}_i - T_i^{\text{PO}}|}_{\substack{\text{when censored, use}\\\text{pseudo ground truth}}}\!\!\!\!\big),
\end{align*}
where they assign weights as follows:
\[
w_i
:=
\begin{cases}
1 & \text{if }\Delta_i = 1, \\
1 - \widehat{S}_{\text{KM}}(Y_i) & \text{if }\Delta_i = 0.
\end{cases}
\]
For a thorough experimental evaluation of these pseudo ground truth evaluation metrics, and also for details on why using (weighted) mean absolute error makes sense in many time-to-event prediction tasks, see the paper by \citet{qi2023effective}.

Meanwhile, \citet{qi2023effective} also showed that MAE with median survival times is a proper scoring rule for \emph{uncensored} datasets. However, this theoretical result is unsatisfying in that the main technical hurdle in survival analysis is censoring.

\subsection{Additional Remarks on Classification and Regression}
\label{sec:connections-regression-classification}

Although we already discussed how the time-to-event prediction in discrete time relates to binary classification at different time indices (Section~\ref{sec:setup-discrete}), the models we had derived using maximum likelihood (DeepHit, Nnet-survival, Kaplan-Meier and Nelson Aalen estimators) were not just existing off-the-shelf binary classifiers. In this section, we discuss an approach called \emph{survival stacking} \citep{craig2021survival} that converts any time-to-event prediction problem with raw input space $\mathcal{X}=\mathbb{R}^d$ into a binary classification problem such that we can use any off-the-shelf probabilistic binary classifier for prediction such as logistic regression or random forests (Section~\ref{sec:survival-stacking}). This conversion fundamentally models time to be discrete and can be quite expensive: with $n$ training points that each have $d$ features, the input training feature matrix (each row is a feature vector) could be viewed as a 2D table that is $n$-by-$d$. After converting the problem using survival stacking, the training feature matrix for the binary classifier could have as many as $\mathcal{O}(n^2)$ rows and $\mathcal{O}(d+n)$ columns.

Separately, we relate time-to-event prediction to the classical regression setup (Section~\ref{sec:conditional-CDF-estimation}). This relationship is more straightforward and considers what happens if censoring did not happen.

\subsection[Survival Stacking: Converting Time-to-Event\texorpdfstring{\\}{} Prediction to Binary Classification]{Survival Stacking: Converting Time-to-Event Prediction to Binary Classification}
\label{sec:survival-stacking}

To explain survival stacking, we follow \citet{craig2021survival} and provide an example of how survival stacking converts a small toy time-to-event prediction problem with $n=3$ and $\mathcal{X}=\mathbb{R}^d$ with $d=2$ into a binary classification problem (we use their same toy example, although we use the notation that we have introduced in this monograph).

Specifically, suppose that we have three training points $(X_1,Y_1,\Delta_1),(X_2,Y_2,\Delta_2),(X_3,Y_3,\Delta_3)$, where for ease of exposition, we assume that these points are sorted so that $Y_1<Y_2<Y_3$. Suppose that $\Delta_1=1$, $\Delta_2=0$, and $\Delta_3=1$. In terms of notation, each $X_i$ is in $\mathbb{R}^2$, for which we write $X_i=(X_{i,1},X_{i,2})\in\mathbb{R}^2$. Thus, we could view our training data as follows:
\[
\begin{blockarray}{ccc}
& \text{\phantom{M}} & \text{\phantom{M}} \\
& \text{feature 1} & \text{feature 2} \\
\begin{block}{c[cc]}
\text{point 1} & X_{1,1} & X_{1,2} \\
\text{point 2} & X_{2,1} & X_{2,2} \\
\text{point 3} & X_{3,1} & X_{3,2} \\
\end{block}
\end{blockarray}
\qquad
\begin{blockarray}{cc}
\text{observed} & \text{event} \\
\text{time} & \text{indicator} \\
\begin{block}{[cc]}
Y_1 & 1 \\
Y_2 & 0 \\
Y_3 & 1 \\
\end{block}
\end{blockarray}
\]
\Paragraph{Enumerate the unique times of death}
To begin the survival stacking conversion process, we enumerate the unique times in which death occurred, which in this case is $\tau_{(1)} = Y_1$ and $\tau_{(2)} = Y_3$ (we are intentionally reusing our notation from earlier for modeling discrete time, where the time grid points are denoted $\tau_{(1)}<\cdots<\tau_{(L)}$). In this case, there are~$L=2$ unique times of death.

\paragraph{Collect information from time index~1}
At time index 1 (corresponding to time $\tau_{(1)}=Y_1$), we list all the data points that are ``at risk'' (could still possibly die) at that time. In this case, since $Y_1<Y_2<Y_3$, at time $\tau_{(1)}=Y_1$, all three data points are at risk. What we do then is we create the following 2D table:
\[
\mathbf{X}_{(1)}
:=
\begin{blockarray}{ccccc}
& \text{feature 1} & \text{feature 2} & \text{time index 1} & \text{time index 2} \\
\begin{block}{c[cc|cc]}
\text{point 1} & X_{1,1} & X_{1,2} & 1 & 0 \\
\text{point 2} & X_{2,1} & X_{2,2} & 1 & 0 \\
\text{point 3} & X_{3,1} & X_{3,2} & 1 & 0 \\
\end{block}
\end{blockarray}
\]
In particular, the number of rows of $\mathbf{X}_{(1)}$ is the number of points at risk at time index~1 (corresponding to time $\tau_{(1)}$) while the number of columns is $d+L = 2+2 = 4$. We have added new columns that simply indicate which time index we are currently looking at. Next, we also create the following vector that indicates which of the points at risk actually died at time index~1, which would only be training point 1:
\[
\mathbf{y}_{(1)}
:=
\begin{blockarray}{cc}
& \text{death at time index 1} \\
\begin{block}{c[c]}
\text{point 1} & 1 \\
\text{point 2} & 0 \\
\text{point 3} & 0 \\
\end{block}
\end{blockarray}
\]

\paragraph{Collect information from time index~2}
Now we proceed to time index~2 (corresponding to time $\tau_{(2)}=Y_3$) and repeat the same idea. We list all the data points at risk, which would just be data point~3. We create the following 2D table:
\[
\mathbf{X}_{(2)}
:=
\begin{blockarray}{ccccc}
& \text{feature 1} & \text{feature 2} & \text{time index 1} & \text{time index 2} \\
\begin{block}{c[cc|cc]}
\text{point 3} & X_{3,1} & X_{3,2} & 0 & 1 \\
\end{block}
\end{blockarray}
\]
We also create a vector indicating that point 3 did experience death at time index~2:
\[
\mathbf{y}_{(2)}
:=
\begin{blockarray}{cc}
& \text{death at time index 2} \\
\begin{block}{c[c]}
\text{point 3} & 1 \\
\end{block}
\end{blockarray}
\]

\paragraph{Stack information vertically to get training data for classifier}
At this point, we have gone through all the unique times of death. We vertically stack $\mathbf{X}_{(1)}$ and $\mathbf{X}_{(2)}$, and similarly we vertically stack $\mathbf{y}_{(1)}$ and $\mathbf{y}_{(1)}$ to get:
\begin{align*}
\mathbf{X}
&:=
  \begin{bmatrix}
  \mathbf{X}_{(1)} \\[-.5em]
  \rule[.5ex]{5ex}{0.5pt} \\[-.5em]
  \mathbf{X}_{(2)}
  \end{bmatrix}
 =\begin{bmatrix}
  X_{1,1} & X_{1,2} & 1 & 0 \\
  X_{2,1} & X_{2,2} & 1 & 0 \\
  X_{3,1} & X_{3,2} & 1 & 0 \\
  X_{3,1} & X_{3,2} & 0 & 1
  \end{bmatrix}
\\
\mathbf{y}
&:=
  \begin{bmatrix}
  \mathbf{y}_{(1)} \\[-.5em]
  \rule[.5ex]{5ex}{0.5pt} \\[-.5em]
  \mathbf{y}_{(2)}
  \end{bmatrix}
 =\begin{bmatrix}
  1 \\
  0 \\
  0 \\
  1
  \end{bmatrix}
\end{align*}
Then we treat the rows of $\mathbf{X}$ as feature vectors, where the $i$-th row has binary classification label given by the $i$-th entry of $\mathbf{y}$, and we use these to train a probabilistic binary classifier of our choosing, such as logistic regression or a random forest.

\paragraph{The binary classification task}
To help make sense of what this binary classifier is predicting, consider the fourth training feature vector being fed in, \ie, the last row of $\mathbf{X}$: $(X_{3,1},X_{3,2},0,1)$. The classifier is being told what the original feature vector is for this data point (namely $X_3 = (X_{3,1},X_{3,2})$) along with which time index we are making a prediction for (time index~2, encoded as the vector $(0,1)$), which of course implies that the data point is still alive at this particular time step. We are thus predicting the probability of death, assuming that the point is still at risk. In other words, we are predicting the discrete time hazard probability at time index~2.

More generally, what is happening is that we are discretizing time using the unique times of death $\tau_{(1)}<\tau_{(2)}<\cdots<\tau_{(L)}$. Let $e_\ell\in\{0,1\}^L$ denote an $L$-dimensional vector that is all zeros except for a single~1 at the $\ell$-th entry, where $\ell\in[L]$. For any test feature vector $x\in\mathbb{R}^d$ in the original space, let $x\oplus e_\ell$ denote the concatenation of vectors $x$ and $e_\ell$ (so $x\oplus e_\ell$ is in $\mathbb{R}^{d+L}$). Then the binary classifier, given input feature vector $x\oplus e_\ell$, predicts the hazard probability corresponding to the original feature vector $x$ at time index $\ell\in[L]$, \ie, what we had denoted as $h[\ell|x]$ in our earlier coverage. When the binary classifier is logistic regression, \citet[Section~2.3]{craig2021survival} show that the resulting classifier closely approximates the original Cox proportional hazards model~\citep{cox1972regression}. We discuss the Cox model later in Section~\ref{sec:cox-semiparametric}.

\paragraph{Training dataset size for the binary classifier} In the toy example above, we ended up with a classification training feature matrix $\mathbf{X}$ that is 4-by-4 even though the original time-to-event prediction feature matrix was only 3-by-2. How much larger could the classification training feature matrix be?

We can compute the exact size of $\mathbf{X}$. In general, the number of columns of $\mathbf{X}$ is $d+L$ where, as a reminder, $L$ is the number of unique times of death. As for the number of rows of $\mathbf{X}$, we can determine this by adding up the number of rows of $\mathbf{X}_{(1)}$, $\mathbf{X}_{(2)}$, and so forth (which we had vertically stacked to form $\mathbf{X}$). The number of rows in each $\mathbf{X}_{(\ell)}$ (for $\ell\in[L]$) is precisely the number of points at risk at time index~$\ell$, given by $N[\ell]=\ind\{Y_j \ge \tau_{(\ell)}\}$ from \eqref{eq:at-risk-counter}. Thus, in general,
\[
\text{number of rows in }\mathbf{X}
=
\sum_{\ell=1}^L N[\ell] = \sum_{\ell=1}^L \sum_{j=1}^n \ind\{Y_j \ge \tau_{(\ell)}\}.
\]
In the worst case, every observed time is unique and is a time of death (so that at each time index, exactly one point dies). In this case, one could see that at the first time index there are $n$ points at risk (so $\mathbf{X}_{(1)}$ would have $n$ rows), at the second time index there are $n-1$ points at risk (so $\mathbf{X}_{(2)}$ would have $n-1$ rows), \etc. In this case, the number of rows of $\mathbf{X}$ is $n+(n-1)+(n-2)+\cdots+1 = \frac{n(n+1)}{2}$. Meanwhile, if every observed time is unique, then it means that $L=n$, so the number of columns of $\mathbf{X}$ is $d+L=d+n$. Thus, the worst case size of $\mathbf{X}$ is $\frac{n(n+1)}{2} = \mathcal{O}(n^2)$ rows by $d+n$ columns, which of course is larger than the time-to-event prediction feature matrix size of $n$-by-$d$.

In practice, to prevent the stacked matrix $\mathbf{X}\in\mathbb{R}^{(\sum_{\ell=1}^L N[\ell])\times (d+L)}$ from being too large, what one could do is discretize the time grid to be coarser: instead of taking $L$ to be the number of unique times of death, we could manually specify $L$ to be much smaller than $n$ and discretize $Y_i$ values to $L$ time indices (we presented some ways of doing this in Section~\ref{sec:how-to-discretize}). In doing so, we would directly control the number of columns of $\mathbf{X}$ (since we would set the value of $L$ and the number of columns is $d+L$), while also decreasing the number of rows of $\mathbf{X}$. Note that dealing with the number of rows of $\mathbf{X}$ being large is somewhat less of an issue in that many classifiers are designed to scale to large datasets (\eg, XGBoost \citep{chen2016xgboost}). Classifiers that support minibatch training could avoid looking at all rows of $\mathbf{X}$ at once.

\subsubsection{Connection to Regression: Conditional CDF Estimation}
\label{sec:conditional-CDF-estimation}

The connection to regression is straightforward: consider the case where censoring happens with probability~0 (which could be thought of as an extreme case where the censoring time $C$ is deterministically $+\infty$ in the generative procedure from Section~\ref{sec:setup}). Then our training data would be the same as that of standard regression (we could ignore the event indicator $\Delta_i$ variables since they would all be equal to~1) although all the regression labels (the $Y_i$ variables) are guaranteed to be nonnegative (whereas in, for instance, linear regression, the regression labels are not constrained to be nonnegative). The prediction task of estimating the conditional survival function without censoring would simply amount to conditional CDF estimation for a regression label, which has previously been studied (\eg, \citealt{chagny2014adaptive}).

\begin{subappendices}
\subsection{Technical Details}
\label{sec:setup-technical}

\subsubsection{Definition of the Raw Input Space}
\label{sec:feature-support}

As stated in Section~\ref{sec:setup}, we assume that the raw input space $\mathcal{X}$ is the ``support'' of distribution $\mathbb{P}_X$. Roughly, the support of $\mathbb{P}_X$ consists of all possible values that we could sample from $\mathbb{P}_X$. We now build up to formally defining what the support of a distribution is.

First, to motivate why $\mathcal{X}$ cannot be defined arbitrarily, consider the following toy example. Suppose that we set~$\mathcal{X}:=\mathbb{R}$, and that~$\mathbb{P}_X$ is uniform over the unit interval~$[0,1]$. However, we aim to be able to make predictions (\ie, to estimate one of the target functions that fully characterize $\mathbb{P}_{T|X}(\cdot|x)$) for all $x\in\mathcal{X}$. In the toy example given, $\mathbb{P}_{T|X}(\cdot|x)$ would not actually be defined when~$x$ is outside of $[0,1]$ (\eg, we cannot condition on $X=2$). The fix is simple in this case: we should instead define~$\mathcal{X}:=[0,1]$.

In general, we should set~$\mathcal{X}$ to be the \emph{support} of distribution~$\mathbb{P}_X$, which we denote as $\text{supp}(\mathbb{P}_X)$. As concrete examples:
\begin{itemize}
\item If $X$ is a discrete random vector in $\mathbb{R}^d$, then
\[
\text{supp}(\mathbb{P}_X):=\{x\in\mathbb{R}^d:\mathbb{P}(X=x)>0\}.
\]
\item If $X$ is a continuous random vector over $\mathbb{R}^d$ with PDF $f_X(\cdot)$, then
\[
\text{supp}(\mathbb{P}_X):=\overline{\{x\in\mathbb{R}^d:f_X(x)>0\}},
\]
where the line over the set indicates that we are taking its closure.
\end{itemize}
However, even if $X$ resides in $\mathbb{R}^d$ so that it is a fixed-length feature vector, in real applications, $X$ could consist of a mix of discrete and continuous features. In this case, the above definitions that require all features to be discrete or all features to be continuous are not adequate. A general definition of the support of $\mathbb{P}_X$ that works whenever $X$ takes on a value in $\mathbb{R}^d$ is as follows:
\[
\text{supp}(\mathbb{P}_X)
:= \{ x\in\mathbb{R}^d : \mathbb{P}( \| X - x \| \le r) > 0\text{ for all }r>0\},
\]
where $\|\cdot\|$ denotes Euclidean distance.

Since neural networks can accommodate input spaces $\mathcal{X}$ that are not $\mathbb{R}^d$, we now substantially generalize the definition of $\text{supp}(\mathbb{P}_X)$. Specifically, suppose that $\mathbb{P}_X$ is defined over a separable metric space $(\mathcal{X}, \rho)$, where the function $\rho:\mathcal{X}\times\mathcal{X}\rightarrow[0,\infty)$ is a metric: for any two $x,x'\in\mathcal{X}$, $\rho(x,x')$ gives a distance between $x$ and $x'$ (if $\mathcal{X}$ is Euclidean space, then we could take $\rho$ to be Euclidean distance). Then we define
\[
\text{supp}(\mathbb{P}_X)
:= \{ x\in\mathcal{X} : \mathbb{P}( \rho( X, x ) \le r) > 0\text{ for all }r>0\}.
\]
The reason we assumed that the metric space is separable is to guarantee that $\mathbb{P}\big(X\in\text{supp}(\mathbb{P}_X)\big) = 1$ \citep{cover1967nearest}.

In summary, by defining $\mathcal{X} := \text{supp}(\mathbb{P}_X)$, we can indeed condition on~$X=x$ for all $x\in\mathcal{X}$, and in particular, we can then work with the conditional distribution $\mathbb{P}_{T|X}(\cdot|x)$ for all $x\in\mathcal{X}$.

\subsubsection{Proof of Proposition~\ref{prop:discrete-cumulative-hazard-not-neg-log-surv}: \texorpdfstring{$H[\ell|x]$}{H[l|x]} as a First-Order Taylor Approximation of \texorpdfstring{$-\log S[\ell|x]$}{-log S[l|x]}}
\label{sec:discrete-cumulative-hazard-not-neg-log-surv-pf}

Recall the Taylor series expansion ${-\log(1 - z)} = \sum_{p=1}^\infty \frac{z^p}{p}$ for $z\in[0,1)$. Then assuming that $h[\ell|x]\in[0,1)$ for all $\ell\in[L]$, we have
\begingroup
\allowdisplaybreaks
\begin{align*}
-\log S[\ell|x]
&=-\log\prod_{m=1}^\ell (1-h[m|x]) \\
&=\sum_{m=1}^\ell -\log(1-h[m|x]) \\
&=\sum_{m=1}^\ell \sum_{p=1}^{\infty}\frac{(h[m|x])^p}{p} \\
&=\sum_{m=1}^\ell
    \bigg(
      h[m|x]
      +
      \sum_{p=2}^{\infty}
        \frac{(h[m|x])^p}{p}
    \bigg) \\
&=H[\ell|x]+\sum_{m=1}^\ell \sum_{p=2}^{\infty}\frac{(h[m|x])^p}{p}. \tag*{$\square$}
\end{align*}
\endgroup

\subsubsection{Hazard Function Maximum Likelihood Derivation for the Kaplan-Meier and Nelson-Aalen Estimators}
\label{sec:kaplan-meier-nelson-aalen-hazard-maximum-likelihood-derivation}

In this section, we derive \eqref{eq:kaplan-meier-nelson-aalen-hazard-maximum-likelihood}:
\begin{equation}
\widehat{\theta}_\ell
:= \arg\max_{\theta_\ell} \mathcal{L}_{(\ell)}(\theta_\ell)
 = \frac{D[\ell]}{N[\ell]}
\quad\text{for }\ell\in[L],
\tag*{(\ref{eq:kaplan-meier-nelson-aalen-hazard-maximum-likelihood}, reproduced)}
\end{equation}
where, as a reminder:
\begingroup
\allowdisplaybreaks
\begin{align*}
\mathcal{L}_{(\ell)}(\theta_\ell)
&:=
  \sum_{i=1}^n
    \ind\{\ell\le\kappa(Y_i)\}
    \big\{
      \binaryeventmatrix_{i,\ell} \log \theta_\ell
      + (1 - \binaryeventmatrix_{i,\ell}) \log (1-\theta_\ell)
    \big\},
\tag*{(\ref{eq:log-likelihood-pmf-hazard-alt2}, partially reproduced)} \\
D[\ell]
&:= \sum_{j=1}^n\ind\{Y_j=\tau_{(\ell)}\}\Delta_j = \sum_{j=1}^n \binaryeventmatrix_{j,\ell},
\tag*{(\ref{eq:death-counter}, reproduced)} \\
N[\ell]
&:= \sum_{j=1}^n\ind\{Y_j\ge\tau_{(\ell)}\} = \sum_{j=1}^n \ind\{\kappa(Y_j)\ge\ell\}.
\tag*{(\ref{eq:at-risk-counter}, reproduced)}
\end{align*}
\endgroup
We set the derivative of $\widehat{\theta}_\ell$ with respect to $\theta_\ell$ to 0 (for the moment, we do not worry about the constraint that $\theta_\ell\in[0,1]$; as we shall see shortly, the value of $\theta_\ell$ that achieves derivative 0 is guaranteed to be between 0 and 1). We have
\begingroup
\allowdisplaybreaks
\begin{align*}
0 & =\Big[\frac{\textrm{d}\log\mathcal{L}_{(\ell)}(\theta_\ell)}{\textrm{d}\theta_\ell}\Big]_{\theta_\ell=\widehat{\theta}_\ell}\\
 & =\sum_{i=1}^{n}\ind\{\ell\le\kappa(Y_i)\}\Big[\frac{\binaryeventmatrix_{i,\ell}}{\widehat{\theta}_\ell}-\frac{1-\binaryeventmatrix_{i,\ell}}{1-\widehat{\theta}_\ell}\Big]\\
 & =\frac{\sum_{i=1}^{n}\ind\{\ell\le\kappa(Y_i)\}\binaryeventmatrix_{i,\ell}}{\widehat{\theta}_\ell}-\frac{\sum_{i=1}^{n}\ind\{\ell\le\kappa(Y_i)\}(1-\binaryeventmatrix_{i,\ell})}{1-\widehat{\theta}_\ell}\\
 & =\frac{\sum_{i=1}^{n}\ind\{\ell\le\kappa(Y_i)\}\binaryeventmatrix_{i,\ell}}{\widehat{\theta}_\ell}\\
 & \quad-\frac{\sum_{i=1}^{n}\ind\{\ell\le\kappa(Y_i)\}-\sum_{i=1}^{n}\ind\{\ell\le\kappa(Y_i)\}\binaryeventmatrix_{i,\ell}}{1-\widehat{\theta}_\ell}.
\end{align*}
\endgroup
Rearranging terms, we have
\begin{align*}
&\frac{\sum_{i=1}^{n}\ind\{\ell\le\kappa(Y_i)\}\binaryeventmatrix_{i,\ell}}{\widehat{\theta}_\ell}=\frac{\sum_{i=1}^{n}\ind\{\ell\le\kappa(Y_i)\}-\sum_{i=1}^{n}\ind\{\ell\le\kappa(Y_i)\}\binaryeventmatrix_{i,\ell}}{1-\widehat{\theta}_\ell}\\
&\Longleftrightarrow
  \widehat{\theta}_\ell
  =\frac{\sum_{i=1}^{n}\ind\{\ell\le\kappa(Y_i)\}\binaryeventmatrix_{i,\ell}}
       {\sum_{i=1}^{n}\ind\{\ell\le\kappa(Y_i)\}} \\
&\phantom{\Longleftrightarrow\widehat{\theta}_\ell}
  =\frac{\sum_{i=1}^{n}\ind\{\ell\le\kappa(Y_i)\}\Delta_i\ind\{\kappa(Y_i)=\ell\}}{\sum_{i=1}^{n}\ind\{\ell\le\kappa(Y_i)\}} \\
&\phantom{\Longleftrightarrow\widehat{\theta}_\ell}
  =\frac{\sum_{i=1}^{n}\Delta_i\ind\{\kappa(Y_i)=\ell\}}{\sum_{i=1}^{n}\ind\{\ell\le\kappa(Y_i)\}} \\
&\phantom{\Longleftrightarrow\widehat{\theta}_\ell}
  =\frac{D[\ell]}{N[\ell]}.
\end{align*}
Note that $\widehat{\theta}_\ell=\frac{D[\ell]}{N[\ell]}$ is guaranteed to be between 0 and 1. To verify that $\widehat{\theta}_\ell$ is indeed the maximum and not the minimum, we check that $[\frac{\textrm{d}\mathcal{L}_{(\ell)}(\theta_\ell)}{\textrm{d}\theta_\ell^2}]_{\theta_\ell=\widehat{\theta}_\ell}<0$. In fact,
\[
\frac{\textrm{d}\mathcal{L}_{(\ell)}(\theta_\ell)}{\textrm{d}\theta_\ell^{2}}=-\sum_{i=1}^{n}\ind\{\ell\le\kappa(Y_i)\}\Big[\frac{\binaryeventmatrix_{i,\ell}}{\theta_\ell^{2}}+\frac{1-\binaryeventmatrix_{i,\ell}}{(1-\theta_\ell)^{2}}\Big]<0
\]
for all $\theta_\ell\in(0,1)$. This finishes the proof.$\hfill\square$

\end{subappendices}

\section{Deep Proportional Hazards Models}
\label{chap:proportional-hazards}

In this section, we cover perhaps the most widely used family of time-to-event prediction models used in practice, called proportional hazards models. Our exposition goes over a fairly general formulation that includes, as special cases, the original Cox proportional hazards model \citep{cox1972regression} as well as its deep learning variant called DeepSurv \citep{faraggi1995neural,katzman2018deepsurv}.

When determining what time-to-event prediction model to use in real applications, Cox models (\eg, the original version or DeepSurv) are good baselines to try, similar to how logistic regression is good to try for binary classification and linear regression is good to try for predicting a continuous outcome. However, much like how logistic regression and linear regression make strong assumptions, Cox models do as well, which is why they often do not achieve state-of-the-art prediction accuracy.

In general, proportional hazards models assume that the hazard function factorizes as
\begin{equation}
h(t|x)=\mathbf{h}_0(t;\theta) e^{\mathbf{f}(x;\theta)}
\quad\text{for }t\ge0,x\in\mathcal{X},
\label{eq:prop-hazard-assumption}
\end{equation}
where we have two functions to be learned: the so-called \emph{baseline hazard function} $\mathbf{h}_0(\cdot;\theta):[0,\infty)\rightarrow[0,\infty)$ and the \emph{log partial hazard function} $\mathbf{f}(\cdot;\theta):\mathcal{X}\rightarrow\mathbb{R}$, both of which have parameter variable $\theta$. The factorization implies that regardless of what the input $x$ is, the hazard function $h(\cdot|x)$ must be proportional to the baseline hazard $\mathbf{h}_0(\cdot;\theta)$, which is why \eqref{eq:prop-hazard-assumption} is called the \emph{proportional hazards assumption}. As we shall see in Section~\ref{sec:prop-hazards-implications}, the proportional hazards assumption imposes a strict constraint on the shapes of survival functions $S(\cdot|x)$. \emph{The proportional hazard assumption often does not hold in real data, which is why proportional hazards models often do not work as well as more flexible time-to-event prediction models.}

The log partial hazard function $\mathbf{f}(\cdot;\theta)$ maps each input $x\in\mathcal{X}$ to a single real number that could be thought of as a risk score. A higher value of $\mathbf{f}(x;\theta)$ implies that the survival time tends to be lower. Since all inputs share the same dependence on time that is captured by the baseline hazard $\mathbf{h}_0(\cdot;\theta)$, under the proportional hazard assumption, the only difference between inputs $x,x'\in\mathcal{X}$ is captured entirely in comparing their log partial hazard function values $\mathbf{f}(x;\theta)$ and $\mathbf{f}(x';\theta)$. Thus, proportional hazards models could fundamentally be viewed as providing a way to rank data points based on the ``scoring'' function $\mathbf{f}(\cdot;\theta)$. We refer to $\mathbf{f}(\cdot;\theta)$ as the log partial hazard function because
\[
\log h(t|x) = \log \mathbf{h}_0(t;\theta) + \mathbf{f}(x;\theta),
\]
so $\mathbf{f}(x;\theta)$ only captures part of the full log hazard.

As a reminder, we saw two examples of proportional hazards models in Section~\ref{chap:setup}, where we had assumed that $\mathcal{X}=\mathbb{R}^d$:
\begin{itemize}

\item In the exponential time-to-event prediction model (Examples~\ref{ex:parametric-hazard-exp-survival} and~\ref{ex:parametric-hazard-exp-survival-maximum-likelihood}), we had $\theta=(\beta,\psi)\in\mathbb{R}^d\times\mathbb{R}$, $\mathbf{h}_0(t;\theta)=e^\psi$, and $\mathbf{f}(x;\theta)=\beta^\top x$.

\item In the Weibull time-to-event prediction model (Example~\ref{ex:parametric-hazard-weibull-survival}), we instead had $\theta=(\beta,\psi,\phi)\in\mathbb{R}^d\times\mathbb{R}\times\mathbb{R}$, $\mathbf{h}_0(t;\theta) = t^{e^\phi - 1} e^{\psi + \phi}$, and $\mathbf{f}(x;\theta) = e^\phi \beta^\top x$.

\end{itemize}
These models are considered parametric proportional hazards models because the baseline hazard $\mathbf{h}_0(\cdot;\theta)$ and the log partial hazard $\mathbf{f}(\cdot;\theta)$ are assumed to have parametric forms. We had already shown in Section~\ref{chap:setup} how these two models could be learned via maximum likelihood (\ie, how to obtain an estimate $\widehat{\theta}$ of $\theta$) and how to subsequently predict hazard, cumulative hazard, and survival functions for these models. Basically, after learning the model parameters, we could estimate the baseline hazard with $\mathbf{h}_0(\cdot;\widehat{\theta}\hspace{1.5pt})$ and the log partial hazard with $\mathbf{f}(\cdot;\widehat{\theta}\hspace{1.5pt})$.

The rest of this section is organized as follows:
\begin{itemize}

\item (Section~\ref{sec:prop-hazards-implications}) First, assuming that the $\mathbf{h}_0(\cdot;\theta)$ and $\mathbf{f}(\cdot;\theta)$ are known (or alternatively that we have estimates for these), we show how the proportional hazards assumption restricts the shapes of $S(\cdot|x)$ that are possible across different $x\in\mathcal{X}$.

\item (Section~\ref{sec:cox-parametric}) Next, we go over the general procedure for learning $\mathbf{h}_0(\cdot;\theta)$ and $\mathbf{f}(\cdot;\theta)$ for parametric proportional hazards models as well as how to make predictions after model training. Our coverage here generalizes what we saw in Section~\ref{chap:setup} for the exponential and Weibull time-to-event prediction models.

\item (Section~\ref{sec:cox-semiparametric}) We then discuss \emph{semiparametric} proportional hazards models, where $\mathbf{h}_0(\cdot;\theta)$ is learned nonparametrically while $\mathbf{f}(\cdot;\theta)$ is learned parametrically. Note that semiparametric proportional hazards models are commonly referred to as \emph{Cox proportional hazards models} (which we just abbreviate as \emph{Cox models}). The original Cox model \citep{cox1972regression} leaves $\mathbf{h}_0(\cdot;\theta)$ unspecified and assumes that $\mathbf{f}(\cdot;\theta) = \theta^\top x$, where $\theta\in\mathbb{R}^d$ and $x\in\mathcal{X}\subseteq\mathbb{R}^d$. The deep learning version (DeepSurv) \citep{faraggi1995neural,katzman2018deepsurv} also leaves $\mathbf{h}_0(\cdot;\theta)$ unspecified and replaces $\mathbf{f}(\cdot;\theta)$ with a neural network.

\item (Section~\ref{sec:cox-time})
Finally, we discuss an extension of the Cox model called Cox-Time \citep{kvamme2019time} that removes the proportional hazards assumption.

\end{itemize}
A major selling point of the original Cox model \citep{cox1972regression} (where $\mathbf{f}(\cdot;\theta) = \theta^\top x$ with both $\theta$ and $x$ belonging to $\mathbb{R}^d$) is that it is straightforward to interpret if the $d$ input features themselves are interpretable (as is commonly the case for tabular data). In this setting, parameter vector $\theta=(\theta_1,\theta_2,\dots,\theta_d)$ simply says how to weight each feature: the $\ell$-th feature has weight $\theta_\ell$. In fact, a standard quantity used for model interpretation is called the \emph{hazard ratio}, which is defined for the $\ell$-th feature to be $\exp(\theta_\ell)$. The basic idea is as follows. Consider a feature vector $x\in\mathbb{R}^d$. Now consider a second feature vector $\widetilde{x}\in\mathbb{R}^d$ that is the same as $x$ except that the $\ell$-th entry is larger by 1, \ie, $\widetilde{x}_j=x_j$ for every index $j\ne\ell$ whereas $\widetilde{x}_\ell = x_\ell + 1$. In this case, it turns out that the second feature vector $\widetilde{x}$ has a hazard value that is a multiplicative factor of $\exp(\theta_\ell)$ larger than that of $x$. To see this, note that
\begin{align*}
\frac{h(t|\widetilde{x})}{h(t|x)}
&= \frac{\cancel{\mathbf{h}_0(t;\theta)} \exp(\theta^\top \widetilde{x})}
       {\cancel{\mathbf{h}_0(t;\theta)} \exp(\theta^\top x)} \\
&= \frac{\exp\big(\theta_\ell (x_\ell + 1) + \sum_{j\ne\ell} \theta_j x_j\big)}
        {\exp\big(\sum_{j=1}^d \theta_j x_j\big)} \\
&= \frac{\exp\big(\theta_\ell + \sum_{j=1}^d \theta_j x_j\big)}
        {\exp\big(\sum_{j=1}^d \theta_j x_j\big)} \\
&= \exp(\theta_\ell).
\end{align*}
To recap, under the proportional hazard assumption for the standard Cox model, an increase in the $\ell$-th feature by 1 unit (with all other features being held the same) is associated with an increase in risk by a multiplicative factor given by the hazard ratio $\exp(\theta_\ell)$.

By replacing $\mathbf{f}(x;\theta) = \theta^\top x$ (a linear function of $x$) with a potentially highly nonlinear function, the DeepSurv model is more flexible than the original Cox model, but interpreting how the DeepSurv model makes predictions is less straightforward. In more detail, consider again the setting where raw inputs are feature vectors in $\mathbb{R}^d$. Reusing the earlier notation where $x\in\mathbb{R}^d$ and $\widetilde{x}\in\mathbb{R}^d$ differ only in the $\ell$-th feature, with $\widetilde{x}_\ell = x_\ell + 1$, the hazard ratio would be
\[
\frac{h(t|\widetilde{x})}{h(t|x)}
= \frac{\cancel{\mathbf{h}_0(t;\theta)} \exp(\mathbf{f}(\widetilde{x};\theta))}
       {\cancel{\mathbf{h}_0(t;\theta)} \exp(\mathbf{f}(x;\theta))}.
\]
When $\mathbf{f}$ is highly nonlinear, then this ratio does not, in general, simplify ``nicely'' and could depend on many parameters, unlike in the linear setting where the hazard ratio ends up depending only on a single parameter $\theta_\ell$. The Cox-Time model is even more flexible than the DeepSurv model and can be even less straightforward to interpret.

\subsection{Constraint on Survival Function Shapes}
\label{sec:prop-hazards-implications}

\begin{figure}[!t]
\begin{centering}
\subfloat[]{%
\includegraphics[scale=0.62]{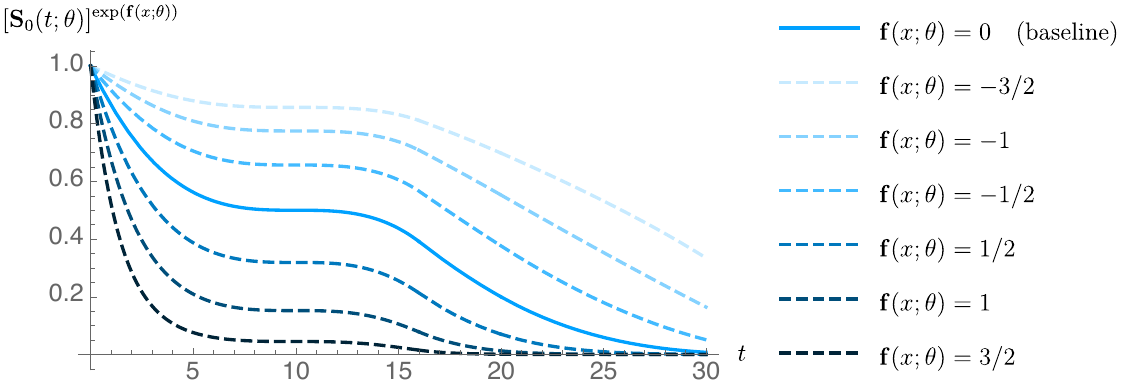}

}\\
\subfloat[]{\includegraphics[scale=0.62]{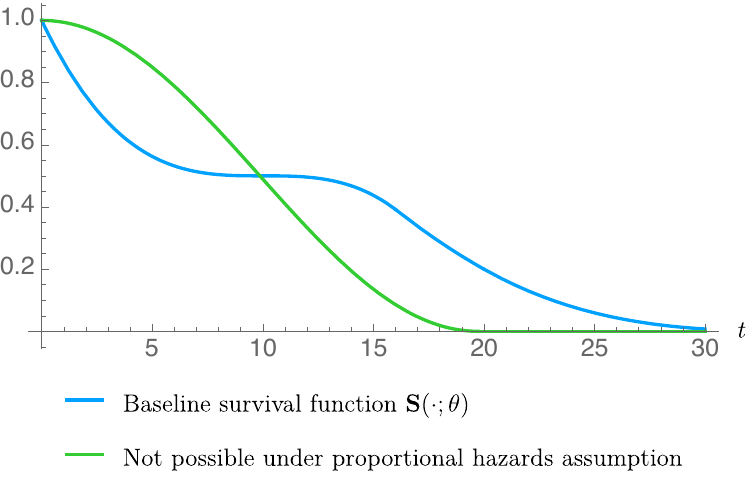}

}
\par\end{centering}
\caption{Under the proportional hazards assumption (\eqref{eq:prop-hazard-assumption}), possible survival functions are all powers of the baseline survival function $\mathbf{S}_0(\cdot;\theta)$ as shown in panel (a); note that we can always unambiguously order these functions based on the log partial hazard function $\mathbf{f}(\cdot;\theta)$. In contrast, the green curve shown in panel (b) is not possible under a proportional hazards model and is neither uniformly better nor uniformly worse than the baseline survival function.}
\label{fig:survival-curve-powers}
\end{figure}

In this section, we assume that we already know $\theta$ (or have an estimate for it), which means that we also know $\mathbf{h}_0(\cdot;\theta)$ and $\mathbf{f}(\cdot;\theta)$. Let's look at what the proportional hazards assumption implies about the shapes of survival functions that are possible. Recall from Summary~\ref{sum:conversions} that by knowing the hazard function $h(t|x)$, we can recover $H(t|x)=\int_{0}^{t}h(u|x)\textrm{d}u$ and $S(t|x)=e^{-H(t|x)}$. Thus, under the proportional hazards assumption (\eqref{eq:prop-hazard-assumption}), the cumulative hazard function is
\begin{equation}
H(t|x)
=\int_0^t h(u|x) \textrm{d}u
=\int_0^t \mathbf{h}_0(u;\theta)e^{\mathbf{f}(x;\theta)} \textrm{d}u
=e^{\mathbf{f}(x;\theta)}
 \underbrace{
   \int_0^t \mathbf{h}_0(u;\theta) \textrm{d}u
 }_{=:\mathbf{H}_0(t;\theta)},
\label{eq:cox-cumulative-hazard}
\end{equation}
where the newly defined $\mathbf{H}_0(\cdot;\theta)$ is referred to as the \emph{baseline cumulative hazard function}. Then we recover the survival function
\begin{equation}
S(t|x)
=e^{-H(t|x)}
=e^{-e^{\mathbf{f}(x;\theta)} \mathbf{H}_0(t;\theta)}
=[\!~\underbrace{e^{-\mathbf{H}_0(t;\theta)}}_{=:\mathbf{S}_0(t;\theta)}\!~]^{e^{\mathbf{f}(x;\theta)}},
\label{eq:cox-survival}
\end{equation}
where we have the newly defined \emph{baseline survival function} $\mathbf{S}_0(\cdot;\theta)$. \Eqref{eq:cox-survival} tells us that all possible survival functions under the proportional hazards assumption must be powers of $\mathbf{S}_0(\cdot;\theta)$---see Figure~\ref{fig:survival-curve-powers}(a) for an illustration. This is a strong assumption! \emph{Survival functions that are not powers of $\mathbf{S}_0(\cdot;\theta)$ are impossible under the proportional hazards assumption (such as the green curve that crisscrosses the baseline survival function in Figure~\ref{fig:survival-curve-powers}(b).}

As shown in Figure~\ref{fig:survival-curve-powers}(a), the allowed survival functions that are closer to the origin---which have higher $\mathbf{f}(x;\theta)$ value---are uniformly worse than ones farther away from the origin, regardless of what time~$t$ we look at (as a reminder from Section~\ref{sec:estimands-continuous}, the area under a survival function is the mean survival time and the time at which the survival function crosses the y-axis value of 1/2 is the median survival time). In particular, under a proportional hazards assumption, whether a data point with  $x$ is likely to have a shorter or longer survival time is entirely determined by the log partial hazard function value $\mathbf{f}(x;\theta)$ (which we previously pointed out could be interpreted as a risk score): higher values of $\mathbf{f}(x;\theta)$ correspond to shorter survival times.

\subsection{Parametric Proportional Hazards Models}
\label{sec:cox-parametric}

In Section~\ref{sec:likelihood}, we had already shown how exponential and Weibull time-to-event prediction models could be learned via maximum likelihood and also how to subsequently make predictions. We now generalize both of these cases. Consider when $\mathbf{h}_0(\cdot;\theta)$ and $\mathbf{f}_0(\cdot;\theta)$ are parametric functions differentiable with respect to~$\theta$ (where we assume that $\mathbf{h}_0(\cdot;\theta)$ is specified in continuous time just like in the exponential and Weibull examples), and we have some closed-form expression for
\[
\mathbf{H}_0(t;\theta)
:= \int_0^t \mathbf{h}_0(u;\theta)\textrm{d}u,
\]
which is also differentiable with respect to~$\theta$.\footnote{For the exponential time-to-event prediction model, recall that $\theta=(\beta,\psi)\in\mathbb{R}^{d}\times\mathbb{R}$ and $\mathbf{h}_{0}(t;\theta)=e^{\psi}$, so that 
\[
\mathbf{H}_{0}(t;\theta)=\int_{0}^{t}e^{\psi}\textrm{d}u=te^{\psi}.
\]
For the Weibull time-to-event prediction model, recall that $\theta=(\beta,\psi,\phi)\in\mathbb{R}^{d}\times\mathbb{R}\times\mathbb{R}$ and $\mathbf{h}_{0}(t;\theta)=t^{e^{\phi}-1}e^{\psi+\phi}$, so that
\[
\mathbf{H}_{0}(t;\theta)=\int_{0}^{t}u^{e^{\phi}-1}e^{\psi+\phi}\textrm{d}u=t^{e^{\phi}}e^{\psi}.
\]} Then learning $\theta$ and making predictions works similarly to what is described in Examples~\ref{ex:parametric-hazard-exp-survival-maximum-likelihood} and~\ref{ex:parametric-hazard-weibull-survival}. We now state the general procedure.

\paragraph{Training}
Model training amounts to learning the parameters $\theta$, which we do by writing the loss function (that depends on training data $(X_1,Y_1,\Delta_1),\dots,(X_n,Y_n,\Delta_n)$) and then optimizing:
\begin{enumerate}

\item We write the negative log likelihood loss (\eqref{eq:nll-loss-hazard-form}) but replace $\mathbf{h}(t|x;\theta)$ with $\mathbf{h}_0(t;\theta) e^{\mathbf{f}(x;\theta)}$:
\begin{align*}
&\textbf{L}_{\text{Hazard-NLL}}(\theta) \\
&=
 -
 \frac{1}{n}
 \sum_{i=1}^n
    \bigg\{
      \Delta_i \log \mathbf{h}(Y_i|X_i;\theta)
      -\int_0^{Y_i} \mathbf{h}(u|X_i;\theta)\textrm{d}u
    \bigg\} \\
&=
 -
 \frac{1}{n}
 \sum_{i=1}^n
    \bigg\{
      \Delta_i \log(\mathbf{h}_0(Y_i;\theta) e^{\mathbf{f}(X_i;\theta)})
      -\int_0^{Y_i} \mathbf{h}_0(u;\theta) e^{\mathbf{f}(X_i;\theta)} \textrm{d}u
    \bigg\} \\
&=
 -
 \frac{1}{n}
 \sum_{i=1}^n
    \bigg\{
      \Delta_i [\mathbf{f}(X_i;\theta) + \log\mathbf{h}_0(Y_i;\theta)]
      -e^{\mathbf{f}(X_i;\theta)} \int_0^{Y_i} \mathbf{h}_0(u;\theta) \textrm{d}u
    \bigg\} \\
&=
 -
 \frac{1}{n}
 \sum_{i=1}^n
    \big\{
      \Delta_i [\mathbf{f}(X_i;\theta) + \log\mathbf{h}_0(Y_i;\theta)]
      -e^{\mathbf{f}(X_i;\theta)} \mathbf{H}_0(Y_i; \theta)
    \big\}.
\end{align*}

\item We then use a standard neural network optimizer to solve
\[
\widehat{\theta}:=\arg\min_\theta \textbf{L}_{\text{Hazard-NLL}}(\theta).
\]
\end{enumerate}

\paragraph{Prediction}
To predict the hazard function, we simply use \eqref{eq:prop-hazard-assumption}, where we plug in $\widehat{\theta}$ in place of $\theta$:
\begin{equation*}
\widehat{h}(t|x)
:= \mathbf{h}_0(t; \widehat{\theta}\hspace{1.5pt}) e^{\mathbf{f}(x;\widehat{\theta}\hspace{1.5pt})}
\quad\text{for }t\ge0,x\in\mathcal{X}.
\end{equation*}
Predicting the cumulative hazard and survival functions then just amount to using the conversions from Summary~\ref{sum:conversions}. Specifically, we could estimate the cumulative hazard function using
\begin{equation*}
\widehat{H}(t|x)
:= \int_0^t \widehat{h}(u|x)\textrm{d}u
 = e^{\mathbf{f}(x;\widehat{\theta}\hspace{1.5pt})} \int_0^t \mathbf{h}_0(u; \widehat{\theta}\hspace{1.5pt})\textrm{d}u
 = e^{\mathbf{f}(x;\widehat{\theta}\hspace{1.5pt})} \mathbf{H}_0(t; \widehat{\theta}\hspace{1.5pt}),
\end{equation*}
and the survival function using
\begin{equation*}
\widehat{S}(t|x)
:= \exp(-\widehat{H}(t|x))
 = \exp\big(-e^{\mathbf{f}(x;\widehat{\theta}\hspace{1.5pt})} \mathbf{H}_0(t; \widehat{\theta}\hspace{1.5pt})\big).
\end{equation*}

\subsection{Semi-Parametric Proportional Hazards Models: DeepSurv}
\label{sec:cox-semiparametric}

We now turn to the more complicated case where $\mathbf{h}_0(\cdot;\theta)$ is left unspecified (but we shall see that it still depends on parameter variable~$\theta$) while $\mathbf{f}(\cdot;\theta)$ remains parametric, resulting in a model commonly referred to as DeepSurv \citep{faraggi1995neural,katzman2018deepsurv}. We state the standard procedure for learning $\mathbf{f}(\cdot;\theta)$ and $\mathbf{h}_0(\cdot;\theta)$ followed by how the hazard, cumulative hazard, and survival functions are predicted. The training procedure does maximum likelihood estimation (for a derivation, see Section~\ref{sec:cox-mle-derivation}).

\paragraph{Training} Model training consists of two steps:
\begin{enumerate}

\item We estimate $\theta$ by numerically solving the optimization problem
\begin{equation}
\widehat{\theta}
:=\arg\min_\theta
     \frac{1}{n}
     \sum_{i=1}^n
       \Delta_i
       \bigg[
         -\mathbf{f}(X_i;\theta)
         +
         \log\bigg(
           \sum_{j=1}^n
             \ind\{Y_j\ge Y_i\}
             e^{\mathbf{f}(X_j;\theta)}
         \bigg)
       \bigg]
\label{eq:cox-step1}
\end{equation}
using a standard neural network optimizer.\footnote{In the classical setting where $\mathcal{X}\subseteq\mathbb{R}^d$ and $\mathbf{f}(x;\theta) = \theta^\top x$ with $\theta\in\mathbb{R}^d$, this problem is convex and can be solved using, for instance, Newton-Raphson. Moreover, in the classical setting, researchers have explored adding lasso \citep{tibshirani1997lasso} or elastic net regularization \citep{zou2005regularization} on $\theta$ to encourage the estimated $\theta$ to, for instance, be sparse. In the neural network setting that we consider, standard tricks can be used for regularization, such as using dropout or weight decay.}

\item Next, we estimate $\mathbf{h}_0(\cdot;\theta)$ using the method by \citet{breslow1972discussion}: consider the discrete time grid given by the unique times of death $\tau_{(1)}<\tau_{(2)}<\cdots<\tau_{(L)}$. We define $\tau_{(0)}:=0$. Breslow assumes that $\mathbf{h}_0(\cdot;\theta)$ (defined in continuous time) is piecewise constant, where the possible changes in $\mathbf{h}_0(\cdot;\theta)$ happen at the discrete time grid points. Specifically, the Breslow estimator of $\mathbf{h}_0(\cdot;\theta)$ is
\begin{equation}
\widehat{\mathbf{h}}_0(t)
:=\begin{cases}
    \frac{D[\ell]}
         {(\tau_{(\ell)}-\tau_{(\ell-1)})
          \sum_{j=1}^n
            \ind\{Y_j\ge\tau_{(\ell)}\} e^{\mathbf{f}(X_j;\widehat{\theta}\hspace{1.5pt})}}
    & \begin{array}{r}
      \text{if }\tau_{(\ell-1)}<t\le\tau_{(\ell)} \\
      \text{ for }\ell\in[L],
      \end{array} \\
  0 & \hspace{5pt}\text{if }t > \tau_{(L)}.
  \end{cases}
\label{eq:breslow-hazard}
\end{equation}
Note that $\widehat{\mathbf{h}}_0(\cdot)$ depends on $\widehat{\theta}$.
\end{enumerate}

\paragraph{Prediction}
To predict the hazard function $h(t|x) =  \mathbf{h}_0(t;\theta) e^{\mathbf{f}(x;\theta)}$, we simply plug in the estimated $\widehat{\theta}$ and $\widehat{h}_0(\cdot)$:
\[
\widehat{h}(t|x)
:= \widehat{\mathbf{h}}_0(t) e^{\mathbf{f}(x;\widehat{\theta}\hspace{1.5pt})}
\quad\text{for }t\ge0,x\in\mathcal{X}.
\]
To predict the cumulative hazard function, we use \eqref{eq:cox-cumulative-hazard}: $H(t|x) = e^{\mathbf{f}(x;\theta)} \mathbf{H}_0(t;\theta)$. In particular, we first estimate the baseline cumulative hazard with
\begin{equation}
\widehat{\mathbf{H}}_0(t)
:=\int_0^t \widehat{\mathbf{h}}_0(u) \textrm{d}u
 =\sum_{m=1}^L
    \frac{\ind\{\tau_{(m)}\le t\}D[m]}
         {\sum_{j=1}^n \ind\{Y_j\ge\tau_{(m)}\} e^{\mathbf{f}(X_j;\widehat{\theta}\hspace{1.5pt})}}
 \quad\text{for }t\ge0.
\label{eq:breslow}
\end{equation}
Then we predict $H(\cdot|x)$ using the estimator
\begin{equation}
\widehat{\mathbf{H}}(t|x)
:=e^{\mathbf{f}(x;\widehat{\theta}\hspace{1.5pt})} \widehat{\mathbf{H}}_0(t)
\quad\text{for }t\ge0,x\in\mathcal{X}.
\label{eq:cox-semiparametric-cumulative-hazard}
\end{equation}
Finally, to predict $S(\cdot|x)$ using the estimator
\begin{equation}
\widehat{S}(t|x)
:= e^{-H(t|x)}
 = \exp\big(-e^{\mathbf{f}(x;\widehat{\theta}\hspace{1.5pt})} \widehat{\mathbf{H}}_0(t)\big)
\quad\text{for }t\ge0,x\in\mathcal{X}.
\label{eq:cox-semiparametric-survival}
\end{equation}
We provide a Jupyter notebook that shows how to implement \mbox{DeepSurv}.\footnote{\texttt{\url{https://github.com/georgehc/survival-intro/blob/main/S3.3_DeepSurv.ipynb}}}
\begin{fremark}[Relationship to ranking]
The loss function in~\eqref{eq:cox-step1} is equal to
\[
\frac{1}{n}\sum_{i=1}^{n}\Delta_{i}\log\bigg(\sum_{j=1}^{n}\ind\{Y_{j}\ge Y_{i}\}e^{\mathbf{f}(X_{j};\theta)-\mathbf{f}(X_{i};\theta)}\bigg).
\]
By carefully staring at this, notice the following. Minimizing this loss means that when $\Delta_i=1$ and also $Y_i\le Y_j$ (\ie, the $i$-th point died and has an observed time before that of the $j$-th point), we would like risk score $\mathbf{f}(X_j;\theta)$ to be lower than risk score $\mathbf{f}(X_i;\theta)$. This directly corresponds to the intuition for the concordance index metric (Definition~\ref{def:c-index}). In this sense, the Cox model's loss function could be interpreted as a ranking-based loss. For a more detailed explanation, see the paper by \citet{raykar2007ranking}.
\end{fremark}
\begin{fremark}[The Nelson-Aalen estimator as a special case of the Cox model]
\label{rem:nelson-aalen-special-case-of-cox}
\citet{breslow1972discussion} pointed out that in the special case where $\mathbf{f}(x;\theta)=0$ for all $x\in\mathcal{X}$, then $\widehat{\mathbf{H}}_0(\cdot)$ from \eqref{eq:breslow} actually just becomes the Nelson-Aalen estimator (\eqref{eq:nelson-aalen}). In this case, the cumulative hazard estimate (\eqref{eq:cox-semiparametric-cumulative-hazard}) would also just become the Nelson-Aalen estimator. However, the survival function estimator (\eqref{eq:cox-semiparametric-survival}) would not be exactly equal to (but would approximate) the Kaplan-Meier estimator due to the result of Proposition~\ref{prop:discrete-cumulative-hazard-not-neg-log-surv}.
\end{fremark}
\begin{fremark}[Piecewise constant functions in continuous time]\label{rem:piecewise-constant}
Note that $\widehat{h}_0(\cdot)$ in \eqref{eq:breslow-hazard} could be thought of as a discrete time object since it only has $L$ nonzero values. Thus, it could be stored on a computer in a 1D array/table. However, we want to emphasize that $\mathbf{h}_0(\cdot)$ really is a continuous time hazard function and \emph{not} a forward-filled version of a discrete time hazard function like the ones we see in Section~\ref{sec:setup-discrete}.

As a reminder, in Section~\ref{sec:setup-discrete}, when we were working directly in discrete time, we required that the discrete time hazard function $h[\cdot|x]$ to have values that are at most~1 since they are probabilities. We used this to, for instance, derive that $S[\ell|x]=\prod_{m=1}^\ell (1 - h[m|x])$.

In contrast, one can easily check that the Breslow estimator $\widehat{\mathbf{h}}_0(\cdot)$ could take on values that are arbitrarily large (possibly larger than 1) since $\mathbf{f}(\cdot; \theta)$ is unconstrained (so that the nonnegative term $e^{\mathbf{f}(X_j;\widehat{\theta}\hspace{1.5pt})}$ could be arbitrarily large or small).

Later on in Section~\ref{sec:discrete-to-continuous}, we will show that we can convert any discrete time model like the ones in Section~\ref{sec:setup-discrete} to continuous time. However, what we are saying here is that for a hazard function that is defined originally in continuous time, even if it is piecewise constant, we cannot in general convert it to be a discrete time hazard function of the form we saw in Section~\ref{sec:setup-discrete}.
\end{fremark}

\subsection{Removing the Proportional Hazards Assumption: Cox-Time}
\label{sec:cox-time}

A generalization of the Cox model called \mbox{Cox-Time} \citep{kvamme2019time} replaces $\mathbf{f}(x;\theta)$ with the neural network $\mathbf{g}(x,t;\theta)$ that depends on both input $x\in\mathcal{X}$ and time $t\ge0$. As a result, we have the factorization
\begin{equation}
h(t|x) = \mathbf{h}_0(t;\theta)e^{\mathbf{g}(x,t;\theta)}
\quad\text{for }t\ge0,x\in\mathcal{X}.
\label{eq:coxtime-factorization}
\end{equation}
The earlier proportional hazards factorization from \eqref{eq:prop-hazard-assumption} has time $t$ and input $x$ contribute to different factors. Now, in \eqref{eq:coxtime-factorization}, time $t$ and input~$x$ could interact within the function~$\mathbf{g}(x,t;\theta)$. Thus, whereas previously, the proportional hazards factorization constrained the survival function shapes to be powers of a baseline survival function (as discussed in Section~\ref{sec:prop-hazards-implications}), \emph{this constraint is no longer guaranteed to hold in general for Cox-Time}.\footnote{We point out that the idea of replacing $\mathbf{f}(x;\theta)$ with a function that depends on both raw input $x$ and time $t$ is an innovation that predates the paper by \citet{kvamme2019time}. For example, Chapter~9 of the textbook by \citet{klein2003survival} suggests setting $\mathbf{f}$ to still have a linear form but where there are some newly added features that capture interactions between some of the original features and a pre-specified function of time (\eg, see equation~(9.2.2) of \citet{klein2003survival}). It is possible to specify such an $\mathbf{f}$ that still enables statistical inference (see Example 9.2 of \citet{klein2003survival}). Cox-Time was not designed with statistical inference in mind and aims to simply replace $\mathbf{f}$ with an arbitrarily complex neural network that depends on both $x$ and $t$, with the hope that after model training, the neural network will encode interactions between~$x$ and~$t$ that are relevant for prediction.}

One might wonder why there needs to even be a baseline hazard function $\mathbf{h}_0(t;\theta)$ if $\mathbf{g}(x,t;\theta)$ already depends on time~$t$. This has to do with how the Cox-Time model is trained. \citet{kvamme2019time} proposed simply using the DeepSurv training and prediction procedures with minor modifications to train the Cox-Time model.

\paragraph{Training} Model training consists of two steps:
\begin{enumerate}

\item We define the loss
\begin{align}
&\mathbf{L}_{\text{Cox-Time}}(\theta) \nonumber\\
&:=
  \frac{1}{n}
  \sum_{i=1}^n
    \Delta_i
    \bigg[
      -\mathbf{g}(X_i,Y_i;\theta)
      +
      \log\bigg(
        \sum_{j=1}^n
          \ind\{Y_j\ge Y_i\}
          e^{\mathbf{g}(X_j,Y_i;\theta)}
      \bigg)
    \bigg]. \label{eq:cox-time-step1}
\end{align}
Using a neural network optimizer, we compute $\widehat{\theta} := \arg\min_\theta \mathbf{L}_{\text{Cox-Time}}(\theta)$. The summation inside the log intentionally uses time $Y_i$ and not $Y_j$ (notice the expression ``$\mathbf{g}(X_j,Y_i;\theta)$'').

\item Denote the unique times of death by $\tau_{(1)}<\tau_{(2)}<\cdots<\tau_{(L)}$, and define $\tau_{(0)}:=0$. We compute:
\begin{align*}
&\widehat{\mathbf{h}}_0(t) \nonumber\\
&:=
  \begin{cases}
    \frac{D[\ell]}
         {(\tau_{(\ell)}-\tau_{(\ell-1)})
          \sum_{j=1}^n
            \ind\{Y_j\ge\tau_{(\ell)}\} e^{\mathbf{g}(X_j,\tau_{(\ell)};\widehat{\theta}\hspace{1.5pt})}}
    & \begin{array}{r}
        \text{if }\tau_{(\ell-1)}<t\le\tau_{(\ell)} \\
        \text{ for }\ell\in[L],
        \end{array} \\
    0 & \hspace{5pt}\text{if }t > \tau_{(L)}.
  \end{cases}
\end{align*}
\end{enumerate}
The reason why we still need to estimate the baseline hazard function $\mathbf{h}_0(\cdot;\theta)$ is as follows \citep[Section~3.4]{kvamme2019time}: if $\mathbf{g}(x,t;\theta) = \mathbf{a}(x,t;\theta) + \mathbf{b}(t;\theta)$ for some functions $\mathbf{a}(\cdot,\cdot;\theta)$ and $\mathbf{b}(\cdot;\theta)$, then the function $\mathbf{b}(\cdot;\theta)$ would actually be cancelled out in the loss $\mathbf{L}_{\text{Cox-Time}}(\theta)$. Thus, minimizing $\mathbf{L}_{\text{Cox-Time}}(\theta)$ would not be able to learn $\mathbf{b}(\cdot;\theta)$. Instead, we learn $\mathbf{b}(\cdot;\theta)$ as part of the baseline hazard function.

\paragraph{Prediction}
We predict the hazard function $h(\cdot|x)$ using
\[
\widehat{h}(t|x)
:= \widehat{\mathbf{h}}_0(t) e^{\mathbf{g}(x,t;\widehat{\theta}\hspace{1.5pt})}
\quad\text{for }t\ge0,x\in\mathcal{X}.
\]
To predict the cumulative hazard function, we first compute
\[
\widehat{\mathbf{H}}_0(t)
:=\int_0^t \widehat{\mathbf{h}}_0(u) \textrm{d}u
 =\sum_{m=1}^L
    \frac{\ind\{\tau_{(m)}\le t\}D[m]}
         {\sum_{j=1}^n \ind\{Y_j\ge\tau_{(m)}\} e^{\mathbf{g}(X_i,\tau_{(m)};\widehat{\theta}\hspace{1.5pt})}}
 \quad\text{for }t\ge0.
\]
Then we predict $H(\cdot|x)$ using
\[
\widehat{H}(t|x)
:= e^{\mathbf{g}(x,t;\widehat{\theta}\hspace{1.5pt})} \widehat{\mathbf{H}}_0(t)
\quad\text{for }t\ge0,x\in\mathcal{X}.
\]
We predict the survival function $S(\cdot|x)$ using
\[
\widehat{S}(t|x)
:= e^{-H(t|x)}
 = \exp\big(-e^{\mathbf{g}(x,t;\widehat{\theta}\hspace{1.5pt})} \widehat{\mathbf{H}}_0(t)\big)
\quad\text{for }t\ge0,x\in\mathcal{X}.
\]

\noindent
Overall, the training and prediction procedures for Cox-Time are heuristically justified. It is unclear whether there is a more theoretically sound manner for jointly estimating $\mathbf{h}_0(\cdot;\theta)$ and $\mathbf{g}(\cdot,\cdot;\theta)$, where $\mathbf{h}_0(\cdot;\theta)$ is left unspecified and $\mathbf{g}(\cdot,\cdot;\theta)$ is within, say, some reasonably wide family of neural networks.

In our companion code repository, we provide a Jupyter notebook that implements Cox-Time.\footnote{\texttt{\url{https://github.com/georgehc/survival-intro/blob/main/S3.4_Cox-Time.ipynb}}} Our notebook uses the original Cox-Time code by \citet{kvamme2019time} for which the loss $\mathbf{L}_{\text{Cox-Time}}(\theta)$ in \eqref{eq:cox-time-step1} is \emph{not} computed exactly. To speed up computation, in \eqref{eq:cox-time-step1}, the summation (over index $j$) inside the log is approximated by randomly sampling one of the nonzero terms of the summation. This is referred to as a ``case-control'' approximation, with the idea that for each training point $i$ (the outer summation of \eqref{eq:cox-time-step1}) that we call the ``case'' data point, we are randomly choosing a single other training point (called the ``control'') to be used inside the log.

\begin{subappendices}
\subsection{Technical Details: Derivation of the Cox Model's Two-Step Maximum Likelihood Estimator}
\label{sec:cox-mle-derivation}

The derivation we present here spells out the steps of the terse derivation given by \citet{breslow1972discussion} that was for the original Cox model (specifically, the first few paragraphs of his discussion outlines how the derivation works). Breslow's derivation trivially can be adapted to the DeepSurv setting, where the log partial hazard function is specified by a neural network and is not simply a linear function.

As a reminder, we discretize time so that $\tau_{(1)}<\tau_{(2)}<\cdots<\tau_{(L)}$ are the unique times of death, and we set $\tau_{(0)}:=0$. For a noncensored data point $i$, let $\kappa(Y_i)\in[L]$ denote the time index corresponding to time $Y_i$. Extremely importantly, as \citet{breslow1972discussion} states in his derivation, he follows Kalbfleisch and Prentice's convention and takes a censored data point's observed time to be the \emph{preceding} observed time of death. In other words, for a censored data point $i$, we take $\kappa(Y_i)\in[L]$ to be the time index of the largest time of death that is before $Y_i$ (if there is no death prior to $Y_i$, then we take $\kappa(Y_i)=0$).

Next, suppose that $\mathbf{h}_0(t;\theta)$ is piecewise constant so that
\[
\mathbf{h}_0(t;\theta):=\begin{cases}
\lambda_{\ell} & \begin{array}{r}
\text{if }\tau_{(\ell-1)}<t\le\tau_{(\ell)}\text{ for }\ell\in[L],\end{array}\\
0 & \hspace{5pt}\text{if }t>\tau_{(L)},
\end{cases}
\]
where $\lambda:=(\lambda_{1},\lambda_{2},\dots,\lambda_{L})\in[0,\infty)^{L}$. Thus, the hazard function is equal to
\[
\mathbf{h}(t|x;\theta):=\mathbf{h}_0(t;\theta)e^{\mathbf{f}(x;\theta)}.
\]
The hazard form of the log likelihood (\eqref{eq:log-likelihood-pdf-hazard-theta}) is (where we emphasize both the dependence on $\theta$ and $\lambda$):
\begingroup
\allowdisplaybreaks
\begin{align}
 & \log\mathcal{L}(\theta,\lambda) \nonumber\\
 & =\sum_{i=1}^{n}\bigg\{\Delta_i\log\mathbf{h}(Y_i|X_i;\theta)-\int_{0}^{Y_i}\mathbf{h}(u|X_i;\theta)\textrm{d}u\bigg\} \nonumber\\
 & =\sum_{i=1}^{n}\bigg\{\Delta_i\log(\mathbf{h}_0(Y_i;\theta)e^{\mathbf{f}(X_i;\theta)})-\int_{0}^{Y_i}\mathbf{h}_0(u;\theta)e^{\mathbf{f}(X_i;\theta)}\textrm{d}u\bigg\} \nonumber\\
 & =\sum_{i=1}^{n}\bigg\{\Delta_i\log(\mathbf{h}_0(Y_i;\theta)e^{\mathbf{f}(X_i;\theta)})-e^{\mathbf{f}(X_i;\theta)}\int_{0}^{Y_i}\mathbf{h}_0(u;\theta)\textrm{d}u\bigg\} \nonumber\\
 & =\sum_{i=1}^{n}\bigg\{\Delta_i\log(\lambda_{\kappa(Y_{i})}e^{\mathbf{f}(X_i;\theta)})-e^{\mathbf{f}(X_i;\theta)}\sum_{m=1}^{\kappa(Y_{i})}(\tau_{(m)}-\tau_{(m-1)})\lambda_{m}\bigg\} \nonumber\\
 & =\sum_{i=1}^{n}\bigg\{\Delta_i\log\lambda_{\kappa(Y_{i})}+\Delta_{i}\mathbf{f}(X_i;\theta)-e^{\mathbf{f}(X_i;\theta)}\sum_{m=1}^{\kappa(Y_{i})}(\tau_{(m)}-\tau_{(m-1)})\lambda_{m}\bigg\} \nonumber\\
 & =\sum_{i=1}^{n}\Delta_i\log\lambda_{\kappa(Y_{i})}+\sum_{i=1}^{n}\Delta_{i}\mathbf{f}(X_i;\theta)-\sum_{i=1}^{n}e^{\mathbf{f}(X_i;\theta)}\sum_{m=1}^{\kappa(Y_{i})}(\tau_{(m)}-\tau_{(m-1)})\lambda_{m} \nonumber\\
 & =\sum_{m=1}^{L}D[m]\log\lambda_{(m)}
    +\sum_{i=1}^{n}\Delta_{i}\mathbf{f}(X_i;\theta) \nonumber\\
 & \phantom{=}
    -\sum_{m=1}^{L}(\tau_{(m)}-\tau_{(m-1)})\lambda_{m}\sum_{j=1}^{n}\ind\{Y_{j}\ge m\}e^{\mathbf{f}(X_{j};\theta)}.
 \label{eq:cox-mle-derivation-helper1}
\end{align}
\endgroup
Setting the derivative with respect to $\lambda_{(\ell)}$ to 0, we get
\[
0=\Big[\frac{\textrm{d}\log\mathcal{L}(\theta)}{\textrm{d}\lambda_{(\ell)}}\Big]_{\lambda_{(\ell)}=\widehat{\lambda}_{(\ell)}}=\frac{D[\ell]}{\widehat{\lambda}_{(\ell)}}-(\tau_{(\ell)}-\tau_{(\ell-1)})\sum_{j=1}^{n}\ind\{Y_{j}\ge\ell\}e^{\mathbf{f}(X_{j};\theta)}.
\]
Rearranging terms, we get
\begin{equation}
\widehat{\lambda}_{(\ell)}=\frac{D[\ell]}{(\tau_{(\ell)}-\tau_{(\ell-1)})\sum_{j=1}^{n}\ind\{Y_{j}\ge\ell\}e^{\mathbf{f}(X_{j};\theta)}},
\label{eq:cox-mle-derivation-helper2}
\end{equation}
which is strictly positive. One can verify that $\big[\frac{\textrm{d}^{2}\log\mathcal{L}(\theta)}{\textrm{d}\lambda_{(\ell)}^{2}}\big]_{\lambda_{(\ell)}=\widehat{\lambda}_{(\ell)}}<0$ so that indeed we are looking at a maximum. Plugging in the optimal choice of $\widehat{\lambda}_{(\ell)}$ back into \eqref{eq:cox-mle-derivation-helper1}, we get
\begin{align*}
 & \log\mathcal{L}(\theta,\widehat{\lambda})\\
 & \quad=\sum_{m=1}^{L}D[m]\log\frac{D[\ell]}{(\tau_{(m)}-\tau_{(m-1)})\sum_{j=1}^{n}\ind\{Y_{j}\ge m\}e^{\mathbf{f}(X_{j};\theta)}}\\
 & \phantom{\quad=}
   +\sum_{i=1}^{n}\Delta_{i}\mathbf{f}(X_i;\theta)-\sum_{m=1}^{L}D[m] \\
 & \quad=\sum_{i=1}^{n}\Delta_{i}\mathbf{f}(X_i;\theta)-\sum_{m=1}^{L}D[m]\log\sum_{j=1}^{n}\ind\{Y_{j}\ge m\}e^{\mathbf{f}(X_{j};\theta)}\\
 & \phantom{\quad=}
   +\underbrace{\sum_{m=1}^{L}D[m]\log\frac{D[\ell]}{(\tau_{(m)}-\tau_{(m-1)})}-\sum_{m=1}^{L}D[m]}_{\text{constant (with respect to }\theta\text{)}} \\
 & \quad=\sum_{i=1}^{n}\Delta_{i}\mathbf{f}(X_i;\theta)-\sum_{i=1}^{n}\Delta_i\log\sum_{j=1}^{n}\ind\{Y_{j}\ge Y_{i}\}e^{\mathbf{f}(X_{j};\theta)}+\text{constant}\\
 & \quad=\sum_{i=1}^{n}\Delta_{i}\bigg[\mathbf{f}(X_i;\theta)-\sum_{i=1}^{n}\log\sum_{j=1}^{n}\ind\{Y_{j}\ge Y_{i}\}e^{\mathbf{f}(X_{j};\theta)}\bigg]+\text{constant}.
\end{align*}
Then
\begin{align*}
\widehat{\theta} & :=\arg\max_{\theta}\log\mathcal{L}(\theta,\widehat{\lambda})\\
 & =\arg\min_{\theta}-\frac{1}{n}\log\mathcal{L}(\theta,\widehat{\lambda})\\
 & =\arg\min_{\theta}-\frac{1}{n}\sum_{i=1}^{n}\Delta_{i}\bigg[\mathbf{f}(X_i;\theta)-\sum_{i=1}^{n}\log\sum_{j=1}^{n}\ind\{Y_{j}\ge Y_{i}\}e^{\mathbf{f}(X_{j};\theta)}\bigg],
\end{align*}
where we have dropped the constant as it does not affect the argument achieving the minimum. This finishes the derivation of the first step of the Cox training procedure (namely, \eqref{eq:cox-step1}). The second step of the Cox training procedure simply plugs in the optimal choice of $\widehat{\theta}$ in place of $\theta$ in \eqref{eq:cox-mle-derivation-helper2}.
\end{subappendices}

\section{Deep Conditional Kaplan-Meier Estimators}
\label{chap:deep-kaplan-meier}

In Section~\ref{chap:setup}, we encountered the Kaplan-Meier estimator \citep{kaplan1958nonparametric}, which estimates the population-level survival function~$S_{\text{pop}}(t):=\mathbb{P}(T>t)$. In this section, we describe a wide class of deep learning variants of the Kaplan-Meier estimator that estimate survival function~$S(\cdot|x)$.\footnote{Backing out estimates of the hazard and cumulative hazard functions is also possible, but for simplicity, we focus just on predicting $S(\cdot|x)$ in this section.} Specifically, we cover what are called \emph{deep kernel survival analysis} (DKSA) models \citep{chen2020deep,chen2024survival}. As we shall see, these models provide a couple different notions of interpretability. A special case of these models also has a theoretical accuracy guarantee. In experiments on standard datasets, these models have been shown to be competitive with various deep time-to-event prediction models.

As a reminder, classically, the Kaplan-Meier estimator discretizes time using the unique times of death $\tau_{(1)}<\tau_{(2)}<\cdots<\tau_{(L)}$. At each time index $\ell\in[L]$, we keep track of the number of deaths at time index~$\ell$ (denoted as $D[\ell]$) and the number of points at risk at time index~$\ell$ (denoted as $N[\ell]$):
\begingroup
\allowdisplaybreaks
\begin{align*}
D[\ell]
&:= \sum_{j=1}^n\ind\{Y_j=\tau_{(\ell)}\}\Delta_j,
\tag*{(\ref{eq:death-counter}, partially reproduced)} \\
N[\ell]
&:= \sum_{j=1}^n\ind\{Y_j\ge\tau_{(\ell)}\}.
\tag*{(\ref{eq:at-risk-counter}, partially reproduced)}
\end{align*}
\endgroup
Then the Kaplan-Meier estimator is given by
\[
\widehat{S}_{\text{KM}}(t)
:=\prod_{m=1}^L
    \Big(
      1 - \frac{D[m]}{N[m]}
    \Big)^{\ind\{\tau_{(m)}\le t\}}
  \quad\text{for }t\ge0.
\tag*{(\ref{eq:kaplan-meier}, partially reproduced)}
\]
We start from this classical nonparametric estimator and build our way to increasingly more sophisticated methods:
\begin{itemize}

\item (Section~\ref{sec:conditional-kaplan-meier}) We begin by presenting conditional Kaplan-Meier estimators \citep{beran1981nonparametric}, which estimate $S(\cdot|x)$ instead of $S_{\text{pop}}(\cdot)$. Conditional Kaplan-Meier estimators build on a simple idea: to predict a survival function specific to $x\in\mathcal{X}$, first determine which of the training raw inputs $X_1,\dots,X_n$ are the $k$ closest ones to~$x$. We then compute a Kaplan-Meier survival function using only these $k$ training points' ground truth labels. In this manner, we just constructed a survival function that depends on~$x$. Moreover, any prediction comes with ``evidence'' as we would know exactly which $k$ training points contributed to the prediction.

A more elaborate ``kernel'' version is to weight the contribution of each training point based on how similar it is to~$x$. Specifically, the user pre-specifies a kernel function (also called a similarity function) $K:\mathcal{X}\times\mathcal{X}\rightarrow[0,\infty)$ that measures how similar any two inputs $x,x'\in\mathcal{X}$ are (\eg, $K(x,x'):=\exp(-\|x-x'\|^2)$). We explain how this so-called \emph{kernel Kaplan-Meier estimator}~works.

\item (Section~\ref{sec:dksa}) The problem with the kernel Kaplan-Meier estimator is that how well it works in practice depends heavily on the kernel function used. To address this problem, we present the DKSA framework by \citet{chen2020deep} that automatically learns the kernel function in a neural network framework. Prediction is still done exactly the same way as the kernel Kaplan-Meier estimator, just with an automatically learned kernel function.

\item (Section~\ref{sec:kernets}) The kernel Kaplan-Meier estimator is computationally expensive for large datasets. When making predictions, a naive implementation would need to compute a similarity score (using the kernel function) of each test raw input $x$ with every training input $X_i$. We go over a compression strategy by \citet{chen2024survival} that results in a class of models called \emph{survival kernets}. Conceptually, survival kernets could be viewed as representing any data point as a combination of a few clusters, each of which could be visualized in terms of how it relates to raw input features and also to time-to-event outcomes. A special case of survival kernets has a theoretical accuracy guarantee.
\end{itemize}

\subsection{Conditional Kaplan-Meier Estimators: \texorpdfstring{$\bm{k}$}{k} Nearest Neighbor and Kernel Variants}
\label{sec:conditional-kaplan-meier}

The basic idea of conditional Kaplan-Meier estimators \citep{beran1981nonparametric} is that we could compute the Kaplan-Meier estimator restricted to (or ``conditioned on'') using only a subset of our training dataset and not necessarily all of it. Beran suggested both $k$ nearest neighbor and kernel Kaplan-Meier estimators.

\paragraph{$\bm{k}$ nearest neighbor Kaplan-Meier estimator}
How the $k$ nearest neighbor Kaplan-Meier estimator works is that we first specify a distance function $\rho:\mathcal{X}\times\mathcal{X}\rightarrow[0,\infty)$ between any two raw inputs (for example, if $\mathcal{X}=\mathbb{R}^d$, then $\rho$ could be Euclidean distance). The distance between test raw input $x$ and training raw input $X_i$ is thus $\rho(x,X_i)$. Let $(X_{(1)},Y_{(1)},\Delta_{(1)}),(X_{(2)},Y_{(2)},\Delta_{(2)}),\dots,(X_{(n)},Y_{(n)},\Delta_{(n)})$ denote the training raw inputs sorted so that $X_{(1)}$ is the closest training raw input to $x$ (according to distance function $\rho$), $X_{(2)}$ is the second closest, and so forth. Then the $k$ nearest neighbor Kaplan-Meier estimator simply computes the standard Kaplan-Meier estimator only using the $k$ ground truth outcome labels $(Y_{(1)},\Delta_{(1)}),(Y_{(2)},\Delta_{(2)})\dots,(Y_{(k)},\Delta_{(k)})$. We recover the standard Kaplan-Meier estimator by choosing $k=n$.

\paragraph{Kernel Kaplan-Meier estimator}
The more general kernel version that Beran suggested assigns weights to different training points according to kernel function $K:\mathcal{X}\times\mathcal{X}\rightarrow[0,\infty)$, where $K(x,X_i)$ having a higher value means that $x$ and $X_i$ are ``more similar''. Then we generalize the definitions for the number of deaths $D[\ell]$ and number of data points at risk $N[\ell]$ at time $\tau_{(\ell)}$ from equations~(\ref{eq:death-counter}) and~(\ref{eq:at-risk-counter}), respectively, into kernel versions. Specifically, for $\ell\in[L]$ and $x\in\mathcal{X}$, we define
\begin{align*}
D_{\text{kernel}}[\ell|x]
&:= \sum_{j=1}^n K(x, X_j)\ind\{Y_j=\tau_{(\ell)}\}\Delta_j,
\\
N_{\text{kernel}}[\ell|x]
&:= \sum_{j=1}^n K(x, X_j)\ind\{Y_j\ge \tau_{(\ell)}\}.
\end{align*}
Here, $D_{\text{kernel}}[\ell|x]$ could be thought of as the number of deaths at time~$\tau_{(\ell)}$ among training points who ``look like''~$x$, where we have weighted each training point $j\in[n]$ by its similarity to~$x$ (the nonnegative weight $K(x,X_j)$). Likewise, $N_{\text{kernel}}[\ell|x]$ could be thought of as the number of training points at risk at time~$\tau_{(\ell)}$ among those who ``look like''~$x$.

Then the kernel Kaplan-Meier estimator is given by
\begin{equation*}
\widehat{S}_{\text{kernel-KM}}(t|x)
:=\prod_{m=1}^{L}
    \Big(1-\frac{D_{\text{kernel}}[m|x]}{N_{\text{kernel}}[m|x]}\Big)^{\ind\{\tau_{(m)}\le t\}}
\quad\text{for }t\ge0,x\in\mathcal{X},
\end{equation*}
with the convention that in the product, we ignore any time index $m$ such that $N_{\text{kernel}}[m|x] = 0$, and if the product is ``empty'' (there are no terms to multiply), then the output is just 1.

The standard Kaplan-Meier estimator could be obtained by just setting $K(x, x') = 1$ for all $x,x'\in\mathcal{X}$. The $k$-nearest neighbor Kaplan-Meier estimator corresponds to the case where $K(x,X_j) = 1$ if $X_j$ is one of the $k$ nearest neighbors of $x$ (among training raw inputs $X_1,X_2,\dots,X_n$) according to some distance function $\rho:\mathcal{X}\times\mathcal{X}\rightarrow[0,\infty)$ (such as Euclidean distance), and $K(x,X_j) = 0$ otherwise.

Similar to how the Kaplan-Meier estimator can be viewed as having a corresponding discrete time hazard function (\eqref{eq:kaplan-meier-hazard-discrete}), so too does the kernel Kaplan-Meier estimator. Specifically, we can define the kernel hazard function estimate to be
\begin{equation}
\widehat{h}_{\text{kernel}}[\ell|x]
:= \frac{D_{\text{kernel}}[\ell|x]}{N_{\text{kernel}}[\ell|x]}
 = \frac{\sum_{j=1}^n
           K(x, X_j)
           \ind\{Y_j=\tau_{(\ell)}\}\Delta_j}
        {\sum_{j=1}^n
           K(x, X_j)
           \ind\{Y_j\ge \tau_{(\ell)}\}},
\label{eq:kernel-hazard}
\end{equation}
which the convention that if the denominator is 0, then we just output $\widehat{h}_{\text{kernel}}[\ell|x]=0$. In particular,
\[
\widehat{S}_{\text{kernel-KM}}(t|x)
=\prod_{m=1}^L (1 - \widehat{h}_{\text{kernel}}[\ell|x])^{\ind\{\tau_{(m)}\le t\}}
\quad\text{for }t\ge0,x\in\mathcal{X}.
\]

\Paragraph{Interpreting predictions}
An appealing aspect of the kernel Kaplan-Meier estimator is that when predicting $S(\cdot|x)$, we know how much any training point $j\in[n]$ contributes to the prediction since it is just given by the similarity score $K(x,X_j)$. Thus, in some sense, we have ``evidence'' for every prediction made, as we could always look at which training points contributed the most to the prediction. How interpretable this evidence is depends on whether it is straightforward for a user to make sense of these ``most similar'' training points to~$x$.

\paragraph{Theoretical guarantees}
Meanwhile, we point out that theory for $k$ nearest neighbor and Kaplan Meier estimators is well-understood. Much like how as $n\rightarrow\infty$, the Kaplan-Meier estimator, under fairly general settings, converges to $S_{\text{pop}}(t):=\mathbb{P}(T>t)$ for all times $t$ that are not too large \citep{foldes1981strong}, a similar result holds for the $k$ nearest neighbor and kernel Kaplan-Meier estimators provably converging to $S(t|x)$ for all times $t$ that are not too large (\citet{chen2019nearest} provides rates of convergence in the case where raw inputs could reside in separable metric spaces, of which Euclidean space is a special case).

\subsection{Learning the Kernel Function: Deep Kernel Survival Analysis}
\label{sec:dksa}

In practice, how well the kernel Kaplan-Meier estimator works heavily depends on the choice of the kernel function $K$ \citep{chen2019nearest}. To automatically learn $K$, \citet{chen2020deep} proposed an approach called deep kernel survival analysis (DKSA). Specifically, Chen set $K$ equal to
\begin{equation}
\mathbf{K}(x,x'; \theta)
:= \exp(-\|\mathbf{f}(x; \theta) - \mathbf{f}(x'; \theta)\|^2)
\quad\text{for }x,x'\in\mathcal{X},
\label{eq:dksa-kernel}
\end{equation}
where $\mathbf{f}(\cdot;\theta)$ is a neural network with parameter variable $\theta$ that maps from the raw input space $\mathcal{X}$ to a latent embedding space $\mathbb{R}^{d_\text{emb}}$ for a user-specified embedding space dimension $d_\text{emb}$.\footnote{Traditionally, the kernel Kaplan-Meier estimator would be used in a simplistic manner where the user specifies $\mathbf{K}(x,x';\theta)$ in terms of a single real-valued ``bandwidth'' parameter. For instance, we could have $\mathbf{f}(x;\theta) = \frac{x}{\sqrt{2\sigma^2}}$ (with $\theta=\sigma$) so that $\mathbf{K}(\cdot,\cdot;\theta)$ is a Gaussian kernel with standard deviation parameter~$\sigma$ (that is taken to be the ``bandwidth''). Other ways to parameterize $\mathbf{K}(x,x';\theta)$ in terms of $\mathbf{f}(\cdot;\theta)$ are possible, \eg, $\mathbf{K}(x,x';\theta) = 1/[e^{\|\mathbf{f}(x;\theta) - \mathbf{f}(x';\theta)\|}+2+e^{-\|\mathbf{f}(x;\theta) - \mathbf{f}(x';\theta)\|}]$.}

We now state how prediction works as it is fairly straightforward and, moreover, the training procedure (that learns $\theta$) depends on the prediction procedure.

\paragraph{Prediction} We use the kernel Kaplan-Meier estimator with the learned kernel function from~\eqref{eq:dksa-kernel}. Specifically, we set $\widehat{h}_{\text{kernel}}[\ell|x]$ from \eqref{eq:kernel-hazard} equal to
\begin{align}
\mathbf{h}_{\text{DKSA}}[\ell|x;\theta]
&:=
  \frac{\sum_{j=1}^n
          \mathbf{K}(x, X_j; \theta)
          \ind\{Y_j=\tau_{(\ell)}\}\Delta_j}
       {\sum_{j=1}^n
          \mathbf{K}(x, X_j; \theta)
          \ind\{Y_j\ge \tau_{(\ell)}\}} \nonumber\\
&\phantom{:}=
  \frac{\sum_{j=1}^n
          \exp(-\|\mathbf{f}(x; \theta) - \mathbf{f}(X_j; \theta)\|^2)
          \ind\{Y_j=\tau_{(\ell)}\}\Delta_j}
       {\sum_{j=1}^n
          \exp(-\|\mathbf{f}(x; \theta) - \mathbf{f}(X_j; \theta)\|^2)
          \ind\{Y_j\ge \tau_{(\ell)}\}},
\label{eq:dksa-hazard}
\end{align}
where, in terms of notation, on the left-hand side, we have dropped the hat ``\hspace{1.5pt}$\widehat{~}$\hspace{1.5pt}'' over the function name to emphasize that at this point, the function has not actually been estimated yet since we still need to learn~$\theta$. Meanwhile, the survival function is given by
\[
\mathbf{S}_{\text{DKSA}}(t|x;\theta)
:=\prod_{m=1}^L (1 - \mathbf{h}_{\text{DKSA}}[\ell|x;\theta])^{\ind\{\tau_{(m)}\le t\}}
\quad\text{for }t\ge0,x\in\mathcal{X}.
\]
If we can come up with an estimate $\widehat{\theta}$ of $\theta$, then we could then predict $S(\cdot|x)$ using
\begin{equation}
\widehat{S}_{\text{DKSA}}(t|x):= \mathbf{S}_{\text{DKSA}}(t|x;\widehat{\theta}\hspace{1.5pt})
\quad\text{for }t\ge0,x\in\mathcal{X}.
\label{eq:dksa-prediction}
\end{equation}

\paragraph{Training}
The basic idea of how to learn $\theta$ is to plug in the hazard function estimate (\eqref{eq:dksa-hazard}) into the hazard form of the discrete time negative log likelihood loss (\eqref{eq:nnet-survival-loss-helper}), \ie, we could use the loss function
\begin{align*}
\mathbf{L}_{\text{DKSA-NLL-naive}}(\theta)
&:=
  -
  \frac{1}{n}
  \sum_{i=1}^n
    \Bigg\{
      \Delta_i \log(\mathbf{h}_{\text{DKSA}}[\kappa(Y_i)|X_i;\theta]) \\
&\phantom{=-\frac{1}{n}\sum_{i=1}^n\Bigg\{}
      +
      (1 - \Delta_i) \log(1-\mathbf{h}_{\text{DKSA}}[\kappa(Y_i)|X_i;\theta]) \\
&\phantom{=-\frac{1}{n}\sum_{i=1}^n\Bigg\{}
      +
      \sum_{m=1}^{\kappa(Y_i)-1}
        \log(1-\mathbf{h}_{\text{DKSA}}[m|X_i;\theta])
    \Bigg\},
\end{align*}
where, as a reminder, $\kappa(Y_i)\in[L]$ is the time index that $Y_i$ corresponds to. However, it turns out that minimizing $\mathbf{L}_{\text{DKSA-NLL-naive}}(\theta)$ works poorly in practice due to overfitting issues. Specifically note that
\begin{equation*}
\mathbf{h}_{\text{DKSA}}[\ell|X_i;\theta]
= \frac{\sum_{j=1}^n
          \exp(-\|\mathbf{f}(X_i; \theta) - \mathbf{f}(X_j; \theta)\|^2)
          \ind\{Y_j=\tau_{(\ell)}\}\Delta_j}
       {\sum_{j=1}^n
          \exp(-\|\mathbf{f}(X_i; \theta) - \mathbf{f}(X_j; \theta)\|^2)
          \ind\{Y_j\ge \tau_{(\ell)}\}},
\end{equation*}
which means that to predict the hazard function for $X_i$, on the right-hand side, we use ground truth outcome labels (namely $Y_i$ and $\Delta_i$) from the $i$-th training point itself. Thus, \citet{chen2020deep} suggested instead using a leave-one-out hazard estimator during model training:
\begin{align}
\mathbf{h}_{\text{DKSA-train}}[\ell|i;\theta]
\!:=\!
  \frac{\sum_{j \ne i}
          \exp(-\|\mathbf{f}(X_i; \theta) - \mathbf{f}(X_j; \theta)\|^2)
          \ind\{Y_j=\tau_{(\ell)}\}\Delta_j}
       {\sum_{j \ne i}
          \exp(-\|\mathbf{f}(X_i; \theta) - \mathbf{f}(X_j; \theta)\|^2)
          \ind\{Y_j\ge \tau_{(\ell)}\}} \nonumber\\
\text{for }\ell\in[L],i\in[n], \nonumber\\
\label{eq:dksa-leave-one-out}
\end{align}
which predicts the hazard function for the $i$-th training point only using the other training points' ground truth information. On the left-hand side, our notation now intentionally makes it clear that $\mathbf{h}_{\text{DKSA-train}}[\ell|i;\theta]$ is not meant to be evaluated at an arbitrary raw input~$x$. Instead, it is only used to predict the hazard function for each training point $i\in[n]$.

The final negative log likelihood training loss used by \citet{chen2020deep}~is
\begin{align*}
\mathbf{L}_{\text{DKSA-NLL}}(\theta)
&:=
  -
  \frac{1}{n}
  \sum_{i=1}^n
    \Bigg\{
      \Delta_i \log(\mathbf{h}_{\text{DKSA-train}}[\kappa(Y_i)|i;\theta]) \nonumber\\
&\phantom{=-\frac{1}{n}\sum_{i=1}^n\Bigg\{}
      +
      (1 - \Delta_i) \log(1-\mathbf{h}_{\text{DKSA-train}}[\kappa(Y_i)|i;\theta]) \nonumber\\
&\phantom{=-\frac{1}{n}\sum_{i=1}^n\Bigg\{}
      +
      \sum_{m=1}^{\kappa(Y_i)-1}
        \log(1-\mathbf{h}_{\text{DKSA-train}}[m|i;\theta])
    \Bigg\}.
\end{align*}
This loss could be numerically minimized using a neural network optimizer to estimate $\widehat{\theta}:=\arg\min_\theta \mathbf{L}_{\text{DKSA-NLL}}(\theta)$. We emphasize here that using a variant of minibatch gradient descent is important rather than using full batch gradient descent (that uses the entire training data per optimization step). In particular, the loss function would require computation time that scales as $\mathcal{O}(n^2)$ when using full batch gradient descent. By using minibatches of batch size $b$ that the user specifies (where $b<n$), the loss function would only be computed for training data in each minibatch, with computation time scaling as $\mathcal{O}(b^2)$.

We mention several refinements that are important in practice to getting DKSA to work well:
\begin{itemize}

\item When coming up with an initial guess for $\theta$ during neural network optimizer, standard random parameter initialization (\eg, \citealt{he2015delving}) tends to work poorly compared to a random initialization strategy based on tree ensembles \citep{chen2020deep,chen2024survival}. In particular, \citet[Section~4]{chen2024survival} showed how to randomly initialize $\theta$ with the help of XGBoost; this initialization strategy can scale to large datasets. The rough idea is that an already trained XGBoost model comes with a kernel function $\widetilde{K}:\mathcal{X}\times\mathcal{X}\rightarrow[0,1]$ that we could easily evaluate for any pair of inputs \citep[Section~7.1.3]{chen2018explaining}. We initialize~$\theta$ by first minimizing (in minibatches) a mean squared error loss between $\mathbf{K}(\cdot,\cdot;\theta)$ and $\widetilde{K}$ (evaluated on data for a single minibatch at a time) for some number of iterations.

\item How time is discretized during training and how time is interpolated for prediction matter in practice \citep{chen2020deep,chen2024survival}. For time discretization, for some datasets, it could be helpful to use all unique times of death whereas for other datasets, using a coarser grid is helpful (which requires some preprocessing that discretizes the $Y_i$ values seen in training data into some user-specified number of bins; we gave examples of ways to do this in Section~\ref{sec:how-to-discretize}). As for time interpolation, even though for simplicity we have presented the kernel Kaplan-Meier estimator using forward filling interpolation, in practice, using a more sophisticated interpolation strategy such as constant density or constant hazard interpolation is better (an explanation of how these work is provided by \citet{kvamme2021continuous}). In short, we advise against using forward filling interpolation.

\item \citet[Section~6]{chen2020deep} discussed how to incorporate the DeepHit model's ranking loss \citep{lee2018deephit} although he did not implement it in the original DKSA paper. The subsequent work by \citet{chen2024survival} does include this ranking loss (so that the overall training loss is the sum of the negative log likelihood loss $\mathbf{L}_{\text{DKSA-NLL}}(\theta)$ and the ranking loss), which improves the achieved time-dependent concordance index of the resulting model.

\end{itemize}
These refinements, however, do not remedy the issue that after model training, when we predict $S(\cdot|x)$ for a single test raw input $x\in\mathcal{X}$, computing the prediction $\widehat{S}_{\text{DKSA}}(\cdot|x)$ using \eqref{eq:dksa-prediction} would involve computing $\mathbf{h}_{\text{DKSA}}[\ell|x;\widehat{\theta}\hspace{1.5pt}]$ using \eqref{eq:dksa-hazard}, which iterates through all $n$ training points. When $n$ is large, this computation is impractical. We resolve this issue in the next section. The resolution turns out to also help with model interpretability and comes with an accuracy guarantee.

\subsection{Scalable Deep Kernel Survival Analysis: Survival Kernets}
\label{sec:kernets}

To accelerate prediction for a single raw input, we want to avoid having to look at all $n$ training points. \citet{chen2024survival} proposed a method for doing this that uses two key ideas:
\begin{itemize}

\item First, after training a DKSA model (which can scale to large datasets using the ideas we presented in the previous section), we cluster the training points using an exemplar-based clustering method. In other words, each cluster has an ``exemplar'' that is an actual training point that represents the cluster. When we make predictions, we only compute the similarity between test raw input $x$ and these exemplars. This idea could be thought of as us compressing training data into a few clusters.

There will be a hyperparameter $\varepsilon\ge0$ that controls how much compression happens, where $\varepsilon=0$ means that there is no compression (in which case prediction will be slower) whereas when $\varepsilon\rightarrow\infty$, then we maximally compress the data (so that there's only a single cluster), but this will mean that we predict the exact same survival function regardless of what the test raw input~$x$ is.

\item Second, we ignore any exemplar that is ``too far'' from~$x$. A key idea here is to exploit existing fast nearest neighbor search algorithms that enable us to quickly find only the exemplars that are close enough to~$x$.

There will be a distance threshold hyperparameter $\tau>0$ that controls what it means to be too far. Having $\tau\rightarrow\infty$ will mean that there is no distance restriction (so prediction will be slower).

\end{itemize}
To combine these ideas in a manner that can come with a theoretical guarantee, Chen extended an existing approach designed for classification and regression called \emph{kernel netting} \citep{kpotufe2017time} to time-to-event prediction. Chen referred to the resulting class of models as \emph{survival kernets} (the latter word abbreviates ``kernel netting'').

We now state the training and prediction procedures for survival kernets, which depends on the hyperparameters $\varepsilon\ge0$ and $\tau>0$ mentioned above. Recall that we have discretized time to the user-specified grid $\tau_{(1)}<\tau_{(2)}<\cdots<\tau_{(L)}$. Let $\mathcal{D}_1\subseteq[n]$ and $\mathcal{D}_2\subseteq[n]$ be subsets of the training data (\eg, we randomly divide the $n$ training data into two halves $\mathcal{D}_1$ and $\mathcal{D}_2$).

\paragraph{Training} We proceed as follows:
\begin{enumerate}

\item (Learn kernel function) Train a DKSA model on dataset $\mathcal{D}_1$ by using a neural network optimizer to minimize $\mathbf{L}_{\text{DKSA-NLL}}(\theta)$ (restricted to using only training data in $\mathcal{D}_1$). Denote the resulting estimate of $\theta$ as $\widehat{\theta}$.

\item (Compute embedding vectors) Recall that $\mathbf{K}(x,x';\theta) = \exp(-\|\mathbf{f}(x;\theta)-\mathbf{f}(x';\theta)\|^2)$ (\eqref{eq:dksa-kernel}). In particular, the distance ${\|\mathbf{f}(x;\theta)-\mathbf{f}(x';\theta)\|}$ is computed in the latent embedding space. We compute the embedding vectors $\widetilde{X}_i=\mathbf{f}(X_i;\widehat{\theta}\hspace{1.5pt})\in\mathbb{R}^{d_{\text{emb}}}$ for each $i\in\mathcal{D}_2$.

\item (Cluster embedding vectors) Run an exemplar-based clustering method on the embedding vectors $\{\widetilde{X}_i : i\in\mathcal{D}_2\}$ to obtain a set of exemplars $\mathcal{Q}\subseteq\mathcal{D}_2$.

While there are different clustering methods that are possible here, to get a theoretical guarantee, the clustering method used by \citet{chen2024survival} is the standard $\varepsilon$-net method (see, for instance, the textbook by \citet{vershynin2018high}), which has a hyperparameter $\varepsilon\ge0$ (where choosing $\varepsilon = 0$ results in every point being its own cluster, and as $\varepsilon\rightarrow\infty$, we will have all points in the same cluster):
\begin{enumerate}

\item Initialize $\mathcal{Q}:=0$.

\item For $i\in\mathcal{D}_2$:
\begin{enumerate}

\item If $\mathcal{Q}$ is empty, then add $i$ to the set of exemplars $\mathcal{Q}$. This means that now point $i$ forms a new cluster.

\item Otherwise:
\begin{enumerate}

\item Let $j$ be the point in $\mathcal{Q}$ whose embedding vector is closest (in Euclidean distance) to $\widetilde{X}_i$.

\item If $\| \widetilde{X}_i - \widetilde{X}_j \| > \varepsilon$: Add $i$ to the set of exemplars~$\mathcal{Q}$, so that point $i$ forms a new cluster.

\smallskip
Otherwise: assign point $i$ to the cluster of exemplar~$j$.

\end{enumerate}

\end{enumerate}

\end{enumerate}

\item (Compute summary information per cluster) For each exemplar $j\in\mathcal{Q}$, let $\mathcal{C}_j\subseteq\mathcal{D}_2$ denote the points that have been assigned to the same cluster as~$j$. Per exemplar~$j$, we compute two summary functions:
\begin{align*}
D_{\text{cluster}}[\ell|j]
&:= \sum_{i\in\mathcal{C}_j} \ind\{Y_i=\tau_{(\ell)}\}\Delta_i,
\\
N_{\text{cluster}}[\ell|j]
&:= \sum_{i\in\mathcal{C}_j} \ind\{Y_i\ge \tau_{(\ell)}\}.
\end{align*}
These are just the death and at-risk counts from equations~(\ref{eq:death-counter}) and~(\ref{eq:at-risk-counter}) restricted to points in the cluster of exemplar~$j$.

\item (Optional summary information fine-tuning) As an optional step, \citet[Section~3.3]{chen2024survival} described a method that fine-tunes the summary functions $D_{\text{cluster}}$ and $N_{\text{cluster}}$ in a neural network framework. For simplicity, we do not go over this fine-tuning step in this monograph.

\end{enumerate}

\paragraph{Prediction} After model training, for a specific test raw input~$x$, we predict $S(\cdot|x)$ as follows.
\begin{enumerate}

\item Compute $\widetilde{x} := \mathbf{f}(x;\widehat{\theta}\hspace{1.5pt})$.

\item Use a nearest neighbor search algorithm to find all exemplars in~$\mathcal{Q}$ whose embedding vectors are within distance $\tau$ from $\widetilde{x}$. Denote this resulting set of exemplars as $\mathcal{Q}(x;\tau)$.

\item Compute the weighted number of deaths and the weighted number of points at risk of dying at each time index $\ell\in[L]$ as follows:
\begin{align*}
D_{\text{kernet}}[\ell|x]
:= \sum_{j\in\mathcal{Q}(x;\tau)} \mathbf{K}(x, X_j; \widehat{\theta}) D_{\text{cluster}}[\ell|j], \\
N_{\text{kernet}}[\ell|x]
:= \sum_{j\in\mathcal{Q}(x;\tau)} \mathbf{K}(x, X_j; \widehat{\theta}) N_{\text{cluster}}[\ell|j].
\end{align*}

\item Finally, we predict $S(\cdot|x)$ using
\[
\widehat{S}_{\text{kernet}}(t|x)
:= \prod_{m=1}^L
     \Big(
       1
       -
       \frac{D_{\text{kernet}}[\ell|x]}
            {N_{\text{kernet}}[\ell|x]}
     \Big)^{\ind\{\tau_{(m)}\le t\}}
  \quad\text{for }t\ge0.
\]

\end{enumerate}
If we set $\mathcal{D}_1=\mathcal{D}_2=[n]$ (\ie, $\mathcal{D}_1$ and $\mathcal{D}_2$ are both set to be the full training data), $\varepsilon=0$, and $\tau=\infty$, and we do not use the optional summary information fine-tuning step that is mentioned, then the resulting survival kernet model would make the same prediction as the original DKSA model (using \eqref{eq:dksa-prediction}). Meanwhile, if $\varepsilon\rightarrow\infty$ (so that there is only a single cluster), $\tau\rightarrow\infty$ (so that for any test point, we predict using the only cluster present), and we do not use the optional summary information fine-tuning, then the resulting survival kernet model would just become the classical Kaplan-Meier estimator.

In his experiments, \citet{chen2024survival} found that setting $\mathcal{D}_1=\mathcal{D}_2=[n]$ and using the optional summary information fine-tuning tends to result in the survival kernet model that achieves the highest time-dependent concordance index in practice.

We provide a Jupyter notebook that implements survival kernets.\footnote{\texttt{\url{https://github.com/georgehc/survival-intro/blob/main/S4.3_Survival_Kernets.ipynb}}} Note that our notebook also shows how to warm-start neural network training with the help of XGBoost, using the strategy by \citet[Section~4]{chen2024survival}.

\paragraph{Model interpretation}
Ignoring the optional summary information fine-tuning step, when a survival kernet model makes a prediction for a test raw input~$x$, the only training data that contribute to the prediction are the ones in $\mathcal{Q}(x;\tau)$. In particular, suppose for a moment that $\mathcal{Q}(x;\tau)$ consists of only a single point $q\in[n]$ (which we could interpret as~$x$ being ``purely explained'' by exemplar~$q$'s cluster according to the learned survival kernet model). Then we would have
\begin{align*}
D_{\text{kernet}}[\ell|x]
&= \mathbf{K}(x, X_q; \widehat{\theta}\hspace{1.5pt}) D_{\text{cluster}}[\ell|q], \\
N_{\text{kernet}}[\ell|x]
&= \mathbf{K}(x, X_q; \widehat{\theta}\hspace{1.5pt}) N_{\text{cluster}}[\ell|q].
\end{align*}
This means that
\begin{align*}
\widehat{S}_{\text{kernet}}(t|x)
&=\prod_{m=1}^L
    \Big(
      1
      -
      \frac{\mathbf{K}(x, X_q; \widehat{\theta}\hspace{1.5pt})
            D_{\text{cluster}}[\ell|q]}
           {\mathbf{K}(x, X_q; \widehat{\theta}\hspace{1.5pt})
            N_{\text{cluster}}[\ell|q]}
    \Big)^{\ind\{\tau_{(m)}\le t\}} \\
&=\prod_{m=1}^L
    \Big(
      1
      -
      \frac{D_{\text{cluster}}[\ell|q]}
           {N_{\text{cluster}}[\ell|q]}
    \Big)^{\ind\{\tau_{(m)}\le t\}},
\end{align*}
which is just the Kaplan-Meier estimator restricted to the points assigned to the cluster of exemplar~$q$.

\begin{figure}[!t]
\centering
\includegraphics[scale=.55]{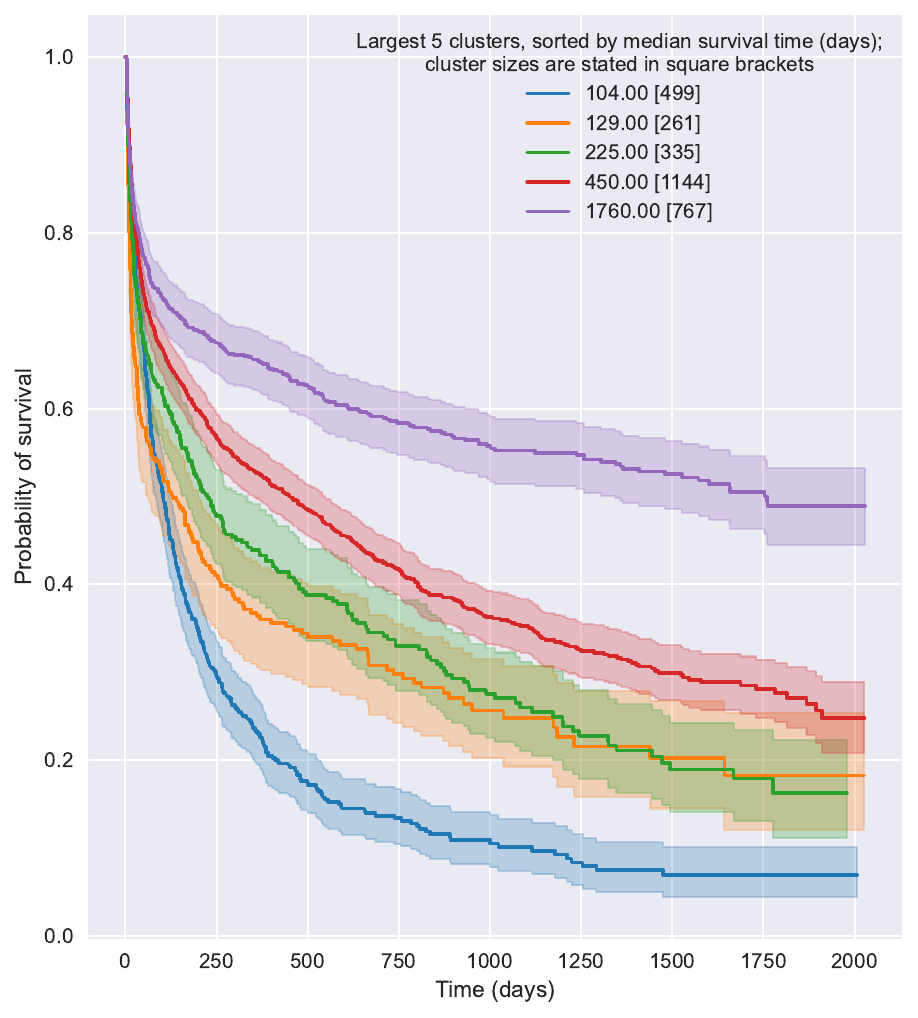}\vspace{-.25em}
\caption{(Figure source: \citealt[Figure~1]{chen2024survival}) For the largest 5 clusters found by a survival kernet model on the SUPPORT dataset \citep{knaus1995support}, these are the clusters' Kaplan-Meier survival function plots overlaid over each other.}
\label{fig:kernet-km-example}
\end{figure}

Thus, since any data point purely explained by a single cluster has a prediction given by the Kaplan-Meier for that cluster, a straightforward way to visualize the different clusters is to just overlay their Kaplan-Meier survival functions over each other (we could make such a visualization for any subset of clusters, such as the five largest clusters found). As an example of this, see Figure~\ref{fig:kernet-km-example}, where we also display: (i) 95\% confidence intervals per Kaplan-Meier survival function estimate (computed using the standard exponential Greenwood formula \citep{kalbfleisch1980statistical}), (ii) an estimate of each cluster's median survival time (the time when the survival function crosses probability~1/2, as discussed in Section~\ref{sec:estimands-continuous}), and (iii) each cluster's~size.

\begin{figure}[!p]
\centering
\includegraphics[scale=.55]{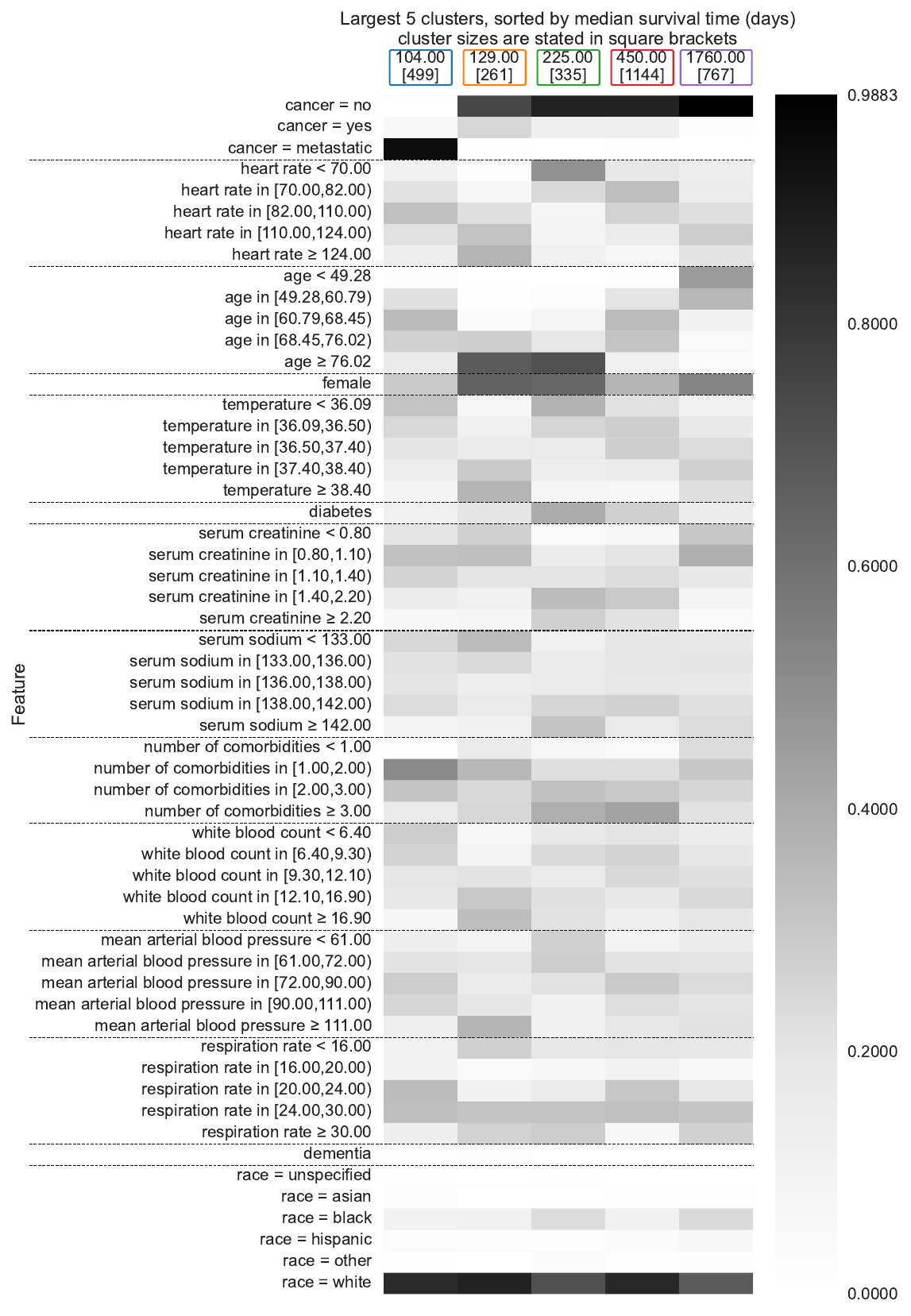}\vspace{-.25em}
\caption{(Figure source: \citealt[Figure~1]{chen2024survival}) For the same clusters as the ones in Figure~\ref{fig:kernet-km-example}, this heatmap shows the prevalence of raw feature values per cluster.}
\label{fig:kernet-heatmap-example}
\end{figure}

In the case where that raw input space corresponds to fixed-length feature vectors (\eg, $\mathcal{X}=\mathbb{R}^d$), \citet{chen2024survival} also proposed a heat map visualization that shows, for each cluster, what values each feature tends to take on. An example of this is shown in Figure~\ref{fig:kernet-heatmap-example}, where rows correspond to values that the features can take on (note that continuous features have been discretized for visualization purposes) and the columns correspond to different clusters.

When making a prediction for test raw input~$x$, we could determine which clusters contribute to the prediction for~$x$ (which are precisely the clusters corresponding to the exemplars in $\mathcal{Q}(x,\tau)$). For only these clusters, we could make visualizations like the ones in Figures~\ref{fig:kernet-km-example} and~\ref{fig:kernet-heatmap-example} as well as report the similarity scores $\mathbf{K}(x,X_j;\widehat{\theta}\hspace{1.5pt})$ for each $j\in\mathcal{Q}(x,\tau)$.

\paragraph{Theory}
Under fairly general conditions, \citet{chen2024survival} showed that the survival kernet model's predicted survival curve converges to $S(t|x)$ for times~$t$ that are not too large. To show this result, Chen's analysis requires the training data subsets $\mathcal{D}_1$ and $\mathcal{D}_2$ to be independent of each other (\eg, two disjoint halves of the full training data) and the optional summary information fine-tuning step cannot be used. Unfortunately, this theory does not help with analyzing the best-performing variant of survival kernets that Chen found in his experimental results, which sets $\mathcal{D}_1=\mathcal{D}_2=[n]$ and uses the summary information fine-tuning step.

\section{Neural Ordinary Differential Equation Formulation of Time-to-Event Prediction}
\label{chap:ode}

As we discussed in Section~\ref{chap:setup}, modeling time-to-event outcomes in continuous time can be challenging in that computing likelihoods involves evaluating integrals. In particular, we can write the likelihood in \eqref{eq:likelihood-pdf} as
\begin{align*}
\mathcal{L}
&=\prod_{i=1}^n
    \big\{f(Y_i|X_i)^{\Delta_i}
          S(Y_i|X_i)^{1-\Delta_i}\big\} \\
&=\prod_{i=1}^n
    \bigg\{
      f(Y_i|X_i)^{\Delta_i}
      \bigg[
        \int_{Y_i}^\infty f(u|X_i)\textrm{d}u
      \bigg]^{1-\Delta_i}
    \bigg\},
\end{align*}
or if we specify the likelihood using the hazard function, then
\[
\mathcal{L}
= \prod_{i=1}^n
    \bigg\{h(Y_i|X_i)^{\Delta_i}
          \exp\Big(-\int_0^{Y_i} h(u|X_i)\textrm{d}u\Big)\bigg\}.
\tag*{(\ref{eq:likelihood-pdf-hazard}, partially reproduced)}
\]
If $f(\cdot|x)$ or $h(\cdot|x)$ can be integrated in closed form, then computing the likelihood functions is straightforward. However, requiring either of these functions to have a closed-form integral could be a restrictive modeling assumption.

A solution that accommodates functions $f(\cdot|x)$ or $h(\cdot|x)$ that lack closed-form integral expressions and that still models time to be continuous is to use neural ordinary differential equations (ODEs) \citep{chen2018neural}. In a nutshell, by using neural ODEs, we could use the continuous time likelihood expression as written (importantly, we leave the integral expression as is without stating what it explicitly evaluates to, as we shall let an ODE solver compute this integral), and it turns out that it is still possible to use minibatch gradient descent for learning model parameters!

As a point of reference, recall that nonparametric methods like conditional Kaplan-Meier estimators or how the baseline hazard function is estimated for the semiparametric Cox model still fundamentally view time as discrete, so that some interpolation is needed to reason about times that are not along the discrete time grid. For example, we had already mentioned that deep kernel survival analysis models in practice depend heavily on how the modeler chooses to discretize time (during training) and to interpolate time (during prediction). Neural ODE time-to-event prediction models move these time discretization and interpolation steps ``under the hood'' so that the modeler does not have to worry about time discretization issues. Instead, these issues are handled by an ODE solver, which we treat as a black box.

At this point, a number of time-to-event prediction models are available based on neural ODEs (\eg, \citealt{groha2020general,tang2022soden,danks2022derivative,moon2022survlatent}). We cover one example of a neural ODE time-to-event prediction model called SODEN (Survival model through Ordinary Differential Equation Networks), proposed by \citet{tang2022soden}.  In their experiments, Tang \emph{et al.}~found SODEN to significantly outperform DeepSurv and Cox-Time while being competitive with DeepHit. SODEN is very general and encompasses essentially all the models we have covered thus far including the discrete time models (even though we no longer have to manually discretize time when using neural ODEs, we could still manually discretize time as we show later). We say ``essentially all'' rather than ``all'' because as we shall see, the neural ODE versions of many models we have already talked about previously might have some minor differences, and they will be trained differently (with the help of an ODE solver).

As a reminder, in Section~\ref{chap:intro}, we already stated that even though SODEN is very general, we are not always best off using it. As our accompanying code hopefully makes clear, SODEN's training procedure is slow compared to the training procedures we covered in Sections~\ref{chap:setup} to~\ref{chap:deep-kaplan-meier}. Also, from playing with the code, one can occasionally encounter numerical stability issues with the ODE solver used.

The rest of this section is organized as follows:
\begin{itemize}

\item (Section~\ref{sec:soden-formulation}) We first go over the ODE formulation Tang \emph{et al.}~used and show how it can represent all the models we have discussed so far in this monograph as special cases.

\item (Section~\ref{sec:soden-predict-train}) We then explain how training and prediction work with SODEN.

\item (Section~\ref{sec:sumonet}) We briefly mention an alternative to using ODEs for handling the integral of \eqref{eq:likelihood-pdf-hazard}. Specifically, we outline an approach called SurvivalMonotonic-net (\mbox{SuMo-net}) by \citet{rindt2022survival}.

\end{itemize}
The main reason we have chosen to focus our exposition on ODEs is that the ODE formulation can readily be related to other models covered earlier in this monograph.

\subsection{General ODE Formulation: SODEN}
\label{sec:soden-formulation}

We present a special case of SODEN \citep{tang2022soden} that is a little easier to describe and corresponds to our problem setup from Section~\ref{chap:setup}. Recall from Summary~\ref{sum:conversions} that $\frac{\textrm{d}}{\textrm{d} t}H(t|x)=h(t|x)$ and that $H(0|x) = \int_0^0 h(u|x)\textrm{d}u = 0$. We use these two constraints to define the following ODE for any $x\in\mathcal{X}$:
\begin{equation}
\begin{cases}
& \frac{\textrm{d}}{\textrm{d} t}H(t|x) = \mathbf{h}\big((t, H(t|x), x); \theta\big)\qquad\text{for }t > 0, \\
& H(0|x)= 0\qquad\text{(initial condition at time }t=0\text{)},
\end{cases}
\label{eq:SODEN-ODE-constraint}
\end{equation}
where $\mathbf{h}(\cdot;\theta)$ is a neural network with parameter variable $\theta$. Since $\frac{\textrm{d}}{\textrm{d} t} H(t|x)=h(t|x)$, this means that $\mathbf{h}(\cdot;\theta)$ models the hazard function, so its output needs to be a nonnegative number. As our notation indicates, $\mathbf{h}(\cdot;\theta)$ takes three inputs: time~$t$, the cumulative hazard value $H(t|x)$ at time $t$, and a raw input~$x$. The solution to the ODE is precisely the cumulative hazard function $H(\cdot|x)$. Let's look at a concrete example.

\begin{fexample}[Weibull time-to-event prediction model as the solution to an ODE]
\label{ex:weibull-aft-SODEN}
Let $\mathcal{X}=\mathbb{R}^d$. Let
\begin{equation}
\mathbf{h}\big( (t, H(t|x), x); \theta\big)
:= \big(H(t|x)\big)^{1-e^{-\phi}}
   e^{\beta^{\top}x+\phi+\psi e^{-\phi}},
\label{eq:soden-weibull-example}
\end{equation}
where $\beta\in\mathbb{R}^d$, $\psi\in\mathbb{R}$ and $\phi\in\mathbb{R}$ are parameters (\ie, $\theta=(\beta,\psi,\phi)$).
Plugging \eqref{eq:soden-weibull-example} into \eqref{eq:SODEN-ODE-constraint} yields the~ODE
\begin{equation}
\begin{cases}
& \frac{\textrm{d}}{\textrm{d} t} H(t|x)
  =\big(H(t|x)\big)^{1-e^{-\phi}}
   e^{\beta^{\top}x+\phi+\psi e^{-\phi}}
  \qquad\text{for }t > 0, \\
& H(0|x)= 0\qquad\text{(initial condition at time }t=0\text{)}.
\end{cases}
\label{eq:SODEN-ODE-weibull-example}
\end{equation}
This ODE can be solved in closed form (as shown in Section~\ref{sec:weibull-aft-ode-solve}), yielding the solution
\[
H(t|x)= e^{(\beta^{\top}x) e^{\phi}+\psi} t^{e^\phi},
\]
which is precisely the cumulative hazard function that we had derived for the Weibull time-to-event prediction model (see \eqref{eq:cumulative-hazard-weibull}).
\end{fexample}

As a reminder, we had pointed out after we first presented the Weibull time-to-event prediction model (Example~\ref{ex:parametric-hazard-weibull-survival}) that this model is both a proportional hazards model and an accelerated failure time (AFT) model (and moreover, it generalizes the exponential time-to-event prediction model we had presented in Examples~\ref{ex:parametric-hazard-exp-survival} and~\ref{ex:parametric-hazard-exp-survival-maximum-likelihood} which correspond to the case where $\phi=0$). In fact, deep proportional hazards and deep AFT models can both be encoded as special cases of SODEN, as we show shortly.

Note that back in Example~\ref{ex:parametric-hazard-weibull-survival}, we already derived a way to learn parameters of the Weibull time-to-event prediction model using maximum likelihood. We will be able to learn these parameters using maximum likelihood under this neural ODE framework as well, where the difference is that the learning procedure relies on an ODE solver.

We now give some examples of families of time-to-event prediction models that are handled by SODEN to illustrate its level of generality.

\subsubsection{Special Case: Deep Proportional Hazards Models}
\label{sec:deep-ph-soden}

To encode a deep proportional hazards model that is fully parametric in SODEN, it suffices to set
\[
\mathbf{h}\big( (t, H(t|x), x); \theta\big)
:= \mathbf{h}_0(t; \theta) e^{\mathbf{f}(x; \theta)},
\]
where $\mathbf{h}_0(\cdot;\theta):[0,\infty)\rightarrow[0,\infty)$ and $\mathbf{f}(\cdot; \theta):\mathcal{X}\rightarrow\mathbb{R}$ are two different user-specified neural networks with parameter variable~$\theta$. Since $\mathbf{h}(\cdot; \theta)$ models the hazard function, by construction, we have chosen it to satisfy the proportional hazards assumption~(\eqref{eq:prop-hazard-assumption}). Once we have specified $\mathbf{h}\big( (t, H(t|x), x); \theta\big)$, we can then use SODEN's training procedure that depends on an ODE solver (described momentarily in Section~\ref{sec:soden-predict-train}); in fact, this resulting method is called SODEN-PH in the original SODEN paper \citep{tang2022soden}. We emphasize that from a modeling perspective, a fully parametric deep proportional hazards model does not actually need to be encoded in an ODE framework for us to learn the model parameters (since we can use the training procedure described in Section~\ref{sec:cox-parametric} instead). However, we are simply showing that it is also possible to learn this same model class within the SODEN framework with a different training procedure.

To encode a deep proportional hazards model that is semiparametric, we could still use the above formulation, but we instead set $\mathbf{h}_0(\cdot;\theta)$ to be a piecewise constant function. Let $\tau_{(1)}<\tau_{(2)}<\cdots<\tau_{(L)}$ be the unique times of death, and set $\tau_{(0)}:=0$. We introduce unconstrained parameters $\gamma_1,\gamma_2,\dots,\gamma_L\in\mathbb{R}$ (so that $\theta$ now includes $\gamma_1,\dots,\gamma_L$). Then we define
\[
\mathbf{h}_0(t;\theta)
:=\begin{cases}
    g(\gamma_\ell) & \text{if }\tau_{(\ell-1)}<t\le\tau_{(\ell)}\text{ for }\ell\in[L], \\
    0 & \text{if }t > \tau_{(L)},
  \end{cases},
\]
where $g(\cdot)$ is any activation function that outputs a nonnegative number (\eg, ReLU, softplus); recall that in continuous time, hazard function values could be larger than~1, so we do not need to enforce an upper bound constraint (see Remark~\ref{rem:piecewise-constant}). This semiparametric model could then be learned using SODEN's training procedure (Section~\ref{sec:soden-predict-train}) and, in particular, we do not need to use the two-step procedure from Section~\ref{sec:cox-semiparametric}.

Note that even though we motivated the use of neural ODEs as a way to not have to explicitly discretize time, our semiparametric example here emphasizes that we could still intentionally set $\mathbf{h}_0(\cdot;\theta)$ to use a discrete time grid. Separately, we could of course replace $\mathbf{f}(x;\theta)$ with a neural network that depends on both~$x$ and~$t$ to get a model like Cox-Time (Section~\ref{sec:cox-time}).

\subsubsection{Special Case: Deep Accelerated Failure Time Models}
\label{sec:aft}

As another example of a wide family of models that SODEN encompasses, we look at deep AFT models. In general, these assume that the survival function is of the form
\begin{equation}
S(t|x) = \mathbf{S}_0(t e^{\mathbf{f}(x; \theta)}; \theta)
\quad\text{for }t\ge0,x\in\mathcal{X},
\label{eq:aft-deep}
\end{equation}
where $\mathbf{S}_0(\cdot;\theta):[0,\infty)\rightarrow[0,1]$ and $\mathbf{f}(\cdot;\theta):\mathcal{X}\rightarrow\mathbb{R}$ are neural networks with parameter variable~$\theta$. Note that $\mathbf{S}_0(\cdot;\theta)$ is a survival function (so that it monotonically decays from~1 to~0). The above property implies that for all $x\in\mathcal{X}$, the survival function $S(\cdot|x)$ has the same shape (namely that of $\mathbf{S}_0(\cdot;\theta)$) except where we stretch the time axis by a factor of $e^{\mathbf{f}(x;\theta)}$. When $\mathbf{f}(x;\theta)$ is larger, we accelerate how fast death is likely to happen. Note that ``classical'' AFT models \citep{prentice1979hazard} correspond to the case where $\mathcal{X}=\mathbb{R}^d$ and $\mathbf{f}(x;\theta) = \theta^\top x$ for parameter vector~$\theta\in\mathbb{R}^d$, and $\mathbf{S}_0(\cdot;\theta)$ could either be parametric or left unspecified.\footnote{As we pointed out in \cref{foot:classical-aft} on \cpageref{foot:classical-aft}, for more details on classical AFT models, please see Chapter~12 of the textbook by \citet{klein2003survival} or Chapter~3 of the textbook by \citet{box2004event}.}

\begin{fexample}[Weibull time-to-event prediction model is an AFT model]
The Weibull time-to-event prediction model corresponds to the case where $\mathcal{X}=\mathbb{R}^d$, $\mathbf{S}_0(t;\theta) = \exp(-e^\psi t^{e^\phi})$, and $\mathbf{f}(x;\theta) = \beta^\top x$, where $\theta=(\beta,\psi,\phi)\in\mathbb{R}^d\times\mathbb{R}\times\mathbb{R}$.
\end{fexample}

\Eqref{eq:aft-deep} is equivalent to the condition
\begin{equation}
h(t|x) = \mathbf{h}_0(t e^{\mathbf{f}(x;\theta)};\theta)e^{\mathbf{f}(x;\theta)}
\quad\text{for }t\ge0,x\in\mathcal{X},
\label{eq:aft-deep-hazard}
\end{equation}
where $\mathbf{h}_0(t;\theta) := -\frac{\textrm{d}}{\textrm{d}t}\log \mathbf{S}_0(t;\theta)$ (for a derivation of this equivalence, see Section~\ref{sec:aft-deep-hazard-derivation}). Thus, in the SODEN model, we could represent a deep AFT model by setting
\[
\mathbf{h}\big( (t, H(t|x), x); \theta \big)
:= \mathbf{h}_0(t e^{\mathbf{f}(x;\theta)};\theta)e^{\mathbf{f}(x;\theta)},
\]
where the user specifies the neural networks $\mathbf{h}_0(\cdot;\theta)$ and $\mathbf{f}(\cdot;\theta)$.

\citet{tang2022survival} suggested an alternative approach to specifying deep AFT models in an ODE framework (note that they actually suggested it for a classical rather than deep AFT model, but the idea trivially extends to the deep AFT case). In particular, it turns out that if we set
\[
\mathbf{h}\big( (t, H(t|x), x); \theta)
:= \mathbf{g}(H(t|x); \theta) e^{\mathbf{f}(x; \theta)},
\]
for some user-specified neural networks $\mathbf{g}(\cdot;\theta)$ and $\mathbf{f}(\cdot;\theta)$, then the solution to the ODE is a deep AFT model. In fact, we already saw this way of specifying an AFT model in Example~\ref{ex:weibull-aft-SODEN}, where $\mathbf{g}(t; \theta):=e^{\phi + \psi e^{-\phi}} t^{1-e^{-\phi}}$, and $\mathbf{f}(x;\theta) = \beta^\top x$. The function $\mathbf{g}(\cdot;\theta)$ relates to the shape of $\mathbf{S}_0(\cdot;\theta)$ in a nontrivial manner, although it is possible to convert between these two functions (for details, see Section 2.2 of~\citet{tang2022survival}).

\subsubsection{Special Case: Deep Extended Hazard Models}

A family of models that contains both deep proportional hazards and deep AFT models as special cases is called \emph{deep extended hazard models} (DeepEH) \citep{zhong2021deep}. As the basic idea is straightforward, we directly state how to set $\mathbf{h}(\cdot;\theta)$ for SODEN to get a DeepEH model:
\[
\mathbf{h}\big( (t, H(t|x), x); \theta \big)
:= \mathbf{h}_0( te^{\mathbf{f}_1(x;\theta)}; \theta ) e^{\mathbf{f}_2(x;\theta)},
\]
where $\mathbf{h}_0(\cdot;\theta)$, $\mathbf{f}_1(\cdot;\theta)$, and $\mathbf{f}_2(\cdot;\theta)$ are neural networks with parameter variable~$\theta$. When $\mathbf{f}_1(x;\theta) = 0$, then we get a deep proportional hazards model (see \eqref{eq:prop-hazard-assumption}). If instead $\mathbf{f}_1(x;\theta)=\mathbf{f}_2(x;\theta)$, then we get a deep AFT model (see \eqref{eq:aft-deep-hazard}). Note that the way \citet{zhong2021deep} learn a DeepEH model has a known theoretical guarantee, whereas if we learn it using the SODEN learning procedure, there is no known theoretical guarantee that we are aware of.

\subsection[Special Case: Converting Discrete Time Models to Continuous Time]{Special Case: Converting Discrete Time Models to\\ Continuous Time}
\label{sec:discrete-to-continuous}

As we saw for deep proportional hazards models in Section~\ref{sec:deep-ph-soden}, we could set the baseline hazard function $\mathbf{h}_0(\cdot;\theta)$ to be piecewise constant over a discrete time grid. We could use this same idea to directly specify the hazard function $\mathbf{h}(\cdot;\theta)$ as piecewise constant over a user-specified grid $\tau_{(1)}<\tau_{(2)}<\cdots<\tau_{(L)}$ (these need not be the unique times of death and could be chosen by other strategies) with $\tau_{(0)}:=0$. In fact, by making this piecewise constant assumption, we could actually convert any time-to-event prediction model specified in discrete time (such as the ones in Section~\ref{sec:setup-discrete} as well as deep kernel Kaplan-Meier estimators of Section~\ref{chap:deep-kaplan-meier}) into a continuous time model.

To give an example of how to convert a discrete time model to continuous time, consider Nnet-survival \citep{gensheimer2019scalable}, which we covered in Example~\ref{ex:nnet-survival}. For this model, we could set the SODEN parametric hazard function to be
\begin{align*}
&\mathbf{h}\big( (t, H(t|x), x) ;\theta\big) \\
&\quad:=
  \begin{cases}
    \Big(\frac{1}{\tau_{(\ell)}-\tau_{(\ell-1)}}\Big) \mathbf{h}[\ell|x;\theta] & \text{if }\tau_{(\ell-1)}< t\le\tau_{(\ell)}\text{ for }\ell\in[L], \\
    0 & \text{if }t > \tau_{(L)},
  \end{cases}
\end{align*}
where $\mathbf{h}[\cdot|x;\theta]$ is given in \eqref{eq:nnet-survival-hazard}. Note that in this case, the ODE is actually straightforward to solve since $\mathbf{h}\big( (t, H(t|x), x) ;\theta\big)$ does not depend on $H(t|x)$. Then note that by integrating from time~0 to time~$\tau_{(\ell)}$ for $\ell\in[L]$, we get
\[
H(\tau_{(\ell)}|x)
=
\int_0^{\tau_{(\ell)}} \mathbf{h}\big( (u, H(u|x), x) ;\theta\big)\textrm{d}u
= \sum_{m=1}^\ell \mathbf{h}[m|x;\theta].
\]
Perhaps what is more interesting is that the model would now interpolate. Suppose that time $t\in(\tau_{(\ell-1)}, \tau_{(\ell)}]$. Then
\begin{align*}
H(t|x)
&=\int_0^{\tau_{(\ell)}} \mathbf{h}\big( (u, H(u|x), x) ;\theta\big)\textrm{d}u \\
&=\sum_{m=1}^{\ell-1} \mathbf{h}[m|x;\theta]
  +
  \bigg(
    \frac{t - \tau_{(\ell-1)}}
         {\tau_{(\ell)} - \tau_{(\ell-1)}}
  \bigg)
  \mathbf{h}[\ell|x;\theta].
\end{align*}
This is precisely using an interpolation strategy that assumes a piecewise constant hazard function.

In terms of how the prediction targets relate between continuous and discrete time, first note that $H[\ell|x]=H(\tau_{(\ell)}|x)$. However, $h[\ell|x]$ would in general not equal $h(\tau_{(\ell)}|x)=\frac{1}{\tau_{(\ell)}-\tau_{(\ell-1)}}\mathbf{h}[\ell|x;\theta]$ due to the extra multiplicative factor in the latter; instead,
\[
h[\ell|x] = \mathbf{h}[\ell|x;\theta] = h(\tau_{(\ell)}|x)\cdot(\tau_{(\ell)}-\tau_{(\ell-1)})
\quad\text{for }\ell\in[L].
\]
Meanwhile, $S[\ell|x;\theta]$ would also in general not equal $S(\tau_{(\ell)}|x)=\exp(-H(\tau_{(\ell)}|x))=\exp(-H[\ell|x])$ due to Proposition~\ref{prop:discrete-cumulative-hazard-not-neg-log-surv}. However, this proposition shows the precise manner in which $S(\tau_{(\ell)}|x)=\exp(-H[\ell|x])$ approximates~$S[\ell|x]$.

Since deep kernel survival analysis could be viewed as parameterizing a discrete time hazard function (\eqref{eq:dksa-hazard}), converting a deep kernel survival analysis model to continuous time would work the same way as what we just showed for Nnet-survival. The main difference is that we would also have to remember to convert the leave-one-out discrete time hazard functions (\eqref{eq:dksa-leave-one-out}) during model training.

The same conversion strategy could be used with the simplified version of DeepHit \citep{lee2018deephit} that we presented in Example~\ref{ex:deephit}, with just a minor difference: DeepHit parameterizes the distribution~$\mathbb{P}_{T|X}(\cdot|x)$ in terms of the survival time PMF~$f[\cdot|x]$ and not the hazard function~$h[\cdot|x]$. However, we could simply use Summary~\ref{sum:conversions-discrete} to obtain
\[
h[\ell|x] = \frac{f[\ell|x]}{S[\ell-1|x]} = \frac{f[\ell|x]}{\sum_{m=\ell}^L f[m|x]},
\]
which means that by having a parametric form of $f[\cdot|x]$, we have a parametric form for $h[\cdot|x]$.

Overall, converting discrete time models to continuous time models using a neural ODE framework is possible and gives a different way of learning such discrete time models (by using the general learning procedure for SODEN) that automatically also handles interpolation. However, in such cases, solving the maximum likelihood problem directly in discrete time could be faster in practice as there is no need to use an ODE solver. After training a discrete time model, we could also use piecewise constant hazard interpolation to back out continuous time predictions.

\subsection{Prediction and Training with a SODEN Model}
\label{sec:soden-predict-train}

We now give an overview of how to train a SODEN model and how to subsequently make predictions.

\paragraph{Training}
Training neural ODEs (such as the one in \eqref{eq:SODEN-ODE-constraint}) is possible thanks to the landmark paper by \citet{chen2018neural}. Importantly, using any user-specified ODE solver, given any raw input~$x\in\mathcal{X}$ and neural network parameters~$\theta$, we can numerically solve the ODE in \eqref{eq:SODEN-ODE-constraint} (going from time 0 to any user-specified time $t>0$) to obtain an estimate for~$H(t|x)$. We denote the resulting estimate as~$H_{\text{ODE-solve}}(t|x;\theta)$. Then a major result of \citet{chen2018neural} is that the loss function we use can contain the terms~$\mathbf{h}((t,H(t|x),x);\theta)$ and~$H_{\text{ODE-solve}}(t|x;\theta)$, where it is possible to compute the gradient of $H_{\text{ODE-solve}}(t|x;\theta)$ with respect to $\theta$ (Chen \emph{et al.}~provide the software package \texttt{torchdiffeq} for computing such gradients).

Then to train the SODEN model, \citet{tang2022soden} simply use the negative log likelihood loss we had stated in \eqref{eq:nll-loss-hazard-form} except we replace $\mathbf{h}(t|X_i;\theta)$ with $\mathbf{h}((Y_i,H(Y_i|X_i),X_i);\theta)$ and we replace the integral (which is equal to $H(Y_i|X_i)$) with $H_{\text{ODE-solve}}(Y_i|X_i;\theta)$. The resulting loss is
\begin{align*}
&\mathbf{L}_{\text{SODEN-NLL}}(\theta) \\
&\quad:= -\frac{1}{n}\sum_{i=1}^n \{\Delta_i\log \mathbf{h}\big( (Y_i, H(Y_i|X_i), X_i) ;\theta\big) - H_{\text{ODE-solve}}(Y_i|X_i;\theta)\}.
\end{align*}
We can use a neural network optimizer to minimize the loss to obtain an estimate $\widehat{\theta} := \arg\min_\theta \mathbf{L}_{\text{SODEN-NLL}}(\theta)$.

\paragraph{Prediction}
For any test raw input~$x\in\mathcal{X}$, by using the user-specified ODE solver, we can predict the cumulative hazard function using $\widehat{H}(t|x) := H_{\text{ODE-solve}}(t|x;\widehat{\theta}\hspace{1.5pt})$. We can then predict the survival function with $\widehat{S}(t|x):=\exp(-\widehat{H}(t|x))$ and the hazard function with $\widehat{h}(t|x) := \mathbf{h}\big( (t, \widehat{H}(t|x), x); \widehat{\theta}\hspace{1.5pt} \big)$.

Note that to predict the hazard function, we first predict the cumulative hazard function since even though neural network $\mathbf{h}(\cdot; \widehat{\theta}\hspace{1.5pt})$ models the hazard function, recall that it takes in three inputs and in general can depend on the cumulative hazard value at any given time.

We provide a Jupyter notebook that shows how to train a SODEN model and subsequently make predictions with it.\footnote{\texttt{\url{https://github.com/georgehc/survival-intro/blob/main/S5.2_SODEN.ipynb}}} Our notebook makes it clear where calls to the ODE solver happen.

\subsection{An Alternative to ODEs via Monotonic Networks: \mbox{SuMo-net}}
\label{sec:sumonet}

We began Section~\ref{chap:ode} by mentioning that the right-censored likelihood from Section~\ref{chap:setup} has the form
\[
\mathcal{L}
= \prod_{i=1}^n
    \bigg\{h(Y_i|X_i)^{\Delta_i}
          \exp\Big(-\int_0^{Y_i} h(u|X_i)\textrm{d}u\Big)\bigg\},
\tag*{(\ref{eq:likelihood-pdf-hazard}, partially reproduced)}
\]
where the difficulty of working with this likelihood is in evaluating the integral. At a high level, the basic idea of SODEN was that we modeled the hazard function $h$ with a neural network $\mathbf{h}(\cdot;\theta)$, and we relied on the neural ODE framework to take care of automatically integrating $h(\cdot|X_i)$ (specified in terms of $\mathbf{h}(\cdot;\theta)$) to compute $H(Y_i|X_i) = \int_0^{Y_i} h(u|X_i)\textrm{d}u$.

Naturally, this suggests that we could have attempted an alternative parameterization. We can rewrite the likelihood of \eqref{eq:likelihood-pdf-hazard} as
\[
\mathcal{L}
= \prod_{i=1}^n
    \bigg\{\Big[\frac{\textrm{d} H}{\textrm{d} t}(Y_i|X_i)\Big]^{\Delta_i}
          \exp\big(-H(Y_i|X_i) \big)\bigg\}.
\]
Then we could directly model the cumulative hazard function $H(\cdot|x)$ with a neural net $\mathbf{H}(\cdot|x;\theta)$ and instead rely on automatic differentiation software to compute $\frac{d H}{d t}(Y_i|X_i) = h(Y_i|X_i)$ (so that now we leave the derivative unspecified in the loss function, whereas previously in the ODE setup, we left the cumulative hazard function unspecified in the loss). However, we need to add the constraint that $\mathbf{H}(t|x;\theta)$ is nonnegative and monotonically increases with respect to time~$t$. We can take advantage of the fact that there already exist standard neural network architectures for enforcing monotonicity (\eg, \citealt{chilinski2020neural,yanagisawa2022hierarchical}). This approach does not use neural ODEs and simply takes advantage of automatic differentiation, which is already a standard component of all modern neural network software packages. This resulting approach corresponds to the SurvivalMonotonic-Network (\mbox{SuMo-net}) model by \citet{rindt2022survival}.\footnote{Technically, \citet{rindt2022survival} specify \mbox{SuMo-net} in terms of constraining the survival function $S(t|x)$ to monotonically decrease in $t$, but they also explain how to instead directly model the cumulative hazard function $H(t|x)$ (and constrain this to be monotonic), which corresponds to our exposition.}

\citet{rindt2022survival} show that \mbox{SuMo-net} works very well in practice, outperforming neural ODEs (namely, SODEN and also the SurvNODE model by \citet{groha2020general}) on several benchmark datasets in terms of log likelihood scores and also computation time. In terms of computation time, roughly, automatic differentiation is relatively fast (and natively supported by neural network software packages) compared to making many calls to an ODE solver.

The reason that it is straightforward relating SODEN to models we have presented earlier in this monograph is that these models can be specified in terms of a hazard function (continuous or discrete), \ie, there is some way to directly parameterize the hazard function. \mbox{SuMo-net}, on the other hand, could be thought of as asking the modeler to parameterize either the survival or cumulative hazard function directly, but not the hazard function, so that the hazard function is indirectly obtained. Whether this is desirable depends on the use case. Of course, if one just cares about a survival model doing well on a specific evaluation metric and nothing else, then we could just choose whichever model does best on the evaluation metric of interest. However, if there are other design considerations (such as some notion of interpretability), or if the model is not being used purely for prediction but for causal inference, then choosing which model is ``best'' can be more challenging. We discuss some of these issues in Section~\ref{chap:discussion}.

\begin{subappendices}
\subsection{Technical Details}

\subsubsection{Solving the Weibull Time-to-Event Prediction Model's ODE From Example~\ref{ex:weibull-aft-SODEN}}
\label{sec:weibull-aft-ode-solve}

Treating $x$ as fixed, the ODE of \eqref{eq:SODEN-ODE-weibull-example} can be written as
\[
\frac{\textrm{d}}{\textrm{d} t}y(t) = a\cdot b\cdot y(t)^{1-\frac{1}{a}}
\]
subject to the constraint that $y(0) = 0$ (where in our case, $y(t) = H(t|x)$, $a=e^\phi$, and $b=e^{\beta^{\top}x+\psi e^{-\phi}}$). Rearranging terms, we have
\[
y(t)^{\frac{1}{a}-1} \frac{\textrm{d}}{\textrm{d} t}y(t) = a\cdot b.
\]
Integrate both sides with respect to $t$ to get
\[
a (y(t))^{1/a} = a\cdot b\cdot t + c,
\]
where $c$ is a constant to be determined based on the constraint $y(0)=0$. We rearrange terms to get that
\[
y(t) = \Big(b\cdot t + \frac{c}{a}\Big)^a,
\]
where, using the constraint, we get that we must have $c=0$. Hence, $y(t)=(b\cdot t)^a$, \ie,
\[
H(t|x)=(e^{\beta^{\top}x+\psi e^{-\phi}} t)^{e^\phi}=e^{(\beta^{\top}x) e^{\phi}+\psi} t^{e^\phi}.
\]

\subsubsection{Deriving the Hazard Function of Deep AFT Models}
\label{sec:aft-deep-hazard-derivation}

We show that the AFT model as defined in \eqref{eq:aft-deep} (namely that $S(t|x) = \mathbf{S}_0(t e^{\mathbf{f}(x; \theta)}; \theta)$) implies that the hazard function $h(\cdot|x)$ is the one in \eqref{eq:aft-deep-hazard}, which we reproduce here for convenience:
\[
h(t|x) = \mathbf{h}_0(t e^{\mathbf{f}(x;\theta)};\theta)e^{\mathbf{f}(x;\theta)}
\quad\text{for }t\ge0,x\in\mathcal{X}.
\tag*{(\ref{eq:aft-deep-hazard}, reproduced)}
\]
To prove that this factorization holds for the hazard function, we begin by stating yet another equivalent characterization of a deep AFT model that will be helpful.

\begin{fproposition}[Log survival time viewpoint of a deep AFT model]
\label{prop:deep-aft-log-survival-time-viewpoint}
Using the time-to-event prediction setup in Section~\ref{sec:setup} and the key assumptions of~\ref{sec:setup-continuous}, suppose that the random survival time $T$ satisfies the equality
\begin{equation}
\log T = -\mathbf{f}(X;\theta) + W,
\label{eq:aft-deep-alt}
\end{equation}
where the ``noise'' random variable $W$ is independent of everything else, and $e^W$ has a CDF given by $\mathbf{F}_0(t;\theta):=1 - \mathbf{S}_0(t;\theta)$ for $t\ge0$. Then this setup is equivalent to making the assumption that the survival function $S(\cdot|x)$ satisfies the factorization in \eqref{eq:aft-deep}, \ie, this time-to-event prediction model is a deep AFT model.
\end{fproposition}

\begin{proof}[Proof of Proposition~\ref{prop:deep-aft-log-survival-time-viewpoint}]
To see why \eqref{eq:aft-deep-alt} is equivalent to \eqref{eq:aft-deep}, first note that \eqref{eq:aft-deep-alt} can be rearranged as $T = e^{-\mathbf{f}(X;\theta)} e^W$. Then,
\begin{align*}
S(t|x) & =\mathbb{P}(T>t|X=x)\\
 & =\mathbb{P}(e^{-\mathbf{f}(X;\theta)}e^{W}>t|X=x)\\
 & =\mathbb{P}(e^W >te^{\mathbf{f}(X;\theta)})\\
 & =1-\mathbf{F}_0(te^{\mathbf{f}(X;\theta)}; \theta)\\
 & =\mathbf{S}_0(te^{\mathbf{f}(X;\theta)}; \theta),
\end{align*}
which shows that \eqref{eq:aft-deep} holds. We could reverse the steps to show that \eqref{eq:aft-deep} implies \eqref{eq:aft-deep-alt}.
\end{proof}
We now proceed to derive the hazard function of a deep AFT model. Summary~\ref{sum:conversions} tells us that $h(t|x)=\frac{f(t|x)}{S(t|x)}$ which combined with \eqref{eq:aft-deep} yields
\begin{equation}
h(t|x)=\frac{f(t|x)}{\mathbf{S}_0(t e^{\mathbf{f}(x; \theta)}; \theta)}.
\label{eq:aft-hazard-derivation-helper}
\end{equation}
As a reminder, $f(\cdot|x)$ is the PDF of $\mathbb{P}_{T|X}(\cdot|x)$, and the corresponding CDF is $F(t|x) = \int_0^t f(u|x)\textrm{d}u$. We next write $f(\cdot|x)$ in terms of $\mathbf{f}_0(\cdot;\theta)$, the PDF corresponding to CDF $\mathbf{F}_0(\cdot;\theta)$. To do this, we start by writing CDF $F(\cdot|x)$ in terms of~$\mathbf{F}_0(\cdot;\theta)$:
\begin{align*}
F(t|x) & =\mathbb{P}(T\le t|X=x)\\
 & =\mathbb{P}(e^{-\mathbf{f}(X;\theta)+W}\le t|X=x) \quad\qquad (\text{using \eqref{eq:aft-deep-alt}}) \\
 & =\mathbb{P}(e^{W}\le te^{\mathbf{f}(X;\theta)}|X=x)\\
 & =\mathbb{P}(e^{W}\le te^{\mathbf{f}(x;\theta)})\\
 & =\mathbf{F}_0(te^{\mathbf{f}(x;\theta)}; \theta).
\end{align*}
Then using the derivative chain rule,
\begin{equation}
f(t|x) =\frac{\textrm{d}F(t|x)}{\textrm{d}t}
=\frac{\textrm{d}}{\textrm{d}t}\mathbf{F}_0(te^{\mathbf{f}(x;\theta)}; \theta)
=\mathbf{f}_0(te^{f(x;\theta)}; \theta)e^{\mathbf{f}(x;\theta)}.
\label{eq:aft-deep-pdf-derived-distribution}
\end{equation}
Combining equations~(\ref{eq:aft-hazard-derivation-helper}) and~(\ref{eq:aft-deep-pdf-derived-distribution}), we have
\[
h(t|x)=\frac{\mathbf{f}_0(te^{\mathbf{f}(x;\theta)}; \theta)e^{\mathbf{f}(x;\theta)}}{\mathbf{S}_0(te^{\mathbf{f}(x;\theta)};\theta)} = \mathbf{h}_0(te^{\mathbf{f}(x;\theta)};\theta) e^{\beta^{\top}x},
\]
where the last step uses the fact that $\mathbf{h}_0(t;\theta) = \frac{\mathbf{f}_0(t;\theta)}{\mathbf{S}_0(t;\theta)}$ for the same reason $h(t|x) = \frac{f(t|x)}{S(t|x)}$ in Summary~\ref{sum:conversions}. This completes the proof.$\hfill\square$
\end{subappendices}

\section{Beyond the Basic Time-to-Event Prediction Setup: Multiple Critical Events and Time Series as Raw Inputs}
\label{chap:extensions}

In this section, we present two extensions of the basic time-to-event prediction problem setup we described in Section~\ref{chap:setup} that showcase concrete directions where deep learning models have been successful. Specifically, we go over the following two extensions that progressively get more general than the standard time-to-event prediction setup:
\begin{itemize}

\item (Section~\ref{sec:competing-risks}) In previous sections, we focused on modeling the time until a specific critical event (\eg, death) happens. We now consider a more general setup where there are $k$ different critical events that we keep track of. For any data point, we want to reason about the time until the earliest of these events happens, as well as which of the $k$ events it is. This is referred to as the \emph{competing risks} setup since we could think of the $k$ events as competing to see which one happens first. A number of deep learning models have been developed for this setup. We go over one called DeepHit \citep{lee2018deephit}. Note that the $k=1$ case recovers the standard setup from Section~\ref{chap:setup}.

\item (Section~\ref{sec:dynamic}) We then turn to the problem of what happens when each data point is actually a time series. As we see more of a time series over time, we could keep making predictions of the time until the earliest of $k$ critical events happens and which event it is. Each training point is a time series, and different training points could be time series of different lengths. We go over an example model that handles this setup called \mbox{Dynamic-DeepHit} \citep{lee2019dynamic}. In the special case where every time series has length~1 (meaning that per data point, we only see the raw input at a single time point before we aim to make a prediction), we recover the setup of Section~\ref{sec:competing-risks}.

\end{itemize}

\subsection{Time-to-Event Prediction with Multiple Critical Events: The Competing Risks Setup}
\label{sec:competing-risks}

Similar to our exposition in Section~\ref{chap:setup}, we first state the statistical framework (Section~\ref{sec:competing-risks-framework}) and the prediction problem (Section~\ref{sec:competing-risks-prediction}). These lead to a likelihood expression that we could write (Section~\ref{sec:competing-risks-likelihood}). We then give an example model that maximizes the likelihood (Section~\ref{sec:deephit-general}), which is the full version of the DeepHit model we encountered in Example~\ref{ex:deephit}. Because this problem setup is different from earlier sections, how we evaluate model accuracy also is a bit different (Section~\ref{sec:competing-risks-eval}).

Note that DeepHit is at this point a standard baseline to try in time-to-event prediction problems with competing risks (and even in the standard setup without competing risks). More recently, other models that support competing risks have also been developed (\eg, \citealt{nagpal2021dsm,danks2022derivative,jeanselme2023neural}). Similar to what we have seen with the standard time-to-event prediction setup, for this competing risks setup, no single model is best at this point across all datasets. We present DeepHit because it is the original deep competing risks model developed and is fairly straightforward to explain.

There are classical baselines as well although we only mention them now without explaining how they work (as understanding them is not needed for our monograph). Recall that in the standard setup of Section~\ref{chap:setup}, the Kaplan-Meier estimator would give the same population-level predicted survival function regardless of which test point we look at. The analogue in the competing risks setting is called the Aalen-Johansen estimator \citep{aalen1978empirical}. Meanwhile, the Fine-Gray subdistribution hazard model \citep{fine1999proportional} could be thought as the Cox model analogue in the competing risks setting.

\subsubsection{Statistical Framework}
\label{sec:competing-risks-framework}

We keep track of~$k$ different critical events. We still assume that we have~$n$ training points $(X_1,Y_1,\Delta_1),\dots,(X_n,Y_n,\Delta_n)$ like in the standard setup. However, the major difference now is that the event indicator $\Delta_i\in\{0,1,\dots,k\}$ takes on more possible values. When $\Delta_i=0$, then $Y_i$ is the censoring time, just as before. However, if $\Delta_i>0$, then $\Delta_i$ tells us which of the $k$ events happened earliest, and $Y_i$ is equal to the time until this earliest critical event happened.

As before, $X$ denotes the random variable for a generic raw input (with distribution $\mathbb{P}_X$), and $C$ denotes the random variable for the true (possibly unobserved) censoring time corresponding to $X$ (with distribution $\mathbb{P}_{C|X}(\cdot|x)$). However, now the random variable $T$ is a \emph{random vector} taking on values in $[0,\infty)^k$ (with distribution $\mathbb{P}_{T|X}(\cdot|x)$). In particular, $T=(T_1,T_2,\dots,T_k)$ consists of the times until each of the $k$ different critical events happen.

We assume each $(X_i,Y_i,\Delta_i)$ for $i\in[n]$ to be generated i.i.d.~as follows:
\begin{enumerate}
\item Sample raw input $X_i$ from $\mathbb{P}_X$.
\item Sample vector $T_i=(T_{i,1},T_{i,2},\dots,T_{i,k})$ (true times until the $k$ critical events happen) from $\mathbb{P}_{T|X}(\cdot|X_i)$.
\item Sample true censoring time $C_i$ from $\mathbb{P}_{C|X}(\cdot|X_i)$.
\item Set $Y_i = \min\{T_{i,1},T_{i,2},\dots,T_{i,k},C_i\}$.

If $Y_i=C_i$: set $\Delta_i=0$. Otherwise: set $\Delta_i=\arg\min_{\delta\in[k]} T_{i,\delta}$ (if there is a tie for the smallest time, then break the tie arbitrarily).
\end{enumerate}
This generative procedure allows for the times until the critical events happen to potentially depend on each other. However, conditioned on $X_i$, we still assume that $T_i$ is independent of $C_i$ just as in the standard setup (which we see since steps~2 and~3 do not depend on each other). Note that the $k$ critical events are ``exhaustive'' in the sense that they are the only options that could happen (unless none of them happen yet due to censoring), which is implied by step~4.

\subsubsection{Prediction Task}
\label{sec:competing-risks-prediction}

For all $x\in\mathcal{X}$, we assume that $\mathbb{P}_{T|X}(\cdot|x)$ exists. Consider the random variable $T_{\text{test}}:=(T_{\text{test},1},T_{\text{test},2},\dots,T_{\text{test},k})\in[0,\infty)^k$ that is sampled from $\mathbb{P}_{T|X}(\cdot|x)$. Define the random variables
\[
Y_{\text{test}} := \min\{T_{\text{test},1},T_{\text{test},2},\dots,T_{\text{test},k}\},
\quad
\Delta_{\text{test}} := \arg\min_{\delta\in[k]} T_{\text{test},\delta}.
\]
Notice that these are defined without sampling a censoring time (whereas we assumed censoring times to be generated for training data).

A common prediction target is the so-called \emph{cumulative incidence function} (CIF) \citep{gray1988class,fine1999proportional} of each event~$\delta\in[k]$, which is the probability of event~$\delta$ happening by time~$t$ (where $t\ge0$):
\begin{equation}
F_\delta(t|x) := \mathbb{P}(Y_{\text{test}} \le t, \Delta_{\text{test}} = \delta \mid X=x).
\label{eq:cif}
\end{equation}
When the number of critical events is $k=1$, then there would only be a single CIF to estimate corresponding to the single critical event, and it would actually just correspond to the CDF $F(\cdot|x)$ of the survival time distribution $\mathbb{P}_{T|X}(\cdot|x)$ in our problem setup from Section~\ref{chap:setup}.

If we discretize time using the time grid $\tau_{(1)}<\tau_{(2)}<\cdots<\tau_{(L)}$, then the CIF could be written as
\begin{equation*}
F_\delta[\ell|x] := \mathbb{P}(Y_{\text{test}} \le \tau_{(\ell)}, \Delta_{\text{test}} = \delta \mid X=x)
\quad\text{for }\delta\in[k],\ell\in[L],x\in\mathcal{X},
\end{equation*}
from which we can write its PMF version
\begin{equation}
f_\delta[\ell|x] := \mathbb{P}(Y_{\text{test}} = \tau_{(\ell)}, \Delta_{\text{test}} = \delta \mid X=x)
\quad\text{for }\delta\in[k],\ell\in[L],x\in\mathcal{X}.
\label{eq:deephit-cif-pmf}
\end{equation}
By how a CDF and PMF relate, we have $F_\delta[\ell|x] = \sum_{m=1}^\ell f_\delta[m|x]$. Meanwhile, since a PMF sums to 1, we have $\sum_{\delta=1}^k \sum_{\ell=1}^L f_\delta[\ell|x] = 1$.

\subsubsection{Likelihood}
\label{sec:competing-risks-likelihood}

For simplicity, we only present the discrete time likelihood that does not depend on the censoring distribution:
\begin{equation}
\mathcal{L}
:=\prod_{i=1}^n
    \Bigg\{
      (f_{\Delta_i}[\kappa(Y_i)|X_i])^{\ind\{\Delta_i\ne0\}}
      \bigg(
        1 - \sum_{\delta=1}^k F_\delta[\kappa(Y_i)|X_i]
      \bigg)^{\ind\{\Delta_i=0\}}
    \Bigg\},
\label{eq:competing-risks-likelihood}
\end{equation}
where, as a reminder, $\kappa(Y_i)\in[L]$ denotes the time index that $Y_i$ corresponds to. To make sense of the likelihood, note that for the $i$-th point, if it is not censored, then the contribution to the likelihood is the factor $f_{\Delta_i}[\kappa(Y_i)|X_i]$, which is the probability of event $\Delta_i$ happening at time index $\kappa(Y_i)$ for raw input $X_i$. Otherwise, if the $i$-th point is censored, the contribution to the likelihood is
\begin{align*}
1 - \sum_{\delta=1}^k F_\delta[\kappa(Y_i)|X_i]
&=1
  -
  \sum_{\delta=1}^k
    \mathbb{P}(
      Y_{\text{test}} \le \kappa(Y_i), \Delta_{\text{test}} = \delta \mid X=X_i) \\
&=1
  -
  \mathbb{P}(
    Y_{\text{test}} \le \kappa(Y_i) \mid X=X_i) \\
&=\mathbb{P}(
    Y_{\text{test}} > \kappa(Y_i) \mid X=X_i),
\end{align*}
which is the probability that the earliest critical event happens after time index $\kappa(Y_i)$ for raw input $X_i$.

\subsubsection{Example Model: the Full Version of DeepHit}
\label{sec:deephit-general}

We had previously covered a special case of the DeepHit model \citep{lee2018deephit} for when the number of critical events is $k=1$ (Example~\ref{ex:deephit}). We now present the general case that supports multiple critical events. We parameterize the PMF function (\eqref{eq:deephit-cif-pmf}) in terms of a neural network $\mathbf{f}(\cdot;\theta):\mathcal{X}\rightarrow[0,1]^{L\times k}$ as follows:
\begin{align}\!
\scalebox{.9}{
\renewcommand{\arraystretch}{.8}
$
\begin{bmatrix}
f_1[1|x] & f_2[1|x] & \cdots & f_k[1|x] \\
f_1[2|x] & f_2[2|x] & \cdots & f_k[2|x] \\
\vdots   & \vdots   & \ddots & \vdots   \\
f_1[L|x] & f_2[L|x] & \cdots & f_k[L|x]
\end{bmatrix}
$
}\!
&=
\!\!\!
\scalebox{.9}{
\renewcommand{\arraystretch}{.8}
$
\begin{bmatrix}
\mathbf{f}_{1,1}(x;\theta) & \mathbf{f}_{2,1}(x;\theta) & \cdots & \mathbf{f}_{k,1}(x;\theta) \\
\mathbf{f}_{1,2}(x;\theta) & \mathbf{f}_{2,2}(x;\theta) & \cdots & \mathbf{f}_{k,2}(x;\theta) \\
\vdots   & \vdots   & \ddots & \vdots   \\
\mathbf{f}_{1,L}(x;\theta) & \mathbf{f}_{2,L}(x;\theta) & \cdots & \mathbf{f}_{k,L}(x;\theta)
\end{bmatrix}
$
}
\nonumber\\
&=: \mathbf{f}(x;\theta).
\label{eq:deephit-model-full}
\end{align}
Having the output be a 2D table is not required (we can easily flatten the table and it would contain the same information); we have written it this way for clarity of exposition. Recalling that $\sum_{\delta=1}^k \sum_{\ell=1}^L f_\delta[\ell|x] = 1$, we require that the neural network's output always sums to~1, which we can easily get by, for instance, having the neural network output a total of $L\cdot k$ numbers that go through a softmax activation.

By plugging \eqref{eq:deephit-model-full} into \eqref{eq:competing-risks-likelihood}, we obtain
\begin{equation*}
\mathcal{L}(\theta)
:=\prod_{i=1}^n
    \Bigg\{
      (\mathbf{f}_{\Delta_i,\kappa(Y_i)}(X_i;\theta))^{\ind\{\Delta_i\ne0\}}
      \bigg(
        1
        -
        \sum_{\delta=1}^k
          \underbrace{\sum_{m=1}^{\kappa(Y_i)} \mathbf{f}_{\delta,m}(X_i;\theta)}_{F_\delta[\kappa(Y_i)|X_i]}
      \bigg)^{\ind\{\Delta_i=0\}}
    \Bigg\}.
\end{equation*}
Then the negative log likelihood loss averaged across training data is:
\begin{align*}
&\mathbf{L}_{\text{DeepHit-NLL}}(\theta) \\
&\quad:=-\frac{1}{n}\log\mathcal{L}(\theta) \\
&\quad:=-\frac{1}{n}
    \sum_{i=1}^n
      \bigg\{
        \ind\{\Delta_i\ne0\}
        \log(\mathbf{f}_{\Delta_i,\kappa(Y_i)}(X_i;\theta)) \\
&\phantom{\quad:=-\frac{1}{n}\sum_{i=1}^n\Bigg\{}
        +
        \ind\{\Delta_i=0\}
        \log
        \bigg(
          1
          -
          \sum_{\delta=1}^k
            \sum_{m=1}^{\kappa(Y_i)} \mathbf{f}_{\delta,m}(X_i;\theta)
        \bigg)
      \bigg\}.
\end{align*}
Since ranking-based accuracy metrics (Section~\ref{sec:ranking-metrics}) are popular, \citet{lee2018deephit} further introduced a ranking loss term that is motivated by the~$C^{\text{td}}$ index (Definition~\ref{def:c-td-index}). For event $\delta\in[k]$, we define the set of comparable pairs specific to event~$\delta$~as
\begin{equation}
\mathcal{E}_\delta := \{ (i,j)\in[n]\times[n] :
\Delta_i = \delta, Y_i < Y_j\}.
\label{eq:comparable-pairs-competing-risks}
\end{equation}
In particular, this means that if $(i,j)\in\mathcal{E}_\delta$, then training point~$i$ experienced critical event~$\delta$ as the earliest critical event, and training point $j$ has not experienced any critical event yet. This means that at the earlier time $Y_i$ (which corresponds to time index $\kappa(Y_i)$), the model should predict $F_\delta[\kappa(Y_i)|X_i]$ to be \emph{higher} than $F_\delta[\kappa(Y_i)|X_j]$, \ie, we want $F_\delta[\kappa(Y_i)|X_i]-F_\delta[\kappa(Y_i)|X_j]$ to be large. Note that
\begin{equation}
F_\delta[\kappa(Y_i)|X_i] - F_\delta[\kappa(Y_i)|X_j]
= \sum_{m=1}^{\kappa(Y_i)}
    (\mathbf{f}_{\delta,m}(X_i;\theta)
     -
     \mathbf{f}_{\delta,m}(X_j;\theta)).
\label{eq:deephit-want-to-maximize-this-pair}
\end{equation}
Thus, we want this difference to be large for all $(i,j)\in\mathcal{E}_\delta$, for all $\delta\in[k]$.

With the above intuition, for hyperparameters $\eta=(\eta_1,\eta_2,\dots,\eta_k)\in[0,\infty)^k$ and $\sigma>0$, we define the ranking loss term
\begin{align*}
&\mathbf{L}_{\text{DeepHit-ranking}}(\theta;\eta,\sigma) \\
&\quad:=
  \frac{1}{k}
  \sum_{\delta=1}^k
    \frac{\eta_\delta}{|\mathcal{E}_\delta|}
    \sum_{(i,i')\in\mathcal{E}_\delta}
      \exp\bigg(
        \frac{\sum_{m=1}^{\kappa(Y_i)}
                [\mathbf{f}_{\delta,m}(X_{i'};\theta)
                 -
                 \mathbf{f}_{\delta,m}(X_i;\theta)]}
             {\sigma}
      \bigg).
\end{align*}
Having each of the terms being summed as small as possible aims to maximize \eqref{eq:deephit-want-to-maximize-this-pair}. Hyperparameter $\eta_\delta\ge0$ controls how much we care about ranking for critical event $\delta$, and hyperparameter $\sigma>0$ controls how much we care about ranking across all critical events (as $\sigma\rightarrow\infty$, we stop caring about ranking).

Then the full DeepHit loss is:
\begin{equation}
\mathbf{L}_{\text{DeepHit}}(\theta)
:= \mathbf{L}_{\text{DeepHit-NLL}}(\theta) + \mathbf{L}_{\text{DeepHit-ranking}}(\theta;\eta,\sigma).
\label{eq:deephit-loss}
\end{equation}
Note that our presentation of DeepHit is slightly different from the original version by \citet{lee2018deephit} in that for each loss term we use, we have included some normalization constants (the fraction $\frac{1}{n}$ in $\mathbf{L}_{\text{DeepHit-NLL}}$, and the fractions $\frac{1}{k}$ and $\frac{1}{|\mathcal{E}_\delta|}$ in $\mathbf{L}_{\text{DeepHit-ranking}}$). Lee \emph{et al.}~also choose a specific base neural network architecture, whereas we intentionally leave it up to the user to specify.

We then use a neural network optimizer to solve $\widehat{\theta}:=\arg\min_\theta \mathbf{L}_{\text{DeepHit}}(\theta)$. For any test raw input~$x\in\mathcal{X}$, we could then predict the PMF form of the CIFs using $\mathbf{f}(x; \widehat{\theta}\hspace{1.5pt})$ (see \eqref{eq:deephit-model-full}), from which we could readily recover an estimate for all the critical events' CIFs.

In our companion code repository, we provide a Jupyter notebook that implements the full DeepHit model with competing risks.\footnote{\texttt{\url{https://github.com/georgehc/survival-intro/blob/main/S6.1.4_DeepHit_competing.ipynb}}} This notebook builds on the earlier DeepHit Jupyter notebook that we provided a link to in Example~\ref{ex:deephit}, which was for the standard right-censored survival analysis setup. In the competing risks version, instead of using the SUPPORT dataset \citep{knaus1995support}, we now use the PBC dataset \citep{fleming1991counting}, which is on predicting times until death or transplantation of various patients with primary biliary cirrhosis of the liver. Note that the PBC dataset is actually a time series dataset. However, per data point, we only consider the initial time step, so that we reduce it to a tabular dataset.

\subsubsection{Evaluation Metrics}
\label{sec:competing-risks-eval}

We point out a few evaluation metrics that are possible. Note that we state these using the continuous time version of CIFs, which could be converted into discrete time easily.

\paragraph{$\bm{C}^{\text{td}}$ index}
First, the~$C^{\text{td}}$ index (Definition~\ref{def:c-td-index}) generalizes to the competing risks setting by using the set of comparable pairs~$\mathcal{E}_\delta$ (\eqref{eq:comparable-pairs-competing-risks}) so that we now have a $C^{\text{td}}$ index score per critical event~$\delta\in[k]$.
\begin{fdefinition}[$C^{\text{td}}$ index for competing risks]
Let $\delta\in[k]$. Suppose that we have a CIF estimate $\widehat{F}_\delta(\cdot|x)$ for any $x\in\mathcal{X}$. Then using the set of comparable pairs $\mathcal{E}_\delta$ from \eqref{eq:comparable-pairs-competing-risks}, we define the~$C^{\text{td}}$ index for event~$\delta$~as
\[
C_\delta^{\text{td}}
:= \frac{1}{|\mathcal{E_\delta}|}
     \sum_{(i,j)\in\mathcal{E_\delta}}
       \ind\{ \widehat{F}_\delta(Y_i | X_i) > \widehat{F}_\delta(Y_i | X_j) \},
\]
which is between 0 and 1. Higher scores are better.
\end{fdefinition}

\paragraph{Truncated time-dependent concordance index}
We can generalize the truncated time-dependent concordance index $C_t^{\text{td}}$ (Definition~\ref{def:trunc-time-dep-c-index}) to the competing risks setting in the same manner as for the~$C^{\text{td}}$ index. We define the set of comparable pairs specific to event $\delta\in[k]$ and time $t\ge0$~as
\[
\mathcal{E}_\delta(t)
:= \{ (i,j)\in[n]\times[n] : \Delta_i=\delta, Y_i < t, Y_j > Y_i\}.
\]
We then define the following accuracy score.
\begin{fdefinition}[Truncated time-dependent concordance index for competing risks]
Let $\delta\in[k]$ and $t\ge0$. Suppose that we have a CIF estimate $\widehat{F}_\delta(\cdot|x)$ for any $x\in\mathcal{X}$. Then using the set of comparable pairs $\mathcal{E}_\delta(t)$, we define the truncated time-dependent concordance index for event~$\delta$ at time~$t$~as
\begin{equation}
C_{\delta,t}^{\text{td}}
:= \frac{\sum_{(i,j)\in\mathcal{E}_\delta(t)} w_i \ind\{ \widehat{F}_\delta(t | X_i) > \widehat{F}_\delta(t | X_j) \}}
        {\sum_{(i,j)\in\mathcal{E}_\delta(t)} w_i},
\label{eq:truncated-time-dependent-concordance-index}
\end{equation}
where
\[
w_i := \frac{1}{(\widehat{S}_{\text{censor}}(Y_i))^2}
\quad\text{for }i\in[n].
\]
Note that $\widehat{S}_{\text{censor}}(\cdot)$ is trained the same way as we described it in Section~\ref{sec:ranking-metrics}: we fit a Kaplan-Meier estimator to training labels $(Y_1, \ind\{{\Delta_1=0}\}), (Y_2, \ind\{{\Delta_2=0}\}), \dots, (Y_n, \ind\{{\Delta_n=0}\})$. Values of $C_{\delta,t}^{\text{td}}$ are between 0 and 1, where higher is better.
\end{fdefinition}
We could of course integrate $C_{\delta,t}^{\text{td}}$ over time (as done in Definition~\ref{def:integrated-truncated-time-dependent-concordance-index}) to get an integrated $C_{\delta,t}^{\text{td}}$ index for competing risks. As this is straightforward, we omit writing a formal definition. Similarly, it is straightforward to generalize the time-dependent AUC score from Section~\ref{sec:ranking-metrics} to the competing risks setting as well (recall that the time-dependent AUC score we presented was just a slight modification of $C_t^{\text{td}}$).

\paragraph{Brier score}
The Brier score (Definition~\ref{def:brier}) can also be generalized to the competing risks setting \citep[Section 5.4.2]{gerds2021medical}. Just as with the $C^{\text{td}}$ and truncated time-dependent concordance indices stated above, the competing risks version of the Brier score also is specific to a single critical event $\delta\in[k]$. As with the Brier score for the standard survival analysis setup (Section~\ref{sec:brier}), the competing risks version also depends on a specific time of evaluation $t\ge0$. We have the following.
\begin{fdefinition}[Brier score for competing risks]
Let $\delta\in[k]$ and $t\ge0$. Suppose that we have a CIF estimate $\widehat{F}_\delta(\cdot|x)$ for any $x\in\mathcal{X}$. We define the Brier score for event~$\delta$ at time~$t\ge0$ by
\begin{align*}
\text{BS}_\delta(t)
:=
  \frac{1}{n}
    \sum_{i=1}^n
      \bigg[
&       \frac{(1 - \widehat{F}_\delta(t|X_i))^2 \ind\{\Delta_i = \delta\} \ind\{ Y_i \le t\}}
             {\widehat{S}_{\text{censor}}(Y_i)} \\
&       +
        \frac{(\widehat{F}_\delta(t|X_i))^2 \ind\{\Delta_i \ne \delta\text{ and }\Delta_i\ne0\} \ind\{ Y_i \le t\}}
             {\widehat{S}_{\text{censor}}(Y_i)} \\
&       +
        \frac{(\widehat{F}_\delta(t|X_i))^2 \ind\{ Y_i > t \}}
             {\widehat{S}_{\text{censor}}(t)}
      \bigg],
\end{align*}
where $\widehat{S}_{\text{censor}}(\cdot)$ is trained the same way as in Section~\ref{sec:ranking-metrics}. Brier scores are nonnegative, where lower is better.
\end{fdefinition}
Of course, we can integrate the Brier score over time to obtain an integrated Brier score (as done in Definition~\ref{def:ibs}).

\subsection{Dynamic Time-to-Event Prediction with Competing Risks}
\label{sec:dynamic}

We now generalize our setup from Section~\ref{sec:competing-risks} to the setting where every data point is a time series. Each time series could vary in length (in terms of the number of time steps), and the time series could be irregularly sampled, meaning that the amount of time that elapses between consecutive time steps of a time series can vary. Having input data be time series can already be accommodated by the problem setup from Section~\ref{sec:competing-risks}! For example, using the DeepHit model, we could simply set $\mathbf{f}(\cdot;\theta)$ (\eqref{eq:deephit-model-full}) to be a recurrent neural network (RNN), which accepts variable-length time series as inputs. Consequently, training and test data could be variable-length time series.

Thus, we reuse the exact same statistical framework for the training data as in Section~\ref{sec:competing-risks-framework}. However, now we set the raw input space~$\mathcal{X}$ in a particular manner, as we describe in Section~\ref{sec:dynamic-training-data}. The main conceptual difference will be that we phrase the prediction task so that it depends on how much of a test time series we see. We state this new prediction task in Section~\ref{sec:dynamic-prediction}. We then show how DeepHit can be used for this time series prediction task by choosing $\mathbf{f}(\cdot;\theta)$ based on an RNN with an attention module in Section~\ref{sec:dynamic-deephit}. By introducing an RNN, during model training, we add an extra loss term that is meant to help the RNN learn a useful latent representation of time series (\ie, we use the RNN to learn how the time series evolves over time). The resulting model is called \mbox{Dynamic-DeepHit} \citep{lee2019dynamic}. We comment on handling sequences of critical events occurring in Section~\ref{sec:recurrent-critical-events}.

A key takeaway in our exposition is that the time series version of the competing risks problem can just be reduced to the standard competing risks problem of Section~\ref{sec:competing-risks}. How we go from DeepHit to \mbox{Dynamic-DeepHit} is just in how we specify the base neural network~$\mathbf{f}(\cdot;\theta)$ of DeepHit, and the addition of a loss term that learns temporal dynamics. These design steps that enable us to work with variable-length time series as inputs is \emph{not} special to the competing risks setup and works also when the number of critical events is $k=1$. In other words, the high-level idea of how we go from DeepHit to \mbox{Dynamic-DeepHit} can be applied to other deep time-to-event prediction models (such as the ones we covered in Sections~\ref{chap:setup} to~\ref{chap:ode}) to enable them to work with variable-length time series as inputs.

\subsubsection{Variable-length Time Series as Training Data}
\label{sec:dynamic-training-data}

We denote training point $i$'s raw input as
\[
X_i := \Big( \underbrace{(U_i^{(1)},V_i^{(1)})}_{\text{time step 1}},\underbrace{(U_i^{(2)},V_i^{(2)})}_{\text{time step 2}},\dots,\underbrace{(U_i^{(M_i)},V_i^{(M_i)})}_{\text{time step }M_i} \Big),
\]
where $M_i\in\{1,2,\dots\}$ is the number of time steps for point~$i$, and at time step $m\in[M_i]$ (sorted chronologically), $U_i^{(m)}\in\mathcal{\mathcal{U}}$ is the raw input ($\mathcal{U}$ is the raw input space for a single time step and is an input space that standard neural network software can work with), and $V_i^{(m)}\in\mathbb{R}$ is the timestamp. For example, the amount of time between time steps $m$ and $m+1$ is $V_i^{(m+1)} - V_i^{(m)}$. A common assumption is that $V_i^{(1)}:=0$. The raw input space $\mathcal{X}$ consists of all possible time series of the format above. As our notation suggests, different training points~$i$ can have different numbers of time steps~$M_i$. Again, RNNs readily accommodate this sort of time series data that can vary in length. (Transformer models could also be used to accept variable-length time series as inputs.)

In terms of ground truth information, just as in the Section~\ref{sec:competing-risks-framework}, training point~$i$ has an event indicator $\Delta_i\in\{0,1,\dots,k\}$. If $\Delta_i=0$, then the observed time $Y_i$ is a censoring time. Otherwise, $\Delta_i$ is equal to the critical event that happened earliest to point~$i$, and $Y_i$ is the time when this critical event happened. We assume that $Y_i$ starts measuring time starting from the last observed timestamp $V_i^{(M_i)}$. This means that at any time step $m\in[M_i]$, we are at timestamp $V_i^{(m)}$, and the time until the earliest critical event or censoring happens is $Y_i + (V_i^{(M_i)} - V_i^{(m)})$.

Even though this way of specifying the training data is a special case of the framework in Section~\ref{sec:competing-risks-framework}, the notation we introduced here will be important when we talk about prediction next.

\subsubsection{Prediction Task}
\label{sec:dynamic-prediction}

We now write any test raw input $x$ as
\begin{equation*}
x = \Big( (u^{(1)},z^{(1)}),(u^{(2)},z^{(2)}),\dots \Big)
\end{equation*}
using the same format as the training data except where we do not pre-specify a last time step. We denote $x$ truncated to only include its initial $m$ time steps as
\begin{equation*}
x^{(\le m)}
:= \Big( (u^{(1)},v^{(1)}),(u^{(2)},v^{(2)}),\dots,(u^{(m)},v^{(m)}) \Big).
\end{equation*}
As time progresses, we could see more of $x$, similar to what would happen in some real applications (such as a patient in a hospital intensive care unit continuously getting new measurements taken over time).

For any $m\in\{1,2,\dots\}$, just as in the prediction setup from Section~\ref{sec:competing-risks-prediction}, we sample nonnegative durations $T_{\text{test}}:=(T_{\text{test},1},T_{\text{test},2},\dots,T_{\text{test},k})$ from $\mathbb{P}_{T|X}(\cdot|x^{(\le m)})$, and we again define
\[
Y_{\text{test}} := \min\{T_{\text{test},1},T_{\text{test},2},\dots,T_{\text{test},k}\},
\quad
\Delta_{\text{test}} := \arg\min_{\delta\in[k]} T_{\text{test},\delta}.
\]
Then we aim to predict the following dynamic version of the CIF (\eqref{eq:cif}) that depends on the number of time steps revealed~$m$:
\begin{align*}
F_\delta(t|x,m)
:= \mathbb{P}(Y_{\text{test}} \le v^{(m)} + t, \Delta_{\text{test}} = \delta \mid X = x^{(\le m)}, Y_{\text{test}} > v^{(m)}) \nonumber\\
\text{for }\delta\in[k],m\in\{1,2,\dots\},t\ge0,x\in\mathcal{X}.
\end{align*}
This is the probability that the earliest critical event that happens is~$\delta$, and it happens within time duration $t$, starting from timestamp $v^{(m)}$ (the timestamp of the $m$-th time step). Note that the idea that we are starting the prediction from timestamp $v^{(m)}$ (so that this time step is viewed as the ``origin'' of the time-to-event prediction model) is not actually a limitation of this setup.\footnote{It is important to keep in mind that how we have set up the problem, the neural network at any given time step actually knows how much time has elapsed for a data point (which is a time series). Specifically, suppose that we have observed $m$ time steps so far of test raw input $x$, meaning that we have observed
\[
x^{(\le m)}
= \Big( (u^{(1)},v^{(1)}),\dots,(u^{(m)},v^{(m)}) \Big).
\]
The neural net is being asked to make a prediction starting at timestamp $v^{(m)}$, treating timestamp $v^{(m)}$ as the ``origin''. Note that the neural net also has access to $v^{(1)}$ as well. Thus, if desired, the modeler could specify the base neural network so that it explicitly depends on the difference $v^{(m)} - v^{(1)}$, meaning that the neural network knows, as one of its inputs, how much time has elapsed for the current time series since we first started observing it. This idea could be used to set up the base neural network so that it instead views timestamp $v^{(1)}$ as the ``origin'' rather than~$v^{(m)}$.}

The version of the CIF where we discretize duration $t$ to only take on values along the grid $\tau_{(1)}<\tau_{(2)}<\cdots<\tau_{(L)}$ is given by
\begin{align}
F_\delta[\ell|x,m]
:= \mathbb{P}(Y_{\text{test}} \le v^{(m)} + \tau_{(\ell)}, \Delta_{\text{test}} = \delta \mid X = x^{(\le m)}, Y_{\text{test}} > v^{(m)}) \nonumber\\
\text{for }\delta\in[k],m\in\{1,2,\dots\},\ell\in[L],x\in\mathcal{X}.
\label{eq:dynamic-cif-discrete-duration}
\end{align}
Importantly, we do not have to discretize the timestamps $v^{(1)},v^{(2)},\dots$ Even if we do discretize timestamps, how timestamps are discretized does not have to be the same way as how duration $t$ is discretized. When timestamps are not discretized, the PMF version of \eqref{eq:dynamic-cif-discrete-duration} is
\begin{align}
&f_\delta[\ell|x,m] \nonumber\\
&:= \mathbb{P}\Big(Y_{\text{test}} - v^{(m)} \in (\tau_{(\ell-1)}, \tau_{(\ell)}], \Delta_{\text{test}} = \delta ~\Big|~ X = x^{(\le m)}, Y_{\text{test}} > v^{(m)}\Big) \nonumber\\
&\pushright{\text{for }\delta\in[k],m\in\{1,2,\dots\},\ell\in[L],x\in\mathcal{X},} \nonumber\\
\label{eq:dynamic-cif-discrete-duration-pmf}
\end{align}
where $\tau_{(0)}:=0$. We will use this PMF in a moment.

\paragraph{Evaluation metrics}
For this prediction task, the same evaluation metrics as in the standard competing risks setting could be used, although there would be a question of whether we care about reporting evaluation metrics as a function of how much of a test time series we see. For example, for test data, we could just predict starting from each of their final time steps, in which case using existing competing risks evaluation metrics would be straightforward. However, we could instead report evaluation metrics using, for instance, only up to the first hour of test time series. Then we report evaluation metrics using only up to the second hour of test time series, \etc. For a concrete example of this, see Table~1 of~\citet{shen2023neurological}. Note that here we used timestamp thresholds (\eg, 1 hour, 2 hours) rather than time step thresholds (\eg, 1 time step, 2 time steps) in case time steps are highly irregularly sampled across data points.

Overall, from what we can tell, how researchers evaluate models in this dynamic setting has still not become as standardized as in the regular competing risks setting and certainly not as standardized as in the standard right-censored time-to-event prediction setting of Section~\ref{chap:setup}. For a recently proposed ``dynamic c-index'', see the paper by \citet{putzel2021dynamic}.

\paragraph{The standard competing risks setting as a special case}
If all time series in raw input space~$\mathcal{X}$ are restricted to only have one time step, then the entire problem setup would be the same as the standard competing risks setup of Section~\ref{sec:competing-risks}.

\subsubsection{Example Model: \mbox{Dynamic-DeepHit}}
\label{sec:dynamic-deephit}

As we pointed out earlier, DeepHit can already work for this new problem setup provided that we set the neural network $\mathbf{f}(\cdot;\theta)$ in \eqref{eq:deephit-model-full} to accept variable-length time series as inputs (in fact, the training procedure can stay the same although we will add another loss term). We provide details on how to specify $\mathbf{f}(\cdot;\theta)$ shortly. The output of $\mathbf{f}(\cdot;\theta)$ given any time series input is going to still be $L\cdot k$ numbers. We explain how to interpret these numbers first using our new time series notation. In other words, we explain what $\mathbf{f}(\cdot;\theta)$ is predicting, as this will be helpful in explaining the model architecture and training.

\paragraph{Prediction} Given time series $x^{(\le m)}$, the $\ell$-th row, $\delta$-th column of $\mathbf{f}(x^{(\le m)};\theta)\in[0,1]^{L\times k}$ is used to model $f_\delta[\ell|x,m]$ (the PMF in \eqref{eq:dynamic-cif-discrete-duration-pmf}). In other words, DeepHit uses $x^{(\le m)}$ to predict CIFs for the different critical events starting from timestamp $v^{(m)}$. In more detail, the CIFs are giving probabilities of the $k$ critical events happening within time duration $t\in\{\tau_{(1)},\tau_{(2)},\dots,\tau_{(L)}\}$ starting from timestamp~$v^{(m)}$.

\paragraph{How to specify the neural network $\mathbf{f}(\cdot;\theta)$} We give an overview of how \mbox{Dynamic-DeepHit} \citep{lee2019dynamic} specifies $\mathbf{f}(\cdot;\theta)$, deferring details to the original paper. Our exposition will be slightly more general than Lee \emph{et al.}~as we aim to convey the key high-level ideas. For example, Lee \emph{et al.}~explicitly keep track of a missingness vector per time step (that indicates which features are missing) whereas we do not include this (the missingness vector could just be included as part of the raw input space $\mathcal{U}$ for a single time step).

Given time series $x^{(\le m)} = \big( (u^{(1)},v^{(1)}),\dots,(u^{(m)},v^{(m)}) \big)$, \mbox{Dynamic-DeepHit} sets $\mathbf{f}(\cdot;\theta)$ to do the following (see accompanying Figure~\ref{fig:ddh-model}):
\begin{enumerate}

\item We first feed the input time series $x^{(\le m - 1)}$ (we exclude the last time step) into a user-specified RNN (with $d_{\text{hidden}}$ output features per time step, for a user-specified number of dimensions $d_{\text{hidden}}$), where we slightly transform what the input looks like per time step. Specifically at time step $p\in[m-1]$ (again, the last time step is excluded), the input to the RNN is taken to be $(u^{(p)}, v^{(p+1)}-v^{(p)})$, \ie, we also supply a time duration to get to the next time step. The last time step's input $u^{(m)}$ is not used with the RNN, but will be used later on. The RNN's output at time step $p\in[m-1]$ is denoted as $\widetilde{u}^{(p)}\in\mathbb{R}^{d_{\text{hidden}}}$. This first step is shown on the left side of Figure~\ref{fig:ddh-model}.

Note that depending on the format of a single time step's input space $\mathcal{U}$, some additional neural network components may be needed (our diagram and also the original \mbox{Dynamic-DeepHit} treat $\mathcal{U}$ as tabular data, but this need not be the case). For example, if $\mathcal{U}$ corresponds to images of a fixed shape, then prior to feeding each~$u^{(p)}$ into an RNN cell, we could first apply a convolutional neural network or a vision transformer to convert $u^{(p)}$ into a fixed-length feature vector representation that then goes into an RNN~cell.

\begin{figure}[t]
\centering
\includegraphics[width=\linewidth]{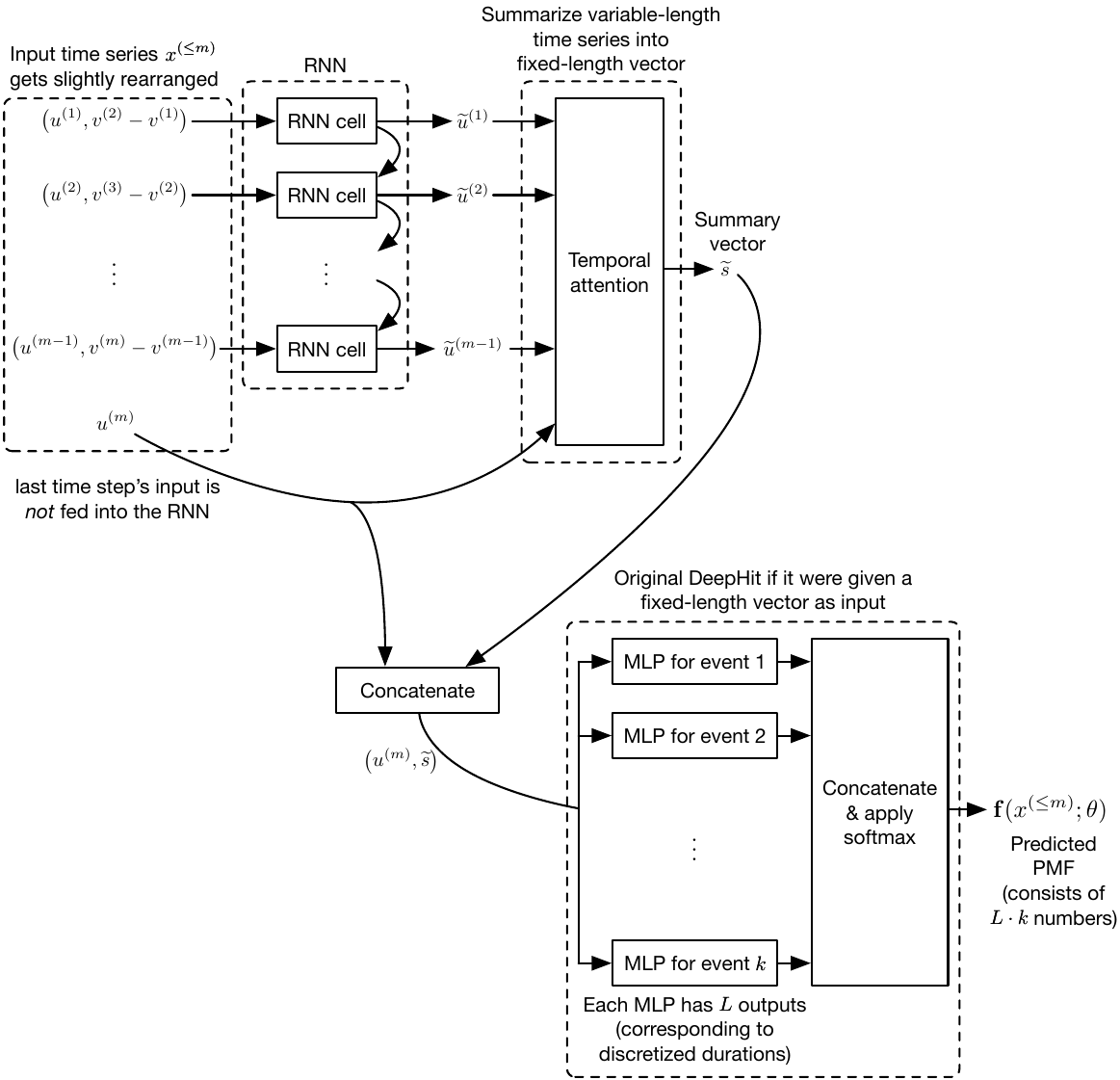}
\vspace{-1em}
\caption{\mbox{Dynamic-DeepHit} neural network architecture.}
\label{fig:ddh-model}
\end{figure}

\item Next, given the RNN outputs $\widetilde{u}^{(1)},\dots,\widetilde{u}^{(m-1)}\in\mathbb{R}^{d_{\text{hidden}}}$ as well as the raw input $u^{(m)}\in\mathcal{U}$, we summarize all this information into a fixed-length summary vector $\widetilde{s}\in\mathbb{R}^{d_{\text{hidden}}}$. To do this, we first let $\mathbf{f}_{\text{attention}}(\cdot;\theta):\mathbb{R}^{d_{\text{hidden}}}\times\mathcal{U}\rightarrow\mathbb{R}$ be a user-specified feed-forward neural network, such as a multilayer perceptron (MLP). Note that for $p\in[m-1]$, the output value $\mathbf{f}_{\text{attention}}\big( (\widetilde{u}^{(p)},u^{(m)});\theta \big)$ is a single number. Then we set the summary vector to be $\widetilde{s} = \sum_{p=1}^{m-1} a_p \widetilde{u}^{(p)}$, where the weights $a_1,a_2,\dots,a_{m-1}$ are given by
\[
\begin{bmatrix}
a_1 \\
a_2 \\
\vdots \\
a_{m-1}
\end{bmatrix}
:= \text{softmax}\!
   \left(
   \begin{bmatrix}
   \mathbf{f}_{\text{attention}}\big((\widetilde{u}^{(1)}, u^{(m)});\theta\big) \\
   \mathbf{f}_{\text{attention}}\big((\widetilde{u}^{(2)}, u^{(m)});\theta\big) \\
   \vdots \\
   \mathbf{f}_{\text{attention}}\big((\widetilde{u}^{(m-1)}, u^{(m)});\theta\big)
\end{bmatrix}
   \right) \in [0,1]^{m-1}.
\]
This step is labeled as the ``temporal attention'' block in the middle of Figure~\ref{fig:ddh-model}.

\item We then combine the last time step's input $u^{(m)}$ with the summary vector $\widetilde{s}$ outputted by the temporal attention block to obtain the concatenated vector $(u^{(m)}, \widetilde{s})$.

\item Lastly, we treat the concatenated vector $(u^{(m)}, \widetilde{s})$ as the input to $k$ different MLPs (one per critical event) that each outputs $L$ numbers (corresponding to different time durations $\tau_{(1)},\tau_{(2)},\dots,\tau_{(L)}$), and the overall output (across the $k$ MLPs) is concatenated and passed through a softmax layer to produce the final neural network output $\mathbf{f}(x^{(\le m)};\theta)$ (the softmax enforces the constraint that the PMF sums to~1). This fourth step is shown on the bottom right of Figure~\ref{fig:ddh-model}. The final output does not have to be reshaped to be $L$-by-$k$ as we had already pointed out when we first stated \eqref{eq:deephit-model-full}.

Note that this last step uses the neural network architecture from the original DeepHit paper \citep{lee2018deephit} that is meant for handling raw inputs that are fixed-length feature vectors (in how we presented DeepHit in Section~\ref{sec:deephit-general}, we intentionally stated it in a more general fashion without assuming raw inputs must be fixed-length vectors).
\end{enumerate}
There is one last neural network component that is not shown in Figure~\ref{fig:ddh-model} as it is not used to compute the output value $\mathbf{f}(x^{(\le m)};\theta)$. In particular, \mbox{Dynamic-DeepHit} also requires that at time step $p\in[m-1]$, the RNN on the left side of Figure~\ref{fig:ddh-model} can output an estimate $\widehat{u}^{(p+1)}$ of the next time step's raw input $u^{(p+1)}$. There are different ways to achieve this. For example:
\begin{itemize}

\item If $\mathcal{U}=\mathbb{R}^d$, then we can choose the RNN in Figure~\ref{fig:ddh-model} to be a type of RNN that already distinguishes between hidden state vectors and output state vectors (such as LSTMs \citep{hochreiter1997long}), in which case we let the hidden state vectors be what we denoted as the $\widetilde{u}^{(p)}$ variables, and we use the output state vectors to predict the next steps' feature vectors (we would set the output state vector to consist of $d$ entries). Thus, we could just denote these output state vectors as $\widehat{u}^{(2)},\widehat{u}^{(3)},\dots,\widehat{u}^{(m)}\in\mathbb{R}^d$.

\item An alternative strategy that also works if $\mathcal{U}$ is not necessarily~$\mathbb{R}^d$ is that we can feed $\widetilde{u}^{(p)}$, along with the time duration to get to the next time step (\ie, $v^{(p+1)}-v^{(p)}$), into a user-specified feed-forward network $\mathbf{f}_{\text{next-time-step}}(\cdot;\theta):\mathbb{R}^{d_{\text{hidden}}}\times[0,\infty)\rightarrow\mathcal{U}$ to produce the estimate $\widehat{u}^{(p+1)}$, \ie,
\[
\widehat{u}^{(p+1)}
:= \mathbf{f}_{\text{next-time-step}}\big( (\widetilde{u}^{(p)}, v^{(p+1)}-v^{(p)}); \theta\big)
\quad\text{for }p\in[m-1].
\]

\end{itemize}
In both of these cases, we would be able to come up with estimate $\widehat{u}^{(p)}$ of $u^{(p)}$ for each $p\in\{2,3,\dots,m\}$.

To summarize, the overall neural network $\mathbf{f}(\cdot;\theta)$ consists of an RNN, an attention network $\mathbf{f}_{\text{attention}}(\cdot;\theta)$, and MLPs for each critical event. Moreover, at the RNN stage of the neural network, we have a neural network component that predicts the next time step's raw input as described above. Note that the variable $\theta$ contains the parameters of all the neural network components involved.

\paragraph{Model training}
Dynamic-DeepHit reuses the same training loss (\eqref{eq:deephit-loss}) as regular DeepHit but adds another loss term that measures how accurate each $\widehat{u}^{(p)}$ predicts $u^{(p)}$ for $p\in\{2,3,\dots,m\}$. This new loss is
\[
\mathbf{L}_{\text{next-time-step}}(\theta)
:= \frac{1}{n}\cdot\frac{1}{m-1} \sum_{i=1}^n \sum_{p=2}^m \zeta\big( \widehat{u}^{(p)}(\theta), u^{(p)}\big),
\]
where we now emphasize that $\widehat{u}^{(p)}$ depends on the parameter variable~$\theta$, and we have a user-specified error function $\zeta:\mathcal{U}\times\mathcal{U}\rightarrow[0,\infty)$. For example, if $\mathcal{U}=\mathbb{R}^d$, then we could use squared Euclidean distance $\zeta(u,u'):=\|u-u'\|^2$. The overall training loss is thus
\begin{align*}
&\mathbf{L}_{\text{Dynamic-DeepHit}}(\theta) \\
&\quad=\mathbf{L}_{\text{DeepHit-NLL}}(\theta) + \mathbf{L}_{\text{DeepHit-ranking}}(\theta;\eta,\sigma) + \gamma \mathbf{L}_{\text{next-time-step}}(\theta),
\end{align*}
where $\gamma\ge0$ is a hyperparameter for how much to weight the new loss (as a reminder, the ranking loss already has a hyperparameter $\eta=(\eta_1,\dots,\eta_k)$ that weights the ranking loss contributions of the different critical events). Note that for the negative log likelihood component of the loss, per training time series $X_i$, we consider prediction only starting at the final time step (\ie, for each training point $i\in[n]$, we predict starting at time $V_i^{(M_i)}$, where the time until the earliest critical event happens is $Y_i$).

We provide a Jupyter notebook that implements Dynamic-DeepHit in our companion code repository.\footnote{\texttt{\url{https://github.com/georgehc/survival-intro/blob/main/S6.2.3_Dynamic-DeepHit.ipynb}}} This notebook builds on the DeepHit Jupyter notebook that we provided a link for in Section~\ref{sec:deephit-general}. We again use the PBC dataset but now treat the data as variable-length time series rather than first converting the dataset to be tabular.

\paragraph{Practical considerations}
After model training, the goal is to use the model to repeatedly make predictions \emph{as we see more and more of an individual data point's time series} (as a reminder, it is helpful to have in mind a real-time application where data keep streaming in, and we keep updating our predictions). During training, it is best to try to mimic the same prediction setup as what we would encounter during model testing. In other words, for the $i$-th training point, which is a time series of length $M_i$, it can be helpful to actually view this individual training point as $M_i$ different ``augmented'' training points (we state each of these with its corresponding prediction task):
\begin{itemize}
\item Given only the first time step of this training time series (so that we observe $(U_i^{(1)},V_i^{(1)})$), predict the CIFs starting at timestamp $V_i^{(1)}$ (\ie, \eqref{eq:dynamic-cif-discrete-duration}), where we treat the ground truth observed time as $Y_i + (V_i^{(M_i)} - V_i^{(1)})$ and the ground truth event indicator as $\Delta_i$.
\item Given the first two time steps of this training time series (so that we observe $(U_i^{(1)},V_i^{(1)}),(U_i^{(2)},V_i^{(2)})$), predict the CIFs starting at timestamp $V_i^{(2)}$, where we treat the ground truth observed time as $Y_i + (V_i^{(M_i)} - V_i^{(2)})$ and the ground truth event indicator as $\Delta_i$.
\item $\dots$
\item Given all $M_i$ observed time steps (so that we observe $(U_i^{(1)},V_i^{(1)}),\dots,(U_i^{(M_i)},V_i^{(M_i)})$), predict the CIFs starting at timestamp $V_i^{(M_i)}$, where we treat the ground truth observed time as $Y_i + (V_i^{(M_i)} - V_i^{(M_i)}) = Y_i$ and the ground truth event indicator as $\Delta_i$. (Note that this actually corresponds to the original $i$-th training point.)
\end{itemize}
We refer to these as augmented training points just to emphasize that these are actually constructed based on an original (or ``non-augmented'') training point (which is in fact just the very last augmented training point listed above). Note that as an alternative to using all $M_i$ of these augmented training points, we could randomly choose one of them (\eg, per neural network training epoch, we randomly choose a different one of the $M_i$ augmented training points above).

Very importantly, we point out that the above augmentation strategy could be essential in practice. If one does \emph{not} use the augmentation strategy and only uses the original training time series (so that for the $i$-th training point, we only use the last augmented training point listed above), then it is possible that the training procedure could, in some sense, be fooled into learning a useless model. A fundamental problem here is one of sampling bias in how the training data are collected.

As an extreme example of sampling bias, suppose that the training time series were collected in a manner where the very last time step always contains a magical feature that deterministically says what critical event happens within the next hour for the data point (but this feature is missing or not helpful at all preceding time steps). If the model is only ever trained in a manner where we always saw this very last time step for every training time series, then the model could learn to just rely on the magical feature (only at the last time step) and nothing else. \emph{However, this model would not have been trained appropriately as to mimic making predictions for a time series as we see more of it!} Instead, if during training, the model was forced to realize that we are predicting starting from a time step that is not necessarily the last one so that the magical feature is not always helpful (note that we could ensure this by using the above augmentation strategy), then the model would be encouraged to learn to predict well even at time steps prior to the last one per training time series. Of course, if the training data were collected in a manner where the length of each training time series is random and independent of the values of the features (or raw inputs) collected over time and also independent of the survival and censoring outcomes, then we would not need to worry about this sort of sampling~bias.

\subsubsection{Sequences of Critical Events}
\label{sec:recurrent-critical-events}

The dynamic setup could readily be extended to the setting where we model a whole sequence of critical events, some of which could happen multiple times. Of course, a critical event such as death would be ``terminal'' so that upon encountering it, there would be no future critical events to consider. Non-terminal critical events (\eg, getting admitted to a hospital) could occur multiple times though. We would simply use the same modeling strategy where the difference in interpretation is that after the earliest of $k$ critical events happens, we can continue to make predictions! We would just be predicting the time until the \emph{next} critical event happens (among the~$k$ options possible). In terms of our raw inputs, we could add history information so that we keep track of what critical events have happened so far and when.

In this extension, we could think of any time series as ``transitioning'' between the $k$ critical events over time, so that the critical events are \emph{states} that a time series is moving between. This setup has been studied extensively under the name of \emph{marked temporal point processes} (\eg, \citealt{daley2003introduction,daley2008introduction,du2016recurrent}), sometimes written as just ``marked point processes''. Note that there is work in this area both with and, separately, without censoring. At this point, common modeling strategies include using attention mechanisms (\eg, \citealt{zhang2020self,zuo2020transformer}) and recurrent neural networks (\eg, \citealt{shchur2020intensity}) to summarize history information. Hawkes processes \citep{hawkes1971spectra} are frequently used to model a continuous time hazard function for when a critical event will occur next. See also the neural ODE multi-state time-to-event model by \citet{groha2020general}. The SurvLatent ODE model \citep{moon2022survlatent}, which handles the dynamic competing risks setting much like Dynamic-DeepHit but with a neural ODE, could also be adapted to this setting of reasoning about a sequence of critical events.

\section{Discussion}
\label{chap:discussion}

A key goal of this monograph has been to introduce time-to-event prediction and survival analysis concepts and how they have been used with modern deep learning tools. We saw some key ideas for deriving deep survival models:
\begin{itemize}

\item We routinely converted between hazard, cumulative hazard, and survival functions. This amounted to using Summaries~\ref{sum:conversions} (for continuous time) and~\ref{sum:conversions-discrete} (for discrete time).

\item Every model we covered in detail involves deriving a likelihood expression that we then turned into a negative log likelihood loss. Depending on the model, additional loss terms may be present.

\item When we work with a ``nonparametric'' function over time, this meant that we discretized the time grid using unique times of death, and then we set the function values at the different time grid points to be parameters. \emph{As an important practical remark: we could intentionally preprocess observed times to, for instance, discrete them into a coarse grid, even before we use a method like the Kaplan-Meier estimator. In other words, even when working with the Kaplan-Meier estimator, we could, if we wanted to, use a time grid that is not the raw unique times of death.}

\item For discrete time models, the base neural network is often set so that there is an output number corresponding to each time step.

\item Many evaluation metrics are ranking-based, with the idea that we can unambiguously rank many (usually not all) pairs of data points even when there is censoring. A key idea is that when the $i$-th point dies ($\Delta_i=1$) and has an observed time before the $j$-th point (so $Y_i\le Y_j$), then we would like the time-to-event prediction model to consider the $i$-th point worse off (at time~$Y_i$) than the $j$-th point (also at time~$Y_i$). The Cox model's loss function relates to ranking. DeepHit has a loss function that's directly about ranking. Including a ranking loss during model training is meant to help with ranking-based accuracy metrics.
\end{itemize}
We now also state some points that could be helpful in practice:
\begin{itemize}
\item For any discrete time model (\eg, DeepHit, Nnet-survival, DKSA, survival kernets), we could intentionally set the time grid to be the unique times of death or, in general, based on the training data. There is no requirement that only nonparametric models are allowed to use time grids defined by the unique times of death. For instance, \citet{chen2024survival} found that sometimes using DeepHit with the time grid set to be all unique times of death resulted in time-dependent concordance indices much higher than what had previously been reported in literature.

\item The use of forward filling interpolation shows up a number of times to convert a discrete time model to continuous time. In practice, we typically would not actually recommend using forward filling interpolation. Constant density or constant hazard interpolation could be used instead (among other interpolation strategies possible).

\item In real applications, it is important to figure out what sort of time resolution one really needs. A major selling point of time-to-event prediction models is their ability to reason about a potentially large window of time (\eg, if we really need to be precise about time and want it to be continuous, then we could use a neural ODE model like SODEN, whereas if we do not need to be too precise and discretizing time is okay, then a discrete time model could suffice).
On the other hand, if one only needs to worry about very few time steps (\eg, even if using a discrete time model, $L$ is small), then there could be less of a benefit to using a full-fledged time-to-event prediction model vs just using survival stacking with an existing probabilistic binary classifier.

\item When choosing a base neural network, we think it is helpful thinking about what happens when parameters of the base neural network go to~0, which we could encourage by regularizing neural network parameters (\eg, by explicitly introducing a loss term or using weight decay). As a concrete example of this, suppose that we are training a DeepSurv model, and we set up a base neural network so that if all its parameters are 0, then the base neural network outputs zero. Then as we pointed out in Remark~\ref{rem:nelson-aalen-special-case-of-cox}, the predicted survival function would approximate the Kaplan-Meier estimator. Thus, we have some intuition for how regularizing neural network parameters or using weight decay would impact the resulting learned model.

\item Working with variable-length time series as inputs has really become easier thanks to recurrent neural networks as well as attention models. These show up in models like Dynamic-DeepHit and SurvLatent ODE. Note that while we did not explicitly cover transformers, one could easily use transformers instead of recurrent neural networks to handle variable-length time series.

\end{itemize}
We discuss a few topics that we either only glossed over earlier in the monograph, or we simply did not cover it at all. We first go over some textbook results of variants of the basic time-to-event prediction setup that we did not yet cover. Afterward, we cover more contemporary topics that are active areas of research.

\subsection{More Variants of the Basic Time-to-Event Prediction Setup: Left and Interval Censoring, Truncation, and Cure Models}
\label{sec:truncation-other-censoring-cure}

We now point out some standard textbook variants of the basic setup in Section~\ref{chap:setup}. As will be apparent, the modifications needed to handle these variants are not difficult and do not dramatically change how we would derive time-to-event prediction models, which is why we have delayed their presentation until now.

We first talk about more kinds of censoring and also a concept called \emph{truncation} (these could, for instance, be found in Chapter~3 of the textbook by \cite{klein2003survival}). We then talk about how to address an issue that could show up in many applications where some data points will actually never experience the critical event and are thus ``cured'' from experiencing the critical event (a book on cure models is provided by \citet{peng2021cure}). For example, in predicting the time until convicted criminals reoffend, it could be that many of them will never reoffend.

\paragraph{Left and interval censoring}
Thus far in the monograph, we have focused on the \emph{right-censored} setup: for training data, when event indicator $\Delta_i=1$, then it means that the observed time $Y_i$ is the true survival time. If instead $\Delta_i=0$, then $Y_i$ is the censoring time, where the true survival time is after $Y_i$. For this setup, we use the likelihood function from \eqref{eq:likelihood-pdf}, which we reproduce below:
\begin{align*}
\mathcal{L}
&:=
  \prod_{i=1}^n
    \big\{f(Y_i|X_i)^{\Delta_i}
          S(Y_i|X_i)^{1-\Delta_i}\big\} \\
&\phantom{:}=
  \prod_{i=1}^n
    \bigg\{
      f(Y_i|X_i)^{\Delta_i}
      \bigg[
        \int_{Y_i}^\infty f(u|X_i)\textrm{d}u
      \bigg]^{1-\Delta_i}
    \bigg\}.
\end{align*}
As a reminder, the likelihood has a simple interpretation: when we observe the $i$-th point's true survival time, then the $i$-th point's contribution to the likelihood is $f(Y_i|X_i)$. When the $i$-th point is right-censored so that the true survival time is after $Y_i$, the point's contribution to the likelihood is
\[
\int_{Y_i}^\infty f(u|X_i)\textrm{d}u
= \mathbb{P}(T > Y_i \mid X = X_i).
\]
Instead, if the $i$-th point is \emph{left-censored}, then this means that the true survival time is \emph{before} the observed time $Y_i$ (instead of \emph{after} as in the right-censored case). As an example of left censoring, suppose that a data point corresponds to a patient and the critical event of interest is the time when a patient gets a disease. If a patient tests positive for a disease at time~$Y_i$, then the actual time $T_i$ when the patient got the disease would be at most $Y_i$, so that $Y_i$ is a censoring time. When the \mbox{$i$-th} point is left-censored, the standard approach is to set its contribution to the likelihood to be
\[
\int_0^{Y_i} f(u|X_i)\textrm{d}u
= \mathbb{P}(T < Y_i \mid X = X_i).
\]
A third possibility is that the $i$-th point is \emph{interval-censored}: continuing off this disease testing example, suppose now that a patient takes the test once at time~$Y_i^{(1)}$, when the test turns up negative for the disease. Then some time later, the patient takes the test a second time at time~$Y_i^{(2)}$, and this second test turns up positive. In this situation, we do not observe the exact time the patient got the disease, but we know that it is within the interval $[Y_i^{(1)}, Y_i^{(2)}]$. This is interval censoring. Importantly, when the $i$-th point is interval-censored, we assume that we observe two times $Y_i^{(1)}$ and $Y_i^{(2)}$ (with $Y_i^{(1)} < Y_i^{(2)}$) rather than only a single time, and the standard approach is to set the contribution of this point to the likelihood to be
\[
\int_{Y_i^{(1)}}^{Y_i^{(2)}} f(u|X_i)\textrm{d}u
= \mathbb{P}\big(T \in [Y_i^{(1)}, Y_i^{(2)}] ~\big|~ X = X_i\big).
\]
It is possible to have a setup where different points experience different censoring types, so that we observe $\Delta_i\in\{\text{not censored},\text{right-censored},\text{left-censored},\text{interval-censored}\}$. Depending on the value of $\Delta_i$, we switch between the different possible contributions to the likelihood that we mentioned above.

\paragraph{Truncation}
Let's return to the right-censored setup (so $\Delta_i=1$ means that we observe the true survival time, and $\Delta_i=0$ means that we observe the censoring time) and consider a different issue that may arise. Suppose that we want to model survival times of people in a city (per person, time~0 would correspond to when they were born). To help with this task, we collect data on people from a retirement home in the city. Suppose that we can get their dates of birth, when they entered the retirement home, when they either died or were last checked up on (corresponding to the censoring event), and other features of these people (that we treat as the raw input per person). In terms of notation, for the $i$-th person, let's denote $A_i$ to be the person's age at the time of entering the retirement home, and $Y_i$ to be the age of the person at either the time of death or the time of censoring (whichever happens earlier, as in the usual right-censored setup). Thus, $Y_i>A_i$.

The issue here is that people could enter the retirement home at different ages, and they must have survived up to that point to even show up in the data. People who died earlier and never had the opportunity to enter the retirement home would not be modeled at all. Overall, what is happening is that there is a selection bias: we are likely to see fewer people in the retirement home who are younger. This issue is called \emph{left truncation}.

The standard way to correct for this sampling bias is to set the $i$-th person's contribution to the likelihood as:
\[
\bigg[\frac{f(Y_i|X_i)}{S(A_i|X_i)}\bigg]^{\Delta_i}
\bigg[
  \int_{Y_i}^\infty \frac{f(u|X_i)}{S(A_i|X_i)}\textrm{d}u
\bigg]^{1-\Delta_i}.
\]
The basic idea is that the PDF used is instead now conditioned on surviving past age $A_i$ (when the $i$-th person entered the retirement home), resulting in the denominator factors above. Roughly, $S(A_i|X_i)$ being smaller could be thought of as corresponding to a person who is less likely to have survived long enough to enter the retirement home; we up-weight this person's contribution so that the overall likelihood (across training individuals) tries to better resemble the city's population rather than only the retirement home's population.

Right and interval truncation are also possible. As the fix for these is similar, we refer the reader to Sections~3.4 and~3.5 of~\citet{klein2003survival} for examples of right and interval truncation in practice, and how to adjust the likelihood.

\paragraph{Cure models}
Next, we discuss what happens if some data points will actually just never experience the critical event of interest, meaning that the population-level survival function satisfies $\lim_{t\rightarrow\infty} S_{\text{pop}}(t) > 0$. There are different ways to model this setup. We describe a simple variant of the \emph{mixture cure model} \citep{boag1949maximum,xu2014nonparametric}.

Suppose that the $i$-th data point could possibly be ``cured'' of ever experiencing the critical event. Let $Z_i\in\{0,1\}$ be a random variable indicating whether the $i$-th data point is cured (if $Z_i=1$, then the $i$-th data point is cured). We assume that $Z_i$ is sampled from some underlying distribution $\mathbb{P}_{Z|X}(\cdot|x)$ that is Bernoulli with probability $\omega(x) := 1 - \lim_{t\rightarrow\infty}S(t|x)$. Note that $\omega(\cdot)$ is called the \emph{cure rate}. We do not get to observe $Z_i$.

Then the main modeling assumption is that
\[
S(t|x) = \omega(x) + (1 - \omega(x)) S_0(t|x),
\]
where $S_0(\cdot|x)$ is a survival function referred to as the \emph{latency}. The interpretation is that with probability $\omega(x)$, the survival probability is exactly~1 across time. Otherwise, the survival function is~$S_0(\cdot|x)$. We could, for instance, set $\omega(\cdot)$ and $S_0(\cdot|x)$ to be neural networks, write the resulting likelihood function, and maximize the likelihood with a neural network optimizer to learn parameters. For more details on various cure models and how explicitly modeling the cure rate can be beneficial in practice, see, for instance, the paper by \citet{ezquerro2023reliability}.

\subsection{Causal Reasoning and Interventions}

We have intentionally kept the scope of this monograph to time-to-event prediction and did not venture into the topics of causal reasoning or of interventions. Classically, assessing whether a treatment has an effect is based on population-level estimates. For example, \citet{mantel1966evaluation} established the now widely used \emph{log-rank} statistical test to assess whether two different groups' time-to-event outcome distributions are different. If the two groups correspond to treatment and control groups that appear the ``same'' aside from the former receiving a treatment (\eg, we randomly assign each individual to one of the two groups with equal probability), then the test would be measuring whether there is a ``treatment effect''. This test is done by comparing hazard function estimates of the two groups.\footnote{Even though the log-rank test pre-dates the classical Cox model \citep{cox1972regression}, the two are intricately related: the log-rank test corresponds to looking at a Cox model with a single feature that indicates which group a data point is (\eg, the feature is equal to 1~for points that received treatment and is~0 otherwise), and we are effectively checking whether the weight for this feature is~0 (which would mean that the time-to-event outcome distribution is the same between the two groups) \citep[Section~17.9]{harrell2015regression}.} Put another way, we are comparing population-level estimates, viewing the two groups as two different~populations.

Now with deep survival models as well as other machine learning survival models capable of predicting time-to-event outcomes at the \emph{individual} data point level rather than only at the population level, causal questions we aim to address could be more fine-grain, such as estimating \emph{individual} treatment effects. For some recent work that uses deep learning, see the papers by \citet{curth2021survite}, \citet{chapfuwa2021enabling}, and \citet{nagpal2022counterfactual}. Non-deep-learning approaches are also possible (\eg, \citealt{xu2023treatment,cui2023estimating}).

Supposing that one knew that an intervention works, there is a separate question of when to intervene, which has been recently studied by \citet{damera2022intervene}. Separately, in Section~\ref{chap:intro}, we already pointed out that time-to-event modeling could be applied to the setting where the time-to-event outcome is actually a number of inventory items to stock \citep{huh2011adaptive}. Huh \emph{et al.}~use the Kaplan-Meier estimator to obtain a simple strategy for how to allocate inventory. This sort of strategy could be implemented by actual businesses.

\subsection{Interpretability}

At this point, interpretability and explainability are widely studied in the machine learning community (\eg, see the book by \citet{molnar2022}). There are two main approaches taken. The first is to build a model that is inherently interpretable. The second is to learn an arbitrarily complex ``black box'' model and then come up with some sort of ``explanation'' of the black box. We discuss both of these in the context of deep survival models and then we separately mention a framework for visualizing any intermediate representation of any deep survival model.

\paragraph{Inherently interpretable deep survival models}
As we had discussed in Section~\ref{chap:proportional-hazards}, the classical Cox model (which is a special case of a deep survival model) is straightforward to interpret. It uses the log partial hazard function $\mathbf{f}(x;\theta) := \theta^\top x$, where $\theta\in\mathbb{R}^d$ and $x\in\mathcal{X}\subseteq\mathbb{R}^d$. Thus, the values in $\theta$ tell us precisely how we weight the different features. A feature with weight~0 would be ignored by the model entirely.
We could also ask for only a subset of the features to be explained by a linear model. For instance, suppose that each raw input can be written as $x = (u,v)$ where $u\in\mathbb{R}^{d_1}$ and $v\in\mathbb{R}^{d_2}$, where we only care about interpreting the model with respect to the features in~$u$ whereas the features in~$v$ are ``nuisance'' variables that are unrelated to~$u$ and that we do not care to make sense of. Then we could set the log partial hazard function to be
\[
\mathbf{f}(x;\theta) := \psi^\top u + \mathbf{g}(z; \phi),
\]
where $\psi\in\mathbb{R}^{d_1}$ is a parameter vector, $\mathbf{g}(\cdot; \phi):\mathbb{R}^{d_2}\rightarrow\mathbb{R}$ is some user-specified neural network with parameter variable $\phi$, and $\theta = (\psi,\phi)$. This partially linear Cox model with neural network $\mathbf{g}(\cdot; \phi)$ has known statistical guarantees \citep{zhong2022deep}.

Meanwhile, in Section~\ref{chap:deep-kaplan-meier}, we already discussed how deep kernel survival analysis \citep{chen2020deep} and survival kernets \citep{chen2024survival} are in some sense interpretable. As a reminder, both provide ``forecast evidence'': deep kernel survival analysis can tell us which training points contribute to predictions, and survival kernets can tell us which exemplar clusters contribute to predictions. For the latter, we also showed how to make visualizations like those in Figures~\ref{fig:kernet-km-example} and~\ref{fig:kernet-heatmap-example}. In the survival kernets paper, \citet{chen2024survival} noted that an interesting direction for future research is determining whether there are better ways to do exemplar-based clustering that, for instance, somehow improves model accuracy or model interpretability (how to define the latter of course is not straightforward). The choice of using $\varepsilon$-net clustering was to make the model amenable to theoretical analysis.

A number of other deep survival models have been developed that ``bake in'' some sort of interpretable component. For example, \citet{chapfuwa2020survival}, \citet{nagpal2021deep}, \citet{manduchi2022deep}, and \citet{jeanselme2022neural} also use clustering models, where clusters are learned in a supervised fashion to help with predicting a time-to-event outcome. Meanwhile, \citet{chen2024neural} propose using neural topic models that are (like the clustering models just mentioned) also supervised so that the topics help with time-to-event prediction.

More recently, \citet{sun2023nsotree} proposed a deep learning approach for training survival trees that are interpretable. We point out that there are also non-deep-learning based methods for training survival trees (\eg, \citealt{ishwaran2008random,bertsimas2022optimal,zhang2024optimal}), for which so long as a learned tree does not have too many leaves, then model interpretation is straightforward. If one uses an ensemble of trees, then if we want the overall model to be straightforward to interpret, then we would have to avoid using too many trees.

\paragraph{Explaining black box models}
As for training an arbitrarily complex deep survival model that is not interpretable and then applying a post hoc explanation tool afterward, here we begin by saying that standard tools that were not originally designed for time-to-event prediction can be used. Specifically, LIME \citep{ribeiro2016should} and SHAP \citep{lundberg2017unified} have already been applied to time-to-event prediction models \citep{kovalev2020survlime,krzyzinski2023survshap}.

We point out two other relatively straightforward strategies that can be applied:
\begin{itemize}

\item We could first train a deep survival model that is not interpretable. After training this model, we take its base neural network (with its learned parameters), and modify the last layer or a few of the last layers so that the modified base neural network can work with one of the inherently interpretable deep survival models (\eg, survival kernets), at which point, we could fine-tune the modified base neural network using the loss function of the interpretable model.

\item As an alternative, we could---just as before---first train a deep survival model that is not interpretable. Let's denote its resulting predicted survival function as $\widehat{S}_{\text{uninterpretable}}(\cdot|x)$. We then train one of the interpretable survival models where instead of (or in additional to) its usual loss function, we use a loss function that tries to match the interpretable model's predicted survival function with $\widehat{S}_{\text{uninterpretable}}(\cdot|x)$. Basically, we try to match the model outputs. We could of course try matching on the hazard function or the cumulative hazard function instead.
\end{itemize}
How to get these ideas to work well for deep survival models would be an interesting direction for future research.

\paragraph{Visualizing an intermediate representation}
To try to make sense of an intermediate embedding representation of any neural network, a standard approach is to apply t-SNE \citep{van2008visualizing} to this embedding representation. Note that t-SNE is \emph{not} designed to explain black box models. However, it can help us understand what semantics are captured by the embedding space (\eg, by showing which raw inputs tend to map close to each other in the embedding space).

In a similar vein, \citet{chen2023general} provided a general framework for visualizing any intermediate embedding representation used by any already trained deep survival model. The framework first identifies so-called \emph{anchor directions} in the embedding space. Afterward, visualizations can be made to relate each anchor direction to raw features and, separately, to time-to-event outcome distributions. In a healthcare dataset where data points are patients, an anchor direction could, for instance, capture the concept of age. Naturally, younger and older patients would have different distributions for how much longer they will live. Chen's framework also comes with statistical tests that could be run to check whether an anchor direction is associated with specific raw features.

\subsection{Fairness}

The machine learning community has now been working on fairness quite extensively. Fairness metrics for time-to-event prediction have only recently been defined \citep{keya2021equitable,rahman2022fair,zhang2022longitudinal}. Roughly, the key ideas of these metrics ask for: (i) similar data points to have similar predicted time-to-event outcomes, (ii) data points from different pre-defined subpopulations to have similar predicted outcomes, or (iii) data points from different pre-defined subpopulations to have similar prediction accuracy.

We point out that some of these metrics do not always make sense depending on the application. For example, in healthcare, age is often highly predictive of various time-to-event outcomes, such as the classical example of time until death. Suppose that we want a time-to-event prediction model to be ``fair'' across age groups. For simplicity, let's say that we just use two age groups (\eg, thresholding based on whether a person is at least 65 years old). Then asking for young people and elderly people to have similar predicted times until death would not make sense. In this case, asking for the model to be equally accurate for the two subpopulations is a better notion of fairness.

In terms of encouraging fairness, the standard approach is simply to add regularization terms \citep{keya2021equitable,rahman2022fair,do2023fair}, which is typically done in practice with the knowledge of which subpopulations we want to account for and which specific fairness metric we care about. However, enumerating all the subpopulations that we want to be fair across is not always straightforward. Minority subpopulations that might be at risk of being treated unfairly by a machine learning model could be defined by the intersection of a variety of different criteria \citep{buolamwini2018gender}. For example, even just considering a few attributes commonly treated as sensitive such as age, gender, and race, we run into the issue that there are many combinations of these three that are possible, in part because we also need to decide on how precisely to discretize age.

It turns out that it is possible to use a training loss function that does \emph{not} require the modeler to enumerate the subpopulations that we want to be fair across, and that still encourages fairness. In particular, \citet{hu2024fairness} introduced a general strategy for converting a wide range of time-to-event prediction models into ones that encourage fairness using \emph{distributionally robust optimization} (DRO) (\eg, \citealt{hashimoto2018fairness,duchi2021learning,li2021evaluating,duchi2022distributionally}). Roughly, DRO minimizes the worst-case error over \emph{all} subpopulations that are large enough (occurring with at least some user-specified probability threshold $\pi_{\min}$), and can be solved tractably. In particular, as $\pi_{\min}$ could be thought of as how ``rare'' of a minority subpopulation that we want the trained model to account for. As $\pi_{\min}\rightarrow0$, we would be training the time-to-event prediction model to have the worst-case error for even an individual data point (which could be an outlier) to be as low as possible. As $\pi_{\min}\rightarrow1$, we switch to simply carrying about the equally weighted average loss across all training data points, disregarding any sort of concern of minority subpopulations. 

The technical complication to applying DRO to time-to-event prediction is that existing DRO theory uses a training loss function that decomposes across contributions of individual data points, \ie, any term that shows up in the loss function depends only on a single training point. This decomposition does not hold for many deep survival models' loss functions, such as those of semiparametric proportional hazards models (\eg, the Cox model, DeepSurv, Cox-Time), DeepHit, deep kernel survival analysis, and survival kernets. \citet{hu2024fairness} address this technical hurdle using a sample splitting DRO strategy, which they also established some theory for. Specifically for classical and deep Cox models, Hu and Chen also derived an exact DRO Cox approach that does not require sample splitting.

\subsection{Statistical Guarantees}

We have tried sprinkling known results of statistical accuracy guarantees as we progressed through the monograph. In this section, we comment more on these accuracy guarantees for deep survival models. We also briefly mention guarantees in terms of producing so-called \emph{prediction intervals} for survival times (\eg, for test patient Alice, we predict that Alice's hospital length of stay is in the interval [0.5, 2.5] days with probability at least 90\%).

\paragraph{Accuracy guarantees} At present, the vast majority of deep survival models have no guarantees whatsoever. Even for the ones that do have guarantees, once we look closely at the fine print as to what the assumptions are, the assumptions made could be impractical.

For example, the theory for deep extended hazard models \citep{zhong2021deep} and deep partially linear Cox models \citep{zhong2022deep} make strong assumptions on the base neural networks used (\eg, on the architecture and on some notion of how large the neural network parameters are), and their theory relies on the neural network optimizer reaching the global minimum of the loss function. The theory for these models also makes some restrictive assumptions on the hazard function (after all, a deep extended hazard model's hazard function still has to satisfy a specific factorization, and the deep partially linear Cox model has an even more restrictive hazard function formulation).

Meanwhile, the theory for survival kernets \citep{chen2024survival} is stated in terms of the intermediate embedding space and how it relates to true survival time and censoring time distributions. In more detail, the theory treats the neural network $\mathbf{f}(\cdot;\theta):\mathcal{X}\rightarrow\mathbb{R}^{d_{\text{emb}}}$ that maps raw inputs to the embedding space as a black box so that it could be trained however the user wants. Instead, the theory requires the embedding space to satisfy ``nice'' properties. Some of these properties are straightforward to enforce or encourage, such as controlling the geometry of the embedding space and its probability distribution (the running example Chen gives is constraining the output of $\mathbf{f}(\cdot;\theta)$ to be on a hypersphere, for which there are known ways of encouraging the embedding vectors to be uniform over the hypersphere \citep{wang2020understanding,liu2021learning}). However, there is no known practical method to our knowledge of verifying the conditions on how the embedding space relates to true survival and censoring time distributions. Even if instead the assumptions were on relating the raw input space (rather than the embedding space) to the true survival and censoring time distributions, we would still not have a practical way to verify such conditions.

Suffice it to say, we think that there is a lot of space for improving our theoretical understanding of deep survival models.
Perhaps making some structural assumptions on the data would be helpful, such as assuming the data to come from a few clusters.

\paragraph{Prediction interval guarantees}
If we can predict a survival function for any data point, then we could also back out a survival time point estimate (e.g., look at when the survival function crosses 1/2 to estimate the median survival time). However, is there any way to estimate error bars for these survival time point estimates? When there is no censoring, this problem has at this point been studied for a long time, where a solution that has become popular is called \emph{conformal prediction} (early work was done by \citet{vovk2005algorithmic}; for an excellent tutorial on the topic, see the monograph by \citet{angelopoulos2023gentle}).

To give a sense of how conformal prediction works, we describe an approach called \emph{split conformal prediction} \citep{papadopoulos2002inductive,lei2015conformal}, which we intentionally phrase in terms of the survival analysis setup of Section~\ref{chap:setup} except where there is no censoring (so that we are looking at a standard regression problem). With the notation we have been using throughout the monograph but now dropping event indicator $\Delta_{i}$ variables (since they are all equal to 1), we take the i.i.d.~training data to be $(X_{1},Y_{1}),\dots,(X_{n},Y_{n})$, where $X_{i}$'s are raw inputs and $Y_{i}$'s are observed times. However, since every observed time $Y_{i}$ is actually a survival time $T_{i}$, for clarity, we write that the training data are $(X_{1},T_{1}),\dots,(X_{n},T_{n})$. (Using this notation will be helpful as we reintroduce censoring later.) We separately suppose that we have access to $n_{\dagger}$ so-called ``calibration'' data points $(X_{1}^{\dagger},T_{1}^{\dagger}),\dots,(X_{n_{\dagger}}^{\dagger},T_{n_{\dagger}}^{\dagger})$ that are i.i.d.~with the same distribution as the training data. Importantly, the calibration data do \emph{not} serve the same purpose as validation data (which are used to help tune hyperparameters; this would be considered as part of the training procedure). The calibration data should not be seen by the training procedure whatsoever.

Again, an example of a survival time prediction interval could be [0.5, 2.5] days. This interval would hold with some probability (since we typically cannot be absolutely certain of survival time prediction intervals unless it is a trivial interval like $[0,\infty)$). Let $\alpha\in(0,1)$ be a user-specified probability tolerance, where we aim to construct a prediction interval that holds with probability at least $1-\alpha$ (so that if we want to construct prediction intervals that hold with probability at least 90\%, we would pick $\alpha=0.1$).

Then split conformal prediction works as follows to produce survival time prediction intervals given the user-specified value for $\alpha$:
\begin{enumerate}

\item We train any regression model of our choosing on the training data $(X_{1},T_{1}),\dots,(X_{n},T_{n})$ to produce a survival time estimator $\widehat{T}:\mathcal{X}\rightarrow[0,\infty)$, meaning that $\widehat{T}(x)$ is the predicted survival time for any raw input $x\in\mathcal{X}$.

\item We compute residuals for the calibration data: $R_{i}=|T_{i}^{\dagger}-\widehat{T}(X_{i}^{\dagger})|$ for $i=1,2,\dots,n_{\dagger}$. We also introduce an additional residual $R_{n_{\dagger}+1}:=\infty$.

\item Note that the residuals $R_{1},\dots,R_{n_{\dagger}+1}$ can be sorted (even when we have $R_{n_{\dagger}+1}=\infty$ ). Let $\widehat{q}$ be the $(1-\alpha)$-th quantile of this empirical distribution. Put another way, if we denoted the sorted residuals as $R_{(1)}\le R_{(2)}\le\cdots\le R_{(n_{\dagger}+1)}=\infty$ (breaking ties randomly), then $\widehat{q}:=R_{(\lceil(1-\alpha)(n_{\dagger}+1)\rceil)}$. (For example, if $\alpha=0.1$, then we would be setting $\widehat{q}$ to be the 90 percentile value of $R_{1},\dots,R_{n_{\dagger}+1}$.)

\item For any $x\in\mathcal{X}$, we output the prediction interval for $x$ to be $\widehat{\mathcal{C}}(x):=[\widehat{T}(x)-\widehat{q},\widehat{T}(x)+\widehat{q}]$. Thus, for any test point $x\in\mathcal{X}$, our regression model's survival time prediction for $x$ is $\widehat{T}(x)$, and we estimate the error bar to be $\widehat{q}$ (this error bar does not depend on~$x$).
\end{enumerate}
A major theoretical result for split conformal prediction is that the prediction intervals it produces are ``statistically valid''. In particular, if we were to sample a data point $(X,T)$ from the same distribution underlying the training (and calibration) data, then with probability at least $1-\alpha$, the true survival time $T$ will be in the prediction interval $\mathcal{C}(X)$ (see Theorem 2.2 of \citet{lei2018distribution}).

Now let's discuss what happens when there is censoring. The training data are now $(X_{1},Y_{1},\Delta_{1}),\dots,(X_{n},Y_{n},\Delta_{n})$, and the calibration data are $(X_{1}^{\dagger},Y_{1}^{\dagger},\Delta_{1}^{\dagger}),\dots,(X_{n_{\dagger}}^{\dagger},Y_{n_{\dagger}}^{\dagger},\Delta_{n_{\dagger}}^{\dagger})$---again, we assume all these data points to be i.i.d. The key observation for how to handle censoring is actually immediate given our discussion in Section~\ref{sec:eval-point-estimates}. When there is censoring, the above split conformal prediction procedure can still be used but with some minor changes:
\begin{itemize}

\item In step 1, we would learn a survival model using the training data. We use this survival model to predict a survival time (again, the survival model chosen might actually predict a survival function from which we back out a median survival time estimate). We still denote this survival time estimator as $\widehat{T}:\mathcal{X}\rightarrow[0,\infty)$.

\item In step 2, we do not have the ground truth survival time for censored calibration points, but we could still compute residuals when there is censoring. For example, close to the start of Section~\ref{sec:eval-point-estimates}, we mentioned that a naive residual that could be computed is called the hinge error; the absolute error version would be
\[
R_{i}=\begin{cases}
|Y_{i}^{\dagger}-\widehat{T}(X_{i}^{\dagger})| & \text{if }\Delta_{i}^{\dagger}=1,\\
(Y_{i}^{\dagger}-\widehat{T}(X_{i}^{\dagger}))\ind\{Y_{i}>\widehat{T}(X_{i}^{\dagger})\} & \text{if }\Delta_{i}^{\dagger}=0.
\end{cases}
\]
Specifically, when a calibration point is censored ($\Delta_{i}^{\dagger}=0$), the residual is nonnegative only when the predicted survival time is less than the observed time (which is known to be a censoring time), and when $Y_{i}\le\widehat{T}(X_{i}^{\dagger})$, we optimistically assume that the error is 0 (even though it could be that the predicted survival time, for instance, way overestimates the unknown true survival time).

A different choice of residual is to use each individual data point's error from \eqref{eq:MAE-PO}, which---phrased in the context of calibration data---would be written as
\[
R_{i}=\Delta_{i}^{\dagger}|\widehat{T}(X_{i}^{\dagger})-Y_{i}|+(1-\Delta_{i})|\widehat{T}(X_{i}^{\dagger})-T_{i}^{\text{PO}}|,
\]
where $T_{i}^{\text{PO}}$ is defined in \eqref{eq:T_i-PO} (we would take the evaluation data points to be the calibration set). Of course, rather than using the pseudo observation approach to impute the unknown true survival time, we could use some other imputation approach (such as the margin method described in Section~\ref{sec:eval-point-estimates}).

\end{itemize}
Once we have made the above changes, the rest of the split conformal prediction procedure stays the same, and the upshot is that we can compute prediction intervals for survival times even when there is censoring. It is possible to come up with statistical guarantees for the resulting prediction intervals, much like how there are guarantees for conformal prediction in the standard regression setting without censoring. Existing such results have been established by \citet{chen2020deep} and \citet{candes2023conformalized}. Also, it is possible to use conformal prediction to make a survival model better calibrated \citep{qi2024conformalized}.

\subsection{Empirical Evaluation}
\label{sec:empirical-eval}

At the time of writing, there is no comprehensive, thorough empirical comparison of different survival models (regardless of whether they use deep learning or not) on a large selection of survival analysis datasets. Understandably, this is an onerous task. Many models have quite a few hyperparameters to tune, and figuring out the ``right'' range of values to try across datasets could be challenging. Also, the availability of publicly available standard datasets is, at present, not remotely at the same level as for classification and regression.

In conducting this sort of benchmark, we think that it is important to try to control for differences that are not related to the actual time-to-event prediction model. For example, if we were to compare two different deep survival models but we supplied them with vastly different base neural networks, then the differences in how the two models perform might be explained more by the differences in the base neural networks than by differences in the two survival models' modeling assumptions.\footnote{Of course, commonly the base neural network cannot be made identical across different time-to-event prediction models (\eg, DeepSurv requires its base neural network to output a single value whereas DeepHit instead asks for many output values, one per discretized time index), but we can still have the base neural networks be as similar as possible (\eg, using multilayer perceptrons that are identical except for the final layer).} As another example, in conducting minibatch gradient descent to train two different deep survival models, whether we use early stopping (\eg, if there is no improvement in some validation set evaluation score after 10 training epochs, then stop training and backtrack to using whichever epoch's model parameters achieved the highest validation set evaluation score) should be the same across models.\footnote{Some models might work better with different learning rates or different numbers of training epochs. We could try to tune these in a similar fashion across models. For example, we can fix some maximum number of training epochs across all models and then use early stopping to automatically tune on the number of training epochs. As for the learning rate, we could pre-specify a grid of learning rates that we try across models. Then per model, we use whichever learning rate achieves the best validation set evaluation score (per learning rate, we use early stopping to decide on the number of training epochs).}

Meanwhile, devising new evaluation metrics for survival models remains an active area of research (see, for instance, the recent paper by \citet{qi2023effective}). The community has focused a lot on ranking-based accuracy metrics, such as time-dependent concordance indices. While these could be useful for ranking data points such as patients in a clinical task (\eg, to help hospitals prioritize which patients to focus on), ranking is not always what we want to do. In some cases, having an actual predicted value of the time-to-event outcome could be useful, in which case, the survival time error metrics in Section~\ref{sec:eval-point-estimates} could be useful (\eg, MAE measured against ground truth survival times for uncensored cases and against imputed survival times for censored cases). However, these error metrics based on pointwise estimates of survival times do not easily generalize to the setting where there are multiple competing events (or also to the setting of cure models mentioned in Section~\ref{sec:truncation-other-censoring-cure}), where the problem is that for a specific critical event, the ground truth could be that this event never happens for a data point.

Separately, it does not help that various metrics require an estimate of the censoring time distribution $\mathbb{P}_C$ (in terms of the function $S_{\text{censor}}(t) = 1 - \mathbb{P}_C(t)$). We had pointed out that when censoring times are independent of the raw inputs, then it suffices to estimate $S_{\text{censor}}$ using the Kaplan-Meier estimator. When this independence assumption does not hold, then there exist versions of some of the evaluation metrics we mentioned that would still work if we instead have a good estimate of the distribution~$\mathbb{P}_{C|X}$, but this distribution could be as difficult to estimate as the target distribution we are trying to predict~$\mathbb{P}_{T|X}$. Regardless, a bad estimate of the censoring time distribution could cause evaluation metrics that depend on this distribution to be unreliable.

Turning toward the dynamic setup of Section~\ref{sec:dynamic} where we see more of a time series over time and can continually make predictions, it is unclear what evaluation metrics make the most sense here. In practice, we are unlikely to need to make predictions after every single time step of new information is collected. When a prediction might actually be useful could depend on whether there is an intervention that might make sense to attempt in the near future. Meanwhile, in practice, it is likely the case that among $k$ critical events, some are more concerning than others so that we would not want to treat all $k$ events equally.

\subsection{Large Language Models and Foundation Models}

Lastly, during the writing of this monograph, large language models (LLMs) and---more generally---foundation models took over the machine learning community by storm \citep{bommasani2021opportunities}. These developments quickly transferred over to the survival analysis setting. After all, at a conceptual level, one could simply set the base neural network to be a foundation model.

There is already a review of how LLMs are used for survival analysis \citep{jeanselme2024review}. As is, in Section~\ref{sec:empirical-eval}, we mentioned that there is no comprehensive experimental benchmark for survival analysis. This is also true for LLMs applied to survival analysis, where differences in experimental setups make ``apples-to-apples'' comparisons between LLM approaches difficult. To address this issue, in their review, Jeanselme \emph{et al.}~propose a framework for evaluating LLMs for survival analysis that aims to standardize various steps of the machine learning pipeline.

Meanwhile, specifically for electronic health records, a foundation model called MOTOR for predicting time durations until the next critical event happens has recently been published \citep{steinberg2024self}. The underlying survival model Steinberg \emph{et al.}~used as the prediction head has been available for a number of years: a piecewise exponential model for hazard functions \citep{fornili2014piecewise}. The base neural network is a transformer. The novelty in the research was in scaling the resulting model to a sizable amount of data (pretraining uses 55M patient records with 9B clinical events), and demonstrating impressive transfer learning results (19 tasks across 3 healthcare datasets). Could using a different survival model as the prediction head have improved their results? What about a different base neural network? What about dramatically more data? For what kinds of time-to-event prediction problems would foundation models be most beneficial, and for what kinds would they be least beneficial---not limited to only the healthcare space? Suffice it to say, there are many open questions related to how best to use foundation models for time-to-event prediction.

\section*{Acknowledgments}

I would like to first thank Jeremy Weiss for introducing me to the topic of survival analysis back when I first started as faculty at Carnegie Mellon University. This monograph is also in some sense a successor to a tutorial Jeremy and I taught on survival analysis at the Conference on Health, Inference, and Learning back in 2020. I would also like to thank a number of collaborators who I have worked on survival analysis problems with aside from Jeremy: Cheng Cheng, Amanda Coston, Jonathan Elmer, Shu Hu, Lihong Li, Zack Lipton, Xiaobin Shen, Helen Zhou, and Ren Zuo. Finally, I would like to thank the editors and reviewers at Foundation and Trends in Machine Learning for providing very helpful feedback. This work is supported by NSF CAREER award \#2047981.

\newpage

\bibliographystyle{abbrvnat}
\bibliography{ref}

\begin{thebibliography}{182}
\providecommand{\natexlab}[1]{#1}
\providecommand{\url}[1]{\texttt{#1}}
\expandafter\ifx\csname urlstyle\endcsname\relax
  \providecommand{\doi}[1]{doi: #1}\else
  \providecommand{\doi}{doi: \begingroup \urlstyle{rm}\Url}\fi

\bibitem[Aalen(1978)]{aalen1978nonparametric}
O.~O. Aalen.
\newblock Nonparametric inference for a family of counting processes.
\newblock \emph{The Annals of Statistics}, 6\penalty0 (4):\penalty0 701--726,
  1978.

\bibitem[Aalen and Johansen(1978)]{aalen1978empirical}
O.~O. Aalen and S.~Johansen.
\newblock An empirical transition matrix for non-homogeneous markov chains
  based on censored observations.
\newblock \emph{Scandinavian Journal of Statistics}, pages 141--150, 1978.

\bibitem[Abadi et~al.(2015)Abadi, Agarwal, Barham, Brevdo, Chen, Citro,
  Corrado, Davis, Dean, Devin, Ghemawat, Goodfellow, Harp, Irving, Isard, Jia,
  Jozefowicz, Kaiser, Kudlur, Levenberg, Man\'{e}, Monga, Moore, Murray, Olah,
  Schuster, Shlens, Steiner, Sutskever, Talwar, Tucker, Vanhoucke, Vasudevan,
  Vi\'{e}gas, Vinyals, Warden, Wattenberg, Wicke, Yu, and
  Zheng]{tensorflow2015-whitepaper}
M.~Abadi, A.~Agarwal, P.~Barham, E.~Brevdo, Z.~Chen, C.~Citro, G.~S. Corrado,
  A.~Davis, J.~Dean, M.~Devin, S.~Ghemawat, I.~Goodfellow, A.~Harp, G.~Irving,
  M.~Isard, Y.~Jia, R.~Jozefowicz, L.~Kaiser, M.~Kudlur, J.~Levenberg,
  D.~Man\'{e}, R.~Monga, S.~Moore, D.~Murray, C.~Olah, M.~Schuster, J.~Shlens,
  B.~Steiner, I.~Sutskever, K.~Talwar, P.~Tucker, V.~Vanhoucke, V.~Vasudevan,
  F.~Vi\'{e}gas, O.~Vinyals, P.~Warden, M.~Wattenberg, M.~Wicke, Y.~Yu, and
  X.~Zheng.
\newblock {TensorFlow}: Large-scale machine learning on heterogeneous systems,
  2015.
\newblock URL \url{https://www.tensorflow.org/}.
\newblock Software available from tensorflow.org.

\bibitem[Allison(1982)]{allison1982discrete}
P.~D. Allison.
\newblock Discrete-time methods for the analysis of event histories.
\newblock \emph{Sociological Methodology}, 13:\penalty0 61--98, 1982.

\bibitem[Angelopoulos and Bates(2023)]{angelopoulos2023gentle}
A.~N. Angelopoulos and S.~Bates.
\newblock A gentle introduction to conformal prediction and distribution-free
  uncertainty quantification.
\newblock \emph{Foundations and Trends{\textregistered} in Machine Learning},
  16\penalty0 (4):\penalty0 494--591, 2023.

\bibitem[Antolini et~al.(2005)Antolini, Boracchi, and
  Biganzoli]{antolini2005time}
L.~Antolini, P.~Boracchi, and E.~Biganzoli.
\newblock A time-dependent discrimination index for survival data.
\newblock \emph{Statistics in Medicine}, 24\penalty0 (24):\penalty0 3927--3944,
  2005.

\bibitem[Avati et~al.(2020)Avati, Duan, Zhou, Jung, Shah, and
  Ng]{avati2020countdown}
A.~Avati, T.~Duan, S.~Zhou, K.~Jung, N.~H. Shah, and A.~Y. Ng.
\newblock Countdown regression: Sharp and calibrated survival predictions.
\newblock In \emph{Uncertainty in Artificial Intelligence}, pages 145--155.
  PMLR, 2020.

\bibitem[Baca{\"e}r(2011)]{bacaer2011short}
N.~Baca{\"e}r.
\newblock \emph{A Short History of Mathematical Population Dynamics}, volume
  618.
\newblock Springer, 2011.

\bibitem[Beran(1981)]{beran1981nonparametric}
R.~Beran.
\newblock Nonparametric regression with randomly censored survival data.
\newblock \emph{Technical report, University of California, Berkeley}, 1981.

\bibitem[Bertsimas et~al.(2022)Bertsimas, Dunn, Gibson, and
  Orfanoudaki]{bertsimas2022optimal}
D.~Bertsimas, J.~Dunn, E.~Gibson, and A.~Orfanoudaki.
\newblock Optimal survival trees.
\newblock \emph{Machine Learning}, 111\penalty0 (8):\penalty0 2951--3023, 2022.

\bibitem[Blanche et~al.(2013)Blanche, Latouche, and Viallon]{blanche2013time}
P.~Blanche, A.~Latouche, and V.~Viallon.
\newblock Time-dependent auc with right-censored data: a survey.
\newblock \emph{Risk Assessment and Evaluation of Predictions}, pages 239--251,
  2013.

\bibitem[Blanche et~al.(2019)Blanche, Kattan, and Gerds]{blanche2019c}
P.~Blanche, M.~W. Kattan, and T.~A. Gerds.
\newblock The c-index is not proper for the evaluation of-year predicted risks.
\newblock \emph{Biostatistics}, 20\penalty0 (2):\penalty0 347--357, 2019.

\bibitem[Blei et~al.(2017)Blei, Kucukelbir, and McAuliffe]{blei2017variational}
D.~M. Blei, A.~Kucukelbir, and J.~D. McAuliffe.
\newblock Variational inference: A review for statisticians.
\newblock \emph{Journal of the American Statistical Association}, 112\penalty0
  (518):\penalty0 859--877, 2017.

\bibitem[Boag(1949)]{boag1949maximum}
J.~W. Boag.
\newblock Maximum likelihood estimates of the proportion of patients cured by
  cancer therapy.
\newblock \emph{Journal of the Royal Statistical Society Series B: Statistical
  Methodology}, 11\penalty0 (1):\penalty0 15--53, 1949.

\bibitem[B{\"o}hmer(1912)]{bohmer1912theorie}
P.~E. B{\"o}hmer.
\newblock Theorie der unabh{\"a}ngigen wahrscheinlichkeiten.
\newblock In \emph{Rapports Memoires et Proces verbaux de Septieme Congres
  International dActuaires Amsterdam}, volume~2, pages 327--343, 1912.

\bibitem[Bommasani et~al.(2021)Bommasani, Hudson, Adeli, Altman, Arora, von
  Arx, Bernstein, Bohg, Bosselut, Brunskill, Brynjolfsson, Buch, Card,
  Castellon, Chatterji, Chen, Creel, Davis, Demszky, Donahue, Doumbouya,
  Durmus, Ermon, Etchemendy, Ethayarajh, Fei-Fei, Finn, Gale, Gillespie, Goel,
  Goodman, Grossman, Guha, Hashimoto, Henderson, Hewitt, Ho, Hong, Hsu, Huang,
  Icard, Jain, Jurafsky, Kalluri, Karamcheti, Keeling, Khani, Khattab, Koh,
  Krass, Krishna, Kuditipudi, Kumar, Ladhak, Lee, Lee, Leskovec, Levent, Li,
  Li, Ma, Malik, Manning, Mirchandani, Mitchell, Munyikwa, Nair, Narayan,
  Narayanan, Newman, Nie, Niebles, Nilforoshan, Nyarko, Ogut, Orr,
  Papadimitriou, Park, Piech, Portelance, Potts, Raghunathan, Reich, Ren, Rong,
  Roohani, Ruiz, Ryan, Ré, Sadigh, Sagawa, Santhanam, Shih, Srinivasan,
  Tamkin, Taori, Thomas, Tramèr, Wang, Wang, Wu, Wu, Wu, Xie, Yasunaga, You,
  Zaharia, Zhang, Zhang, Zhang, Zhang, Zheng, Zhou, and
  Liang]{bommasani2021opportunities}
R.~Bommasani, D.~A. Hudson, E.~Adeli, R.~Altman, S.~Arora, S.~von Arx, M.~S.
  Bernstein, J.~Bohg, A.~Bosselut, E.~Brunskill, E.~Brynjolfsson, S.~Buch,
  D.~Card, R.~Castellon, N.~Chatterji, A.~Chen, K.~Creel, J.~Q. Davis,
  D.~Demszky, C.~Donahue, M.~Doumbouya, E.~Durmus, S.~Ermon, J.~Etchemendy,
  K.~Ethayarajh, L.~Fei-Fei, C.~Finn, T.~Gale, L.~Gillespie, K.~Goel,
  N.~Goodman, S.~Grossman, N.~Guha, T.~Hashimoto, P.~Henderson, J.~Hewitt,
  D.~E. Ho, J.~Hong, K.~Hsu, J.~Huang, T.~Icard, S.~Jain, D.~Jurafsky,
  P.~Kalluri, S.~Karamcheti, G.~Keeling, F.~Khani, O.~Khattab, P.~W. Koh,
  M.~Krass, R.~Krishna, R.~Kuditipudi, A.~Kumar, F.~Ladhak, M.~Lee, T.~Lee,
  J.~Leskovec, I.~Levent, X.~L. Li, X.~Li, T.~Ma, A.~Malik, C.~D. Manning,
  S.~Mirchandani, E.~Mitchell, Z.~Munyikwa, S.~Nair, A.~Narayan, D.~Narayanan,
  B.~Newman, A.~Nie, J.~C. Niebles, H.~Nilforoshan, J.~Nyarko, G.~Ogut, L.~Orr,
  I.~Papadimitriou, J.~S. Park, C.~Piech, E.~Portelance, C.~Potts,
  A.~Raghunathan, R.~Reich, H.~Ren, F.~Rong, Y.~Roohani, C.~Ruiz, J.~Ryan,
  C.~Ré, D.~Sadigh, S.~Sagawa, K.~Santhanam, A.~Shih, K.~Srinivasan,
  A.~Tamkin, R.~Taori, A.~W. Thomas, F.~Tramèr, R.~E. Wang, W.~Wang, B.~Wu,
  J.~Wu, Y.~Wu, S.~M. Xie, M.~Yasunaga, J.~You, M.~Zaharia, M.~Zhang, T.~Zhang,
  X.~Zhang, Y.~Zhang, L.~Zheng, K.~Zhou, and P.~Liang.
\newblock On the opportunities and risks of foundation models.
\newblock \emph{arXiv preprint arXiv:2108.07258}, 2021.

\bibitem[Box-Steffensmeier and Jones(2004)]{box2004event}
J.~M. Box-Steffensmeier and B.~S. Jones.
\newblock \emph{Event History Modeling: A Guide for Social Scientists}.
\newblock Cambridge University Press, 2004.

\bibitem[Bradbury et~al.(2018)Bradbury, Frostig, Hawkins, Johnson, Leary,
  Maclaurin, Necula, Paszke, Vander{P}las, Wanderman-{M}ilne, and
  Zhang]{jax2018github}
J.~Bradbury, R.~Frostig, P.~Hawkins, M.~J. Johnson, C.~Leary, D.~Maclaurin,
  G.~Necula, A.~Paszke, J.~Vander{P}las, S.~Wanderman-{M}ilne, and Q.~Zhang.
\newblock {JAX}: composable transformations of {P}ython+{N}um{P}y programs,
  2018.
\newblock URL \url{http://github.com/jax-ml/jax}.

\bibitem[Breslow(1972)]{breslow1972discussion}
N.~Breslow.
\newblock Discussion of the paper by {D~R}~{C}ox (1972).
\newblock \emph{Journal of the Royal Statistical Society, Series B}.
  34(2):216--217, 1972.

\bibitem[Brown(1975)]{brown1975use}
C.~C. Brown.
\newblock On the use of indicator variables for studying the time-dependence of
  parameters in a response-time model.
\newblock \emph{Biometrics}, 31\penalty0 (4):\penalty0 863--872, 1975.

\bibitem[Buolamwini and Gebru(2018)]{buolamwini2018gender}
J.~Buolamwini and T.~Gebru.
\newblock Gender shades: Intersectional accuracy disparities in commercial
  gender classification.
\newblock In \emph{Conference on Fairness, Accountability and Transparency},
  pages 77--91. PMLR, 2018.

\bibitem[Cand{\`e}s et~al.(2023)Cand{\`e}s, Lei, and
  Ren]{candes2023conformalized}
E.~Cand{\`e}s, L.~Lei, and Z.~Ren.
\newblock Conformalized survival analysis.
\newblock \emph{Journal of the Royal Statistical Society Series B: Statistical
  Methodology}, 85\penalty0 (1):\penalty0 24--45, 2023.

\bibitem[Chagny and Roche(2014)]{chagny2014adaptive}
G.~Chagny and A.~Roche.
\newblock Adaptive and minimax estimation of the cumulative distribution
  function given a functional covariate.
\newblock \emph{Electronic Journal of Statistics}, 8\penalty0 (2):\penalty0
  2352--2404, 2014.

\bibitem[Chapelle(2014)]{chapelle2014modeling}
O.~Chapelle.
\newblock Modeling delayed feedback in display advertising.
\newblock In \emph{ACM SIGKDD International Conference on Knowledge Discovery
  and Data Mining}, pages 1097--1105, 2014.

\bibitem[Chapfuwa et~al.(2018)Chapfuwa, Tao, Li, Page, Goldstein, Duke, and
  Henao]{chapfuwa2018adversarial}
P.~Chapfuwa, C.~Tao, C.~Li, C.~Page, B.~Goldstein, L.~C. Duke, and R.~Henao.
\newblock Adversarial time-to-event modeling.
\newblock In \emph{International Conference on Machine Learning}, pages
  735--744. PMLR, 2018.

\bibitem[Chapfuwa et~al.(2020)Chapfuwa, Li, Mehta, Carin, and
  Henao]{chapfuwa2020survival}
P.~Chapfuwa, C.~Li, N.~Mehta, L.~Carin, and R.~Henao.
\newblock Survival cluster analysis.
\newblock In \emph{Proceedings of the ACM Conference on Health, Inference, and
  Learning}, pages 60--68, 2020.

\bibitem[Chapfuwa et~al.(2021)Chapfuwa, Assaad, Zeng, Pencina, Carin, and
  Henao]{chapfuwa2021enabling}
P.~Chapfuwa, S.~Assaad, S.~Zeng, M.~J. Pencina, L.~Carin, and R.~Henao.
\newblock Enabling counterfactual survival analysis with balanced
  representations.
\newblock In \emph{Proceedings of the Conference on Health, Inference, and
  Learning}, pages 133--145, 2021.

\bibitem[Chen(2019)]{chen2019nearest}
G.~H. Chen.
\newblock Nearest neighbor and kernel survival analysis: Nonasymptotic error
  bounds and strong consistency rates.
\newblock In \emph{International Conference on Machine Learning}, pages
  1001--1010. PMLR, 2019.

\bibitem[Chen(2020)]{chen2020deep}
G.~H. Chen.
\newblock Deep kernel survival analysis and subject-specific survival time
  prediction intervals.
\newblock In \emph{Machine Learning for Healthcare Conference}, pages 537--565.
  PMLR, 2020.

\bibitem[Chen(2023)]{chen2023general}
G.~H. Chen.
\newblock A general framework for visualizing embedding spaces of neural
  survival analysis models based on angular information.
\newblock In \emph{Conference on Health, Inference, and Learning}, pages
  440--476. PMLR, 2023.

\bibitem[Chen(2024)]{chen2024survival}
G.~H. Chen.
\newblock Survival kernets: Scalable and interpretable deep kernel survival
  analysis with an accuracy guarantee.
\newblock \emph{Journal of Machine Learning Research}, 25\penalty0
  (40):\penalty0 1--78, 2024.

\bibitem[Chen and Shah(2018)]{chen2018explaining}
G.~H. Chen and D.~Shah.
\newblock Explaining the success of nearest neighbor methods in prediction.
\newblock \emph{Foundations and Trends{\textregistered} in Machine Learning},
  10\penalty0 (5-6):\penalty0 337--588, 2018.

\bibitem[Chen et~al.(2024)Chen, Li, Zuo, Coston, and Weiss]{chen2024neural}
G.~H. Chen, L.~Li, R.~Zuo, A.~Coston, and J.~C. Weiss.
\newblock Neural topic models with survival supervision: Jointly predicting
  time-to-event outcomes and learning how clinical features relate.
\newblock \emph{Artificial Intelligence in Medicine}, 2024.

\bibitem[Chen et~al.(2018)Chen, Rubanova, Bettencourt, and
  Duvenaud]{chen2018neural}
R.~T. Chen, Y.~Rubanova, J.~Bettencourt, and D.~K. Duvenaud.
\newblock Neural ordinary differential equations.
\newblock In \emph{Advances in Neural Information Processing Systems}, 2018.

\bibitem[Chen and Guestrin(2016)]{chen2016xgboost}
T.~Chen and C.~Guestrin.
\newblock {XGBoost}: A scalable tree boosting system.
\newblock In \emph{ACM SIGKDD International Conference on Knowledge Discovery
  and Data Mining}, pages 785--794, 2016.

\bibitem[Chilinski and Silva(2020)]{chilinski2020neural}
P.~Chilinski and R.~Silva.
\newblock Neural likelihoods via cumulative distribution functions.
\newblock In \emph{Conference on Uncertainty in Artificial Intelligence}, pages
  420--429. PMLR, 2020.

\bibitem[Chung et~al.(1991)Chung, Schmidt, and Witte]{chung1991survival}
C.-F. Chung, P.~Schmidt, and A.~D. Witte.
\newblock Survival analysis: A survey.
\newblock \emph{Journal of Quantitative Criminology}, 7:\penalty0 59--98, 1991.

\bibitem[Ciampi et~al.(1981)Ciampi, Bush, Gospodarowicz, and
  Till]{ciampi1981approach}
A.~Ciampi, R.~S. Bush, M.~Gospodarowicz, and J.~E. Till.
\newblock An approach to classifying prognostic factors related to survival
  experience for non-hodgkin's lymphoma patients: Based on a series of 982
  patients: 1967--1975.
\newblock \emph{Cancer}, 47\penalty0 (3):\penalty0 621--627, 1981.

\bibitem[Collett(2023)]{collett2023modelling}
D.~Collett.
\newblock \emph{Modelling Survival Data in Medical Research, Fourth Edition}.
\newblock Chapman and Hall/CRC, 2023.

\bibitem[Cover and Hart(1967)]{cover1967nearest}
T.~Cover and P.~Hart.
\newblock Nearest neighbor pattern classification.
\newblock \emph{IEEE Transactions on Information Theory}, 13\penalty0
  (1):\penalty0 21--27, 1967.

\bibitem[Cox(1972)]{cox1972regression}
D.~R. Cox.
\newblock Regression models and life-tables.
\newblock \emph{Journal of the Royal Statistical Society: Series B},
  34\penalty0 (2):\penalty0 187--202, 1972.

\bibitem[Cox and Oakes(1984)]{cox1984analysis}
D.~R. Cox and D.~Oakes.
\newblock \emph{Analysis of Survival Data}.
\newblock CRC press, 1984.

\bibitem[Craig et~al.(2021)Craig, Zhong, and Tibshirani]{craig2021survival}
E.~Craig, C.~Zhong, and R.~Tibshirani.
\newblock Survival stacking: casting survival analysis as a classification
  problem.
\newblock \emph{arXiv preprint arXiv:2107.13480}, 2021.

\bibitem[Cui et~al.(2023)Cui, Kosorok, Sverdrup, Wager, and
  Zhu]{cui2023estimating}
Y.~Cui, M.~R. Kosorok, E.~Sverdrup, S.~Wager, and R.~Zhu.
\newblock Estimating heterogeneous treatment effects with right-censored data
  via causal survival forests.
\newblock \emph{Journal of the Royal Statistical Society Series B: Statistical
  Methodology}, 85\penalty0 (2):\penalty0 179--211, 2023.

\bibitem[Curth et~al.(2021)Curth, Lee, and van~der Schaar]{curth2021survite}
A.~Curth, C.~Lee, and M.~van~der Schaar.
\newblock {SurvITE}: Learning heterogeneous treatment effects from
  time-to-event data.
\newblock In \emph{Advances in Neural Information Processing Systems}, 2021.

\bibitem[Daley and Vere-Jones(2003)]{daley2003introduction}
D.~J. Daley and D.~Vere-Jones.
\newblock \emph{An Introduction to the Theory of Point Processes: Volume I:
  Elementary Theory and Methods}.
\newblock Springer, 2003.

\bibitem[Daley and Vere-Jones(2008)]{daley2008introduction}
D.~J. Daley and D.~Vere-Jones.
\newblock \emph{An Introduction to the Theory of Point Processes: Volume II:
  General Theory and Structure}.
\newblock Springer, 2008.

\bibitem[Damera~Venkata and Bhattacharyya(2022)]{damera2022intervene}
N.~Damera~Venkata and C.~Bhattacharyya.
\newblock When to intervene: Learning optimal intervention policies for
  critical events.
\newblock In \emph{Advances in Neural Information Processing Systems}, 2022.

\bibitem[Danks and Yau(2022)]{danks2022derivative}
D.~Danks and C.~Yau.
\newblock Derivative-based neural modelling of cumulative distribution
  functions for survival analysis.
\newblock In \emph{International Conference on Artificial Intelligence and
  Statistics}, pages 7240--7256. PMLR, 2022.

\bibitem[Davidson-Pilon(2019)]{davidson2019lifelines}
C.~Davidson-Pilon.
\newblock lifelines: survival analysis in {P}ython.
\newblock \emph{Journal of Open Source Software}, 4\penalty0 (40):\penalty0
  1317, 2019.

\bibitem[Do et~al.(2023)Do, Chang, Cho, Smyth, and Zhong]{do2023fair}
H.~Do, Y.~Chang, Y.~S. Cho, P.~Smyth, and J.~Zhong.
\newblock Fair survival time prediction via mutual information minimization.
\newblock In \emph{Machine Learning for Healthcare Conference}, 2023.

\bibitem[Downey(2011)]{downey2011think}
A.~B. Downey.
\newblock \emph{Think stats}.
\newblock O'Reilly Media, Inc., 2011.

\bibitem[Du et~al.(2016)Du, Dai, Trivedi, Upadhyay, Gomez-Rodriguez, and
  Song]{du2016recurrent}
N.~Du, H.~Dai, R.~Trivedi, U.~Upadhyay, M.~Gomez-Rodriguez, and L.~Song.
\newblock Recurrent marked temporal point processes: Embedding event history to
  vector.
\newblock In \emph{ACM SIGKDD International Conference on Knowledge Discovery
  and Data Mining}, pages 1555--1564, 2016.

\bibitem[Duchi et~al.(2022)Duchi, Hashimoto, and
  Namkoong]{duchi2022distributionally}
J.~Duchi, T.~Hashimoto, and H.~Namkoong.
\newblock Distributionally robust losses for latent covariate mixtures.
\newblock \emph{Operations Research}, 2022.

\bibitem[Duchi and Namkoong(2021)]{duchi2021learning}
J.~C. Duchi and H.~Namkoong.
\newblock Learning models with uniform performance via distributionally robust
  optimization.
\newblock \emph{The Annals of Statistics}, 49\penalty0 (3):\penalty0
  1378--1406, 2021.

\bibitem[Dybowski and Gant(2001)]{dybowski2001clinical}
R.~Dybowski and V.~Gant.
\newblock \emph{Clinical Applications of Artificial Neural Networks}.
\newblock Cambridge University Press, 2001.

\bibitem[Ebeling(2019)]{ebeling2019introduction}
C.~E. Ebeling.
\newblock \emph{An Introduction to Reliability and Maintainability
  Engineering}.
\newblock Waveland Press, 2019.

\bibitem[Ezquerro et~al.(2023)Ezquerro, Cancela, and
  L{\'o}pez-Cheda]{ezquerro2023reliability}
A.~Ezquerro, B.~Cancela, and A.~L{\'o}pez-Cheda.
\newblock On the reliability of machine learning models for survival analysis
  when cure is a possibility.
\newblock \emph{Mathematics}, 11\penalty0 (19):\penalty0 4150, 2023.

\bibitem[Faraggi and Simon(1995)]{faraggi1995neural}
D.~Faraggi and R.~Simon.
\newblock A neural network model for survival data.
\newblock \emph{Statistics in Medicine}, 14:\penalty0 73--82, 1995.

\bibitem[Fine and Gray(1999)]{fine1999proportional}
J.~P. Fine and R.~J. Gray.
\newblock A proportional hazards model for the subdistribution of a competing
  risk.
\newblock \emph{Journal of the American Statistical Association}, 94\penalty0
  (446):\penalty0 496--509, 1999.

\bibitem[Fleming and Harrington(1991)]{fleming1991counting}
T.~R. Fleming and D.~P. Harrington.
\newblock \emph{Counting Processes and Survival Analysis}.
\newblock John Wiley \& Sons, 1991.

\bibitem[Foekens et~al.(2000)Foekens, Peters, Look, Portengen, Schmitt, Kramer,
  Br{\"u}nner, J{\"a}nicke, Meijer-van Gelder, Henzen-Logmans, van Putten, and
  Klijn]{foekens2000urokinase}
J.~A. Foekens, H.~A. Peters, M.~P. Look, H.~Portengen, M.~Schmitt, M.~D.
  Kramer, N.~Br{\"u}nner, F.~J{\"a}nicke, M.~E. Meijer-van Gelder, S.~C.
  Henzen-Logmans, W.~L.~J. van Putten, and J.~G.~M. Klijn.
\newblock The urokinase system of plasminogen activation and prognosis in 2780
  breast cancer patients.
\newblock \emph{Cancer Research}, 60\penalty0 (3):\penalty0 636--643, 2000.

\bibitem[F\"{o}ldes and Rejt\"{o}(1981)]{foldes1981strong}
A.~F\"{o}ldes and L.~Rejt\"{o}.
\newblock Strong uniform consistency for nonparametric survival curve
  estimators from randomly censored data.
\newblock \emph{The Annals of Statistics}, 9\penalty0 (1):\penalty0 122--129,
  1981.

\bibitem[Fornili et~al.(2014)Fornili, Ambrogi, Boracchi, and
  Biganzoli]{fornili2014piecewise}
M.~Fornili, F.~Ambrogi, P.~Boracchi, and E.~Biganzoli.
\newblock Piecewise exponential artificial neural networks ({PEANN}) for
  modeling hazard function with right censored data.
\newblock In \emph{Computational Intelligence Methods for Bioinformatics and
  Biostatistics: 10th International Meeting, CIBB 2013, Nice, France, June
  20-22, 2013, Revised Selected Papers 10}, pages 125--136. Springer, 2014.

\bibitem[Fotso(2018)]{fotso2018deep}
S.~Fotso.
\newblock Deep neural networks for survival analysis based on a multi-task
  framework.
\newblock \emph{arXiv preprint arXiv:1801.05512}, 2018.

\bibitem[Fotso et~al.(2019)]{pysurvival_cite}
S.~Fotso et~al.
\newblock {PySurvival}: Open source package for survival analysis modeling,
  2019.
\newblock URL \url{https://www.pysurvival.io/}.

\bibitem[Gensheimer and Narasimhan(2019)]{gensheimer2019scalable}
M.~F. Gensheimer and B.~Narasimhan.
\newblock A scalable discrete-time survival model for neural networks.
\newblock \emph{PeerJ}, 7:\penalty0 e6257, 2019.

\bibitem[Gerds and Kattan(2021)]{gerds2021medical}
T.~A. Gerds and M.~W. Kattan.
\newblock \emph{Medical Risk Prediction Models: With Ties to Machine Learning}.
\newblock Chapman and Hall/CRC, 2021.

\bibitem[Glass(1963)]{glass1963john}
D.~V. Glass.
\newblock {J}ohn {G}raunt and his {N}atural and political observations.
\newblock \emph{Proceedings of the Royal Society of London. Series B,
  Biological Sciences}, 159\penalty0 (974):\penalty0 2--37, 1963.

\bibitem[Gneiting and Raftery(2007)]{gneiting2007strictly}
T.~Gneiting and A.~E. Raftery.
\newblock Strictly proper scoring rules, prediction, and estimation.
\newblock \emph{Journal of the American Statistical Association}, 102\penalty0
  (477):\penalty0 359--378, 2007.

\bibitem[Goldstein et~al.(2020)Goldstein, Han, Puli, Perotte, and
  Ranganath]{goldstein2020x}
M.~Goldstein, X.~Han, A.~Puli, A.~Perotte, and R.~Ranganath.
\newblock {X-CAL}: Explicit calibration for survival analysis.
\newblock In \emph{Advances in Neural Information Processing Systems}, 2020.

\bibitem[Goodfellow et~al.(2014)Goodfellow, Pouget-Abadie, Mirza, Xu,
  Warde-Farley, Ozair, Courville, and Bengio]{goodfellow2014generative}
I.~Goodfellow, J.~Pouget-Abadie, M.~Mirza, B.~Xu, D.~Warde-Farley, S.~Ozair,
  A.~Courville, and Y.~Bengio.
\newblock Generative adversarial nets.
\newblock In \emph{Advances in Neural Information Processing Systems}, 2014.

\bibitem[Gordon and Olshen(1985)]{gordon1985tree}
L.~Gordon and R.~A. Olshen.
\newblock Tree-structured survival analysis.
\newblock \emph{Cancer Treatment Reports}, 69\penalty0 (10):\penalty0
  1065--1069, 1985.

\bibitem[Graf(1998)]{graf1998explained}
E.~Graf.
\newblock \emph{Explained variation measures for survival data}.
\newblock PhD thesis, University of Freiburg (in German), 1998.

\bibitem[Graf et~al.(1999)Graf, Schmoor, Sauerbrei, and
  Schumacher]{graf1999assessment}
E.~Graf, C.~Schmoor, W.~Sauerbrei, and M.~Schumacher.
\newblock Assessment and comparison of prognostic classification schemes for
  survival data.
\newblock \emph{Statistics in Medicine}, 18\penalty0 (17-18):\penalty0
  2529--2545, 1999.

\bibitem[Graunt(1662)]{graunt1662natural}
J.~Graunt.
\newblock Natural and political observations mentioned in a following index and
  made upon the bills of mortality, 1662.

\bibitem[Gray(1988)]{gray1988class}
R.~J. Gray.
\newblock A class of {$K$}-sample tests for comparing the cumulative incidence
  of a competing risk.
\newblock \emph{The Annals of Statistics}, pages 1141--1154, 1988.

\bibitem[Groha et~al.(2020)Groha, Schmon, and Gusev]{groha2020general}
S.~Groha, S.~M. Schmon, and A.~Gusev.
\newblock A general framework for survival analysis and multi-state modelling.
\newblock \emph{arXiv preprint arXiv:2006.04893}, 2020.

\bibitem[Haider et~al.(2020)Haider, Hoehn, Davis, and
  Greiner]{haider2020effective}
H.~Haider, B.~Hoehn, S.~Davis, and R.~Greiner.
\newblock Effective ways to build and evaluate individual survival
  distributions.
\newblock \emph{Journal of Machine Learning Research}, 21\penalty0
  (85):\penalty0 1--63, 2020.

\bibitem[Halley(1693)]{halley1693estimate}
E.~Halley.
\newblock An estimate of the degrees of the mortality of mankind; drawn from
  curious tables of the births and funerals at the city of breslaw; with an
  attempt to ascertain the price of annuities upon lives.
\newblock \emph{Philosophical Transactions of the Royal Society of London},
  17:\penalty0 596--610, 1693.

\bibitem[Harrell(2015)]{harrell2015regression}
F.~E. Harrell.
\newblock \emph{Regression Modeling Strategies: With Applications to Linear
  Models, Logistic and Ordinal regression, and Survival Analysis}.
\newblock Spinger, 2015.

\bibitem[Harrell et~al.(1982)Harrell, Califf, Pryor, Lee, and
  Rosati]{harrell1982evaluating}
F.~E. Harrell, R.~M. Califf, D.~B. Pryor, K.~L. Lee, and R.~A. Rosati.
\newblock Evaluating the yield of medical tests.
\newblock \emph{Journal of the American Medical Association}, 247\penalty0
  (18):\penalty0 2543--2546, 1982.

\bibitem[Harrell~Jr et~al.(1996)Harrell~Jr, Lee, and
  Mark]{harrell1996multivariable}
F.~E. Harrell~Jr, K.~L. Lee, and D.~B. Mark.
\newblock Multivariable prognostic models: issues in developing models,
  evaluating assumptions and adequacy, and measuring and reducing errors.
\newblock \emph{Statistics in Medicine}, 15\penalty0 (4):\penalty0 361--387,
  1996.

\bibitem[Hashimoto et~al.(2018)Hashimoto, Srivastava, Namkoong, and
  Liang]{hashimoto2018fairness}
T.~Hashimoto, M.~Srivastava, H.~Namkoong, and P.~Liang.
\newblock Fairness without demographics in repeated loss minimization.
\newblock In \emph{International Conference on Machine Learning}, pages
  1929--1938. PMLR, 2018.

\bibitem[Hawkes(1971)]{hawkes1971spectra}
A.~G. Hawkes.
\newblock Spectra of some self-exciting and mutually exciting point processes.
\newblock \emph{Biometrika}, 58\penalty0 (1):\penalty0 83--90, 1971.

\bibitem[He et~al.(2015)He, Zhang, Ren, and Sun]{he2015delving}
K.~He, X.~Zhang, S.~Ren, and J.~Sun.
\newblock Delving deep into rectifiers: Surpassing human-level performance on
  imagenet classification.
\newblock In \emph{Proceedings of the IEEE International Conference on Computer
  Vision}, pages 1026--1034, 2015.

\bibitem[Hochreiter and Schmidhuber(1997)]{hochreiter1997long}
S.~Hochreiter and J.~Schmidhuber.
\newblock Long short-term memory.
\newblock \emph{Neural Computation}, 9\penalty0 (8):\penalty0 1735--1780, 1997.

\bibitem[Hu and Chen(2024)]{hu2024fairness}
S.~Hu and G.~H. Chen.
\newblock Fairness in survival analysis with distributionally robust
  optimization.
\newblock \emph{Journal of Machine Learning Research}, 25\penalty0
  (246):\penalty0 1--85, 2024.

\bibitem[Hubbard et~al.(2021)Hubbard, Rostykus, Raimond, and
  Jebara]{hubbard2021beta}
D.~Hubbard, B.~Rostykus, Y.~Raimond, and T.~Jebara.
\newblock Beta survival models.
\newblock In \emph{Survival Prediction - Algorithms, Challenges and
  Applications}, pages 22--39. PMLR, 2021.

\bibitem[Huh et~al.(2011)Huh, Levi, Rusmevichientong, and
  Orlin]{huh2011adaptive}
W.~T. Huh, R.~Levi, P.~Rusmevichientong, and J.~B. Orlin.
\newblock Adaptive data-driven inventory control with censored demand based on
  {K}aplan-{M}eier estimator.
\newblock \emph{Operations Research}, 59\penalty0 (4):\penalty0 929--941, 2011.

\bibitem[Hung and Chiang(2010)]{hung2010estimation}
H.~Hung and C.-T. Chiang.
\newblock Estimation methods for time-dependent auc models with survival data.
\newblock \emph{Canadian Journal of Statistics}, 38\penalty0 (1):\penalty0
  8--26, 2010.

\bibitem[Ishwaran et~al.(2008)Ishwaran, Kogalur, Blackstone, and
  Lauer]{ishwaran2008random}
H.~Ishwaran, U.~B. Kogalur, E.~H. Blackstone, and M.~S. Lauer.
\newblock Random survival forests.
\newblock \emph{The Annals of Applied Statistics}, 2\penalty0 (3):\penalty0
  841--860, 2008.

\bibitem[Jeanselme et~al.(2022)Jeanselme, Tom, and
  Barrett]{jeanselme2022neural}
V.~Jeanselme, B.~Tom, and J.~Barrett.
\newblock Neural survival clustering: Non-parametric mixture of neural networks
  for survival clustering.
\newblock In \emph{Conference on Health, Inference, and Learning}, pages
  92--102. PMLR, 2022.

\bibitem[Jeanselme et~al.(2023)Jeanselme, Yoon, Tom, and
  Barrett]{jeanselme2023neural}
V.~Jeanselme, C.~H. Yoon, B.~Tom, and J.~Barrett.
\newblock Neural fine-gray: Monotonic neural networks for competing risks.
\newblock In \emph{Conference on Health, Inference, and Learning}, pages
  379--392. PMLR, 2023.

\bibitem[Jeanselme et~al.(2024)Jeanselme, Agarwal, and
  Wang]{jeanselme2024review}
V.~Jeanselme, N.~Agarwal, and C.~Wang.
\newblock Review of language models for survival analysis.
\newblock In \emph{AAAI 2024 Spring Symposium on Clinical Foundation Models},
  2024.

\bibitem[Kalbfleisch and Prentice(1980)]{kalbfleisch1980statistical}
J.~D. Kalbfleisch and R.~L. Prentice.
\newblock \emph{The Statistical Analysis of Failure Time Data}.
\newblock John Wiley \& Sons, 1980.

\bibitem[Kaplan and Meier(1958)]{kaplan1958nonparametric}
E.~L. Kaplan and P.~Meier.
\newblock Nonparametric estimation from incomplete observations.
\newblock \emph{Journal of the American Statistical Association}, 53\penalty0
  (282):\penalty0 457--481, 1958.

\bibitem[Katzman et~al.(2018)Katzman, Shaham, Cloninger, Bates, Jiang, and
  Kluger]{katzman2018deepsurv}
J.~L. Katzman, U.~Shaham, A.~Cloninger, J.~Bates, T.~Jiang, and Y.~Kluger.
\newblock Deepsurv: personalized treatment recommender system using a cox
  proportional hazards deep neural network.
\newblock \emph{BMC Medical Research Methodology}, 18\penalty0 (24), 2018.

\bibitem[Keya et~al.(2021)Keya, Islam, Pan, Stockwell, and
  Foulds]{keya2021equitable}
K.~N. Keya, R.~Islam, S.~Pan, I.~Stockwell, and J.~Foulds.
\newblock Equitable allocation of healthcare resources with fair survival
  models.
\newblock In \emph{Proceedings of the 2021 SIAM International Conference on
  Data Mining (SDM)}, pages 190--198. SIAM, 2021.

\bibitem[Kingma and Ba(2015)]{kingma2015adam}
D.~P. Kingma and J.~Ba.
\newblock Adam: A method for stochastic optimization.
\newblock In \emph{International Conference for Learning Representations},
  2015.

\bibitem[Klein and Moeschberger(2003)]{klein2003survival}
J.~P. Klein and M.~L. Moeschberger.
\newblock \emph{Survival Analysis: Techniques for Censored and Truncated Data,
  Second Edition}.
\newblock Springer, 2003.

\bibitem[Klein et~al.(2016)Klein, Van~Houwelingen, Ibrahim, and
  Scheike]{klein2016handbook}
J.~P. Klein, H.~C. Van~Houwelingen, J.~G. Ibrahim, and T.~H. Scheike.
\newblock \emph{Handbook of Survival Analysis}.
\newblock CRC Press, 2016.

\bibitem[Kleinbaum and Klein(2012)]{kleinbaum2012survival}
D.~G. Kleinbaum and M.~Klein.
\newblock \emph{Survival Analysis: A Self-Learning Text, Third Edition}.
\newblock Springer, 2012.

\bibitem[Knaus et~al.(1995)Knaus, Harrell, Lynn, Goldman, Phillips, Connors,
  Dawson, Fulkerson, Califf, Desbiens, Layde, Oye, Bellamy, Hakim, and
  Wagner]{knaus1995support}
W.~A. Knaus, F.~E. Harrell, J.~Lynn, L.~Goldman, R.~S. Phillips, A.~F. Connors,
  N.~V. Dawson, W.~J. Fulkerson, R.~M. Califf, N.~Desbiens, P.~Layde, R.~K.
  Oye, P.~E. Bellamy, R.~B. Hakim, and D.~P. Wagner.
\newblock The {SUPPORT} prognostic model: Objective estimates of survival for
  seriously ill hospitalized adults.
\newblock \emph{Annals of Internal Medicine}, 122\penalty0 (3):\penalty0
  191--203, 1995.

\bibitem[Kovalev et~al.(2020)Kovalev, Utkin, and Kasimov]{kovalev2020survlime}
M.~S. Kovalev, L.~V. Utkin, and E.~M. Kasimov.
\newblock {SurvLIME}: A method for explaining machine learning survival models.
\newblock \emph{Knowledge-Based Systems}, 2020.

\bibitem[Koziol and Jia(2009)]{koziol2009concordance}
J.~A. Koziol and Z.~Jia.
\newblock The concordance index {C} and the {M}ann--{W}hitney parameter
  pr($x>y$) with randomly censored data.
\newblock \emph{Biometrical Journal: Journal of Mathematical Methods in
  Biosciences}, 51\penalty0 (3):\penalty0 467--474, 2009.

\bibitem[Kpotufe and Verma(2017)]{kpotufe2017time}
S.~Kpotufe and N.~Verma.
\newblock Time-accuracy tradeoffs in kernel prediction: controlling prediction
  quality.
\newblock \emph{Journal of Machine Learning Research}, 2017.

\bibitem[Krzyzi{\'n}ski et~al.(2023)Krzyzi{\'n}ski, Spytek, Baniecki, and
  Biecek]{krzyzinski2023survshap}
M.~Krzyzi{\'n}ski, M.~Spytek, H.~Baniecki, and P.~Biecek.
\newblock {SurvSHAP(t)}: Time-dependent explanations of machine learning
  survival models.
\newblock \emph{Knowledge-Based Systems}, 2023.

\bibitem[Kvamme and Borgan(2021)]{kvamme2021continuous}
H.~Kvamme and {\O}.~Borgan.
\newblock Continuous and discrete-time survival prediction with neural
  networks.
\newblock \emph{Lifetime Data Analysis}, 27\penalty0 (4):\penalty0 710--736,
  2021.

\bibitem[Kvamme et~al.(2019)Kvamme, Borgan, and Scheel]{kvamme2019time}
H.~Kvamme, {\O}.~Borgan, and I.~Scheel.
\newblock Time-to-event prediction with neural networks and {C}ox regression.
\newblock \emph{Journal of Machine Learning Research}, 20\penalty0
  (129):\penalty0 1--30, 2019.

\bibitem[Lambert and Chevret(2016)]{lambert2016summary}
J.~Lambert and S.~Chevret.
\newblock Summary measure of discrimination in survival models based on
  cumulative/dynamic time-dependent roc curves.
\newblock \emph{Statistical Methods in Medical Research}, 25\penalty0
  (5):\penalty0 2088--2102, 2016.

\bibitem[Lee et~al.(2018)Lee, Zame, Yoon, and Van Der~Schaar]{lee2018deephit}
C.~Lee, W.~Zame, J.~Yoon, and M.~Van Der~Schaar.
\newblock Deephit: A deep learning approach to survival analysis with competing
  risks.
\newblock In \emph{Proceedings of the AAAI Conference on Artificial
  Intelligence}, 2018.

\bibitem[Lee et~al.(2019)Lee, Yoon, and Van Der~Schaar]{lee2019dynamic}
C.~Lee, J.~Yoon, and M.~Van Der~Schaar.
\newblock {Dynamic-DeepHit}: A deep learning approach for dynamic survival
  analysis with competing risks based on longitudinal data.
\newblock \emph{IEEE Transactions on Biomedical Engineering}, 67\penalty0
  (1):\penalty0 122--133, 2019.

\bibitem[Lei et~al.(2015)Lei, Rinaldo, and Wasserman]{lei2015conformal}
J.~Lei, A.~Rinaldo, and L.~Wasserman.
\newblock A conformal prediction approach to explore functional data.
\newblock \emph{Annals of Mathematics and Artificial Intelligence}, 74\penalty0
  (1-2):\penalty0 29--43, 2015.

\bibitem[Lei et~al.(2018)Lei, G’Sell, Rinaldo, Tibshirani, and
  Wasserman]{lei2018distribution}
J.~Lei, M.~G’Sell, A.~Rinaldo, R.~J. Tibshirani, and L.~Wasserman.
\newblock Distribution-free predictive inference for regression.
\newblock \emph{Journal of the American Statistical Association}, 113\penalty0
  (523):\penalty0 1094--1111, 2018.

\bibitem[Li and Ma(2013)]{li2013survival}
J.~Li and S.~Ma.
\newblock \emph{Survival Analysis in Medicine and Genetics}.
\newblock CRC Press, 2013.

\bibitem[Li et~al.(2021)Li, Namkoong, and Xia]{li2021evaluating}
M.~Li, H.~Namkoong, and S.~Xia.
\newblock Evaluating model performance under worst-case subpopulations.
\newblock In \emph{Advances in Neural Information Processing Systems}, 2021.

\bibitem[Liu et~al.(2021)Liu, Lin, Liu, Xiong, Sch{\"o}lkopf, and
  Weller]{liu2021learning}
W.~Liu, R.~Lin, Z.~Liu, L.~Xiong, B.~Sch{\"o}lkopf, and A.~Weller.
\newblock Learning with hyperspherical uniformity.
\newblock In \emph{International Conference On Artificial Intelligence and
  Statistics}, pages 1180--1188. PMLR, 2021.

\bibitem[Lundberg and Lee(2017)]{lundberg2017unified}
S.~M. Lundberg and S.-I. Lee.
\newblock A unified approach to interpreting model predictions.
\newblock In \emph{Advances in Neural Information Processing Systems}, 2017.

\bibitem[Machin et~al.(2006)Machin, Cheung, and Parmar]{machin2006survival}
D.~Machin, Y.~B. Cheung, and M.~Parmar.
\newblock \emph{Survival Analysis: A Practical Approach}.
\newblock John Wiley \& Sons, 2006.

\bibitem[Manduchi et~al.(2022)Manduchi, Marcinkevi{\v{c}}s, Massi, Weikert,
  Sauter, Gotta, M{\"u}ller, Vasella, Neidert, Pfister, Stieltjes, and
  Vogt]{manduchi2022deep}
L.~Manduchi, R.~Marcinkevi{\v{c}}s, M.~C. Massi, T.~Weikert, A.~Sauter,
  V.~Gotta, T.~M{\"u}ller, F.~Vasella, M.~C. Neidert, M.~Pfister, B.~Stieltjes,
  and J.~E. Vogt.
\newblock A deep variational approach to clustering survival data.
\newblock In \emph{International Conference on Learning Representations}, 2022.

\bibitem[Mann et~al.(1974)Mann, Schafer, and Singpurwalla]{mann1974methods}
N.~R. Mann, R.~E. Schafer, and N.~D. Singpurwalla.
\newblock \emph{Methods for Statistical Analysis of Reliability and Life Data}.
\newblock John Wiley \& Sons, 1974.

\bibitem[Mantel(1966)]{mantel1966evaluation}
N.~Mantel.
\newblock Evaluation of survival data and two new rank order statistics arising
  in its consideration.
\newblock \emph{Cancer Chemotherapy Reports}, 50\penalty0 (3):\penalty0
  163--170, 1966.

\bibitem[Molnar(2022)]{molnar2022}
C.~Molnar.
\newblock \emph{Interpretable Machine Learning}.
\newblock 2 edition, 2022.
\newblock URL \url{https://christophm.github.io/interpretable-ml-book}.

\bibitem[Monod et~al.(2024)Monod, Krusche, Cao, Sahiner, Petrick, Ohlssen, and
  Coroller]{monod2024torchsurv}
M.~Monod, P.~Krusche, Q.~Cao, B.~Sahiner, N.~Petrick, D.~Ohlssen, and
  T.~Coroller.
\newblock Torchsurv: A lightweight package for deep survival analysis, 2024.

\bibitem[Moon et~al.(2022)Moon, Groha, and Gusev]{moon2022survlatent}
I.~Moon, S.~Groha, and A.~Gusev.
\newblock {SurvLatent ODE}: A neural ode based time-to-event model with
  competing risks for longitudinal data improves cancer-associated venous
  thromboembolism (vte) prediction.
\newblock In \emph{Machine Learning for Healthcare Conference}, 2022.

\bibitem[Nagpal et~al.(2021{\natexlab{a}})Nagpal, Li, and
  Dubrawski]{nagpal2021dsm}
C.~Nagpal, X.~Li, and A.~Dubrawski.
\newblock Deep survival machines: Fully parametric survival regression and
  representation learning for censored data with competing risks.
\newblock \emph{IEEE Journal of Biomedical and Health Informatics}, 25\penalty0
  (8):\penalty0 3163--3175, 2021{\natexlab{a}}.

\bibitem[Nagpal et~al.(2021{\natexlab{b}})Nagpal, Yadlowsky, Rostamzadeh, and
  Heller]{nagpal2021deep}
C.~Nagpal, S.~Yadlowsky, N.~Rostamzadeh, and K.~Heller.
\newblock Deep {C}ox mixtures for survival regression.
\newblock In \emph{Machine Learning for Healthcare Conference}, pages 674--708.
  PMLR, 2021{\natexlab{b}}.

\bibitem[Nagpal et~al.(2022{\natexlab{a}})Nagpal, Goswami, Dufendach, and
  Dubrawski]{nagpal2022counterfactual}
C.~Nagpal, M.~Goswami, K.~Dufendach, and A.~Dubrawski.
\newblock Counterfactual phenotyping with censored time-to-events.
\newblock In \emph{ACM SIGKDD Conference on Knowledge Discovery and Data
  Mining}, pages 3634--3644, 2022{\natexlab{a}}.

\bibitem[Nagpal et~al.(2022{\natexlab{b}})Nagpal, Potosnak, and
  Dubrawski]{nagpal2022auton}
C.~Nagpal, W.~Potosnak, and A.~Dubrawski.
\newblock auton-survival: An open-source package for regression, counterfactual
  estimation, evaluation and phenotyping with censored time-to-event data.
\newblock In \emph{Machine Learning for Healthcare Conference}, pages 585--608.
  PMLR, 2022{\natexlab{b}}.

\bibitem[Namboodiri and Suchindran(2013)]{namboodiri2013life}
K.~Namboodiri and C.~M. Suchindran.
\newblock \emph{Life Table Techniques and Their Applications}.
\newblock Academic Press, 2013.

\bibitem[Nelson(1969)]{nelson1969hazard}
W.~Nelson.
\newblock Hazard plotting for incomplete failure data.
\newblock \emph{Journal of Quality Technology}, 1:\penalty0 27--52, 1969.

\bibitem[Papadopoulos et~al.(2002)Papadopoulos, Proedrou, Vovk, and
  Gammerman]{papadopoulos2002inductive}
H.~Papadopoulos, K.~Proedrou, V.~Vovk, and A.~Gammerman.
\newblock Inductive confidence machines for regression.
\newblock In \emph{European Conference on Machine Learning}, pages 345--356.
  Springer, 2002.

\bibitem[Paszke et~al.(2019)Paszke, Gross, Massa, Lerer, Bradbury, Chanan,
  Killeen, Lin, Gimelshein, Antiga, Desmaison, Kopf, Yang, DeVito, Raison,
  Tejani, Chilamkurthy, Steiner, Fang, Bai, and Chintala]{paszke2019pytorch}
A.~Paszke, S.~Gross, F.~Massa, A.~Lerer, J.~Bradbury, G.~Chanan, T.~Killeen,
  Z.~Lin, N.~Gimelshein, L.~Antiga, A.~Desmaison, A.~Kopf, E.~Yang, Z.~DeVito,
  M.~Raison, A.~Tejani, S.~Chilamkurthy, B.~Steiner, L.~Fang, J.~Bai, and
  S.~Chintala.
\newblock {PyTorch}: An imperative style, high-performance deep learning
  library.
\newblock In \emph{Advances in Neural Information Processing Systems}, 2019.

\bibitem[Peng and Yu(2021)]{peng2021cure}
Y.~Peng and B.~Yu.
\newblock \emph{Cure Models: Methods, Applications, and Implementation}.
\newblock CRC Press, 2021.

\bibitem[P{\"o}lsterl(2020)]{polsterl2020scikit}
S.~P{\"o}lsterl.
\newblock scikit-survival: A library for time-to-event analysis built on top of
  scikit-learn.
\newblock \emph{Journal of Machine Learning Research}, 21\penalty0
  (1):\penalty0 8747--8752, 2020.

\bibitem[Prentice and Kalbfleisch(1979)]{prentice1979hazard}
R.~L. Prentice and J.~D. Kalbfleisch.
\newblock Hazard rate models with covariates.
\newblock \emph{Biometrics}, pages 25--39, 1979.

\bibitem[Prentice and Zhao(2019)]{prentice2019statistical}
R.~L. Prentice and S.~Zhao.
\newblock \emph{The Statistical Analysis of Multivariate Failure Time Data: A
  Marginal Modeling Approach}.
\newblock CRC Press, 2019.

\bibitem[Putzel et~al.(2021)Putzel, Do, Boyd, Zhong, and
  Smyth]{putzel2021dynamic}
P.~Putzel, H.~Do, A.~Boyd, H.~Zhong, and P.~Smyth.
\newblock Dynamic survival analysis for {EHR} data with personalized parametric
  distributions.
\newblock In \emph{Machine Learning for Healthcare Conference}, pages 648--673.
  PMLR, 2021.

\bibitem[Qi et~al.(2023)Qi, Kumar, Farrokh, Sun, Kuan, Ranganath, Henao, and
  Greiner]{qi2023effective}
S.-A. Qi, N.~Kumar, M.~Farrokh, W.~Sun, L.-H. Kuan, R.~Ranganath, R.~Henao, and
  R.~Greiner.
\newblock An effective meaningful way to evaluate survival models.
\newblock In \emph{International Conference on Machine Learning}, volume 202,
  pages 28244--28276. PMLR, 2023.

\bibitem[Qi et~al.(2024{\natexlab{a}})Qi, Sun, and Greiner]{qi2024survivaleval}
S.-a. Qi, W.~Sun, and R.~Greiner.
\newblock {SurvivalEVAL}: A comprehensive open-source python package for
  evaluating individual survival distributions.
\newblock In \emph{Proceedings of the 2023 AAAI Fall Symposia},
  2024{\natexlab{a}}.

\bibitem[Qi et~al.(2024{\natexlab{b}})Qi, Yu, and Greiner]{qi2024conformalized}
S.-a. Qi, Y.~Yu, and R.~Greiner.
\newblock Conformalized survival distributions: A generic post-process to
  increase calibration.
\newblock In \emph{International Conference on Machine Learning}, volume 235 of
  \emph{Proceedings of Machine Learning Research}, pages 41303--41339. PMLR,
  2024{\natexlab{b}}.

\bibitem[{R Core Team}(2021)]{r2021r}
{R Core Team}.
\newblock \emph{R: A Language and Environment for Statistical Computing}.
\newblock R Foundation for Statistical Computing, Vienna, Austria, 2021.
\newblock URL \url{https://www.R-project.org/}.

\bibitem[Radford et~al.(2021)Radford, Kim, Hallacy, Ramesh, Goh, Agarwal,
  Sastry, Askell, Mishkin, Clark, Krueger, and Sutskever]{radford2021learning}
A.~Radford, J.~W. Kim, C.~Hallacy, A.~Ramesh, G.~Goh, S.~Agarwal, G.~Sastry,
  A.~Askell, P.~Mishkin, J.~Clark, G.~Krueger, and I.~Sutskever.
\newblock Learning transferable visual models from natural language
  supervision.
\newblock In \emph{International Conference on Machine Learning}, pages
  8748--8763. PMLR, 2021.

\bibitem[Rahman and Purushotham(2022)]{rahman2022fair}
M.~M. Rahman and S.~Purushotham.
\newblock Fair and interpretable models for survival analysis.
\newblock In \emph{ACM SIGKDD Conference on Knowledge Discovery and Data
  Mining}, pages 1452--1462, 2022.

\bibitem[Raykar et~al.(2007)Raykar, Steck, Krishnapuram, Dehing-oberije, and
  Lambin]{raykar2007ranking}
V.~C. Raykar, H.~Steck, B.~Krishnapuram, C.~Dehing-oberije, and P.~Lambin.
\newblock On ranking in survival analysis: Bounds on the concordance index.
\newblock In \emph{Advances in Neural Information Processing Systems}, 2007.

\bibitem[Ribeiro et~al.(2016)Ribeiro, Singh, and Guestrin]{ribeiro2016should}
M.~T. Ribeiro, S.~Singh, and C.~Guestrin.
\newblock ``why should {I} trust you?'' {E}xplaining the predictions of any
  classifier.
\newblock In \emph{ACM SIGKDD International Conference on Knowledge Discovery
  and Data Mining}, pages 1135--1144, 2016.

\bibitem[Rindt et~al.(2022)Rindt, Hu, Steinsaltz, and
  Sejdinovic]{rindt2022survival}
D.~Rindt, R.~Hu, D.~Steinsaltz, and D.~Sejdinovic.
\newblock Survival regression with proper scoring rules and monotonic neural
  networks.
\newblock In \emph{International Conference on Artificial Intelligence and
  Statistics}, pages 1190--1205. PMLR, 2022.

\bibitem[Samuel(1959)]{samuel1959some}
A.~L. Samuel.
\newblock Some studies in machine learning using the game of checkers.
\newblock \emph{IBM Journal of Research and Development}, 1959.

\bibitem[Schumacher et~al.(1994)Schumacher, Bastert, Bojar, Huebner,
  Olschewski, Sauerbrei, Schmoor, Beyerle, Neumann, and
  Rauschecker]{schumacher1994randomized}
M.~Schumacher, G.~Bastert, H.~Bojar, K.~Huebner, M.~Olschewski, W.~Sauerbrei,
  C.~Schmoor, C.~Beyerle, R.~L. Neumann, and H.~F. Rauschecker.
\newblock Randomized 2 x 2 trial evaluating hormonal treatment and the duration
  of chemotherapy in node-positive breast cancer patients. german breast cancer
  study group.
\newblock \emph{Journal of Clinical Oncology}, 12\penalty0 (10):\penalty0
  2086--2093, 1994.

\bibitem[Selvin(2008)]{selvin2008survival}
S.~Selvin.
\newblock \emph{Survival Analysis for Epidemiologic and Medical Research}.
\newblock Cambridge University Press, 2008.

\bibitem[Shchur et~al.(2020)Shchur, Bilo{\v{s}}, and
  G{\"u}nnemann]{shchur2020intensity}
O.~Shchur, M.~Bilo{\v{s}}, and S.~G{\"u}nnemann.
\newblock Intensity-free learning of temporal point processes.
\newblock In \emph{International Conference on Learning Representations}, 2020.

\bibitem[Shen et~al.(2023)Shen, Elmer, and Chen]{shen2023neurological}
X.~Shen, J.~Elmer, and G.~H. Chen.
\newblock Neurological prognostication of post-cardiac-arrest coma patients
  using eeg data: A dynamic survival analysis framework with competing risks.
\newblock In \emph{Machine Learning for Healthcare Conference}, pages 667--690.
  PMLR, 2023.

\bibitem[Simon et~al.(2011)Simon, Friedman, Hastie, and
  Tibshirani]{simon2011regularization}
N.~Simon, J.~Friedman, T.~Hastie, and R.~Tibshirani.
\newblock Regularization paths for cox’s proportional hazards model via
  coordinate descent.
\newblock \emph{Journal of Statistical Software}, 39\penalty0 (5):\penalty0 1,
  2011.

\bibitem[Steinberg et~al.(2024)Steinberg, Fries, Xu, and
  Shah]{steinberg2024self}
E.~Steinberg, J.~A. Fries, Y.~Xu, and N.~Shah.
\newblock {MOTOR}: A time-to-event foundation model for structured medical
  records.
\newblock In \emph{International Conference on Learning Representations}, 2024.

\bibitem[Sun and Qiu(2023)]{sun2023nsotree}
X.~Sun and P.~Qiu.
\newblock {NSOTree}: Neural survival oblique tree.
\newblock \emph{arXiv preprint arXiv:2309.13825}, 2023.

\bibitem[Tang et~al.(2022{\natexlab{a}})Tang, He, Xu, and
  Zhu]{tang2022survival}
W.~Tang, K.~He, G.~Xu, and J.~Zhu.
\newblock Survival analysis via ordinary differential equations.
\newblock \emph{Journal of the American Statistical Association},
  2022{\natexlab{a}}.

\bibitem[Tang et~al.(2022{\natexlab{b}})Tang, Ma, Mei, and Zhu]{tang2022soden}
W.~Tang, J.~Ma, Q.~Mei, and J.~Zhu.
\newblock {SODEN}: A scalable continuous-time survival model through ordinary
  differential equation networks.
\newblock \emph{Journal of Machine Learning Research}, 23\penalty0
  (34):\penalty0 1--29, 2022{\natexlab{b}}.

\bibitem[Tibshirani(1997)]{tibshirani1997lasso}
R.~Tibshirani.
\newblock The lasso method for variable selection in the cox model.
\newblock \emph{Statistics in Medicine}, 16\penalty0 (4):\penalty0 385--395,
  1997.

\bibitem[Tutz and Schmid(2016)]{tutz2016modeling}
G.~Tutz and M.~Schmid.
\newblock \emph{Modeling Discrete Time-to-Event Data}.
\newblock Springer, 2016.

\bibitem[Uno et~al.(2007)Uno, Cai, Tian, and Wei]{uno2007evaluating}
H.~Uno, T.~Cai, L.~Tian, and L.-J. Wei.
\newblock Evaluating prediction rules for t-year survivors with censored
  regression models.
\newblock \emph{Journal of the American Statistical Association}, 102\penalty0
  (478):\penalty0 527--537, 2007.

\bibitem[Uno et~al.(2011)Uno, Cai, Pencina, D'Agostino, and Wei]{uno2011c}
H.~Uno, T.~Cai, M.~J. Pencina, R.~B. D'Agostino, and L.-J. Wei.
\newblock On the c-statistics for evaluating overall adequacy of risk
  prediction procedures with censored survival data.
\newblock \emph{Statistics in Medicine}, 30\penalty0 (10):\penalty0 1105--1117,
  2011.

\bibitem[Van~der Maaten and Hinton(2008)]{van2008visualizing}
L.~Van~der Maaten and G.~Hinton.
\newblock Visualizing data using {t-SNE}.
\newblock \emph{Journal of Machine Learning Research}, 9\penalty0 (11), 2008.

\bibitem[Vershynin(2018)]{vershynin2018high}
R.~Vershynin.
\newblock \emph{High-Dimensional Probability: An Introduction with Applications
  in Data Science}.
\newblock Cambridge University Press, 2018.

\bibitem[Virtanen et~al.(2020)Virtanen, Gommers, Oliphant, Haberland, Reddy,
  Cournapeau, Burovski, Peterson, Weckesser, Bright, {van der Walt}, Brett,
  Wilson, Millman, Mayorov, Nelson, Jones, Kern, Larson, Carey, Polat, Feng,
  Moore, {VanderPlas}, Laxalde, Perktold, Cimrman, Henriksen, Quintero, Harris,
  Archibald, Ribeiro, Pedregosa, {van Mulbregt}, and {SciPy 1.0
  Contributors}]{virtanen2020scipy}
P.~Virtanen, R.~Gommers, T.~E. Oliphant, M.~Haberland, T.~Reddy, D.~Cournapeau,
  E.~Burovski, P.~Peterson, W.~Weckesser, J.~Bright, S.~J. {van der Walt},
  M.~Brett, J.~Wilson, K.~J. Millman, N.~Mayorov, A.~R.~J. Nelson, E.~Jones,
  R.~Kern, E.~Larson, C.~J. Carey, {\.I}.~Polat, Y.~Feng, E.~W. Moore,
  J.~{VanderPlas}, D.~Laxalde, J.~Perktold, R.~Cimrman, I.~Henriksen, E.~A.
  Quintero, C.~R. Harris, A.~M. Archibald, A.~H. Ribeiro, F.~Pedregosa, P.~{van
  Mulbregt}, and {SciPy 1.0 Contributors}.
\newblock {{SciPy} 1.0: Fundamental Algorithms for Scientific Computing in
  Python}.
\newblock \emph{Nature Methods}, 17:\penalty0 261--272, 2020.
\newblock \doi{10.1038/s41592-019-0686-2}.

\bibitem[Vovk et~al.(2005)Vovk, Gammerman, and Shafer]{vovk2005algorithmic}
V.~Vovk, A.~Gammerman, and G.~Shafer.
\newblock \emph{Algorithmic Learning in a Random World}.
\newblock Springer Science \& Business Media, 2005.

\bibitem[Wang et~al.(2019)Wang, Li, and Reddy]{wang2019machine}
P.~Wang, Y.~Li, and C.~K. Reddy.
\newblock Machine learning for survival analysis: A survey.
\newblock \emph{ACM Computing Surveys (CSUR)}, 51\penalty0 (6):\penalty0 1--36,
  2019.

\bibitem[Wang and Isola(2020)]{wang2020understanding}
T.~Wang and P.~Isola.
\newblock Understanding contrastive representation learning through alignment
  and uniformity on the hypersphere.
\newblock In \emph{International Conference on Machine Learning}, 2020.

\bibitem[Wiegrebe et~al.(2023)Wiegrebe, Kopper, Sonabend, and
  Bender]{wiegrebe2023deep}
S.~Wiegrebe, P.~Kopper, R.~Sonabend, and A.~Bender.
\newblock Deep learning for survival analysis: A review.
\newblock \emph{arXiv preprint arXiv:2305.14961}, 2023.

\bibitem[Xu and Peng(2014)]{xu2014nonparametric}
J.~Xu and Y.~Peng.
\newblock Nonparametric cure rate estimation with covariates.
\newblock \emph{Canadian Journal of Statistics}, 42\penalty0 (1):\penalty0
  1--17, 2014.

\bibitem[Xu et~al.(2023)Xu, Ignatiadis, Sverdrup, Fleming, Wager, and
  Shah]{xu2023treatment}
Y.~Xu, N.~Ignatiadis, E.~Sverdrup, S.~Fleming, S.~Wager, and N.~Shah.
\newblock Treatment heterogeneity with survival outcomes.
\newblock In \emph{Handbook of Matching and Weighting Adjustments for Causal
  Inference}, pages 445--482. Chapman and Hall/CRC, 2023.

\bibitem[Yanagisawa et~al.(2022)Yanagisawa, Miyaguchi, and
  Katsuki]{yanagisawa2022hierarchical}
H.~Yanagisawa, K.~Miyaguchi, and T.~Katsuki.
\newblock Hierarchical lattice layer for partially monotone neural networks.
\newblock In \emph{Advances in Neural Information Processing Systems}, 2022.

\bibitem[Yu et~al.(2011)Yu, Greiner, Lin, and Baracos]{yu2011learning}
C.-N. Yu, R.~Greiner, H.-C. Lin, and V.~Baracos.
\newblock Learning patient-specific cancer survival distributions as a sequence
  of dependent regressors.
\newblock In \emph{Advances in Neural Information Processing Systems}, 2011.

\bibitem[Zhang et~al.(2023)Zhang, Lipton, Li, and Smola]{zhang2023dive}
A.~Zhang, Z.~C. Lipton, M.~Li, and A.~J. Smola.
\newblock \emph{Dive into Deep Learning}.
\newblock Cambridge University Press, 2023.
\newblock \url{https://D2L.ai}.

\bibitem[Zhang et~al.(2020)Zhang, Lipani, Kirnap, and Yilmaz]{zhang2020self}
Q.~Zhang, A.~Lipani, O.~Kirnap, and E.~Yilmaz.
\newblock Self-attentive {H}awkes process.
\newblock In \emph{International Conference on Machine Learning}, pages
  11183--11193. PMLR, 2020.

\bibitem[Zhang et~al.(2024)Zhang, Xin, Seltzer, and Rudin]{zhang2024optimal}
R.~Zhang, R.~Xin, M.~Seltzer, and C.~Rudin.
\newblock Optimal sparse survival trees.
\newblock In \emph{International Conference on Artificial Intelligence and
  Statistics}, pages 352--360. PMLR, 2024.

\bibitem[Zhang and Weiss(2022)]{zhang2022longitudinal}
W.~Zhang and J.~C. Weiss.
\newblock Longitudinal fairness with censorship.
\newblock In \emph{Proceedings of the AAAI Conference on Artificial
  Intelligence}, 2022.

\bibitem[Zhong et~al.(2021)Zhong, Mueller, and Wang]{zhong2021deep}
Q.~Zhong, J.~W. Mueller, and J.-L. Wang.
\newblock Deep extended hazard models for survival analysis.
\newblock In \emph{Advances in Neural Information Processing Systems}, 2021.

\bibitem[Zhong et~al.(2022)Zhong, Mueller, and Wang]{zhong2022deep}
Q.~Zhong, J.~Mueller, and J.-L. Wang.
\newblock Deep learning for the partially linear {C}ox model.
\newblock \emph{The Annals of Statistics}, 50\penalty0 (3):\penalty0
  1348--1375, 2022.

\bibitem[Zou and Hastie(2005)]{zou2005regularization}
H.~Zou and T.~Hastie.
\newblock Regularization and variable selection via the elastic net.
\newblock \emph{Journal of the Royal Statistical Society Series B}, 67\penalty0
  (2):\penalty0 301--320, 2005.

\bibitem[Zuo et~al.(2020)Zuo, Jiang, Li, Zhao, and Zha]{zuo2020transformer}
S.~Zuo, H.~Jiang, Z.~Li, T.~Zhao, and H.~Zha.
\newblock Transformer {H}awkes process.
\newblock In \emph{International Conference on Machine Learning}, pages
  11692--11702. PMLR, 2020.

\bibitem[Zupan et~al.(2000)Zupan, Dem{\v{s}}ar, Kattan, Beck, and
  Bratko]{zupan2000machine}
B.~Zupan, J.~Dem{\v{s}}ar, M.~W. Kattan, J.~R. Beck, and I.~Bratko.
\newblock Machine learning for survival analysis: a case study on recurrence of
  prostate cancer.
\newblock \emph{Artificial Intelligence in Medicine}, 20\penalty0 (1):\penalty0
  59--75, 2000.

\end{thebibliography}

\end{document}